\documentclass[10pt,journal,romanappendices]{IEEEtran}
\pdfoutput=1
\usepackage{amsfonts}
\usepackage[mathcal]{euscript} 
\usepackage[cmex10]{amsmath}
\usepackage{amssymb}
\usepackage{graphics,subfig,epsfig}
\usepackage{verbatim}
\usepackage{cite,calc} 
\usepackage{algorithm,algorithmicx,algpseudocode}
\usepackage{subdepth}  
\usepackage{bm} 
\usepackage{upgreek}
\usepackage{stackengine} \stackMath

\usepackage{hyperref}
\hypersetup{colorlinks = true, citecolor = blue}

\usepackage{color}

\DeclareMathAlphabet{\mathbbb}{U}{bbold}{m}{n}  

\newtheorem{theorem}{Theorem}
\newtheorem{lemma}[theorem]{Lemma}
\newtheorem{proposition}[theorem]{Proposition}
\newtheorem{corollary}[theorem]{Corollary}
\newtheorem{definition}[theorem]{Definition}
\newtheorem{remark}[theorem]{Remark}

\newtheorem{fact}[theorem]{Fact}

\newtheorem{example}{Example}

\newcommand{\ds}{\displaystyle}
\newcommand{\secref}[1]{Section~\ref{#1}}
\newcommand{\appref}[1]{Appendix~\ref{#1}}

\newcommand{\figref}[1]{Fig.~\ref{#1}}
\newcommand{\lemref}[1]{Lemma~\ref{#1}}

\newcommand{\propref}[1]{Proposition~\ref{#1}}
\newcommand{\corolref}[1]{Corollary~\ref{#1}}
\newcommand{\factref}[1]{Fact~\ref{#1}}
\newcommand{\defref}[1]{Definition~\ref{#1}}
\newcommand{\algoref}[1]{Algorithm~\ref{#1}}

\DeclareMathOperator*{\argmax}{arg\,max}
\DeclareMathOperator*{\argmin}{arg\,min}
\DeclareMathOperator*{\relint}{relint}
\DeclareMathOperator*{\tr}{tr}
\DeclareMathOperator*{\sgn}{sgn}
\DeclareMathOperator*{\cov}{cov}
\DeclareMathOperator*{\diag}{diag}
\DeclareMathOperator*{\rank}{rank}
\DeclareMathOperator*{\nullspace}{null}

\newcommand\tsup[2][2]{%
 \def\useanchorwidth{T}%
  \ifnum#1>1%
    \stackon[-1.3ex]{\tsup[\numexpr#1-1\relax]{#2}}{\mathchar"307E\kern-.5pt}%
  \else%
    \stackon[-1ex]{#2}{\mathchar"307E\kern-.5pt}%
  \fi%
}

\newcommand{\markov}{\leftrightarrow}
\newcommand{\gauss}{\mathtt{N}}
\newcommand{\kron}{\mathbbb{1}}
\newcommand{\ordinal}{\mathbbb{I}}
\newcommand{\diff}{\mathrm{d}}

\newcommand{\eps}{\epsilon}
\newcommand{\epsX}{\eps_X}
\newcommand{\epsY}{\eps_Y}
\newcommand{\epst}{{\tilde{\eps}}}
\newcommand{\la}{\lambda}

\newcommand{\bA}{\mathbf{A}}
\newcommand{\bAt}{\tilde{\bA}}
\newcommand{\bB}{\mathbf{B}}

\newcommand{\cbar}{\bar{c}}

\newcommand{\Db}{\bar{D}}
\newcommand{\bM}{\mathbf{M}}

\newcommand{\bW}{\mathbf{W}}
\newcommand{\bZ}{\mathbf{Z}}
\newcommand{\bZt}{\tilde{\mathbf{Z}}}
\newcommand{\zh}{\hat{z}}

\newcommand{\cE}{\mathcal{E}}

\newcommand{\Hh}{\hat{H}}
\newcommand{\Ih}{\hat{I}}
\newcommand{\ba}{{\mathbf{a}}}

\newcommand{\bc}{{\mathbf{c}}}

\newcommand{\C}{{\mathcal{C}}}
\newcommand{\Cg}{\C} 
\newcommand{\Cgm}{\bar{\C}} 
\newcommand{\st}{{\tilde{s}}}
\newcommand{\St}{{\tilde{S}}}

\newcommand{\Ut}{{\tilde{U}}}

\newcommand{\Vt}{{\tilde{V}}}

\newcommand{\wt}{{\tilde{w}}}
\newcommand{\Wt}{{\tilde{W}}}
\newcommand{\Wc}{{\check{W}}}
\newcommand{\cWt}{{\tilde{\W}}}

\newcommand{\xt}{{\tilde{x}}}
\newcommand{\Xt}{{\tilde{X}}}

\newcommand{\Yh}{{\hat{Y}}}
\newcommand{\yt}{{\tilde{y}}}
\newcommand{\Yt}{{\tilde{Y}}}

\newcommand{\Zh}{{\hat{Z}}}
\newcommand{\Zt}{{\tilde{Z}}}
\newcommand{\At}{{\tilde{A}}}
\newcommand{\Ct}{{\tilde{C}}}
\newcommand{\cT}{{\mathcal{T}}}
\newcommand{\U}{{\mathcal{U}}}
\newcommand{\V}{{\mathcal{V}}}
\newcommand{\W}{{\mathcal{W}}}
\newcommand{\Wg}{{\reals^{\dimW}}}
\newcommand{\X}{{\mathcal{X}}}

\newcommand{\mindim}{K} 
\newcommand{\dimU}{{K_U}}
\newcommand{\dimV}{{K_V}}
\newcommand{\dimW}{{K_W}}
\newcommand{\dimX}{{K_X}}
\newcommand{\dimY}{{K_Y}}
\newcommand{\dimZ}{{K_Z}}
\newcommand{\dimA}{{K_A}}
\newcommand{\Ug}{{\reals^{\dimU}}}
\newcommand{\Vg}{{\reals^{\dimV}}}
\newcommand{\Xg}{{\reals^{\dimX}}}
\newcommand{\Y}{{\mathcal{Y}}}
\newcommand{\Yg}{{\reals^{\dimY}}}
\newcommand{\Z}{{\mathcal{Z}}}
\newcommand{\Zg}{{\reals^{\dimZ}}}
\newcommand{\uE}{{E_-}}
\newcommand{\cA}{\mathcal{A}}
\newcommand{\cB}{\mathcal{B}}
\newcommand{\cF}{\mathcal{F}}
\newcommand{\cG}{\mathcal{G}}
\newcommand{\cL}{\mathcal{L}}
\newcommand{\cM}{\mathcal{M}}
\newcommand{\cN}{\mathcal{N}}
\newcommand{\cP}{\mathcal{P}}
\newcommand{\cPt}{\tilde{\cP}}

\newcommand{\cS}{\mathcal{S}}
\newcommand{\cI}{\mathcal{I}}

\newcommand{\cU}{\U}
\newcommand{\cV}{\V}

\newcommand{\cYh}{\hat{\Y}}
\newcommand{\cYb}{\bar{\Y}}

\newcommand{\ft}{\tilde{f}}
\newcommand{\fb}{\bar{f}}
\newcommand{\fh}{\hat{f}}
\newcommand{\fc}{\check{f}}
\newcommand{\bfe}{\mathbf{e}}
\newcommand{\bff}{\mathbf{f}}
\newcommand{\bfF}{\mathbf{F}}

\newcommand{\bFb}{\bar{\bfF}}
\newcommand{\bFh}{\hat{\bfF}}

\newcommand{\gb}{\bar{g}}
\newcommand{\gh}{\hat{g}}
\newcommand{\gt}{\tilde{g}}
\newcommand{\gc}{\check{g}}
\newcommand{\bg}{\mathbf{g}}

\newcommand{\bGb}{\bar{\bG}}
\newcommand{\bGh}{\hat{\bG}}

\newcommand{\hti}{{\tilde{h}}}
\newcommand{\sigmat}{{\tilde{\sigma}}}
\newcommand{\sigmah}{{\hat{\sigma}}}

\newcommand{\e}{\mathrm{e}}

\newcommand{\ev}{{\mathbb{E}}}
\newcommand{\pr}{\mathbb{P}}
\newcommand{\prob}[1]{\pr\left\{#1\right\}}
\newcommand{\probd}[2]{\pr_{#1}\left\{#2\right\}}
\newcommand{\E}[1]{\ev\left[#1\right]}
\newcommand{\bE}[1]{\ev\bigl[#1\bigr]}

\newcommand{\Ed}[2]{\ev_{#1}\left[#2\right]}
\newcommand{\bEd}[2]{\ev_{#1}\bigl[#2\bigr]}
\newcommand{\pe}{p_\mathrm{e}}

\newcommand{\Eb}{\bar{E}}
\newcommand{\mse}{\la_\mathrm{e}}
\newcommand{\mseb}{\bar{\la}_\mathrm{e}}
\newcommand{\mset}{\tilde{\la}_\mathrm{e}}
\DeclareMathOperator{\var}{var}
\newcommand{\varop}[1]{\var\left[#1\right]}
\newcommand{\varopd}[2]{\var_{#1}\left[#2\right]}
\newcommand{\bvarop}[1]{\var\bigl[#1\bigr]}
\newcommand{\expop}[1]{\exp\left\{#1\right\}}
\newcommand{\reals}{\mathbb{R}}

\newcommand{\card}[1]{|#1|}
\newcommand{\cardS}{{\card{\cS}}}
\newcommand{\cardW}{{\card{\W}}}
\newcommand{\cardZ}{{\card{\Z}}}
\newcommand{\cardX}{{\card{\X}}}
\newcommand{\cardY}{{\card{\Y}}}
\newcommand{\cardU}{{\card{\U}}}
\newcommand{\cardV}{{\card{\V}}}
\newcommand{\detb}[1]{|#1|}
\newcommand{\bdetb}[1]{\bigl|#1\bigr|}

\newcommand{\abs}[1]{{|#1|}}
\newcommand{\babs}[1]{\bigl|#1\bigr|}
\newcommand{\bbabs}[1]{\left|#1\right|}
\newcommand{\norm}[1]{\|#1\|}
\newcommand{\bnorm}[1]{\bigl\|#1\bigr\|}
\newcommand{\bbnorm}[1]{\left\|#1\right\|}

\newcommand{\ip}[2]{\langle#1,#2\rangle}
\newcommand{\bip}[2]{\bigl\langle#1,#2\bigr\rangle}

\newcommand{\alt}{{\tilde{\alpha}}}

\newcommand{\bX}{\mathbf{X}}

\newcommand{\bY}{\mathbf{Y}}
\newcommand{\bF}{\mathbf{F}}
\newcommand{\bG}{\mathbf{G}}
\newcommand{\bH}{\mathbf{H}}
\newcommand{\bHt}{\tilde{\mathbf{H}}}
\newcommand{\bzt}{\boldsymbol{\zeta}}
\newcommand{\bDel}{\boldsymbol{\Delta}}
\newcommand{\bxi}{\boldsymbol{\xi}}

\newcommand{\bXi}{\boldsymbol{\Xi}}

\newcommand{\bpsi}{\boldsymbol{\psi}}
\newcommand{\bPsi}{\boldsymbol{\Psi}}
\newcommand{\psib}{\bar{\psi}{}}
\newcommand{\psih}{\hat{\psi}{}}
\newcommand{\bpsib}{\bar{\bpsi}{}}
\newcommand{\bpsih}{\hat{\bpsi}{}}
\newcommand{\bPsib}{\bar{\bPsi}{}}
\newcommand{\bPsih}{\hat{\bPsi}{}}
\newcommand{\bPsit}{\tilde{\bPsi}{}}

\newcommand{\bphi}{\boldsymbol{\phi}}
\newcommand{\bphit}{\tilde{\bphi}{}}
\newcommand{\bphic}{\check{\bphi}{}}

\newcommand{\bPhi}{\boldsymbol{\Phi}}
\newcommand{\bPhit}{\tilde{\bPhi}{}}

\newcommand{\bPi}{\boldsymbol{\Pi}}
\newcommand{\ups}{\upsilon}
\newcommand{\bups}{\boldsymbol{\upsilon}}
\newcommand{\bUps}{\boldsymbol{\Upsilon}}

\newcommand{\bTh}{\boldsymbol{\Theta}}

\newcommand{\hp}{\pi} 
\newcommand{\pred}{\varphi} 

\newcommand{\nncardX}{\cardX}
\newcommand{\nncardY}{\cardY}
\newcommand{\sigmoid}{\varsigma}

\newcommand{\frob}[1]{\|#1\|_\mathrm{F}}
\newcommand{\bfrob}[1]{\bigl\|#1\bigr\|_\mathrm{F}}
\newcommand{\bbfrob}[1]{\left\|#1\right\|_\mathrm{F}}
\newcommand{\nuclear}[1]{\|#1\|_*}

\newcommand{\kyfank}[1]{\|#1\|_{(k)}}

\newcommand{\spectral}[1]{\|#1\|_\mathrm{s}}
\newcommand{\bspectral}[1]{\bigl\|#1\bigr\|_\mathrm{s}}
\newcommand{\bbspectral}[1]{\left\|#1\right\|_\mathrm{s}}

\newcommand{\defeq}{\triangleq}
\newcommand{\simp}{\cP}
\newcommand{\simpXY}{\simp^{\X\times\Y}}
\newcommand{\dtmset}{\cB^{\X\times\Y}}
\newcommand{\dtmsett}{\cB^{\X\times\Y}_\circ}
\newcommand{\simpZW}{\simp^{\Z\times\W}}

\newcommand{\simpW}{\simp^\W}
\newcommand{\simpX}{\simp^\X}
\newcommand{\simpY}{\simp^\Y}
\newcommand{\simpZ}{\simp^\Z}

\newcommand{\nbhd}{\cN}
\newcommand{\nbhdk}{\bar{\cN}}
\newcommand{\nbhds}{\cS}
\newcommand{\nbhdss}{\cS^\mathrm{s}}
\newcommand{\nbhdsf}{\cS^{\mathrm{F}\!}}

\newcommand{\nbhdbf}{\cM^{\mathrm{F}\!}}
\newcommand{\nbhdbs}{\cM^\mathrm{s}}
\newcommand{\nbhdg}{\cN}
\newcommand{\distgen}{P} 
\newcommand{\Ph}{\hat{P}}
\newcommand{\empgen}{\Ph} 
\newcommand{\refgen}{P_0} 

\newcommand{\fvgen}{\xi}
\newcommand{\fvgena}{\bar{\xi}}
\newcommand{\bfvgen}{\bxi}
\newcommand{\bfVgen}{\bXi}
\newcommand{\fvX}{\fvgen^X}
\newcommand{\fvXa}{\fvgena^X}
\newcommand{\bfvX}{\bfvgen^X}

\newcommand{\bfVX}{\bfVgen^X}
\newcommand{\fvY}{\fvgen^Y}
\newcommand{\fvYa}{\fvgena^Y}
\newcommand{\bfvY}{\bfvgen^Y}
\newcommand{\bfVY}{\bfVgen^Y}
\newcommand{\bfWgen}{\bXi} 

\newcommand{\bfWgent}{\tilde{\bfWgen}{}}
\newcommand{\bfwgeng}{\bxi_\mathrm{G}} 
\newcommand{\bfwalt}{\bzt_\mathrm{G}} 

\newcommand{\bfWX}{\bfWgen^X}
\newcommand{\bfwXg}{\bfwgeng^X}

\newcommand{\bfWY}{\bfWgen^Y}
\newcommand{\bfwYg}{\bfwgeng^Y}

\newcommand{\ivgen}{\phi}
\newcommand{\bivgen}{\bphi}
\newcommand{\ivemp}{\hat{\phi}}  
\newcommand{\ivspacegen}{\cI^\Z} 
\newcommand{\ivspaceX}{\cI^\X}   
\newcommand{\ivspaceY}{\cI^\Y}   

\newcommand{\dtms}{B}
\newcommand{\dtmas}{\bar{\dtms}} 
\newcommand{\dtmts}{\tilde{\dtms}}
\newcommand{\dtmhs}{\hat{\dtms}}
\newcommand{\dtm}{\mathbf{B}}
\newcommand{\dtmh}{\hat{\dtm}}
\newcommand{\dtmt}{\tilde{\dtm}}
\newcommand{\dtmg}{\dtmt}  
\newcommand{\dtmgg}{\dtm_\mathrm{G}} 

\newcommand{\bmin}{B_{\min}}
\newcommand{\cholXg}{\bTh^X}
\newcommand{\cholX}[1]{\bTh^{X,#1}}

\newcommand{\cholYg}{\bTh^Y}
\newcommand{\cholY}[1]{\bTh^{Y,#1}}

\newcommand{\bP}{\mathbf{P}}
\newcommand{\bPh}{\hat{\bP}}
\newcommand{\bQ}{\mathbf{Q}}
\newcommand{\eqd}{\stackrel{\mathrm{d}}{=}}
\newcommand{\cmp}[1]{#1^\mathrm{c}}
\newcommand{\T}{\mathrm{T}}
\newcommand{\bSi}{\boldsymbol{\Sigma}}
\newcommand{\bSit}{\tilde{\bSi}}
\newcommand{\bSih}{\hat{\bSi}}
\newcommand{\bGa}{\boldsymbol{\Gamma}}
\newcommand{\bGat}{\tilde{\bGa}}
\newcommand{\bLa}{\boldsymbol{\Lambda}}
\newcommand{\bLat}{\tilde{\bLa}}
\newcommand{\bLah}{\hat{\bLa}}
\newcommand{\mut}{\tilde{\mu}}
\newcommand{\bmu}{\boldsymbol{\mu}}
\newcommand{\bmuh}{\hat{\bmu}}
\newcommand{\eff}{\nu} 
\newcommand{\Pt}{\tilde{P}}
\newcommand{\phit}{\tilde{\phi}}
\newcommand{\phic}{\check{\phi}}
\newcommand{\bzero}{\mathbf{0}}
\newcommand{\bone}{\mathbf{1}}
\newcommand{\bI}{\mathbf{I}}

\newcommand{\bell}{\boldsymbol{\ell}}
\newcommand{\tell}{\tilde{\ell}}

\newcommand{\tbeta}{\tilde{\beta}}
\newcommand{\betah}{\hat{\beta}}

\newcommand{\bigO}{O}
\newcommand{\mppi}{\dagger} 

\begin{document}

\title{%
On Universal Features for High-Dimensional\\ Learning and Inference%
} \author{Shao-Lun~Huang,~\IEEEmembership{Member, IEEE},
  Anuran~Makur,~\IEEEmembership{Student Member, IEEE}, \\ 
  Gregory~W.~Wornell,~\IEEEmembership{Fellow,~IEEE}, and
  Lizhong~Zheng~\IEEEmembership{Fellow,~IEEE}\thanks{Manuscript
    received September 2019.  This work was supported in part by NSF
    under Grant Nos.~CCF-1717610 and CCF-1816209.  This
    work was presented in part at thte Int.\ Symp.\ Inform.\ Theory
    (ISIT-2017) \cite{hmzw17}, Aachen, Germany, June 2017, the
    Inform.\ Theory and Appl. Workshop (ITA-2018), Feb.\ 2018
    \cite{hmwz18}, and at the Inform.\ Theory Workshop (ITW-2018),
    Guangzhou, China, Nov.\ 2018 \cite{hwz18}.} 
  \thanks{S-L.~Huang is with the Data Science and Information
    Technology Research Center, Tsinghua-Berkeley Shenzhen Institute,
    Shenzhen, China (Email: shaolun.huang@sz.tsinghua.edu.cn).
   A.~Makur, G.~W.~Wornell, and L.~Zheng are with the
    Department of Electrical Engineering and Computer Science,
    Massachusetts Institute of Technology, Cambridge, MA 02139.
    (Email: \{a\_makur, lizhong, gww\}@mit.edu).  }}

\maketitle

\begin{abstract}
We consider the problem of identifying universal low-dimensional
features from high-dimensional data for inference tasks in settings
involving learning.  For such problems, we introduce natural notions
of universality and we show a local equivalence among them.  Our
analysis is naturally expressed via information geometry, and
represents a conceptually and computationally useful analysis.  The
development reveals the complementary roles of the singular value
decomposition, Hirschfeld-Gebelein-R\'enyi maximal correlation, the
canonical correlation and principle component analyses of Hotelling
and Pearson, Tishby's information bottleneck, Wyner's common
information, Ky Fan $k$-norms, and Brieman and Friedman's alternating
conditional expectations algorithm.  We further illustrate how this
framework facilitates understanding and optimizing aspects of learning
systems, including multinomial logistic (softmax) regression and the
associated neural network architecture, matrix factorization methods
for collaborative filtering and other applications, rank-constrained
multivariate linear regression, and forms of semi-supervised learning.
\end{abstract}


\section{Introduction}
\label{sec:intro}

\IEEEPARstart{I}{n} many contemporary and emerging applications of
machine learning and statistical inference, the phenomena of interest
are characterized by variables defined over large alphabets.  Familiar
examples, among many others, include the relationship between
individual consumers and products that may be of interest to them, and
the relationship between images and text in a visual search setting.
In such scenarios, not only are the data high-dimensional, but the
collection of possible inference tasks is also large.  At the same
time, training data available to learn the underlying relationships is
often quite limited relative to its dimensionality.

From this perspective, for a given level of training data, there is a
need to understand which inference tasks can be most effectively
carried out, and, in turn, what features of the data are most relevant
to them.  As we develop in this paper, a natural framework for
addressing such questions rather broadly can be traced back to the
work of Hirschfeld \cite{hoh35}.

As we will develop, the problem can be equivalently expressed as one
of ``universal'' feature extraction, and show that diverse notions of
such universality lead to precisely the same features.  Our
development emphasizes an information theoretic treatment of the
associated questions, and in particular we adopt a convenient
``local'' information geometric analysis that provides useful insight.
In turn, as we describe, the interpetation of such features in terms
of a suitable singular value decomposition (SVD) facilitates their
computation.

An outline of the paper, and summary of its key contributions, is as
follows:

\paragraph*{\secref{sec:modal}} As a foundation, and 
focusing on finite alphabets, we describe the modal decomposition of
bivariate distributions into constituent features that arises out of
Hirschfeld's analysis, developing it in terms of the SVD of a
particular matrix characterization---termed the canonical dependence
matrix (CDM)---of the distribution and the associated conditional
expectation operator.

\paragraph*{\secref{sec:variational}} We describe the variational
characterization of the modal decomposition in terms of standard SVD
analysis, as further developed by Gebelein and R\'enyi, from which we
obtain the resulting Hirchfeld-Gebelein-R\'enyi (HGR) maximal
correlation as the Ky Fan $k$-norm of the CDM.  Via this analysis, the
features defining the modal decomposition are obtained via an
optimization. 

\paragraph*{\secref{sec:geom-simplex}} As a further foundation, we
describe the local geometric analysis on the probability simplex that
is associated with $\chi^2$-divergence.  In the resulting Euclidean
information space, distributions are represented as information
vectors, and features as feature vectors, and we develop an
equivalence between them via log-likelihoods.  Via this geometry, we
develop a suitable notion of weakly dependent variables for which we
obtain a decomposition of mutual information and through which we
interpret truncated modal decompositions as ``information efficient.''
Additionally, we characterize the error exponents in local decision
making in terms of (mismatched) feature projections.

\paragraph*{\secref{sec:unifeature-char}} 
Via the local analysis, we develop several different characterizations
of universal features, all of which coincide with the features that
arise in the modal decomposition of the joint distribution.  As an
initial observation, we note that features characterize a locally
exponential family for the conditional distributions.  For the
remaining characterizations, we introduce latent local attribute
variables.  In particular: \secref{sec:uni-rie} obtains the modal
decomposition features as the solution to a game between system
designer and nature, where the system designer must choose features to
detect attributes that nature chooses at random after these features
are fixed; \secref{sec:uni-coop} obtains the same features as the
solution to a cooperative game in which the system designer and nature
seek the most detectable attributes and locally sufficient statistics
for their detection; \secref{sec:uni-ib} obtains the same features as
the solution to a local symmetric version of Tishby's information
bottleneck problem that seeks mutual information maximizing attributes
and the associated locally sufficient statistics; and
\secref{sec:common} obtains superpositions of these same features
arise as locally sufficient statistics in the solution to a local
version of Wyner's common information, which using variational
analysis we show specializes to the nuclear (trace) norm of the CDM.
In turn, \secref{sec:relate} develops Markov structure relating the
resulting common information variable to the attributes optimizing the
information bottleneck.

\paragraph*{\secref{sec:ace}} We discuss the estimation of
universal features from training data, starting from the statistical
interpretation of the orthogonal iteration method of computing an SVD
as the alternating conditional expectations (ACE) algorithm of Breiman
and Friedman.  We include the relevant analysis of 
sample complexity of feature recovery, which supports the empirical
observation that in practice the dominant modes can typically be
recovered with comparatively little training data.

\paragraph*{\secref{sec:cf}} We use the context of collaborative
filtering to develop matrix factorization perspectives associated with
the modal decomposition.  In particular, we formulate the problem of
collaborative filtering as one of Bayesian attribute matching, and
find that the optimum such filtering is achieved using a truncated
modal decomposition, which corresponds to the optimum low-rank
approximation to the empirical CDM, which differs from some other commonly
used factorizations.

\paragraph*{\secref{sec:nn}} We analyze a local version of multinomial
logistic regression; specifically, under weak dependence we show that
softmax weights correspond to (normalized) conditional expectations,
and that the resulting discriminative model matches, to first order,
that of a Gaussian mixture without any Gaussian assumptions in the
analysis.  We further show that the optimizing features are, again,
those of the modal decomposition, in which case the associated softmax
weights are proportional to the ``dual'' features of the
decomposition.  Our analysis additionally quantifies the performance
limits in this regime in terms of the associated singular values.  As
we discuss, this analysis implies a relationship between the ACE
algorithm and methods used to train at least some classes of neural
networks.

\paragraph*{\secref{sec:gauss}} We provide a treatment
for Gaussian variables that parallels the preceding one for finite
alphabets.  To start, we construct the modal decomposition of
covariance via the SVD of the canonical correlation matrix (CCM), and
obtain the familiar formulation of Hotelling's canonical correlation
analysis (CCA) via the corresponding variational characterization.  We
further define a local Gaussian geometry, the associated notion of
weakly correlated variables, and construct a local modal decomposition
of joint distributions of such variables in terms of the CCA features,
which are linear.  Via Gaussian attribute models, we then show these
CCA features arise in the solution to universal feature problem
formulations.  \secref{sec:uni-rieg} shows they arise in the
solution of an attribute estimation game in which nature chooses the
attribute at random after the system designer chooses the linear
features from which it will be estimated using a minimum mean-square
error (MMSE) criterion, and \secref{sec:uni-coopg} shows they arise in
the solution of the corresponding cooperative MMSE attribute
estimation game; these analyses are global.  \secref{sec:ib-gauss},
shows the CCA features arising in the solution to the local symmetric
version of Tishby's Gaussian information bottleneck problem, and
\secref{sec:common-gauss} describes how superpositions of CCA
features arise in the solution to the (global) Gaussian version of
Wyner's common information problem; locally this common
information is given by the nuclear norm of the CCM.  
\secref{sec:relate-gauss} describes the Markov relationships between
the dominant attributes in the solution to the information bottleneck
and the common information variable.  \secref{sec:pca} interprets the
features arising out of Pearson's principal component analysis (PCA)
as a special case of the preceding analyses in which the underlying
variables are simultaneously diagonalizable, and \secref{sec:ace-g}
discusses the estimation of CCA features, interpreting the associated
SVD computation as a version of the ACE algorithm in which the
features are linearly constrained. \secref{sec:gam} develops Gaussian
attribute matching, and interprets the resulting procedure as one of
optimum rank-constrained linear estimation, and \secref{sec:nn-gauss}
develops a form of rank-constrained linear regression as the
counterpart to softmax regression, and distinguishing it from
classical formulations.

\paragraph*{\secref{sec:semi}} We provide a limited discussion of the
application of universal feature analysis to problems beyond the realm
of fully supervised learning.  \secref{sec:indirect} describes the
problem of ``indirect'' learning in which to carry out clustering on
data, relationships to secondary data are exploited to define an
appropriate measure of distance.  We show, in particular, that our
softmax analysis implies a natural procedure in which Gaussian mixture
modeling is applied to the dominant features obtained from the modal
decomposition with respect to the secondary data.  By contrast,
\secref{sec:MNIST} discusses the problem of partially-supervised
learning in which features are learned in an unsupervised manner, and
labeled data is used only to obtain the classifier based on the
resulting features.  As an illustration of the use of universal
features in this setting, an application to handwritten digit
recognition using the MNIST database is described in which the
relevant features are obtained via the common information between
subblocks of MNIST images.  A simple implementation achieves an error
probability of 3.02\%, close to that of a 3-layer neural net (with
300+100 hidden nodes), which yields an error probability of 3.05\%.

Finally, \secref{sec:conc} contains some concluding remarks.

\section{The Modal Decomposition of Joint Distributions}
\label{sec:modal}

Let $X$ and $Y$ denote random variables over finite alphabets $\X$ and
$\Y$, respectively, with joint distribution $P_{X,Y}$.  Without loss
of generality we assume throughout that the marginals satisfy
$P_X(x)>0$ and $P_Y(y)>0$ for all $x\in\X$ and $y\in\Y$, since
otherwise the associated symbols may be removed from their respective
alphabets.  Accordingly, we let $\simpXY$ denote the set of all such
distributions.

For an arbitrary \emph{feature}\footnote{The literature sometimes
  refers to these as embeddings, referring to functions of embeddings
  as features.  However, our treatment does not require this
  distinction.}  $f\colon
\X\mapsto \reals$, let $g\colon \Y\mapsto \reals$ be the feature
induced by $f$ through conditional expectation with respect to
$P_{X|Y}(\cdot|y)$, i.e.,
\begin{equation}
g(y) = \bE{f(X) \mid Y=y},\qquad y\in\Y.
\label{eq:f-to-g}
\end{equation}
Then we can express \eqref{eq:f-to-g} in the form
\begin{align*} 
g(y) 
&= \frac{1}{P_Y(y)} \sum_{x\in\X} P_{X,Y}(x,y)\, f(x) \\
&= \frac{1}{\sqrt{P_Y(y)}} \sum_{x\in\X}
\frac{P_{X,Y}(x,y)}{\sqrt{P_X(x)}\,\sqrt{P_Y(y)}} \, \sqrt{P_X(x)}\,f(x),
\end{align*}
i.e., 
\begin{equation} 
\fvY(y) = \sum_{x\in\X} \dtms(x,y)\, \fvX(x)
\label{eq:ce-operator}
\end{equation}
where we have defined
\begin{equation}
\dtms(x,y) \defeq \frac{P_{X,Y}(x,y)}{\sqrt{P_X(x)} \, \sqrt{P_Y(y)}},\quad
x\in\X,\ y\in\Y,
\label{eq:dtms-def}
\end{equation}
and 
\begin{subequations} 
\begin{align}
\fvX(x) &\defeq \sqrt{P_X(x)}\, f(x)\\
\fvY(y) &\defeq \sqrt{P_Y(y)}\, g(y)
\end{align}
\label{eq:fvXY-def}%
\end{subequations}

Clearly $\fvX$ and $\fvY$ in \eqref{eq:fvXY-def} are equivalent
representations for $f$ and $g$ respectively.  But $\dtms$ in
\eqref{eq:dtms-def} is also an equivalent representation for
$P_{X,Y}$, as we will verify shortly.  Moreover,
\eqref{eq:ce-operator} expresses that $\dtms$ has an interpretation as
a conditional expectation operator, and thus is 
equivalent to 
$P_{X|Y}$.

Next consider an arbitrary feature  $\gt\colon
\Y\mapsto \reals$, and 
let $\ft\colon \X\mapsto\reals$ be the feature induced by $\gt$
through conditional expectation with respect to 
$P_{Y|X}(\cdot|x)$, i.e. 
\begin{equation}
\ft(x) = \bE{\gt(Y) \mid X=x}.
\label{eq:g-to-f}
\end{equation}
Then using the notation \eqref{eq:dtms-def} and that analogous to
\eqref{eq:fvXY-def}, i.e., 
\begin{subequations} 
\begin{align}
\fvXa(x) &= \sqrt{P_X(x)}\, \ft(x)\\
\fvYa(y) &= \sqrt{P_Y(y)}\, \gt(y),
\end{align}
\label{eq:fvXYt-def}%
\end{subequations}
we can express \eqref{eq:g-to-f} in the form
\begin{equation} 
\fvXa(x) = \sum_{y\in\Y} \underbrace{\dtmas(y,x)}_{\defeq
  \dtms(x,y)}\, \fvYa(y), 
\label{eq:ce-adjoint}
\end{equation}
where $\dtmas$ is the adjoint of $\dtms$.
Likewise $\dtmas$ is an equivalent representation for $P_{X,Y}$ and,
in turn, $P_{Y|X}$.  

It is convenient to represent $\dtms$ as a matrix.  Specifically, we
let $\dtm$ denote the $\cardY\times\cardX$ matrix whose $(y,x)$\/th entry
is $\dtms(x,y)$, i.e.,
\begin{equation} 
\dtm =
\left[\sqrt{\bP_Y}\right]^{-1} \bP_{Y,X}
\left[\sqrt{\bP_X}\right]^{-1},
\label{eq:dtm-def}
\end{equation}
where $\sqrt{\bP_X}$ denotes a $\cardX\times\cardX$ diagonal matrix
whose $x$\/th diagonal entry is $\sqrt{P_X(x)}$, where $\sqrt{\bP_Y}$
denotes a $\cardY\times\cardY$ diagonal matrix whose $y$\/th diagonal
entry is $\sqrt{P_Y(y)}$, and where $\bP_{Y,X}$ denotes the
$\cardY\times\cardX$ matrix whose $(y,x)$\/th entry is $P_{Y,X}(y,x)$.
In \cite{hz12}, $\dtm$ is referred to as the \emph{divergence transfer
  matrix (DTM)} associated with $P_{X,Y}$.\footnote{The work of
  \cite{hz12}, building on \cite{bz08}, focuses on a communication
  network setting.  Subsequently, \cite{mkhz15} recognizes connections
  to learning that motivate aspects of, e.g., the present paper.}

Although for convenience we will generally restrict our attention to
the case in which the marginals $P_X$ and $P_Y$ are strictly positive,
note that extending the DTM definition to arbitrary nonnegative
marginals is straightforward.  In particular, it suffices make the
$x'$\/th column of $\dtm$ all zeros if $P_X(x')=0$ for some $x'\in\X$,
and, similarly, the 
$y'$\/th row of $\dtm$ all zeros if $P_Y(y')=0$ for some $y'\in\Y$,
i.e., \eqref{eq:dtms-def} is extended via
\begin{equation} 
\begin{aligned} 
\dtms(x,y) \defeq 0,\quad 
&\text{all $x\in\X$, $y\in\Y$ such that}\\
&\text{$P_X(x)=0$ or $P_Y(y)=0$}.
\end{aligned}
\label{eq:dtms-extend}
\end{equation}

Useful alternate forms of $\dtms$ and $\dtmas$ are
[cf.\ \eqref{eq:dtms-def}]
\begin{align*}
\dtms(x,y) &=
\frac{P_{Y|X}(y|x)}{\sqrt{P_Y(y)}}\,\sqrt{P_X(x)} 
\\
\dtmas(x,y) &=
\frac{P_{X|Y}(x|y)}{\sqrt{P_X(x)}}\,\sqrt{P_Y(y)}, 
\end{align*}
from which we obtain the alternate matrix representations
\begin{align}
\dtm &= \left[\sqrt{\bP_Y}\right]^{-1} \, \bP_{Y|X} \,
\left[\sqrt{\bP_X}\right] \label{eq:dtm-alt} \\
\dtm^\T &= \left[\sqrt{\bP_X}\right]^{-1} \, \bP_{X|Y} \,
\left[\sqrt{\bP_Y}\right] \label{eq:dtma-alt},
\end{align}
where $\bP_{Y|X}$ denotes the $\cardY\times\cardX$ left (column) stochastic
transition probability matrix whose $(y,x)$\/th entry is
$P_{Y|X}(y|x)$, and where, similarly,
$\bP_{X|Y}$ denotes the $\cardX\times\cardY$ left (column) stochastic
transition probability matrix whose $(x,y)$\/th entry is
$P_{X|Y}(x|y)$.

The SVD of $\dtm$ takes the form
\begin{subequations} 
\begin{equation} 
\begin{aligned}
\dtm &= \sum_{i=0}^{K-1} \sigma_i\, \bpsi^Y_i \bigl(\bpsi^X_i\bigr)^\T \\
\text{i.e.,}\quad \dtms(x,y) &= \sum_{i=0}^{K-1} \sigma_i\, \psi^X_i(x)\, \psi^Y_i(y),
\end{aligned}
\label{eq:dtm-svd-form}
\end{equation}
with
\begin{equation}
K\defeq\min\{\cardX,\cardY\},
\label{eq:K-def}
\end{equation}
where $\sigma_i$ denotes the $i$\/th singular value and where
$\bpsi^Y_i$ and $\bpsi^X_i$ are the corresponding left and right singular
vectors, and where by convention we order the singular values
according to
\begin{equation}
\sigma_0 \ge \sigma_1 \ge \dots \ge \sigma_{K-1}.
\label{eq:dtm-sv}
\end{equation}
\label{eq:dtm-svd}%
\end{subequations}

The following proposition establishes that $\dtms$ (and thus $\dtmas$)
is a contractive
operator, a proof of which is provided in \appref{app:contractive}.
\begin{proposition}
\label{prop:contractive}
For $\dtm$ defined via
\eqref{eq:dtm-def}
we have
\begin{equation}
\spectral{\dtm} = 1,
\end{equation}
where $\spectral{\cdot}$ denotes the spectral (i.e., operator) norm
of its matrix argument.\footnote{The spectral norm of an
  arbitrary matrix $\bA$ is
\begin{equation*}
\spectral{\bA}=\max_i \sigma_i(\bA),
\end{equation*}
where $\sigma_i(\bA)$ denotes the $i$\/th singular value of $\bA$.}
Moreover, in \eqref{eq:dtm-svd}, the 
left and right singular vectors 
$\bpsi^X_0$ and $\bpsi^Y_0$ 
associated with singular value
\begin{subequations}%
\begin{equation}
\sigma_0=1
\label{eq:sval0-def}
\end{equation}%
have elements
\begin{equation}
\psi^X_0(x) \defeq \sqrt{P_X(x)}\qquad\text{and}\qquad
\psi^Y_0(y) \defeq \sqrt{P_Y(y)}.
\label{eq:svec0-def}
\end{equation}
\label{eq:sv0-def}%
\end{subequations}%
\end{proposition}

It follows immediately from the second part of
\propref{prop:contractive} that $\dtm$ is an equivalent representation
for $P_{X,Y}$.  Indeed, given $\dtm$, we can compute the singular
vectors $\psi^X_0$ and $\psi^Y_0$, from which we obtain $P_X$ and
$P_Y$ via \eqref{eq:svec0-def}. In turn, using these marginals
together with $\dtm$, whose $(y,x)$\/th entry is \eqref{eq:dtms-def},
yields $P_{X,Y}(x,y)=\dtms(x,y)\,\sqrt{P_X(x)}\,\sqrt{P_Y(y)}$.  We
provide a more complete characterization of the class of DTMs, i.e.,
$\dtm(\simpXY)$ in \appref{app:dtm-char}.  In so doing, we extend the
equivalence result above, establishing the continuity of bijective
mapping between $P_{X,Y}$ and $\dtm$.

The SVD \eqref{eq:dtm-svd} provides a key expansion of the joint
distribution $P_{X,Y}(x,y)$.  In particular, we have the following
result.
\begin{proposition}
\label{prop:modal}
Let $\X$ and $\Y$ denote finite alphabets.  Then for any
$P_{X,Y}\in\simpXY$, there exist features $f_i^*\colon \X\mapsto\reals$
and $g_i^*\colon \Y \mapsto \reals$, for $i=1,\dots,K-1$, such that
\begin{equation}
P_{X,Y}(x,y) = P_X(x)\,P_Y(y)
\left[ 1 + \sum_{i=1}^{K-1} \sigma_i \, f_i^*(x)\,g_i^*(y) \right],
\label{eq:modal}
\end{equation}
where $\sigma_1,\dots,\sigma_{K-1}$ are as defined in
\eqref{eq:dtm-svd}, and where\footnote{We use
  the Kronecker notation 
\begin{equation*}
\kron_{\cA} = \begin{cases} 1 & \text{$\cA$ is true} \\ 0 &
  \text{otherwise} 
\end{cases}
\end{equation*}}
\begin{subequations} 
\begin{align} 
\E{f_i^*(X)} &=0,\quad i\in\{1,\dots,K-1\} \label{eq:fi-0mean}\\
\E{g_i^*(Y)} &=0,\quad i\in\{1,\dots,K-1\} \label{eq:gi-0mean}\\
\E{f_i^*(X)\,f_j^*(X)} &= \kron_{i=j},\quad
i,j\in\{1,\dots,K-1\} \label{eq:fi-uncorr}\\ 
\E{g_i^*(Y)\,g_j^*(Y)} &= \kron_{i=j},\quad
i,j\in\{1,\dots,K-1\}.\label{eq:gi-uncorr}
\end{align}%
\label{eq:fgi-moments}%
\end{subequations}
Moreover, $f_i^*$ and $g_i^*$ are related to the singular vectors in
\eqref{eq:dtm-svd} according to
\begin{subequations} 
\begin{align} 
f_i^*(x) &\defeq \frac{\psi^X_i(x)}{\sqrt{P_X(x)}},\quad i=1,\dots,K-1
\label{eq:fi-def}\\
g_i^*(y) &\defeq \frac{\psi^Y_i(y)}{\sqrt{P_Y(y)}},\quad i=1,\dots,K-1, \label{eq:gi-def}
\end{align}%
\label{eq:fgi-def}%
\end{subequations}
where $\psi^X_i(x)$ and $\psi^Y_i(y)$ are the
$x$\/th and $y$\/th entries of $\bpsi^X_i$ and $\bpsi^Y_i$,
respectively.  
\end{proposition}

\begin{IEEEproof}
It suffices to note that
\begin{align}
\dtms(x,y) 
&= \frac{P_{X,Y}(x,y)}{\sqrt{P_X(x)}\sqrt{P_Y(y)}} \label{eq:use-dtms-def}\\
&= \sqrt{P_X(x)}\,\sqrt{P_Y(y)} + \sum_{i=1}^{K-1}
\sigma_i\, \psi^X_i(x)\, \psi^Y_i(y) \label{eq:use-sv0}\\
&= \sqrt{P_X(x)}\,\sqrt{P_Y(y)} \notag\\
&\qquad{}+ \sum_{i=1}^{K-1}
\sigma_i\, \sqrt{P_X(x)}\, f_i^*(x)\, \sqrt{P_Y(y)}\,
g_i^*(y) \label{eq:use-fgi-def} \\
&= \sqrt{P_X(x)}\,\sqrt{P_Y(y)}\left[ 1 + \sum_{i=1}^{K-1}
\sigma_i\, f_i^*(x)\, g_i^*(y) \right],
\end{align}
where to obtain \eqref{eq:use-dtms-def} we have used
\eqref{eq:dtms-def}, to obtain \eqref{eq:use-sv0} we have used
\eqref{eq:dtm-svd-form} with \eqref{eq:sv0-def}, and where to obtain
\eqref{eq:use-fgi-def} we have made the choices \eqref{eq:fgi-def},
which we note satisfy the constraints \eqref{eq:fgi-moments}.  In
particular, \eqref{eq:fi-0mean} follows from the fact that $\psi^X_0$
and $\psi^X_i$ are orthogonal, for $i=1,\dots,K-1$, and, likewise,
\eqref{eq:gi-0mean} follows from the fact that $\psi^Y_0$ and
$\psi^Y_i$ are orthogonal, for $i=1,\dots,K-1$.  Finally,
\eqref{eq:fi-uncorr} and \eqref{eq:gi-uncorr} follow from the
remaining orthogonality relations among the $\psi^X_i$ and $\psi^Y_i$,
respectively.
\end{IEEEproof}

The expansion \eqref{eq:modal} in \propref{prop:modal} effectively
forms the basis of what is sometimes referred to as ``correspondence
analysis,'' which was originated by Hirschfeld \cite{hoh35} and sought
to extend the applicability of the methods of Pearson
\cite{kp01,kp04}.  Such analysis was later independently developed and
further extended by Lancaster \cite{hol53,hol58}, and yet another
independent development began with the work of Gebelein \cite{hg41},
upon which the work of R\'enyi \cite{ar59} was based.\footnote{The
  associated analysis was expressed in terms of eigenvalue
  decompositions of $\dtm^\T\dtm$ instead of the SVD of $\dtm$, since
  the latter was not widely-used at the time.}  This analysis was
reinvented again and further developed in \cite{jpb73,jpb92}, which
established the correspondence analysis terminology, and further
interpreted in \cite{mg84,lmw84}.\footnote{This terminology includes that of ``inertia,'' which is also
  adopted in a variety of works, an example of which is
  \cite{calmon17}, which refers to ``principal inertial components.''}
Subsequent developments appear in \cite{ag90,mdl98}, and more recent
expositions and developments include \cite{lrr04,sn06}, and the
practical guide \cite{mg16}.

The features \eqref{eq:fgi-def} in \eqref{eq:modal} can be
interpreted as suitably normalized sufficient statistics for
inferences involving $X$ and $Y$.  Indeed, since
\begin{subequations} 
\begin{align} 
P_{Y|X}(y|x) &= P_Y(y)\, \left[ 1 + \sum_{i=1}^{K-1} \sigma_i \,
  f_i^*(x)\,g_i^*(y) \right] \label{eq:Y|X-form}\\
P_{X|Y}(x|y) &= P_X(x)\, \left[ 1 + \sum_{i=1}^{K-1} \sigma_i \,
  f_i^*(x)\,g_i^*(y) \right],\label{eq:X|Y-form} %
\end{align}
\end{subequations}
it follows that\footnote{Throughout, we use the convenient sequence notation
  $a^l\defeq(a_1,\dots,a_l)$.} 
\begin{equation*}
f^{K-1}_*(x) \defeq \bigl(f_1^*(x),\dots,f_{K-1}^*(x)\bigr)
\end{equation*}
is a sufficient statistic for inferences about $y$ based on $x$, i.e.,
we have the Markov structure 
\begin{equation*}
Y\markov f^{K-1}_*(X) \markov X.
\end{equation*}
Analogously, 
\begin{equation*}
g^{K-1}_*(y) \defeq \bigl(g_1^*(y),\dots,g_{K-1}^*(y)\bigr)
\end{equation*}
is a sufficient statistic for inferences about $x$ based on $y$, i.e.,
we have the Markov structure 
\begin{equation*} 
X\markov g^{K-1}_*(Y) \markov Y.
\end{equation*}
Combining these results, we have
\begin{equation}
X\markov \bigl(f^{K-1}_*(X),g^{K-1}_*(Y)\bigr) \markov Y.
\end{equation}

Additionally, note that \propref{prop:modal} has additional
consequences that are direct result of its connection to the SVD of
$\dtm$.  In particular, since the left and right singular vectors are
related according to
\begin{subequations} 
\begin{align} 
\sigma_i\, \bpsi^Y_i &= \dtm \,  \bpsi^X_i \\
\sigma_i\, \bpsi^X_i &= \dtm^\T \bpsi^Y_i,
\end{align}%
\label{eq:psiXY-rel}%
\end{subequations}
it follows from \eqref{eq:fgi-def} that the $f_i^*$ and $g_i^*$ are
related according to
\begin{subequations} 
\begin{align}
\sigma_i\,f_i^*(x) &= \E{g_i^*(Y) \mid X=x}  \label{eq:fi-condexp}\\
\sigma_i\,g_i^*(y) &= \E{f_i^*(X) \mid Y=y}, \label{eq:gi-condexp}
\end{align}%
\label{eq:fgi-condexp}%
\end{subequations}
for $i=1,\dots,K-1$.   Moreover, in turn, we obtain, for 
$i,j\in\{1,\dots,K-1\}$,
\begin{align} 
\E{f_i^*(X)\,g_j^*(Y)} 
&= \E{ \E{f_i^*(X) \mid Y=y} g_j^*(Y)} \notag\\
&= \E{ \sigma_i\, g_i^*(Y) \, g_j^*(Y)} \notag\\
&= \sigma_i \, \kron_{i=j}.
\label{eq:fgi-uncorr}
\end{align}

\subsection*{The Canonical Dependence Matrix}

In our development, it is convenient for our analysis to remove the
zeroth mode from $\dtm$.   We do this by defining the matrix $\dtmt$
whose $(y,x)$\/th
entry is
\begin{align} 
\dtmts(y,x)
&\defeq \frac{P_{X,Y}(x,y)-P_X(x)\,P_Y(y)}{\sqrt{P_X(x)} \,
  \sqrt{P_Y(y)}} \label{eq:dtmts-def}\\
&= \sum_{i=1}^{K-1} \sigma_i\, \psi^X_i(x)\, \psi^Y_i(y), \notag
\end{align}
where in the last equality we have expressed its SVD in terms of that
for $\dtm$, and from which we see that $\dtmt$ has singular values
\begin{equation*}
1\ge \sigma_1 \ge \sigma_2 \ge \cdots \ge \sigma_{K-1} \ge \sigma_K = 0,
\end{equation*}
where we have defined the zero singular value $\sigma_K$ as a
notational convenience.  Note that we can intrepret $\dtmts$ as the
conditional expectation operator $\E{\cdot|Y=y}$ restricted to the
(sub)space of zero-mean features $f(X)$, which produces a
corresponding zero-mean features $g(Y)$.  We refer to $\dtmt$, which
we can equivalently write in the form\footnote{As first used in
  \appref{app:contractive}, we use $\bone$ to denote a vector of all
  ones (with dimension implied by context).}
\begin{align} 
\dtmt &= \left[\sqrt{\bP_Y}\right]^{-1} \, \left[ \bP_{Y|X} -
  \bP_Y\,\bone\,\bone^\T \right]\,
\left[\sqrt{\bP_X}\right]
\label{eq:dtmt-def} \\
&= \sum_{i=1}^{K-1} \sigma_i\, \bpsi^Y_i \bigl(\bpsi^X_i\bigr)^\T,
\label{eq:dtmt-svd}
\end{align}
as the \emph{canonical dependence matrix (CDM)}.  Some additional
perspectives on this representation of the conditional expectation
operator---and thus the particular choice of SVD---are provided in
\appref{app:ce-rep}.

It is worth emphasing that restricting attention to features of $X$
and $Y$ that are zero-mean is without loss of generality, as
there is an invertible mapping between any set of 
features and their zero-mean counterparts.  As a result, we will
generally impose this constraint.

\section{Variational Characterization of the Modal Decomposition}
\label{sec:variational}

The feature functions $(f^*_i,g^*_i)$, $i=1,2,\dots K-1$, in
\propref{prop:modal} can be equivalently obtained from a variational
characterization, via which the key connection to the correlation
maximization problem considered (in turn) by Hirschfeld \cite{hoh35},
Gebelein \cite{hg41}, and R\'{e}nyi \cite{ar59} is obtained, as we now
develop.

\subsection{Variational Characterizations of the SVD}

We begin by summarizing some classical variational results on
the SVD that will be useful in our analysis.
First, we have the following lemma
(see, e.g., \cite[Corollary~4.3.39, p.~248]{hj12}).
\begin{lemma}
\label{lem:eig-k}
Given an arbitrary $k_1\times k_2$ matrix $\bA$ and any
$k\in\bigl\{1,\dots,\min\{k_1,k_2\}\bigr\}$, we have\footnote{We use
  $\bI$ to denote the identity matrix of appropriate dimension.}
\begin{equation}
\max_{\left\{\bM\in\reals^{k_2\times
      k}\colon \bM^\T\bM=\bI\right\}} 
\bfrob{\bA \bM}^2 = \sum_{i=1}^k \sigma_i(\bA)^2,
\label{eq:var-eig-k}
\end{equation}
where $\frob{\cdot}$ denotes the Frobenius norm
of its matrix argument,\footnote{Specifically, the Frobenius norm of an
arbitrary matrix $\bA$ is
\begin{equation*}
\frob{\bA} \defeq \tr\bigl(\bA^\T\bA\bigr) = \sum_i \sigma_i(\bA)^2,
\end{equation*}
where $\sigma_i(\bA)$ denotes the $i$\/th singular value of $\bA$, and
were $\tr(\cdot)$ denotes the trace of its matrix argument.}
and where
$\sigma_1(\bA) \ge \dots \ge \sigma_{\min\{k_1,k_2\}}(\bA)$ denote the
(ordered) singular values of $\bA$.  Moreover, the maximum in
\eqref{eq:var-eig-k} is achieved by 
\begin{equation}
\bM = \begin{bmatrix} \bpsi_1(\bA) & \cdots & \bpsi_k(\bA) 
\end{bmatrix},
\end{equation}
with $\bpsi_i(\bA)$ denoting the right singular vector of $\bA$
corresponding to $\sigma_i(\bA)$, for $i=1,\dots,\min\{k_1,k_2\}$.
\end{lemma}

Second, the following lemma, essentially due to von Neumann (see,
e.g., \cite{asl95} \cite[Theorem~7.4.1.1]{hj12}), will also be useful
in our analysis, and can be obtained using \lemref{lem:eig-k} in
conjunction with the Cauchy-Schwarz inequality. 
\begin{lemma}
\label{lem:svd-k}
Given an arbitrary $k_1\times k_2$ matrix $\bA$, we have
\begin{align}  
\smash[b]{\max_{\substack{\left\{\bM_1\in\reals^{k_1\times
      k},\ \bM_2\in\reals^{k_2\times
      k}\colon \right.\\\left. \bM_1^\T\bM_1=\bM_2^\T\bM_2=\bI
    \vphantom{\reals^{k_1\times k}} \right\}}}}\vphantom{\max_M}
&\tr \bigl(\bM_1^\T \bA \bM_2\bigr) \notag\\
&\qquad= \sum_{i=1}^k \sigma_i(\bA),
\label{eq:var-svd-k}
\end{align}
with $\sigma_1(\bA) \ge \dots \ge \sigma_{\min\{k_1,k_2\}}(\bA)$
denoting the
(ordered) singular values of $\bA$.
Moreover, the maximum in \eqref{eq:var-svd-k} is achieved by
\begin{equation} 
\bM_j = \begin{bmatrix} \bpsi_1^{(j)}(\bA) & \cdots &
  \bpsi_k^{(j)}(\bA) \end{bmatrix},\quad j=1,2,
\end{equation}
with $\bpsi_i^{(1)}(\bA)$ and $\bpsi_i^{(2)}(\bA)$ denoting the left
and right singular vectors, respectively, of $\bA$ corresponding to
$\sigma_i(\bA)$, for $i=1,\dots,\min\{k_1,k_2\}$.
\end{lemma}

\subsection{Maximal Correlation Features}

We now have the following result, which relates the modal
decomposition and correlation maximization, and reveals the role of
the Ky Fan $k$-norms (as defined in, e.g., \cite[Section~7.4.8]{hj12})
in the analysis.
\begin{proposition}
\label{prop:var-modal}
For any $k\in\{1,\dots,K-1\}$, 
the dominant $k$ features \eqref{eq:fgi-def}
in \propref{prop:modal}, i.e., 
\begin{equation} 
f^k_* \defeq (f_1^*,\dots,f_k^*) \quad\text{and}\quad
g^k_* \defeq (g_1^*,\dots,g_k^*),
\label{eq:fgk*-def}
\end{equation}
are obtained via\footnote{We use $\norm{\cdot}$ to denote the Euclidean
  norm, i.e., $\bnorm{a^k}=\sqrt{\sum_{i=1}^k a_i^2}$ for any $k$ and $a^k$.}
\begin{subequations} 
\begin{align} 
(f^k_*,g^k_*)
&= \argmin_{(f^k,g^k)\in\cF_k\times\cG_k} \bE{\bnorm{f^k(X)-g^k(Y)}^2} \notag\\
&= \argmax_{(f^k,g^k)\in\cF_k\times\cG_k} \sigma(f^k,g^k),
\label{eq:hgr-def}
\end{align}
where
\begin{equation}
\sigma(f^k,g^k) \defeq \E{\bigl(f^k(X)\bigr)^\T g^k(Y)} 
\label{eq:hgr-k-veccor}
\end{equation}
and
\begin{align} 
\cF_k &\defeq \Bigl\{ f^k\colon
\E{f^k(X)} = \bzero,\quad
\E{f^k(X)\,f^k(X)^\T} = \bI \Bigr\}
\label{eq:cF-def} \\
\cG_k &\defeq \Bigl\{ g^k\colon 
\E{g^k(Y)} = \bzero, \quad
\E{g^k(Y)\,g^k(Y)^\T} = \bI \Bigr\}.
\label{eq:cG-def}
\end{align}%
\label{eq:hgr-k}%
\end{subequations}
Moreover, the resulting maximal correlation is
\begin{equation}
\sigma(f_*^k,g_*^k) = \E{\bigl(f_*^k(X)\bigr)^\T g_*^k(Y)} =
\sum_{i=1}^k \sigma_i, 
\label{eq:hgr-opt}
\end{equation}
which we note is the Ky Fan $k$-norm of $\dtmt$.\footnote{We use
  $\kyfank{\cdot}$ to denote the Ky Fan $k$-norm of its argument,
  i.e., for $\bA\in\reals^{k_1\times, k_2}$,  
\begin{equation} 
\kyfank{\bA} \defeq \sum_{i=1}^k \sigma_i(\bA),
\label{eq:kyfank-def}
\end{equation}
with $\sigma_1(\bA)\ge \cdots \ge \sigma_k(\bA)$ denoting its
singuar values.}
\end{proposition}
The quantity \eqref{eq:hgr-opt} is often referred to as the
Hirschfeld-Gebelein-R\'enyi (HGR) maximal correlation associated
with the distribution $P_{X,Y}$ (particularly in the special case
$k=1$).

\begin{IEEEproof}
First, note that the constraints \eqref{eq:cF-def} and
\eqref{eq:cG-def} express \eqref{eq:fgi-moments} in
\propref{prop:modal}.  Next, to facilitate our development, we define
        [cf. \eqref{eq:fvXY-def}]
\begin{subequations} 
\begin{align}
\fvX_i(x) &\defeq \sqrt{P_X(x)} \, f_i(x),\quad x\in\X \\
\fvY_i(y) &\defeq \sqrt{P_Y(y)} \, g_i(y), \quad y\in\Y.
\end{align}
\label{eq:fvXY-i-def}%
\end{subequations}
for $i=1,\dots,K$
We refer to $\fvX_i$ and $\fvY_i$ as the \emph{feature vectors}
associated with the feature functions $f_i$ and $g_i$, respectively,
and we further use $\bfvX_i$ and $\bfvY_i$ to denote column
vectors whose $x$\/th and $y$\/th entries are $\fvX_i(x)$ and
$\fvY_i(y)$, respectively.  Then 
\begin{subequations} 
\begin{equation}
\sigma(f^k,g^k) = \sum_{i=1}^k \sigma_i(f_i,g_i)
\end{equation}
with 
\begin{equation}
\sigma_i(f_i,g_i) = \E{f_i(X)\,g_i(Y)}
= \bigl(\bfvY_i\bigr)^\T \dtm \, \bfvX_i
= \bigl(\bfvY_i\bigr)^\T \dtmt \, \bfvX_i,
\label{eq:sigma-i-def}
\end{equation}
\label{eq:sigma-decomp}%
\end{subequations}
where the last equality in \eqref{eq:sigma-i-def} follows from the
mean constraints in \eqref{eq:cF-def} and \eqref{eq:cG-def}, which imply,
for $i=1,\dots,K$,
\begin{equation*}
\sum_{x\in\X} \sqrt{P_X(x)} \, \fvX_i(x) = 
\sum_{y\in\Y} \sqrt{P_Y(y)} \, \fvY_i(y) = 0.
\end{equation*}
In turn, from \eqref{eq:sigma-decomp} we have
\begin{equation} 
\sigma(f^k,g^k) 
= \sum_{i=1}^k \bigl(\bfvY_i\bigr)^\T \dtmt \, \bfvX_i
= \tr\Bigl( \bigl(\bfVY\bigr)^\T \dtmt \, \bfVX \Bigr) ,
\label{eq:sigma-tr}
\end{equation}
where 
\begin{subequations} 
\begin{align}
\bfVX &\defeq \begin{bmatrix} \bfvX_1 & \cdots & \bfvX_k \end{bmatrix}
\label{eq:bfVX-k-def} \\
\bfVY &\defeq \begin{bmatrix} \bfvY_1 & \cdots & \bfvY_k \end{bmatrix}.
\label{eq:bfVY-k-def}
\end{align}
\label{eq:bfV-k-def}%
\end{subequations}
Moreover, from the covariance constraints in \eqref{eq:cF-def} and
\eqref{eq:cG-def} we have 
\begin{equation}
\bigl(\bfVX\bigr)^\T \bfVX = \bigl(\bfVY\bigr)^\T \bfVY = \bI.
\label{eq:fvXY-orthon}
\end{equation}
Hence, applying \lemref{lem:svd-k} we immediately obtain that
\eqref{eq:sigma-tr} is maximized subject to \eqref{eq:fvXY-orthon} by
the feature vectors
\begin{subequations} 
\begin{align} 
\bfVX &= \bPsi^X_{(k)} \label{eq:bfVX-opt}\\
\bfVY &= \bPsi^Y_{(k)},\label{eq:bfVY-opt}
\end{align}
\label{eq:bfVXY-opt}%
\end{subequations}
with 
\begin{subequations} 
\begin{align} 
\bPsi^X_{(k)} &\defeq 
\begin{bmatrix} \bpsi^X_1 & \cdots  & \bpsi^X_k \end{bmatrix} \label{eq:bPsiX-k-def}\\
\bPsi^X_{(k)} &\defeq 
\begin{bmatrix} \bpsi^X_1 & \cdots  & \bpsi^X_k \end{bmatrix},
\label{eq:bPsiY-k-def}
\end{align}%
\label{eq:bPsi-k-def}%
\end{subequations}
whence $f_i^*$ and $g_i^*$ as given by \eqref{eq:fgi-def}, for
$i=1,\dots,k$.   The final statement of the proposition follows
immediately from the properties of the SVD; specifically,
\begin{equation*}
\bigl(\bpsi^Y_i\bigr)^\T \dtmt\, \bpsi^X_i = \sigma_i,\quad i=1,\dots,k.
\end{equation*}
\end{IEEEproof}

\section{Local Information Geometry}
\label{sec:geom-simplex}

Further interpretation of the features $f_*^{K-1}$ and $g_*^{K-1}$
arising out of the modal decomposition of \secref{sec:modal} benefits
from developing the underlying inner product space.  More
specifically, a local analysis of information geometry leads to key
information-theoretic interpretations of \eqref{eq:fgi-def} as
\emph{universal} features.  Accordingly, we begin with a
foundation for such analysis.

\subsection{Basic Concepts, Terminology, and Notation}
\label{sec:local-concepts}

Let $\simpZ$ denote the space of distributions on some finite alphabet
$\Z$, where $\cardZ<\infty$, and let $\relint(\simpZ)$ denote the
relative interior of $\simpZ$, i.e., the subset of strictly positive
distributions.  

\begin{definition}[$\eps$-Neighborhood]
\label{def:nhbd}
For a given $\eps>0$, the $\eps$-neighborhood of a reference
distribution $\refgen \in \relint(\simpZ)$ is the set of distributions
in a (Neyman) $\chi^2$-divergence \cite{jn49} ball of radius $\eps^2$ about
$\refgen$, i.e.,
\begin{subequations} 
\begin{equation}
\nbhd_\eps^\Z(\refgen) 
\defeq \biggl\{ P' \in \simpZ \colon
D_{\chi^2}(P'\|\refgen) \le \eps^2 \biggr\},
\end{equation}
where for $P\in\simpZ$ and $Q\in\relint(\simpZ)$, 
\begin{equation} 
D_{\chi^2}(P\|Q) \defeq \sum_{z\in\Z} \frac{\bigl(Q(z)-P(z)\bigr)^2}{Q(z)}.
\label{eq:chi2-def}
\end{equation}
\label{eq:nbhd-def}%
\end{subequations}
\end{definition}

In the sequel, we assume that all the distributions of interest,
including all empirical distributions that may be observed, lie in
such an $\eps$-neighborhood of the prescribed $\refgen$.  While we
don't restrict $\eps$ to be small, most of our information-theoretic
insights arise from the asymptotics corresponding to $\eps\to0$.   

An equivalent representation for a distribution
$P\in\nbhd^\Z_\eps(\refgen)$ is in terms of its \emph{information
  vector}
\begin{equation}
\ivgen(z) \defeq \frac{P(z) -
  \refgen(z)}{\eps\sqrt{\refgen(z)}},
\label{eq:ivgen-def}
\end{equation}
which we note satisfies
\begin{equation}
\norm{\phi} \le 1,
\label{eq:phi-norm-bnd}
\end{equation}
with $\norm{\cdot}$ denoting the usual Euclidean
norm.\footnote{Specifically, for $\phi$ defined on $\Z$,
\begin{equation*}
\norm{\phi}^2 \defeq \sum_{z\in\Z} \ivgen(z)^2.
\end{equation*}
} We will sometimes find it convenient to express $\phi=\ivgen(\cdot)$
as a $\cardZ$-di\-men\-sion\-al column vector $\bphi$, according to some
arbitrarily chosen but fixed ordering of the elements of $\Z$.

Hence, we can equivalently interpret the $(\cardZ-1)$-di\-men\-sion\-al
neighborhood $\nbhd_\eps^\Z(\refgen)$ as the set of distributions whose
corresponding information vectors lie in the unit Euclidean ball about
the origin.  Note that since
\begin{equation} 
\sum_{z\in\Z} \sqrt{\refgen(z)} \, \ivgen(z) = 0,
\label{eq:iv-constr}
\end{equation}
the $(\cardZ-1)$-di\-men\-sion\-al vector space subset
\begin{equation} 
\ivspacegen(\refgen) = \left\{ \ivgen \colon \bip{\sqrt{\refgen}}{\phi} = 0
\text{ and } \norm{\ivgen}\le 1 \right\},
\label{eq:ivspace-def}
\end{equation}
with $\ip{\cdot}{\cdot}$ denoting the usual Euclidean inner
product,\footnote{Specifically, for $\phi_1$ and $\phi_2$ defined on $\Z$,
\begin{equation*} 
\ip{\phi_1}{\phi_2} \defeq \sum_{z\in\Z} \phi_1(z)\,\phi_2(z).
\end{equation*}
} characterizes all the possible information vectors:
$\ivgen\in\ivspacegen(\refgen)$ if and only if
$P\in\nbhd_\eps^\Z(\refgen)$, for all $\eps$ sufficiently small.  It is convenient to refer to
$\ivspacegen(\refgen)$ as \emph{information space}.   When the
relevant reference distribution $\refgen$ is clear from context we
will generally omit it from our notation, and simply use $\ivspacegen$
to refer to this space. 

For a feature function $h$, we let
\begin{equation} 
\fvgen(z) \defeq \sqrt{P_0(z)}\, h(z)
\label{eq:fvgen-def}
\end{equation}
denote its associated feature vector.  As with information vectors, we
will sometimes find it convenient to express $\fvgen=\fvgen(\cdot)$ as
a $\cardZ$-dimensional column vector $\bfvgen$, according to the
chosen ordering of the elements of $\Z$.  Moreover, there is an
effective equivalence of feature vectors and information vectors,
which the following proposition establishes.  A proof is provided in
\appref{app:h-P-equiv}.
\begin{proposition}
\label{prop:h-P-equiv}
Let $\refgen\in\relint(\simpZ)$ be an arbitrary reference
distribution, and $\eps$ a positive constant.  Then 
for any distribution $P\in\simpZ$, 
\begin{equation}
h(z) = \frac1\eps \left(\frac{P(z)}{\refgen(z)}-1\right)
\label{eq:P-to-h}
\end{equation}
is a feature function satisfying 
\begin{equation}
\bEd{\refgen}{h(Z)}=0,
\label{eq:fvgen-0mean}
\end{equation}
and has as
its feature vector the information vector of $P(z)$, i.e.,
\begin{equation}
\fvgen(z) = \ivgen(z) = \frac{P(z)-\refgen(z)}{\eps\sqrt{P_0(z)}}.
\label{eq:phi-to-fvgen}
\end{equation}
Conversely, for any feature function $h\colon \Z\mapsto\reals$ such
that \eqref{eq:fvgen-0mean} holds,
\begin{equation}
P(z) = \refgen(z)\bigl(1+\eps\, h(z)\bigr)
\label{eq:h-to-P}
\end{equation}
is a valid distribution for all $\eps$ sufficiently small, and has as its
information vector the feature vector of $h$, i.e., 
\begin{equation}
\ivgen(z) = \fvgen(z) = \sqrt{P_0(z)} \, h(z).
\end{equation}
\end{proposition}

The following corollary of \propref{prop:h-P-equiv} specific to the
case of (relative) \emph{log-likelihood} feature functions is further
useful in our analysis.  A proof is provided in \appref{app:ll-equiv}.
\begin{corollary}
\label{corol:ll-equiv}
Let $\refgen\in\relint(\simpZ)$ be an arbitrary reference distribution
and $\eps$ a positive constant.  Then for any distribution
$P\in\nbhd_\eps^\Z(\refgen)$ with associated information vector
$\phi$, the feature vector $\fvgen_\mathrm{LL}$ associated with the
relative log-likelihood feature function\footnote{Throughout, all
  logarithms are base $e$, i.e., natural.}
\begin{equation} 
h_\mathrm{LL}(z) 
\defeq \frac1\eps \left( \log \frac{P(z)}{\refgen(z)} -
\Ed{\refgen}{\log\frac{P(Z)}{\refgen(Z)}}\right),\quad z\in\Z
\label{eq:L-def}
\end{equation}
satisfies\footnote{Note that the $o(1)$ term has zero mean with
  respect to $\refgen$, consistent with $\fvgen_\mathrm{LL}\in\ivspacegen(\refgen)$.} 
\begin{equation}
\fvgen_\mathrm{LL}(z) = \ivgen(z) + o(1),\quad \eps\to0,\ z\in\Z.
\label{eq:LL-phi}
\end{equation}
Conversely, every feature function $h\colon \Z\mapsto\reals$
satisfying $\bEd{\refgen}{h(Z)}=0$ can be interpreted to first order
as a (relative) log-likelihood, i.e., can be expressed in the form
\begin{equation}
h(z) = \frac1{\eps} \left( \log\frac{P(z)}{\refgen(z)} -
  \Ed{\refgen}{\log\frac{P(Z)}{\refgen(Z)}} \right) + o(1),\quad z\in\Z
\label{eq:h-LL}
\end{equation}
as $\eps\to0$ for some $P\in\simpZ$
\end{corollary}

A consequence of \propref{prop:h-P-equiv} is that we do not need to
distinguish between feature vectors and information vectors in the
underlying inner product space.  Indeed, note that when without loss
of generality we normalize a feature $h$ so that both
\eqref{eq:fvgen-0mean} and
\begin{equation*}
\bEd{\refgen}{h(Z)^2} = 1,
\end{equation*}
are satisfied, then we have $\fvgen\in\ivspacegen(\refgen)$, where $\fvgen$ is
the feature vector associated with $h$, as defined in \eqref{eq:fvgen-def}.

The following lemma, verified in \appref{app:Eh-char}, interprets
inner products between feature vectors and information vectors.
\begin{lemma}
\label{lem:Eh-char}
For any $\refgen\in\relint(\simpZ)$, let $h$ be a feature function
satisfying \eqref{eq:fvgen-0mean} with associated feature vector
$\fvgen\in\ivspacegen(\refgen)$.  Then for any $\eps>0$ and 
$\distgen\in\nbhd_\eps^\Z(\refgen)$ with associated information
vector $\ivgen\in\ivspacegen(\refgen)$,
\begin{equation*}
\Ed{\distgen}{h(Z)} = \eps\, \bip{\ivgen}{\fvgen}.
\end{equation*}
\end{lemma}

The squared-norm of a feature vector is its variance; specifically,
for a feature function $h$ satisfying \eqref{eq:fvgen-0mean} so
$\fvgen\in\ivspacegen(\refgen)$,
\begin{equation}
\bEd{\refgen}{h(Z)^2} = \norm{\fvgen}^2.
\end{equation}
However, it is natural to interpret the squared-norm of an information
vector in terms of Kullback-Leibler (KL) divergence\footnote{For
  $P,Q\in\simpZ$, we use the usual
\begin{equation*} 
D(P\|Q)=\sum_{z\in\Z} P(z)\,\log\frac{P(z)}{Q(z)}
\end{equation*}
to denote KL divergence of $Q$ from $P$.}  with respect to $\refgen$,
which follows as a special case of the following more general lemma.
A proof is provided in \appref{app:div-deriv}.  
\begin{lemma}
\label{lem:D-char}
For a given $\refgen\in\relint(\simpZ)$ and $\eps>0$,
let $P_1,P_2\in\nbhd_\eps^\Z(\refgen)$ be arbitrary, and let $\phi_1$
and $\phi_2$ denote the corresponding information vectors,
respectively.  Then
\begin{equation} 
D(P_1\|P_2) \defeq \sum_{z\in\Z} P_1(z) \log\frac{P_1(z)}{P_2(z)}
= \frac{\eps^2}{2} \, \| \phi_1 - \phi_2\|^2 + o(\eps^2),
\label{eq:div-euclid}
\end{equation}
a special case of which is, for $P\in\nbhd_\eps^\Z(\refgen)$ and with
$\phi$ denoting its information vector,\footnote{Note, for comparison,
  that $D_{\chi^2}(P\|\refgen) = \eps^2 \norm{\phi}^2$.}
\begin{equation}
D(P\|\refgen) = \frac{\eps^2}2 \, \norm{\phi}^2 + o(\eps^2),
\label{eq:div-euclid-ref}
\end{equation}
since $\phi_0\equiv0$ is the information vector associated with
$\refgen$.
\end{lemma}
Note that as \eqref{eq:div-euclid} reflects, divergence is symmetric in $P_1$
and $P_2$ to first order in $\eps^2$. 

Additionally, in \eqref{eq:div-euclid} we recognize $\phi_1-\phi_2$ as,
to first order, the information vector associated with the
log-likelihood ratio feature function
\begin{subequations}
\begin{equation}
h_\mathrm{LLR}(z) \defeq \frac1{\eps} \left( \log\frac{P_1(z)}{P_2(z)}
- \Ed{\refgen}{\log\frac{P_1(z)}{P_2(z)}} \right) .
\label{eq:llr-def}
\end{equation}
In particular, 
since
\begin{equation*}
\log \frac{P_1(z)}{P_2(z)} = \log \frac{P_1(z)}{P_0(z)} - \log
\frac{P_2(z)}{P_0(z)}, 
\end{equation*}
it follows from the first part of
\corolref{corol:ll-equiv} that \eqref{eq:llr-def} has feature vector
\begin{equation}
\fvgen_\mathrm{LLR}(z) = \phi_1(z)-\phi_2(z) + o(1),\quad \eps\to
0,\quad z\in\Z. 
\label{eq:fvgen-llr}
\end{equation}
\label{eq:llr-form}%
\end{subequations}

It is also important to appreciate that \eqref{eq:div-euclid} is
invariant to the choice of reference distribution within the
neighborhood, which is an immediate consequence of the following
result, verified in \appref{app:refdist-invar}.
\begin{lemma}
\label{lem:refdist-invar}
For a given $\refgen\in\relint(\simpZ)$ and $\eps>0$ sufficiently small
that $\nbhd_\eps^\Z(\refgen)\subset\relint(\simpZ)$, let
$P_1,P_2\in\nbhd_\eps^\Z(\refgen)$ be arbitrary, and let $\phi_1$ and
$\phi_2$ be the corresponding information vectors.  Then for any
$\Pt_0\in\nbhd_\eps^\Z(\refgen)$, the information vectors
\begin{equation*}
\phit_1(z) \defeq \frac{P_1(z)-\Pt_0(z)}{\eps\,\sqrt{\Pt_0(z)}}
\quad\text{and}\quad
\phit_2(z) \defeq \frac{P_2(z)-\Pt_0(z)}{\eps\,\sqrt{\Pt_0(z)}}
\end{equation*}
satisfy, for each $z\in\Z$,
\begin{equation}
\phit_1(z) - \phit_2(z) = \bigl(\phi_1(z) -
\phi_2(z)\bigr) \bigl(1+o(1)\bigr),\quad \eps\to0. 
\label{eq:refdist-invar}
\end{equation}
\end{lemma}

\subsection{Weakly Dependent Variables}
\label{sec:eps-dep}

An instance of local analysis corresponds to weak dependence between
variables, a concept we formally define as follows. 
\begin{definition}[$\eps$-Dependence]
\label{def:eps-dep}
Let $Z$ and $W$ be defined over alphabets $\Z$ and $\W$,
respectively, and distributed according to $P_{Z,W}\in\simpZW$,
where $\simpZW$ is the (usual) restriction of the simplex to
distributions with strictly positive marginals.  
Then $Z$ and $W$
are $\eps$-dependent if there exists an $\eps>0$ such
that\footnote{Note that the condition \eqref{eq:eps-dep-ZW} is
  equivalent to 
\begin{equation*} 
D_{\chi^2}\bigl(P_{Z,W} \bigm\| P_{Z}P_{W}\bigr) \le \eps^2,
\end{equation*}
the left-hand side of which defines mutual information with respect to
$\chi^2$-divergence.  This mutual information was historically
referred to as ``mean-square contingency'' \cite{hoh35}, a concept
introduced by Pearson.}
\begin{equation}
P_{Z,W}\in\nbhd_\eps^{\Z\times\W}(P_{Z}P_{W}),
\label{eq:eps-dep-ZW}
\end{equation}
where
$P_{Z}$ and $P_{W}$ are the marginal distributions associated
with  $P_{Z,W}$. 
\end{definition}

As related notions of $\eps$-dependence, we can replace
\eqref{eq:eps-dep-ZW} with one of 
\begin{align}
P_{W|Z}(\cdot|z) &\in\nbhd_\eps^{\W}(P_{W}),\quad\text{all
  $z\in\Z$} \label{eq:eps-dep-W|Z}\\ 
P_{Z|W}(\cdot|w) &\in\nbhd_\eps^{\Z}(P_{Z}),\quad\text{all
  $w\in\W$}. \label{eq:eps-dep-Z|W} 
\end{align}
Asymptotically, all these notions are equivalent, however, as the
following lemma establishes; a proof is provided in \appref{app:eps-dep-equiv}
\begin{lemma}
\label{lem:eps-dep-equiv}
Let $Z$ and $W$ be defined over alphabets $\Z$ and $\W$,
respectively, and distributed according to $P_{Z,W}\in\simpZW$,
where $\simpZW$ is the (usual) restriction of the simplex to
distributions with strictly positive marginals.   When
\begin{subequations} 
\begin{align} 
\liminf_{\eps\to0} P_Z(z) &>0,\quad\text{all $z\in\Z$}\\
\liminf_{\eps\to0} P_W(w) &>0,\quad\text{all $w\in\W$},
\end{align}
\label{eq:PZ-PW-const}%
\end{subequations}
the following statements are equivalent (as $\eps\to0$):
\begin{subequations} 
\begin{align}
P_{Z,W} &\in\nbhd_{O(\eps)}^{\Z\times\W}(P_{Z}P_{W}) \label{eq:Oeps-dep-ZW}\\
P_{W|Z}(\cdot|z) &\in\nbhd_{O(\eps)}^{\W}(P_{W}),\quad\text{all
  $z\in\Z$} \label{eq:Oeps-dep-W|Z}\\ 
P_{Z|W}(\cdot|w) &\in\nbhd_{O(\eps)}^{\Z}(P_{Z}),\quad\text{all
  $w\in\W$}. \label{eq:Oeps-dep-Z|W} 
\end{align}
\label{eq:Oeps-dep}%
\end{subequations}
\end{lemma}

Accordingly, any of \eqref{eq:Oeps-dep} can be used to characterize
$O(\eps)$-dependence.  In the sequel, except where the distinction is
needed, with some abuse of terminology we will use 
$\eps$-dependence and $O(\eps)$-dependence interchangeably.

A further asymptotic equivalence between the notion of
$\eps$-dependence based on $\chi^2$-divergence and one based on KL
divergence is established by the following lemma, whose proof
is provided in \appref{app:weak-dep-KL}.
\begin{lemma}
\label{lem:weak-dep-KL}
Under the hypotheses of \lemref{lem:eps-dep-equiv},
\begin{equation}
I(Z;W)\! =\! O(\eps^2) 
\quad \text{if and only if}\quad
P_{Z,W}\!\in\!\nbhd_{O(\eps)}^{\Z\times\W}(P_ZP_W).
\end{equation}
\end{lemma}

Finally, for completeness, we
have the following asymptotic equivalences among notions of
$\eps$-dependence based on KL divergence, analogous to
\lemref{lem:eps-dep-equiv}.  A proof is provided in
\appref{app:KL-eps-dep-equiv}. 
\begin{lemma}
\label{lem:KL-eps-dep-equiv}
Under the hypotheses of \lemref{lem:eps-dep-equiv},
the following statements are equivalent (as $\eps\to0$):
\begin{subequations} 
\begin{align}
I(Z;W) &= O(\eps^2) \label{eq:KL-eps-dep-ZW}\\
D\bigl(P_{W|Z}(\cdot|z)\|P_W\bigr) &= O(\eps^2),\quad\text{all
  $z\in\Z$} \label{eq:KL-eps-dep-W|Z}\\ 
D\bigl(P_{Z|W}(\cdot|w)\|P_Z\bigr) &= O(\eps^2),\quad\text{all
  $w\in\W$}. \label{eq:KL-eps-dep-Z|W} 
\end{align}
\label{eq:KL-eps-dep}%
\end{subequations}
\end{lemma}

We will exploit the various equivalences
\eqref{eq:Oeps-dep}--\eqref{eq:KL-eps-dep} in our analysis.

\subsection{The Modal Decomposition of Mutual Information}
\label{sec:svd-mi}

The modal decomposition \eqref{eq:modal} of $P_{X,Y}$ leads directly
to a corresponding decomposition of mutual information when $X$ and
$Y$ are weakly dependent.     In particular, we have the following result.
\begin{lemma}
\label{lem:I-modal}
Let $X\in\X$ and $Y\in\Y$ with $P_{X,Y}\in\simpXY$ be $\eps$-dependent
random variables, and let $\dtmt$ denote the associated CDM.  Then
\begin{equation}
I(X;Y) 
= \frac12 \bfrob{\dtmt}^2 + o(\eps^2) 
= \frac12 \sum_{i=1}^{K-1} \sigma_i^2 + o(\eps^2),
\label{eq:I-modal}
\end{equation}
where the summation is $O(\eps^2)$.
\end{lemma}
\begin{IEEEproof}
It suffices to make the choices $P=P_{X,Y}$ and $P_0=P_XP_Y$ in 
\eqref{eq:div-euclid-ref} of \lemref{lem:D-char}, and recognize
that the corresponding information vector---which is convenient to
express as a matrix in this case---has elements
\begin{equation}
\phi(x,y) =
\frac{P_{X,Y}(x,y)-P_X(x)\,P_Y(y)}{\eps\sqrt{P_X(x)}\,\sqrt{P_Y(y)}} =
\frac1\eps\, \dtmts(x,y).
\label{eq:phixy-def}
\end{equation}
Then, since the Frobenius norm of an information vector in matrix form
coincides with its Euclidean norm, and since for any matrix $\bA$
whose singular values are $\sigma_1(\bA),\dots,\sigma_l(\bA)$ for some
$l$, we have $\frob{\bA}^2 = \sum_{i=1}^l \sigma_i(\bA)^2$,
\eqref{eq:I-modal} follows.  Finally, that the first term on the
right-hand side of \eqref{eq:I-modal} is at most $O(\eps^2)$ follows
from applying the constraint \eqref{eq:phi-norm-bnd} to the
information vector defined via \eqref{eq:phixy-def}.
\end{IEEEproof}

A key interpetation of the decomposition \eqref{eq:I-modal} is as
follows.  For each $1\le k\le K-1$, the bivariate function
\begin{subequations} 
\begin{equation}
P_{X,Y}^{(k)}(x,y)\defeq P_X(x)\,P_Y(y) \left( 1 + \sum_{i=1}^k \sigma_i \,
f_i^*(x)\,g_i^*(y) \right)
\label{eq:Pxyk-exp}
\end{equation}
obtained by truncating \eqref{eq:modal} sums to unity and, for all
$\eps$ sufficiently small, is nonnegative for all
$(x,y)\in\X\times\Y$, so has the interpretation as a joint
distribution for new variables $\bigl(X^{(k)},Y^{(k)}\bigr)$, i.e.,
$P_{X,Y}^{(k)} = P_{X^{(k)},Y^{(k)}}$, having the same (original)
marginals $P_X$ and $P_Y$ for all such $k$.  Moreover, these new
variables have mutual information 
\begin{equation}
I\bigl(X^{(k)};Y^{(k)}\bigr) = \frac12 \sum_{i=1}^k \sigma_i^2 +
o(\eps^2). 
\end{equation}
\label{eq:PIk-def}%
\end{subequations}
Hence, the $k$\/th term in the expansion contributes an increment of
$\sigma_k^2/2+o(\eps^2)$ to the mutual information.  From this
perspective, the chosen ordering captures the largest proportion of
mutual information from the fewest number of terms.   Valuable
complementary perspectives on these order-$k$ distributions will
become apparent later in our development.

\subsection{The Local Geometry of Decision Making}
\label{sec:geom-inf}

In our development, it will be useful to exploit a geometric
interpretation of traditional binary hypothesis testing, which we now
describe.  In particular, suppose we observe $m$ samples $z_1^m=(z_1,
\dots, z_m)$ drawn in an independent, identically distributed (i.i.d.)
manner from either distribution $P_1$ or
distribution $P_2$, where $P_1,P_2\in\nbhd_\eps^\Z(\refgen)$.  As in 
\secref{sec:local-concepts}, let $\phi_1$ and $\phi_2$ denote the
associated information vectors.

For this problem, for some $1\le k\le K-1$, consider a sequence of
$k$-dimensional statistics
\begin{subequations} 
\begin{equation}
\ell^k = (\ell_1,\dots,\ell_k)
\end{equation}
with
\begin{equation}
\ell_l = \frac1m \sum_{j=1}^m h_l(z_j), \quad l = 1, \ldots, k,. 
\label{eq:ell-l}
\end{equation}%
\label{eq:dec-stat-k}%
\end{subequations}
for some feature functions $h^k=(h_1,\dots,h_k)$ with associated
feature vectors $\fvgen^k=(\fvgen_1,\dots,\fvgen_k)$. 

Without loss of generality we restrict our attention to normalized
feature functions such that the statistics
\begin{equation*}
h^k(Z) = \bigl(h_1(Z),\dots,h_k(Z)\bigr)
\end{equation*}
are zero mean, unit-variance, and uncorrelated with respect to
$\refgen$, i.e.,
\begin{subequations} 
\begin{align} 
\bEd{\refgen}{h_i(Z)} &= 0,\quad i\in\{1,\dots,k\} \label{eq:h-0mean}\\
\bEd{\refgen}{h_i(Z)\,h_j(Z)} &= \kron_{i=j},\quad i,j\in\{1,\dots,k\}.
\label{eq:h-uncorr}
\end{align}%
\label{eq:h-norm}%
\end{subequations}
Indeed, if $h^k(Z)$ had any other mean and (nonsingular)
covariance structure, then we could apply an invertible transformation
to $\ell^k$ to generate an equivalent statistic $\tilde{\ell}^k$ with
the desired structure.\footnote{In particular, with $\bell$ denoting the
vector representation of $\ell^k$, if $\bell$ has mean vector
$\bmu_{\bell}$ and covariance matrix is $\bLa_{\bell}$, then
$\smash{\tilde{\bell} \defeq \bLa_{\bell}^{-1/2}\,(\bell-\bmu_{\bell})}$,
with $\bLa_{\bell}^{1/2}$ denoting any square root matrix of
$\bLa_{\bell}$, has mean $\bmu_{\tilde{\bell}}=\bzero$ and covariance
matrix $\bLa_{\tilde{\bell}}=\bI$ as desired.}

\begin{figure}[t]
\centering
\includegraphics[width=.58\columnwidth]{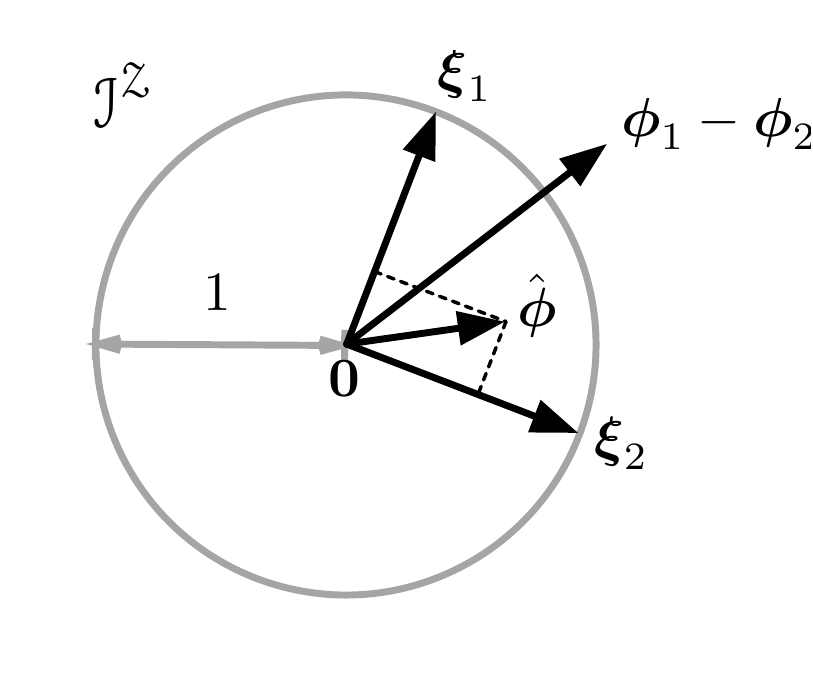}
\caption{The local geometry of decision making for distingishing
  i.i.d.\ samples $z^m$ over alphabet $\Z$ from one of
  $P_1,P_2\in\simpZ$ based on a statistic $\ell = (1/m) \sum_{i=1}^m
  h^k(z_i)$ involving feature functions $h^k$.  In information space
  $\ivspacegen(\refgen)$, where the reference distribution $\refgen$ maps to
  the origin, $\ell$ corresponds to the projection of the information
  vector $\hat{\bphi}$ for $\Ph$ onto subspace spanned by the feature
  vectors $\bfvgen^k$ for $h^k$.  The optimum decision rule projects
  $\hat{\bphi}$ directly onto $\bphi_1-\bphi_2$, the feature vector
  associated with the log-likehood ratio $h_\mathrm{LLR}$.
\label{fig:projection}}
\end{figure}

With $\empgen$ denoting the empirical distribution of the data
$z_1^m$, we can express \eqref{eq:ell-l} in the form
\begin{equation*} 
\ell_l
= \sum_{z\in\Z} \empgen(z) \, h_l(z) \notag\\
= \eps\, \bip{\ivemp}{\fvgen_l}
\end{equation*}
where we have used \lemref{lem:Eh-char} with
$\distgen = \empgen$, and
where
\begin{equation}
\ivemp(z) = 
\frac{1}{\eps} \frac{\empgen(z) -
  \refgen(z)}{\sqrt{\refgen(z)}}.
\label{eq:ivemp-def}
\end{equation}
is the \emph{observed} information vector [cf.\ \eqref{eq:ivgen-def}].
The associated geometry is depicted in \figref{fig:projection}.

Our main result is as follows, a proof of
which is provided in \appref{app:mismatch-k}. 
\begin{lemma}
\label{lem:mismatch-k}
Given a reference distribution $\refgen\in\relint(\simpZ)$ a constant
$\eps>0$ and integers $m$ and $k$, let $z_1,\dots,z_m$ denote
i.i.d.\ samples from one of $P_1$ or $P_2$, where
$P_1,P_2\in\nbhd_\eps^\Z(\refgen)$ and both strictly positive.  Then
the error probability $\pe$ for deciding whether $P_1$ or $P_2$ is the
generating distribution, based on a statistic $\ell^k$ of the form
\eqref{eq:dec-stat-k} with normalized feature functions $h^k$, decays
exponentially in $m$ as $m\to\infty$, with (Chernoff) exponent
\begin{subequations} 
\begin{equation}
\lim_{m\to\infty}\frac{-\log\pe}{m} \defeq E_{h^k} = \sum_{l=1}^k E_{h_l},
\end{equation}
where
\begin{equation}
E_{h_l} \defeq \frac{\eps^2}{8}  \ip{\ivgen_1-\ivgen_2}{\fvgen_l}^2 +
o(\eps^2). 
\end{equation}%
\label{eq:mm-exp-k}%
\end{subequations}
\end{lemma}

The 
$k$-fold local efficiency $\eff(h^k)$ of the
rule defined by $h^k$ quantifies the
goodness of the exponent \eqref{eq:mm-exp-k} in
\lemref{lem:mismatch-k} relative to the ideal exponent
\begin{equation*}
E \defeq \frac{\eps^2}{8} \norm{\ivgen_1-\ivgen_2}.
\end{equation*}
Specifically, 
\begin{equation}
\eff(h^k) \defeq \lim_{\eps\to0} \frac{E_{h^k}}{E} = 
\frac{\sum_{l=1}^k\ip{\ivgen_1-\ivgen_2}{\fvgen_l}^2}{\norm{\ivgen_1-\ivgen_2}^2}.
\end{equation}
It follows from Bessel's inequality that $0\le
\eff(h^k)\le 1$, and from
\eqref{eq:llr-form} that the upper bound is achieved by the choices
\begin{equation}
h_1 = h_\mathrm{LLR},\quad\text{and}\quad h_i\equiv0,\ i\in\{2,\dots,k\},
\end{equation}
i.e., the log-likelihood ratio is an optimum statistic, as expected.
In the sequel, we focus on inference scenarios in which such a statistic
cannot be used directly.

\section{Universal Feature Characterizations}
\label{sec:unifeature-char}

We now develop several complementary formulations of the universal
feature selection problem, and show that they yield the same features,
and that these features are directly related to the modal
decomposition of $P_{X,Y}$.  

As an initial viewpoint, when $X$ and $Y$ are weakly dependent, their
conditional distributions are exponential families in which the
features in the modal decomposition \eqref{eq:modal} are natural
statistics.  Specifically, when $X$ and $Y$ are $\eps$-dependent
according to \defref{def:eps-dep}, 
we have, starting from \eqref{eq:Y|X-form},
\begin{equation} 
P_{Y|X}(y|x) 
= P_Y(y) \expop{\sum_{i=1}^{K-1} \sigma_i \, f_i^*(x)\,g_i^*(y) + o(\eps)},
\label{eq:Y|X-expfam}
\end{equation}
where we have used the Taylor series approximation $\e^\omega= 1+
\omega + o(\omega)$ and that the exponent in the first term in
\eqref{eq:Y|X-expfam} is $O(\eps)$.  We recognize the posterior
\eqref{eq:Y|X-expfam} as an exponential family with natural parameters
$g_*^{K-1}(y)$ and natural statistics $f_*^{K-1}(x)$.  Moreover, by
symmetry we have, or equivalently via \eqref{eq:X|Y-form},
\begin{equation} 
P_{X|Y}(x|y) 
= P_X(x) \expop{\sum_{i=1}^{K-1} \sigma_i \, f_i^*(x)\,g_i^*(y) + o(\eps)},
\label{eq:X|Y-expfam}
\end{equation}
from which we see that for inferences about $X$ from $Y$, the roles of
the features are reversed in the associated posterior: $f_*^{K-1}(x)$
are the natural parameters and $g_*^{K-1}(y)$ are the natural
statistics.  

Since exponential families with such structure are widely used in
discriminative models for learning, we can interpret the
\eqref{eq:Y|X-expfam} and \eqref{eq:X|Y-expfam} as indicating
universal feature choices.  Moreover, dimensionally-reduced families
of the form
\begin{align*}
P^{(k)}_{Y|X}(y|x) 
&= P_Y(y) \expop{\sum_{i=1}^k \sigma_i \, f_i^*(x)\,g_i^*(y) + o(\eps)}\\
P^{(k)}_{X|Y}(x|y) 
&= P_X(x) \expop{\sum_{i=1}^k \sigma_i \, f_i^*(x)\,g_i^*(y) + o(\eps)}
\end{align*}
for some $1\le k\le K-1$ represent approximations that maximize the
retained mutual information, as per the discussion in \secref{sec:svd-mi}
surrounding \eqref{eq:PIk-def}.  We develop and interpret
these posterior distributions further in \secref{sec:nn}.  However,
there are other senses in which the features in \eqref{eq:modal} are
universal, which we develop first, and which require a data model we
now introduce.

\subsection{Latent Attribute and Statistic Model}
\label{sec:latent-model}

In the sequel, we further develop universal features from the modal
decomposition \eqref{eq:modal} via the introduction of latent
(auxiliary) variables.  Latent variable models have a long history in
facilitating both the interpretion and exploitation of relationships
in data.  While the original focus was on linear relationships,
corresponding to factor analysis as introduced by Spearman
\cite{cs1904}, the modern view is considerably broader; see, e.g.,
\cite{cmb99} for a discussion.

As we now describe, our treatment models scenarios in which the
inference task involving $X$ and $Y$ is not known in advance through
the introduction of latent \emph{attribute} variables whose values we
seek to determine.  We emphasize at the outset that in this model, we
treat $P_{X,Y}$ as known or, equivalently, to have been sufficiently
reliably estimated from training samples, an efficient process for
which we will later discuss.

We begin by formalizing the notion of an attribute.\footnote{More
  generally, we use \emph{attribute} to refer to an $\eps$-attribute in
  which there is no restriction on $\eps$, i.e., it can be arbitrarily
  large.} 
\begin{definition}[$\eps$-Attribute]
\label{def:attribute}
Given $\eps>0$ and $P_Z\in\relint(\simpZ)$ for some $\Z$, then $W$ on
some alphabet $\W$ with $2\le\cardW\le\cardZ$ and having distribution
$P_W\in\relint(\simpW)$ is an $\eps$-attribute of $Z$ if $W$ is
$\eps$-dependent on $Z$, i.e.,
\begin{equation*}
P_{Z|W}(\cdot|w)\in\nbhd_\eps^\Z(P_Z), \quad w\in\W,
\end{equation*}
$P_{Z|W}(\cdot|w)\not\in\nbhd_0^\Z(P_Z)$ for all $w\in\W$,
and $W$ is conditionally independent of all other variables in the
model given $Z$.
\end{definition}

Such attributes are specified by a collection of parameters.
In particular, we have the following.
\begin{definition}[$\eps$-Attribute Configuration]
\label{def:config}
Given $\eps>0$ and $P_Z\in\relint(\simpZ)$ for some $\Z$, then $\eps$-attribute
$W$ of $Z$ is characterized by its \emph{configuration}
\begin{align} 
\C^\Z_\eps(P_Z) &\defeq \biggl\{\,
\W,\, \{P_W(w),\, w\in\W\},\, \{P_{Z|W}(\cdot|w),\, w\in\W \} \colon \notag\\
&\quad\qquad P_{Z|W}(\cdot|w)\in\nbhd_\eps^\Z(P_Z),\ w\in\W, \notag\\
&\,\ \quad\qquad \sum_{w\in\W} P_W(w)\, P_{Z|W}(z|w)  =
P_Z(z),\ z\in\Z \biggr\}.
\label{eq:Cz-def}
\end{align}
which can be equivalently expressed in the form
\begin{align} 
\C^\Z_\eps(P_Z) &= \biggl\{\,
\W,\, \{P_W(w),\, w\in\W\},\, \{\bphi^{Z|W}_w,\, w\in\W \} \colon \notag\\
&\qquad\qquad \bphi^{Z|W}_w\in\ivspacegen(P_Z),\ w\in\W, \notag\\
&\qquad\qquad\qquad \sum_{w\in\W} P_W(w)\, \phi^{Z|W}_w(z) = 0,\ z\in\Z
\biggr\},
\label{eq:Cz-phi}
\end{align}
where
\begin{equation}
\phi^{Z|W}_w(z) \defeq
\frac{P_{Z|W}(z|w)-P_Z(z)}{\eps\,\sqrt{P_Z(z)}},\quad z\in\Z,\ w\in\W
\label{eq:iv-attribute-def}
\end{equation}
define the information vectors associated with the $\eps$-attribute $W$.
\end{definition}
In \defref{def:config}, we note that the equivalent form \eqref{eq:Cz-phi} is a consequence of
the fact the constraint
\begin{equation*}
\sum_{w\in\W} P_W(w)\, P_{Z|W}(z|w) = P_Z(z),
\end{equation*}
implies the information vectors must satisfy
\begin{equation}
\sum_{w\in\W} P_W(w)\, \ivgen^{Z|W}_w(z) = 0.
\label{eq:pzw-constr} 
\end{equation}

In the context of a given model $P_{X,Y}$, the attribute variables
$U$ and $V$ for $X$ and $Y$, respectively, are characterized by the
Markov structure
\begin{equation}
U \markov X \markov Y \markov V.
\label{eq:markov-uxyv}
\end{equation}
More generally, in the case of $m$ samples drawn from $P_{X,Y}$, our
model has the Markov structure
\begin{subequations} 
\begin{equation} 
U\markov X^m
\markov Y^m \markov V
\label{eq:full-markov-uxyv}
\end{equation}
with the conditional independence and memoryless structure
\begin{align} 
P_{X^m|U}(x^m|u) &= \prod_{i=1}^m P_{X|U}(x_i|u) \label{eq:full-markov-xu}\\
P_{Y^m|V}(y^m|v) &= \prod_{i=1}^m P_{Y|V}(y_i|v) \label{eq:full-markov-yv}\\
P_{X^m,Y^m}(x^m,y^m) &= \prod_{i=1}^m P_{X,Y}(x_i,y_i).\label{eq:full-markov-xy}
\end{align}%
\label{eq:full-markov}%
\end{subequations}
The attributes $U$ and $V$
can be interpreted as instances of \emph{class} variables, whose
values correspond to different categories of $X$ and $Y$ behavior,
respectively.

Our development focuses on the case where $U$ and $V$ depend only
weakly on $X$ and $Y$.  Specifically, we consider the
$\eps$-dependence
\begin{align*} 
P_{X|U}(\cdot|u) &\in\nbhd_\eps^\X(P_X), \quad\text{for all $u\in\U$} \\
P_{Y|V}(\cdot|v) &\in\nbhd_\eps^\Y(P_Y),\, \quad\text{for all $v\in\V$}.
\end{align*}
Via \lemref{lem:D-char}, $\eps$-dependence can be equivalently
expressed as the condition
\begin{subequations} 
\begin{align} 
D(P_{X|U}(\cdot|u)\|P_X) 
&\le \frac{\eps^2}{2}\bigl(1+o(1)\bigr), \quad\text{for all $u\in\U$}
\label{eq:XU-weak}\\
D(P_{Y|V}(\cdot|v)\|P_Y) 
&\le \frac{\eps^2}{2}\bigl(1+o(1)\bigr), \quad\text{for all $v\in\V$},
\label{eq:YV-weak}
\end{align}
\end{subequations}
which, in turn, of course implies
\begin{subequations} 
\begin{align} 
I(X;U) &= \sum_{u\in\U} P_U(u)\, D(P_{X|U}(\cdot|u)\| P_X) \le
\frac{\eps^2}{2}\bigl(1+o(1)\bigr) \label{eq:IXU-weak}\\
I(Y;V) &= \sum_{v\in\V} P_V(v)\, D(P_{Y|V}(\cdot|v)\| P_Y) \le
\frac{\eps^2}{2}\bigl(1+o(1)\bigr) \label{eq:IYV-weak}.%
\end{align}
\end{subequations}

For inferences about attributes $U$ and $V$, we will generally
consider statistics of the form
\begin{equation}
S^k \defeq \frac1m \sum_{i=1}^m f^k(X_i)\quad\text{and}\quad
T^k \defeq \frac1m \sum_{i=1}^m g^k(Y_i), 
\label{eq:stmk-def}
\end{equation}
for some $k\in\{1,\dots,K-1\}$ and feature choices $f^k\colon
\X\mapsto \reals^k$ and $g^k\colon \Y\mapsto \reals^k$.  Moreover, in
accordance with our earlier discussion, without loss of generality we
restrict our attention to normalized features,
i.e. $(f^k,g^k)\in\cF_k\times\cG_k$ with $\cF_k$ and $\cG_k$ as defined in
\eqref{eq:cF-def} and \eqref{eq:cG-def}, respectively.  As we will
develop, the particular choices
\begin{equation}
S^k_* \defeq \frac1m \sum_{i=1}^m f^k_*(X_i)\quad\text{and}\quad
T^k_* \defeq \frac1m \sum_{i=1}^m g^k_*(Y_i), 
\label{eq:stmk*-def}
\end{equation}
with $f^k_*$ and $g^k_*$ as defined in \eqref{eq:fgk*-def} play a
special role.

Finally, we will sometimes extend the model \eqref{eq:full-markov} to
the case of multidimensional $U$ and $V$ with special structure, which
we term \emph{multi-attributes}.\footnote{As in the case of
  attributes, we more generally use \emph{multi-attribute} to refer to
  an $\eps$-multi-attribute in which there is no restriction on
  $\eps$, i.e., it can be arbitrarily large.}
\begin{definition}[$\eps$-Multi-Attribute]
\label{def:orthog-attribute}
Given $\eps>0$, $l$, and $Z$ over some alphabet $\Z$, then an
attribute $W$ of $Z$ with configuration $\C^\Z_{l\eps}(P_Z)$ over
alphabet $\W$ is an $l$-dimensional $\eps$-multi-attribute $W^l$ over
alphabet $\W=\W_1\times\cdots\times\W_l$ if the variables $W^l$
are: \newline 1) such that
\begin{equation*}
\card{\W_i}\ge2\quad\text{and}\quad
  P_{W_i}\in\relint(\cP^{\W_i}),\quad i\in\{1,\dots,l\};
\end{equation*}
2) $\eps$-dependent on $Z$, i.e.,
\begin{equation*}
\begin{aligned}
P_{Z|W_i}(\cdot|w_i) &\in\nbhd_\eps^\Z(P_Z),\\
P_{Z|W_i}(\cdot|w_i) &\not\in\nbhd_0^\Z(P_Z),
\end{aligned}
\quad\text{all $w_i\in\W_i$
  and $i=1,\dots,l$;} 
\end{equation*}
3) conditionally independent given $Z$, i.e.,
\begin{equation*}
P_{W^l|Z}(w^l|z) = \prod_{i=1}^l P_{W_i|Z}(w_i|z),\quad\text{all
  $w^l\in\W$, $z\in\Z$};
\end{equation*}
and \newline
4) (marginally) independent, i.e., 
\begin{equation*}
P_{W^l}(w^l) = \prod_{i=1}^l P_{W_i}(w_i),\quad\text{all
  $w^l\in\W$}.
\end{equation*}
We use $\C^{\Z,l}_\eps(P_Z)$ to denote the configuration of such a
$\eps$-multi-attribute variable.
\end{definition}

Multi-attribute variables have the following key orthogonality
property.  A proof is provided in \appref{app:orthog}.  
\begin{lemma}
\label{lem:orthog}
For some $\eps>0$ and integer $l\ge1$, let $W^l$ be an
$\eps$-multi-attribute of $Z\in\Z$ over alphabet
$\W=\W_1\times\cdots\times\W_l$.  Then with the information vector
notation
\begin{equation}
\ivgen^{Z|W_i}_{w_i}(z) \defeq
\frac{P_{Z|W_i}(z|w_i)-P_Z(z)}{\eps\sqrt{P_Z(z)}},\quad i=1,\dots,l,
\label{eq:phiZWi-def}
\end{equation}
we have, for $i,j\in\{1,\dots,l\}$,
\begin{equation*}
\bip{\ivgen^{Z|W_i}_{w_i}}{\ivgen^{Z|W_j}_{w_j}} =0,\
\text{for all $i\ne j$, $w_i\in\W_i$ and $w_j\in\W_j$}.
\end{equation*}
\end{lemma}

In addition, multi-attributes admit the following information vector
decomposition.\footnote{Note that this decomposition implies that an
  $\eps$-multi-attribute is an $l\eps$-attribute.}
\begin{lemma}
\label{lem:ivma-sum}
For some $\eps>0$ and integer $l\ge1$, let $W^l$ be an $\eps$-multi-attribute
of $Z\in\Z$ over alphabet $\W=\W_1\times\cdots\times\W_l$.
Then with the information vector notation
\begin{equation} 
\ivgen^{Z|W^l}_{w^l}(z)
\defeq \frac{P_{Z|W^l}(z|w^l)-P_Z(z)}{\eps\sqrt{P_Z(z)}},
\label{eq:pxuk-def}
\end{equation}
and $\phi^{Z|W_i}_{w_i}$ as defined in \eqref{eq:phiZWi-def}, we have
\begin{equation}
\bphi^{Z|W^l}_{w^l} = \sum_{i=1}^l \bphi^{Z|W_i}_{w_i} + o(1).
\label{eq:ivma-sum}
\end{equation}
\end{lemma}
A proof is provided in \appref{app:ivma-sum}, and
exploits the following simple approximation.
\begin{fact}
\label{fact:prod-sum}
For any integer $l\ge 1$ and constants $\eps$ and $a_1,\dots,a_l$,
then as $\eps\to0$
\begin{equation} 
\prod_{i=1}^l (1 + \eps\, a_i ) = 1 + \eps \sum_{i=1}^l a_i + o(\eps).
\label{eq:prod-sum}
\end{equation}
\end{fact}

For multi-attributes $U^k$ and $V^k$ of
$X$ and $Y$, respectively, we use
\begin{subequations} 
\begin{align} 
\phi^{X|U_i}_{u_i}(x)
&\defeq \frac{P_{X|U_i}(x|u_i)-P_X(x)}{\eps\sqrt{P_X(x)}}
\label{eq:phiXUi-def} \\
\phi^{Y|V_i}_{v_i}(y)
&\defeq \frac{P_{Y|V_i}(y|v_i)-P_Y(y)}{\eps\sqrt{P_Y(y)}}
\label{eq:phiYVi-def} 
\end{align}
\label{eq:phiXUYVi-def}%
\end{subequations}
to denote the information vectors corresponding to $P_{X|U_i}(\cdot|u_i)$ and
$P_{Y|V_i}(\cdot|v_i)$, respectively.  

Note that for the extended Markov model \eqref{eq:full-markov},
orthogonality for multi-attribute $U^k$ of $X^m$ further implies that the
$U^k$ are conditionally independent given $X_j$, each
$j\in\{1,\dots,m\}$, and $X^m$ are conditionally independent given
$U_i$, each $i\in\{1,\dots,k\}$, i.e., 
\begin{subequations} 
\begin{align}
P_{U^k|X_j}(u^k|x_j) &= \prod_{i=1}^k
P_{U_i|X_j}(u_i|x_j) \label{eq:Uk-suborthog} \\
P_{X^m|U_i}(x^m|u_i) &= \prod_{j=1}^m
P_{X_j|U_i}(x_j|u_i), \label{eq:Xm-subprod}%
\end{align}
\end{subequations}
and the orthogonality for multi-attribute $V^k$ of $Y^m$ implies
$V^k$ are conditionally independent given $Y_j$, each
$j\in\{1,\dots,m\}$, and $Y^m$ are conditionally independent given
$V_i$, each $i\in\{1,\dots,k\}$, i.e., 
\begin{subequations} 
\begin{align}
P_{V^k|Y_j}(v^k|y_j) &= \prod_{i=1}^k
P_{V_i|Y_j}(v_i|y_j) \label{eq:Vk-suborthog} \\
P_{Y^m|V_i}(y^m|v_i) &= \prod_{j=1}^m
P_{Y_j|V_i}(y_j|v_i). \label{eq:Ym-subprod}%
\end{align}
\end{subequations}

\subsection{Induced Local Geometries of Attribute Variables}

We now express the relationships between
$U,V$ and $X,Y$ geometrically.  In particular, we
show that the local geometry of $P_{X|U}(\cdot|u)$ in the simplex
$\simpX$ induces a corresponding local geometry for $P_{Y|U}(\cdot|u)$
in the simplex $\simpY$ via the operator $\dtms$, and, likewise, the
local geometry of $P_{Y|V}(\cdot|v)$ in the simplex $\simpY$ induces a
corresponding local gometry for $P_{X|V}(\cdot|v)$ in the simplex
$\simpX$ via the adjoint.

Indeed, the Markov relation $U\markov X\markov Y$ implies
\begin{align*}
P_Y(y) &= \sum_{x \in \X} P_{Y|X}(y|x) \, P_X(x) \\
P_{Y|U}(y|u) &= \sum_{x \in \X} P_{Y|X}(y|x) \, P_{X|U}(x|u),
\end{align*}
from which we conclude that a neighborhood of $P_X$ in the simplex $\simpX$
maps to a neighborhood of $P_Y$ in the simplex $\simpY$.    In
particular, with $P_X$ and $P_Y$ as the reference distributions in
$\simpX$ and $\simpY$, respectively, the
information vectors 
\begin{subequations} 
\begin{align} 
\phi^{X|U}_u(x) &= \frac{P_{X|U}(x|u)-P_X(x)}{\eps \,
  \sqrt{P_X(x)}} \label{eq:phiXU-def} \\
\phi^{Y|U}_u(y) &= \frac{P_{Y|U}(y|u)-P_Y(y)}{\eps \, \sqrt{P_Y(y)}}
\label{eq:phiYU-def}%
\end{align}
\end{subequations}
associated with the distributions
$P_{X|U}(\cdot|u)$ and $P_{Y|U}(\cdot|u)$, respectively, satisfy
\begin{align}
\phi^{Y|U}_u(y) = \frac{1}{\sqrt{P_Y(y)}} \ \sum_{x\in\X} P_{Y|X}(y|x) 
\sqrt{P_X(x)} \, \phi^{X|U}_u(x). 
\label{eq:linmap-u}
\end{align}
With $\bphi^{X|U}_u$ and $\bphi^{Y|U}_u$ denoting the associated
column vectors, using \eqref{eq:dtm-alt} we can equivalently express
\eqref{eq:linmap-u} in the matrix form
\begin{equation} 
\bphi^{Y|U}_u = \dtm \, \bphi^{X|U}_u.
\label{eq:X|U-to-Y|U}
\end{equation}

\begin{figure}
\centering
\includegraphics[width=.9\columnwidth]{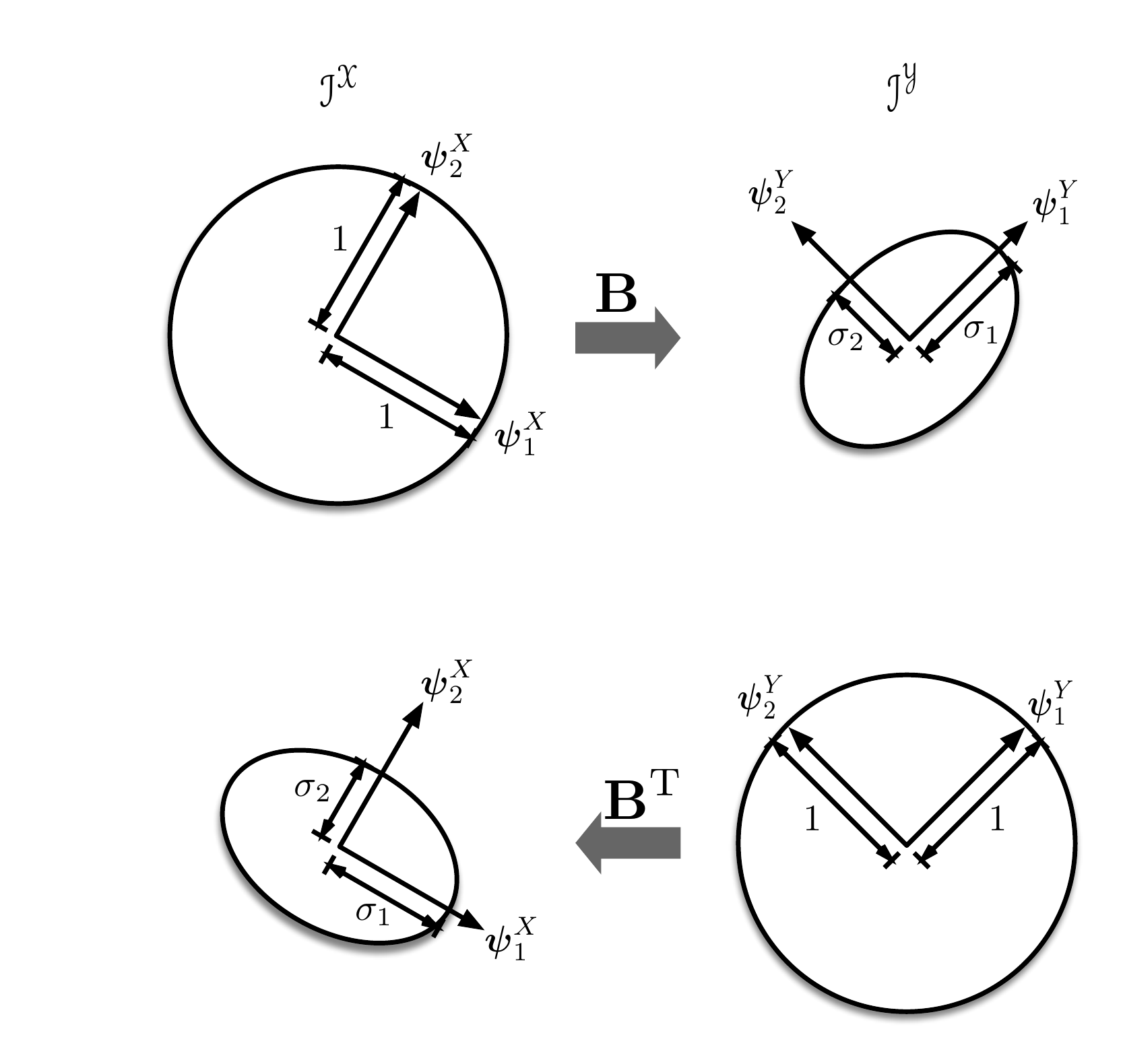}
\caption{The information geometry associated with the DTM $\dtm$.  For
  $i=1,\dots,K-1$, the unit information vector $\psi^X_i$ in
  $\ivspaceX$ maps via $\dtm$ to the shorter information vector
  $\sigma_i \psi^Y_i$ in $\ivspaceY$, and the unit information vector
  $\psi^Y_i$ in $\ivspaceY$ maps via $\dtm^\T$ to the shorter
  information vector $\sigma_i \psi^X_i$ in
  $\ivspaceX$. \label{fig:dtm}}
\end{figure}

Evidently, $\dtm$ maps a local divergence sphere in
$\simpX$ to a local divergence ellipsoid in $\simpY$ whose
principal axes correspond to the left singular vectors of $\dtm$, as
\figref{fig:dtm} depicts.  

Analogously,
the Markov relation $X\markov Y\markov V$ implies
\begin{align*}
P_X(x) &= \sum_{y\in\Y} P_{X|Y}(x|y) \, P_Y(y) \\
P_{X|V}(x|v) &= \sum_{y\in\Y} P_{X|Y}(x|y) \, P_{Y|V}(y|v),
\end{align*}
from which we conclude that a neighborhood of $P_Y$ in the simplex $\simpY$
maps to a neighborhood of $P_X$ in the simplex $\simpX$.    In
particular, with the same reference distributions, and using 
\eqref{eq:dtma-alt}, we obtain that the
information vectors 
\begin{subequations}
\begin{align} 
\phi^{Y|V}_v(y) = \frac{P_{Y|V}(y|v)-P_Y(y)}{\eps \, \sqrt{P_Y(y)}} \label{eq:phiYV-def}\\
\phi^{X|V}_v(x) = \frac{P_{X|V}(x|v)-P_X(x)}{\eps \, \sqrt{P_X(x)}}
\label{eq:phiXV-def}%
\end{align}
\end{subequations}
associated with the distributions $P_{Y|V}(\cdot|v)$ and
$P_{X|V}(\cdot|v)$ are related according to
\begin{equation} 
\bphi^{X|V}_v = \dtm^\T \bphi^{Y|V}_v,
\label{eq:Y|V-to-X|V}
\end{equation}
where $\bphi^{Y|V}_v$ and $\bphi^{X|V}_v$ denote the corresponding
column vectors.  In this case, as \figref{fig:dtm} also depicts,
$\dtm^\T$ maps a local divergence sphere in $\simpY$ to a local
divergence ellipsoid in $\simpX$ whose principal axes correspond to
the right singular vectors of $\dtm$.

Finally, we note that there are constraints on the distributions
governing the attributes $U$ and $V$.  In particular, we have
[cf.\ \eqref{eq:pzw-constr}]
\begin{subequations}
\begin{align} 
\sum_{u\in\U} P_U(u)\, \phi^{X|U}_u(x) = 0,\quad x\in\X
\label{eq:pxu-constr} \\
\sum_{v\in\V} P_V(v)\, \phi^{Y|V}_v(y) = 0,\quad y\in\Y.
\label{eq:pyv-constr}
\end{align}
\label{eq:pxyuv-constr}%
\end{subequations}

\subsection{Minimum Error Probability Universal Features}
\label{sec:uni-rie}

In this section, we model the universal feature selection problem as
the following game between a system designer and nature.  First,
nature chooses the distribution for latent attribute variables $(U,V)$ in
the Markov chain \eqref{eq:markov-uxyv} at random.  Next, before
nature reveals its chosen distributions, the system designer chooses
feature functions $f^k$ and $g^k$ knowing $P_{X,Y}$ and the
probability law according to which nature chooses its distribution.
Finally, after revealing its chosen distributions, the system designer
implements a test for determining $(U,V)$ with minimum error
probability from statistics formed from its chosen features applied to
samples of $(X,Y)$.  The details are as follows.

Let $\C^\X_\eps(P_X)$ and $\C^\Y_\eps(P_Y)$ denote configurations for
attributes $U$ and $V$, respectively, in the sense of
\defref{def:config}, i.e., 
\begin{subequations} 
\begin{align} 
\C^\X_\eps(P_X) 
&= \biggl\{\,
\U,\, \{P_U(u),\, u\in\U\},\, \{P_{X|U}(\cdot|u),\, u\in\U \} \colon \notag\\
&\quad\qquad P_{X|U}(\cdot|u)\in\nbhd_\eps^\X(P_X),\ u\in\U, \notag\\
&\,\ \quad\qquad \sum_{u\in\U} P_U(u)\,P_{X|U}(x|u)  =
P_X(x),\ x\in\X \biggr\} \label{eq:Cx-def} \\
&= \biggl\{\,
\U,\, \{P_U(u),\, u\in\U\},\, \{\bphi^{X|U}_u,\, u\in\U \} \colon \notag\\
&\qquad\qquad \bphi^{X|U}_u\in\ivspaceX,\ u\in\U, \notag\\
&\qquad\qquad\qquad \sum_{u\in\U} P_U(u)\, \phi^{X|U}_u(x) = 0,\ x\in\X
\biggr\} \notag\\ 
\C^\Y_\eps(P_Y) 
&= \biggl\{\,
\V,\, \{P_V(v),\, v\in\V\},\, \{P_{Y|V}(\cdot|v),\, v\in\V\} \colon
\notag \\
&\quad\qquad P_{Y|V}(\cdot|v)\in\nbhd_\eps^\Y(P_Y),\ v\in\V, \notag\\
&\,\ \quad\qquad \sum_{v\in\V} P_V(v)\,P_{Y|V}(y|v)  = P_Y(y),\ y\in\Y
\biggr\}, \label{eq:Cy-def} \\
&= \biggl\{\, \V,\, \{P_V(v),\, v\in\V\},\, \{\bphi^{Y|V}_v,\,
v\in\V\} \colon \notag \\
&\qquad\qquad \bphi^{Y|V}_v\in\ivspaceY,\ v\in\V, \notag\\
&\qquad\qquad\qquad \sum_{v\in\V} P_V(v)\, \phi^{Y|V}_v(y) = 0,\ y\in\Y
\biggr\}.
\notag 
\end{align}
\label{eq:CxCy-def}%
\end{subequations}

In choosing a configuration pair, nature uses a probability law in
which the ensemble for each attribute is characterized by rotational
invariance, our definition for which relies on the following concept
of spherical symmetry \cite{apd77,mac81}.
\begin{definition}[Spherical Symmetry]
\label{def:sphere-sym}
A  $k_1\times k_2$ random matrix $\bZ$ is
spherically symmetric if 
for any orthogonal $k_1\times k_1$ and $k_2\times
k_2$ matrices $\bQ_1$ and $\bQ_2$, respectively, we have 
\begin{equation}
\bZ\eqd \bQ_1^\T\bZ\,\bQ_2,
\label{eq:sphere-sym}
\end{equation}
where $\eqd$ denotes equality in distribution.
\end{definition}
Moreover, the following consequence of spherical symmetry is useful in
our analysis; a proof is provided in \appref{app:rie}.\footnote{The
  proof introduces the notation $\bfe_i$ for the (elementary) vector
  whose $i$\/th element is 1, and all other elements are 0, which we
  will more generally find convenient in our analysis.}
\begin{lemma}
\label{lem:rie}
Let $\bZ$ be a $k_1\times k_2$ spherically symmetric 
random matrix.  Then if $\bA_1$ and $\bA_2$ are any fixed
matrices of compatible dimensions, then
\begin{equation}
\E{\bfrob{\bA_1^\T\bZ\bA_2}^2 } = \frac{1}{k_1 k_2}\,
\bfrob{\bA_1}^2\, \bfrob{\bA_2}^2\, \E{\bfrob{\bZ}^2}. 
\label{eq:rie-prop}
\end{equation}
\end{lemma}

Our ensemble of interest is defined in terms of spherical symmetry as
follows.
\begin{definition}[Rotation Invariant Ensemble]
\label{def:rie}
Given $\eps>0$, a rotationally invariant ensemble (RIE) for an
$\eps$-attribute $W$ of a variable $Z$ is a collection of attribute
configurations of the form \eqref{eq:Cz-phi}, together with a
probability measure over the collection such that $\bPhi^{Z|W}$, the
$\cardZ\times\cardW$ matrix whose columns are $\bphi^{Z|W}_w$,
$w\in\W$,\footnote{With some abuse of terminology and
  notation, we will say the $w$\/th column of $\bPhi^{Z|W}$ is
  $\bPhi^{Z|W}\bfe_w=\bphi^{Z|W}_w$, to avoid cumbersome
  exposition.  More precisely, given an (arbitrary) bijective function
  $\ordinal_\W\colon\W\mapsto\{1,\dots,\cardW\}$ with inverse
  $\ordinal_\W^{-1}$,
\begin{equation}
\bPhi^{Z|W} \defeq  \begin{bmatrix} \bphi^{Z|W}_{\ordinal_\W^{-1}(1)} & \cdots &
  \bphi^{Z|W}_{\ordinal_\W^{-1}(\cardW)} \end{bmatrix},
\end{equation}
i.e., the $\ordinal_\W(w)$\/th column of $\bPhi^{Z|W}$ is $\bPhi^{Z|W}\bfe_{\ordinal_\W(w)}=\bphi^{Z|W}_w$.
\label{fn:abuse} }
 is
spherically symmetric.
\end{definition}

In what follows, we denote the error probability in decisions for $U$
and $V$ based on $S^k$, respectively, via
\begin{subequations} 
\begin{align}
\pe^{U|S}\bigl(\C^\X_\eps(P_X),f^k\bigr)\quad\text{and}\quad
\pe^{V|S}\bigl(\C^\Y_\eps(P_Y),f^k\bigr),
\end{align}
and those for the decisions based on $T^k$ via, respectively,
\begin{align}
\pe^{U|T}\bigl(\C^\X_\eps(P_X),g^k\bigr)\quad\text{and}\quad
\pe^{V|T}\bigl(\C^\Y_\eps(P_Y),g^k\bigr),
\end{align}
\label{eq:pe-def}%
\end{subequations}
where $S^k$ and $T^k$ are as defined in \eqref{eq:stmk-def}
for feature choices $f^k\colon \X\mapsto \reals^k$ and $g^k\colon
\Y\mapsto \reals^k$.  In turn, we define the error exponents
\begin{subequations} 
\begin{align} 
\Eb^{U|S}\bigl(f^k\bigr)
&\defeq \lim_{m\to\infty} -\frac{
\bEd{\mathrm{RIE}}{\log\pe^{U|S}\bigl(\C^\X_\eps(P_X),f^k\bigr)}}{m}\\
\Eb^{V|S}\bigl(f^k\bigr)
&\defeq \lim_{m\to\infty} -\frac{
\bEd{\mathrm{RIE}}{\log\pe^{V|S}\bigl(\C^\Y_\eps(P_Y),f^k\bigr)}}{m}\\
\Eb^{U|T}\bigl(g^k\bigr)
&\defeq \lim_{m\to\infty} -\frac{
\bEd{\mathrm{RIE}}{\log\pe^{U|T}\bigl(\C^\X_\eps(P_X),g^k\bigr)}}{m}\\
\Eb^{V|T}\bigl(g^k\bigr)
&\defeq \lim_{m\to\infty} -\frac{
\bEd{\mathrm{RIE}}{\log\pe^{V|T}\bigl(\C^\Y_\eps(P_Y),g^k\bigr)}}{m},
\end{align}
\label{eq:Eb-def}%
\end{subequations}
where $\bEd{\mathrm{RIE}}{\cdot}$ denotes expectation with respect to
the RIEs for $\C^\X_\eps(P_X)$ and $\C^\Y_\eps(P_Y)$.

Our main result is the following proposition, which identifies the
features the system designer should choose, and the exponent of the
resulting error probability.  A proof is provided in
\appref{app:mpe-rie}. 
\begin{proposition}
\label{prop:mpe-rie}
Given $P_{X,Y}\in\simpXY$ and attributes $U$ and $V$ of $X$ and $Y$,
respectively, each drawn from a RIE for some $\eps>0$, then for any
dimension $k\in\{1,\dots,K-1\}$,\footnote{For arbitrary sequences $a^l$
  and $b^l$ of arbitrary length $l$, we use 
  $a^l\le b^l$ to denote that $a_i\le b_i$ for
  $i\in\{1,\dots,l\}$.} 
\begin{align} 
&\Bigl(
  \Eb^{U|S}(f^k),\Eb^{V|S}(f^k),\Eb^{U|T}(g^k),\Eb^{V|T}(g^k)\Bigr) \notag\\
&\qquad \le \Bigl( 
\bar{E}_0^{X|U} \, \eps^2\,k,\ 
\bar{E}_0^{Y|V} \, \eps^2 \sum_{i=1}^k \sigma_i^2,\notag\\
&\qquad\qquad\qquad\bar{E}_0^{X|U} \, \eps^2 \sum_{i=1}^k \sigma_i^2,\  
\bar{E}_0^{Y|V} \, \eps^2\,k \Bigr) + o(\eps^2)
\label{eq:pe-rie-pareto}
\end{align}
where $\bar{E}^{X|U}_0$ and $\bar{E}^{Y|V}_0$ are positive constants
that do not depend on $\eps$, $k$, or $P_{X,Y}$.  Moreover, all the
inequalities in \eqref{eq:pe-rie-pareto} simultaneously hold with
equality for the choices $\bigl(f^k_*,g^k_*\bigr)$ as defined in
\eqref{eq:fgk*-def}, i.e., the associated multi-objective maximization has a
unique\footnote{Note that while the optimized multi-objective function
  is unique, the features that achieve them need not be, as is the
  case when there are repeated singular values.}  Pareto-optimal
solution.  
\end{proposition}

We emphasize that the result does not depend on any details of
probability law governing nature's choice other than the RIE property.
Hence, $f_*^k,g_*^k$ are optimum no matter what priors we might place
over various parameters of the configurations
$\C^\X_\eps(P_X),\C^\Y_\eps(P_Y)$ generating the RIE.  In this sense,
their optimality is fairly strong.

\subsection{Universal Features via a Cooperative Game} 
\label{sec:uni-coop}

In this section, we show how the same universal features arise as the
solution to a \emph{cooperative game}, which further reveals the
latent variable configurations for which these features are
effectively sufficient statistics.  In this game, for a given
$k\in\{1,\dots,K-1\}$, nature chooses
configurations $\C^{\X,k}_\eps(P_X)$ and $\C^{\Y,k}_\eps(P_Y)$ of
multi-attribute 
variable collections $U=U^k$ and $V=V^k$ in \eqref{eq:full-markov},
and the system designer chooses the features $f^k$ and $g^k$.  Their
shared goal is to identify variables $(U^k,V^k)$ that are, in an
appropriate sense, most detectable from the statistics $(s^k,t^k)$ as
defined in  \eqref{eq:stmk-def} in terms of these features.

The specific shared goal of nature and the system designer is to
maximize the probability that the least detectable of $U_1,\dots,U_i$
and the least detectable of $V_1,\dots,V_i$, for $i=1,\dots, k$, are
correctly detected, as $m\to\infty$.  

For the analysis of this game, the following min-max characterization
of singular values is useful.  \cite[Theorem~4.2.6]{hj12}.
\begin{lemma}[Courant-Fischer]
\label{lem:maxmin}
Let $\bA$ be a $k_1\times k_2$ matrix with singular values
$\sigma_1(\bA)\ge \cdots \ge \sigma_k(\bA)$ where $k=\min\{k_1,k_2\}$. Then
for every $i\in\{1,\dots,k\}$, 
\begin{equation} 
\sigma_i(\bA) = \max_{\{\cS\subset\reals^k\colon \dim(\cS)=i\}}\ 
\min_{\{\bphi\in\cS\colon \vphantom{\reals^k}
  \norm{\bphi}=1\}}  \ \bnorm{\bA\bphi},
\end{equation}
where $\cS$ denotes a subspace, and the maximum is achieved by
$\bphi=\bpsi_i^\mathrm{R}$, a right singular vector of $\bA$
corresponding to $\sigma_i(\bA)$.
\end{lemma}
In addition, in our development the following well-known inequality,
which follows from the fact that $\spectral{\cdot}$ is
the matrix norm induced by the (Euclidean) vector norm $\norm{\cdot}$,
is convenient.
\begin{fact}
\label{fact:frob-submult}
For any compatible matrices $\bA_1$ and $\bA_2$, we have
$\frob{\bA_1\bA_2} \le \spectral{\bA_1}\,\frob{\bA_2}$.
\end{fact}

In the sequel, we denote the error probabilities in decisions based
on $S^k$ about each of constituent elements of $U^k$ and $V^k$,
respectively, via
\begin{subequations} 
\begin{align}
\pe^{U_i|S}\bigl(\C^{\X,k}_\eps(P_X),f^k\bigr)\quad\text{and}\quad
\pe^{V_i|S}\bigl(\C^{\Y,k}_\eps(P_Y),f^k\bigr),
\end{align}
and those based based on $T^k$ via, respectively, 
\begin{align}
\pe^{U_i|T}\bigl(\C^{\X,k}_\eps(P_X),g^k\bigr)\quad\text{and}\quad
\pe^{V_i|T}\bigl(\C^{\Y,k}_\eps(P_Y),g^k\bigr),
\end{align}
\label{eq:pek-def}%
\end{subequations}
for $i\in\{1,\dots,k\}$, where $S^k$ and $T^k$ are as defined in
\eqref{eq:stmk-def} for feature choices $f^k\colon \X\mapsto \reals^k$
and $g^k\colon \Y\mapsto \reals^k$.  In turn, we define the error
exponents
\begin{subequations} 
\begin{align} 
E^{U_i|S}\bigl(\C^{\X,k}_\eps(P_X),f^k\bigr)
&\defeq \lim_{m\to\infty} \frac{-\log
\pe^{U_i|S}\bigl(\C^{\X,k}_\eps(P_X),f^k\bigr)}{m}\\
E^{V_i|S}\bigl(\C^{\Y,k}_\eps(P_Y),f^k\bigr)
&\defeq \lim_{m\to\infty} \frac{-\log
\pe^{V_i|S}\bigl(\C^{\Y,k}_\eps(P_Y),f^k\bigr)}{m}\\
E^{U_i|T}\bigl(\C^{\X,k}_\eps(P_X),g^k\bigr)
&\defeq \lim_{m\to\infty} \frac{-\log
\pe^{U_i|T}\bigl(\C^{\X,k}_\eps(P_X),g^k\bigr)}{m}\\
E^{V_i|T}\bigl(\C^{\Y,k}_\eps(P_Y),g^k\bigr)
&\defeq \lim_{m\to\infty} \frac{-\log
\pe^{V_i|T}\bigl(\C^{\Y,k}_\eps(P_Y),g^k\bigr)}{m}.
\end{align}
\label{eq:Ek-def}%
\end{subequations}

Our main result is as follows.  A proof is provided in \appref{app:mpe-coop}.
\begin{proposition}
\label{prop:mpe-coop}
Given $k\in\{1,\dots,K-1\}$ and $P_{X,Y}\in\simpXY$, let
$\C^{\X,k}_\eps(P_X)$ and $\C^{\Y,k}_\eps(P_Y)$ denote configurations
of $\eps$-multi-attribute variables $U^k$ and $V^k$ of $X$ and $Y$,
respectively, for some $\eps>0$.  Then
\begin{align} 
\Bigl(
&\min_{j\le i} E^{U_j|S}\bigl(\C^{\X,k}_\eps(P_X),f^k\bigr),\ i\in\{1,\dots,k\}, \notag\\
&\quad \min_{j\le i} E^{V_j|S}\bigl(\C^{\Y,k}_\eps(P_Y),f^k\bigr),\ i\in\{1,\dots,k\}, \notag\\
&\quad \quad \min_{j\le i} E^{U_j|T}\bigl(\C^{\X,k}_\eps(P_X),g^k\bigr),\ i\in\{1,\dots,k\}, \notag\\
&\quad \quad \quad \min_{j\le i} E^{V_j|T}\bigl(\C^{\Y,k}_\eps(P_Y),g^k\bigr),\ i\in\{1,\dots,k\}
\Bigr) \notag\\
&\qquad\qquad\le \biggl( \frac{\eps^2}{2},\ i\in\{1,\dots,k\},\notag\\
&\quad\qquad\qquad\qquad\frac{\eps^2}{2} \sigma_i^2,\ i\in\{1,\dots,k\},\notag\\ 
&\qquad\qquad\qquad\qquad\frac{\eps^2}{2} \sigma_i^2,\ i\in\{1,\dots,k\},\notag\\
&\quad\qquad\qquad\qquad\qquad\frac{\eps^2}{2},\ i\in\{1,\dots,k\}
\biggr)  + o(\eps^2).
\label{eq:pe-coop-pareto}
\end{align}
Moreover, the inequalities in \eqref{eq:pe-coop-pareto} all hold with
equality when $\bigl(f^k,g^k\bigr)$ are chosen to be
$\bigl(f^k_*,g^k_*\bigr)$ as defined in \eqref{eq:fgk*-def}, and 
$\C^{\X,k}_\eps(P_X)$ and $\C^{\Y,k}_\eps(P_Y)$ as chosen to be, respectively,
\begin{subequations} 
\begin{align} 
&\C^{\X,k}_{\eps,*}(P_X) \notag\\
&\ \defeq \Bigl\{ \U_i=\{+1,-1\},\ \bigl\{
P_{U_i}(u_i)=1/2,\ u_i\in\U_i \bigr\},\notag\\ 
&\qquad\qquad \bigl\{P_{X|U_i}(x|u_i) =
P_X(x)\bigl( 1+ \eps\, u_i \, f_i^*(x) \bigr),\notag\\
&\qquad\qquad\qquad\qquad  u_i\in\U_i,\ x\in\X\bigr\},\ i=1,\dots,k\Bigr\}
\label{eq:CXk-opt}
\end{align}
and
\begin{align} 
&\C^{\Y,k}_{\eps,*}(P_Y) \notag\\
&\ \defeq \Bigl\{ \V_i=\{+1,-1\},\quad \bigl\{
P_{V_i}(v_i)=1/2,\ v_i\in\V_i \bigr\},\notag\\ 
&\qquad\qquad \bigl\{P_{Y|V_i}(y|v_i) =
P_Y(y)\bigl( 1+ \eps\, v_i \, g_i^*(y) \bigr),\notag\\
&\qquad\qquad\qquad\qquad  v_i\in\V_i,\ y\in\Y\bigr\},\ i=1,\dots,k\Bigr\},
\label{eq:CYk-opt}
\end{align}
\label{eq:CXYk-opt}%
\end{subequations}
i.e.,  the associated multi-objective 
maximization has a unique Pareto-optimal solution.
\end{proposition}

In addition, from the Markov structure \eqref{eq:markov-uxyv} and the
modal structure \eqref{eq:fgi-condexp}, we immediately obtain the following
corollary.
\begin{corollary}
\label{corol:mpe-coop}
The optimizing multi-attribute variables $U^k$ and $V^k$ in
\propref{prop:mpe-coop} have the property that 
\begin{subequations} 
\begin{align}
P_{X|V_i}(x|v_i) &=
P_X(x)\bigl( 1+ \eps\, v_i \, \sigma_i\, f_i^*(x) \bigr)
\label{eq:PXVi-opt}\\
P_{Y|U_i}(y|u_i) &= 
P_Y(y)\bigl( 1+ \eps\, u_i\, \sigma_i\, g_i^*(y)
\bigr). \label{eq:PYUi-opt}
\end{align}%
\end{subequations}
for $i=1,\dots,k$.
\end{corollary}

Given data $(x^m,y^m)$ from the extended Markov model 
\eqref{eq:full-markov}, it further follows that 
$(S^k_*,T^k_*)$ defined via \eqref{eq:stmk*-def}
is, as $\eps\to0$, a sufficient statistic for inferences about the
optimizing multi-attributes $(U^k,V^k)$, i.e., we have the Markov
chains
\begin{equation}
(U^k,V^k) \markov (S_*^k,T_*^k) \markov (X^m,Y^m),\quad \eps\to0
\end{equation}
and
\begin{equation}
U^k \markov S_*^k \markov T_*^k \markov V^k,\quad\eps\to0.
\end{equation}
In particular, we have the
following result, a proof of which is provided in
\appref{app:suffstat}.  
\begin{corollary}
\label{corol:suffstat}
In the solution to the optimization in \propref{prop:mpe-coop} for the
extended model \eqref{eq:full-markov},
\begin{align} 
&P_{U^k,V^k|X^m,Y^m}(u^k,v^k|x^m,y^m) \notag\\
&\qquad\qquad= \frac1{4^k} \left( 1 + \eps\, m \sum_{i=1}^k
  \bigl(u_i\, s_i^* + v_i\, t_i^*\bigr) \right) + o(\eps),
\label{eq:Puv|xy-form}
\end{align}
with, consistent with \eqref{eq:stmk*-def},
\begin{equation}
s_i^* = \frac1m \sum_{j=1}^m f_i^*(x_j)\quad\text{and}\quad
t_i^* = \frac1m \sum_{j=1}^m g_i^*(y_j).
\label{eq:sti*-def}
\end{equation}
Moreover, 
\begin{subequations} 
\begin{align} 
P_{U^k|S_*^k,T_*^k,V^k}(u^k|s_*^k,t_*^k,v^k) &= 
\frac1{2^k}\!
\left( 1 + \eps\, m \sum_{i=1}^k u_i\, s_i^* \right)\! +\!
o(\eps) \label{eq:U|STV}\\ 
P_{V^k|S_*^k,T_*^k,U^k}(v^k|s_*^k,t_*^k,u^k) &= 
\frac1{2^k}\!
\left( 1 + \eps\, m \sum_{i=1}^k u_i\, t_i^* \right)\! +\! o(\eps). \label{eq:V|STU}
\end{align}
\label{eq:USTV-chain}%
\end{subequations}
\end{corollary}

In essence, \corolref{corol:suffstat} shows that in making inferences
about the (optimizing) attributes $U^k$ and $V^k$ from
high-cardinality data $(X^m,Y^m)$, it is sufficient to extract a
low-dimensional real-valued sufficient statistic $(S^k,T^k)$.   Morover,
it is clear that it is sufficient to extract a statistic of dimension
$k$ corresponding to the number of ``significant'' singular values of
$\dtmt$.  As importantly, we emphasize that the sufficient statistic
pair $(S_*^k,T_*^k)$ involves \emph{separate} processing of $X^m$ and
$Y^m$, and as such implies significantly lower computational
complexity than would generally be required for statistics that
require joint processing of $(X^m,Y^m)$.

In addition, as suggested by
\corolref{corol:suffstat} and revealed in its proof,
\begin{subequations} 
\begin{multline}
P_{U^k,V^k|X^m,Y^m}(u^k,v^k|x^m,y^m) \\
= \prod_{i=1}^k 
P_{U_i,V_i|X^m,Y^m}(u_i,v_i|x^m,y^m)
\label{eq:PUV|XYk-fact}
\end{multline}
with
\begin{align} 
&P_{U_i,V_i|X^m,Y^m}(u_i,v_i|x^m,y^m) \notag\\
&\qquad\qquad= \frac14\, \bigl( 1 + \eps\, m 
  (u_i\, s_i^* + v_i\, t_i^*) \bigr) + o(\eps),
\label{eq:PUVi|XY}
\end{align}
\end{subequations}
from which we see that to achieve the optimum exponents given by
\eqref{eq:pe-coop-pareto}, it is sufficient for decisions about the
attribute pair $(U_i,V_i)$ to be made based on the statistic
$(S_i^*,T_i^*)$ alone.  Moreover,
\eqref{eq:PUV|XYk-fact} reveals that although not imposed as a
constraint, the optimizing configuration is such that the $U^k$ are
conditionally independent of $Y^m$ (and the $V^k$ are conditionally
independent of $X^m$).

\subsection{Universal Features via an Information Bottleneck}
\label{sec:uni-ib}

In this section, we show that the same configurations
$\C^{\X,k}_{\eps,*}(P_X)$ and $\C^{\Y,k}_{\eps,*}(P_Y)$ that are
optimum in the cooperative game of \secref{sec:uni-coop} are the
solution to a natural mutual information maximization problem, which
provides a third viewpoint from which to interpret $f_*^k$ and $g_*^k$
as universal features.

Our main result is as follows.    A proof is provided in
\appref{app:ib-double}. 
\begin{proposition}
\label{prop:ib-double}
Given $\eps>0$, $P_{X,Y}\in\simpXY$, and $\eps$-multi-attribute
variables $U=U^k$ and $V=V^k$ for some $k$ in the Markov chain
\eqref{eq:markov-uxyv}, then
\begin{equation}
I(U^k;V^k) \le \frac{\eps^4}{2} \, \sum_{i=1}^k \sigma_i^2 + o(\eps^4).
\label{eq:ib-bnd}
\end{equation}
Moreover, the inequality in \eqref{eq:ib-bnd} holds with equality when
the configurations $\C^{\X,k}_{\eps,*}(P_X)$ and $\C^{\Y,k}_{\eps,*}(P_Y)$ of
$U^k$ and $V^k$, respectively, are given by \eqref{eq:CXYk-opt}, in
which case 
\begin{align} 
&P_{U^k,V^k}(u^k,v^k) \notag\\
&\ = \frac1{4^k}\! \left(1+ \eps^2\, \sum_{i=1}^k
\sigma_i \,u_i \,v_i\right) + o(\eps^2),
\quad u^k,v^k\in\{+1,-1\}^k.
\label{eq:PUV-k}
\end{align}
\end{proposition}

We interpret the optimizing multi-attribute pair $(U^k,V^k)$ in
\propref{prop:ib-double}, with joint distribution \eqref{eq:PUV-k}, as
expressing the dominant components of the dependency in the
relationship between $X$ and $Y$ as determined by $P_{X,Y}$.  This is
reflected in the fact that
\begin{equation*}
P_{U^k,V^k}(u^k,v^k) = \prod_{i=1}^k P_{U_i,V_i}(u_i,v_i)
\end{equation*}
with, for $i,j=1,\dots,k$,
\begin{equation}
P_{U_i,V_j}(u_i,v_j) = \frac14 \bigl(1+ \eps^2\, \sigma_i
\,u_i\,v_j\,\kron_{i=j}\bigr) + o(\eps^2),
\label{eq:PUiVj}
\end{equation}
for which [cf.\ \eqref{eq:I-modal}]
\begin{equation}
I(U_i;V_j) = \frac{\eps^4}{2} \, \sigma_i^2 \, \kron_{i=j} + o(\eps^4).
\end{equation}

An immediate consequence of \propref{prop:ib-double} is that for
observations $X^m,Y^m$ from the model \eqref{eq:full-markov}, we have
that $(S^k_*,T^k_*)$ is a sufficient statistic for inferences about
$U^k,V^k$, i.e., \corolref{corol:suffstat} applies.  This sufficiency
can be equivalently expressed in the form
\begin{equation}
\lim_{\eps\to0}\frac{I(U^k,V^k;X^m,Y^m)}{I(U^k,V^k;S^k_*,T^k_*)} = 1.
\end{equation}

Note too that \eqref{eq:PUV-k} provides a higher-order
characterization of $P_{U^k,V^k}$ than that derived from
\eqref{eq:Puv|xy-form}.  Indeed, from the latter (setting $m=1$ for
convenience) we obtain only
\begin{align*}
&P_{U^k,V^k}(u^k,v^k)\notag\\
&\ = \sum_{x\in\X,y\in\Y} P_{X,Y}(x,y)\, P_{U^k,V^k|X,Y}(u^k,v^k,x,y) \\
&\ = \frac1{4^k} \sum_{x\in\X,y\in\Y} P_{X,Y}(x,y)\notag\\
&\qquad\qquad{} \cdot \left( 1 + \eps\, m \sum_{i=1}^k
  \bigl(u_i\, f_i^*(x) + v_i\, g_i^*(y)\bigr) \right) + o(\eps) \\
&\ = \frac1{4^k} + o(\eps),\quad 
\end{align*}

Additionally, it follows from the discussion in \secref{sec:eps-dep}
that when $\U_i=\V_i=\{+1,-1\}$ and $P_{U_i}=P_{V_i}\equiv1/2$, then
the $\eps$-multi-attribute constraints
$P_{X|U_i}(\cdot|u_i)\in\nbhd^\X_\eps(P_X)$ and
$P_{Y|V_i}(\cdot|u_i)\in\nbhd^\Y_\eps(P_Y)$ are equivalent to
$I(X,U_i)\le \eps^2/2$ and $I(Y;V_i)\le \eps^2/2$ as $\eps\to0$, for
$i=1,\dots,k$.  As a result, \propref{prop:ib-double} can be
equivalently expressed in the form of a solution to a information
bottleneck problem \cite{tpb99} in the weak dependence
regime.\footnote{For an early application of the use of information
  bottleneck techniques in learning, see \cite{st00,ts00}.}  In
particular, we have the following immediate corollary.\footnote{As the
  proof reveals, sufficiently weak pairwise dependence will suffice
  for condition 3, but we impose mutual independence for convenience.
  Moreover, while condition 4 alone implies a degree of weak marginal
  dependence, it is insufficient.  Finally, conditions 3 and 4
  together can be viewed, in some sense, as ``entropy maximizing''
  conditions.  }
\begin{corollary}
\label{corol:ib-double}
Given $\eps>0$, $P_{X,Y}\in\simpXY$ and variables
$U=U^k$ and $V=V^k$ in the Markov chain \eqref{eq:markov-uxyv}, then
\begin{equation*}
\max_{U^k,V^k} I(U^k;V^k) =
\frac{\eps^4}{2} \, \sum_{i=1}^k \sigma_i^2 + o(\eps^4),
\end{equation*}
where the maximization is over all configurations of $(U^k,V^k)$ such
that: constituent variables $(U_i,V_i)$, $i=1,\dots,k$ satisfy: 1)
$\max\bigl\{I(U_i;X), I(V_i;Y)\bigr\}\le \eps^2/2$; 2) they are binary
and equiprobable, i.e., $U_i,V_i\in\{+1,-1\}$ and
$P_{U_i}=P_{V_i}\equiv1/2$; 3) $U^k$ and $V^k$ are each collections of
independent variables; and 4) the $U^k$ and $V^k$ are each collections
of conditionally independent variables given $X$ and $Y$,
respectively.  Moreover, the maximum is achieved by the
configurations \eqref{eq:CXYk-opt}.
\end{corollary}

As an aside, different but related one-sided information bottleneck
problems can also be analyzed within the same framework of analysis.
For example, the following result is proved in
\appref{app:ib-single}.\footnote{Other variations of this result
  correspond to avoiding the binary, equi\-probable and 
  mutual information constraits and instead using
  $D(P_{X|U}(\cdot|u)\|P_X)\le\eps^2/2$ for all $u\in\U$.
  Alternatively, by the equidistant property of capacity-achieving
  output distributions, we can equivalently express this divergence
  constraint as $\max_{P_U} I(X;U)\le\eps^2/2$. }
\begin{proposition}
\label{prop:ib-single}
Given $\eps>0$, $P_{X,Y}\in\simpXY$ and variables
$U=U^k$ and $V=V^k$ in the Markov chain \eqref{eq:markov-uxyv}, then
\begin{equation*}
\max_{V^k} I(V^k;X) =\max_{U^k} I(U^k;Y) =
\frac{\eps^2}{2} \, \sum_{i=1}^k \sigma_i^2 + o(\eps^2),
\end{equation*}
where the maximization is over all configurations of $(U^k,V^k)$ such
that: constituent variables $U_i,V_i$, $i=1,\dots,k$ satisfy: 1)
$\max\bigl\{I(U_i;X),I(V_i;Y)\bigr\}\le \eps^2/2$; 2) they are binary
and equiprobable, i.e., $U_i,V_i\in\{+1,-1\}$ and
$P_{U_i}=P_{V_i}\equiv1/2$; 3) $U^k$ and $V^k$ are each collections of
independent variables; and 4) the $U^k$ and $V^k$ are collections of
conditionally independent variables given $X$ and $Y$, respectively.
Moreover, the maximum is achieved by the configurations
\eqref{eq:CXYk-opt}.
\end{proposition}

As a further comment, work on aspects the more general information
bottleneck problem and the associated role of hyper-contractivity
analysis in its treatment includes
\cite{AhlswedeG76,Biswal11,ka12,kc13,agkn13a,agkn13,agkn14,mz15,mz18,pw17}. 
From this perpective, the development of this section reveals that a
number of the subtleties that complicate such analysis and lead to
anomalous behavior are avoided by the restriction to local
variables.  

\subsection{Universal Features via Common Information} 
\label{sec:common}

As we develop in this section, there is a key relationship between the
optimizing multi-attributes $(U^k,V^k)$ in \secref{sec:uni-ib} (and
\secref{sec:uni-coop}), and the common information associated with the
pair $(X,Y)$ characterized by a given joint distribution $P_{X,Y}$.
Recall that Wyner\footnote{Although we do not make use of them in
  our treatment, related but different notions of common information
  include those due to G\'acs and K\"orner \cite{gk73} and
  Witsenhausen \cite{hsw75}.}  \cite{adw75} defines common information
via an auxiliary variable $W$ as
\begin{equation} 
C(X,Y) = \min_{P_{W|X,Y}\colon X\markov W\markov Y}
I(W;X,Y).
\label{eq:wci-def}
\end{equation}

The results we obtain are for the case where $X$ and $Y$ are
extra-weakly dependent; specifically, for some $\eps>0$,
\begin{equation}
\nuclear{\dtmt}\le\eps,
\label{eq:sub-eps-prop}
\end{equation}
where $\nuclear{\cdot}$ denotes the nuclear norm of its
argument.\footnote{Specifically, the nuclear norm of an arbitrary
  matrix $\bA$ is
\begin{equation*}
\nuclear{\bA} \defeq \tr\biggl(\sqrt{\bA^\T\bA}\,\biggr) =
\sum_i \sigma_i(\bA),
\end{equation*}
where $\sigma_i(\bA)$ denotes the $i$\/th singular value of $\bA$.
Note that the nuclear norm is the Ky Fan $k$-norm with $k=\rank(\bA)$,
i.e., $\nuclear{\bA}=\norm{\bA}_{(\rank(\bA))}$.}
We refer to $X$ and $Y$ as sub-$\eps$ dependent in this
case, since by standard norm inequalities \cite{gvl12}
\begin{equation}
\frob{\dtmt} \le \nuclear{\dtmt},
\label{eq:dtmt-frob-bnd}
\end{equation}
whence $P_{X,Y}\in\nbhd_\eps^{\X\times\Y}(P_XP_Y)$, i.e., sub-$\eps$
dependence implies $\eps$-dependence.\footnote{Of course, since
  $\nuclear{\bA}\le \sqrt{\rank(\bA)} \frob{\bA}$ for any $\bA$, we also have
  that $\eps$-dependence implies sub-$(K\eps)$ dependence.  In turn,
  $O(\eps)$-dependence and sub-$O(\eps)$ dependence are equivalent.}
 We further use
$\nbhdk_\eps^{\X\times\Y}(P_XP_Y)$ to denote the joint
distributions in $\simpXY$ with sub-$\eps$ dependence given marginals
$P_X$ and $P_Y$, via which \eqref{eq:dtmt-frob-bnd} expresses
\begin{equation}
\nbhdk_\eps^{\X\times\Y}(P_XP_Y) \subset
\nbhd_\eps^{\X\times\Y}(P_XP_Y).
\label{eq:sub-eps-nbhd}
\end{equation}

Under sub-$\eps$ dependence, we define the following restricted common
information. 
\begin{definition}[$\eps$-Common Information]
\label{def:eci}
Given $P_{X,Y}\in\nbhdk_\eps^{\X\times\Y}(P_XP_Y)$ for $\eps>0$, the
$\eps$-common information is
\begin{equation} 
C_\eps(X,Y) = \min_{P_{W|X,Y}\in\cP_\eps} I(W;X,Y),
\label{eq:eps-ci-def}
\end{equation}
where
\begin{align}
\cP_\eps &\defeq 
\Bigl\{\, P_{W|X,Y}\in\cP^\W,\ \text{some $\W$} \colon X\markov W\markov Y \ \text{and} \notag\\
&\qquad
P_{X|W}(\cdot|w)\!\in\!\smash{\nbhd_{\sqrt{\smash[b]{\delta(\eps)}}}^\X(P_X)},\
P_{Y|W}(\cdot|w)\!\in\!\smash{\nbhd_{\sqrt{\smash[b]{\delta(\eps)}}}^\Y(P_Y)},\notag\\
&\, \qquad  \text{for all $w\!\in\!\W$ and $\delta(\cdot)>0$ such that
  $\lim_{\eps\to0}\delta(\eps)\!\to\!0$.} \ \Bigr\}. 
\label{eq:cPe-def}
\end{align}
\end{definition}

In \defref{def:eci}, a configuration of $W$ such that
$P_{W|X,Y}\in\cP_\eps$ takes the form
\begin{align} 
\C^{\X,\Y}_\eps(P_{X,Y}) 
&\defeq \smash[b]{\Bigl\{}\, \W,\ \{P_W(w),\ w\in\W\},\notag\\ 
&\quad\qquad\{P_{X|W}(\cdot|w),\ w\in\W\},\notag\\
&\qquad\qquad\{P_{Y|W}(\cdot|w),\ w\in\W\} \smash[t]{\Bigr\}}
\label{eq:Cw-def}
\end{align}
subject to the constraints
\begin{subequations} 
\begin{align}
P_{X|W}(\cdot|w) &\in\nbhd^\X_{\sqrt{\smash[b]{\delta(\eps)}}}(P_X),\ w\in\W, \\
P_{Y|W}(\cdot|w) &\in\nbhd^\Y_{\sqrt{\smash[b]{\delta(\eps)}}}(P_Y),\ w\in\W, 
\end{align}
\label{eq:PXWYW-nbhd}%
\end{subequations}
for some $\delta$ such that $\delta(\eps)\to0$ as $\eps\to0$,
and 
\begin{equation}
P_{X|W}(x|w)\, P_{Y|W}(y|w) = 
P_{X,Y|W}(x,y|w).
\label{eq:P_XY|W-form}
\end{equation}
In turn, \eqref{eq:P_XY|W-form} implies the constraint
\begin{equation} 
\sum_{w\in\W} P_W(w)\, P_{X|W}(x|w)\, P_{Y|W}(y|w) = P_{X,Y}(x,y),
\label{eq:config-joint}
\end{equation}
which further implies the constraints
\begin{subequations} 
\begin{align}
\sum_{w\in\W} P_W(w)\, P_{X|W}(x|w) &= P_X(x) \\
\sum_{w\in\W} P_W(w)\, P_{Y|W}(y|w) &= P_Y(y).
\end{align}
\label{eq:config-marg}%
\end{subequations}

Defining the information vectors
\begin{subequations} 
\begin{align} 
\phi^{X|W}_w(x) &\defeq
\frac{P_{X|W}(x|w)-P_X(x)}{\sqrt{\smash[b]{\delta(\eps)}}\sqrt{P_X(x)}} \label{eq:phiXW-def} \\
\phi^{Y|W}_w(y) &\defeq
\frac{P_{Y|W}(y|w)-P_Y(y)}{\sqrt{\smash[b]{\delta(\eps)}}\sqrt{P_Y(y)}}, \label{eq:phiYW-def} 
\intertext{and}
\phit^{X,Y|W}_w(x,y) &\defeq
\frac{P_{X,Y|W}(x,y|w)-P_X(x)\,P_Y(y)}{\sqrt{\smash[b]{2\delta(\eps)}}\sqrt{P_X(x)\,P_Y(y)}},
 \label{eq:phiXYW-def}
\end{align}
\label{eq:phiXWYW-def}%
\end{subequations}
we can equivalently express \eqref{eq:Cw-def} in the form
\begin{align} 
\C^{\X,\Y}_\eps(P_{X,Y}) 
&\defeq \smash[b]{\Bigl\{}\, \W,\ \{P_W(w),\ w\in\W\},\notag\\ 
&\qquad\qquad \{\bphi^{X|W}_w,\ w\in\W\} \notag\\
&\qquad\qquad\qquad \{\bphi^{Y|W}_w,\ w\in\W\} \smash[t]{\Bigr\}}
\label{eq:Cw-alt}
\end{align}
subject to the constraints
\begin{subequations} 
\begin{align}
\bphi^{X|W}_w\in\ivspaceX,\quad w\in\W, \\
\bphi^{Y|W}_w\in\ivspaceY,\quad w\in\W, 
\end{align}
\label{eq:phiXWYW-ivspace}%
\end{subequations}
which correspond to \eqref{eq:PXWYW-nbhd},
and
\begin{subequations} 
\begin{align} 
&\phit^{X,Y|W}_w(x,y) \notag\\
&\qquad= \phic^{X,Y|W}_w(x,y)
+ \sqrt{\frac{\delta(\eps)}{2}}\,
\phi^{X|W}_w(x)\, \phi^{Y|W}_w(y)
\end{align}
with
\begin{align} 
&\phic^{X,Y|W}_w(x,y) \notag\\
&\qquad\defeq \frac1{\sqrt{2}} \left( \sqrt{P_Y(y)} \, \phi^{X|W}_w(x) +
\sqrt{P_X(x)} \, \phi^{Y|W}_w(y) \right),
\label{eq:ivcXY|W-def}
\end{align}
\label{eq:ivXY|W}%
\end{subequations}
as well as
\begin{equation} 
\delta(\eps)\!\! \sum_{w\in\W} P_W(w)\,
  \phi^{X|W}_w\!(x)\,\phi^{Y|W}_w\!(y) 
 = \dtmts(y,x),\ x\in\X,\ y\in\Y,
\label{eq:ivXY}%
\end{equation}
and
\begin{subequations}
\begin{align} 
\sum_{w\in\W} P_W(w)\,\phi^{X|W}_w(x) &= 0,\quad x\in\X, \\
\sum_{w\in\W} P_W(w)\,\phi^{Y|W}_w(y) &= 0,\quad y\in\Y,
\end{align}
\label{eq:phi-marg}%
\end{subequations}
which correspond to \eqref{eq:P_XY|W-form}--\eqref{eq:config-marg},
respectively.  In particular, we obtain \eqref{eq:phi-marg} from
\eqref{eq:config-marg} using
\eqref{eq:phiXW-def}--\eqref{eq:phiYW-def}, and 
we obtain \eqref{eq:ivXY|W} from
\begin{align}
&P_{X,Y|W}(x,y|w) \\
&\quad= P_{X|W}(x|w)\,P_{Y|W}(y|w) \\
&\quad= \bigl(P_X(x)+\sqrt{\smash[b]{\delta(\eps)}}\,\sqrt{P_X(x)}\,\phi^{X|W}_w(x)\bigr) \notag\\
&\qquad\qquad\qquad\qquad{}\cdot
  \bigl(P_Y(y)+\sqrt{\smash[b]{\delta(\eps)}}\,\sqrt{P_Y(y)}\,\phi^{Y|W}_w(y)\bigr)
  \notag\\
&\quad= P_X(x)\,P_Y(y) + \sqrt{\smash[b]{\delta(\eps)}} \sqrt{P_X(x)\,P_Y(y)} \notag\\
&\quad\qquad{} \cdot
\Bigl[
    \sqrt{P_Y(y)}\,\phi^{X|W}_w(x)+\sqrt{P_X(x)}\,\phi^{Y|W}_w(y) \notag\\
&\ \ \,\quad\qquad\qquad\qquad\qquad\qquad{} +
        \sqrt{\smash[b]{\delta(\eps)}}\,
        \phi^{X|W}_w(x)\,\phi^{Y|W}_w(y)\Bigr] , 
\label{eq:PXY|W-int}
\end{align}
where we have used \eqref{eq:P_XY|W-form} and
\eqref{eq:phiXW-def}--\eqref{eq:phiYW-def}, and where we recognize the
term in brackets as $\sqrt{2}\,\phit^{X,Y|W}_w(x)$ according to
\eqref{eq:phiXYW-def}.  Finally, we obtain \eqref{eq:ivXY} from the
expectation of 
\eqref{eq:PXY|W-int} with respect to $P_W$, yielding
\begin{align} 
&P_{X,Y}(x,y) \notag\\
&\quad= \sum_{w\in\W} P_W(w) \, P_{X|W}(x|w)\, P_{Y|W}(y|w) \notag\\
&\quad= P_X(x) \, P_Y(y) \notag\\
&\qquad\qquad{} + \sqrt{P_X(x)\,P_Y(y)} \notag\\
&\quad\qquad\qquad\qquad{}\cdot \delta(\eps) \!
\sum_{w\in\W} P_W(w)\, \phi^{X|W}_w(x)\,\phi^{Y|W}_w(y),
\label{eq:PXY-iv-exp}
\end{align}
where we have used \eqref{eq:phi-marg}, and where we recognize
$\dtmts(y,x)$ as defined in \eqref{eq:dtmts-def} as the final factor
in \eqref{eq:PXY-iv-exp}.

The following variational characterization of the nuclear (i.e.,
 trace) norm (see, e.g., \cite{rfp10,bmp08}) is useful in our
development.
\begin{lemma}
\label{lem:kyfan}
Given an arbitrary $k_1\times k_2$ matrix $\bA$, we have
\begin{equation}
\min_{\substack{\left\{k,\ \bM_1\in\reals^{k_1\times
      k},\ \bM_2\in\reals^{k\times k_2}\colon
    \right.\\ \left. \bM_1\bM_2=\bA \vphantom{\reals^{k_1\times
      k}} \right\}}}
\left( \frac12 \frob{\bM_1}^2 + \frac12 \frob{\bM_2}^2 \right) = \nuclear{\bA}.
\end{equation}
\end{lemma}

In particular, we obtain that the $\eps$-common information is given
by the nuclear norm of $\dtmt$.  A proof is provided in
\appref{app:common}.
\begin{proposition}
\label{prop:common}
Given $P_{X,Y}\in\nbhdk_\eps^{\X\times\Y}(P_XP_Y)$ for $\eps>0$, we
have\footnote{ Since $I(W;X,Y)\ge \max\{I(W;X),I(W;Y)\}$ by the chain
  rule, it follows that our result does not change if we further
  include in \eqref{eq:cPe-def} of \defref{def:eci} all distributions
  $P_{X|W}(\cdot|w)$ and $P_{Y|W}(\cdot|w)$ that for all $w\in\W$ do
  not depend on $\eps$, since they will give rise to nonvanishing
  $I(W;X)$ and $I(W;Y)$.  In essence, the configurations our
  definition omits are those for which the $P_W$ is increasingly
  severely imbalanced as $\eps\to0$. }
\begin{subequations} 
\begin{equation}
C(X,Y) \le C_\eps(X,Y) = \nuclear{\dtmt} + o(\eps),
\label{eq:CXY}
\end{equation}
where
\begin{equation}
\nuclear{\dtmt} = \sum_{i=1}^{K-1} \sigma_i,
\end{equation}
\label{eq:CXY-full}%
\end{subequations}
which is achieved by the configuration
\begin{align}
&C^{\X,\Y}_{*}(P_{X,Y}) \notag\\
&\ = \biggl\{
\W = \{\pm1,\dots,\pm(K-1)\}, \notag\\
&\ \, \qquad P_W(w) = \frac{\sigma_\abs{w}}{2\nuclear{\dtmt}},\notag\\
&\ \, \qquad P_{X|W}(x|w) = P_X(x)
\left(1+\sgn(w)\,\nuclear{\dtmt}^{1/2} \, f_\abs{w}^*(x)\right),\notag\\
&\ \, \qquad P_{Y|W}(y|w) = P_Y(y)
\left(1+\sgn(w)\,\nuclear{\dtmt}^{1/2} \, g_\abs{w}^*(y)\right)
 \biggr\}.
\label{eq:W-config-opt}
\end{align}%
and $\delta(\eps)=\eps$ in \eqref{eq:cPe-def}.
\end{proposition}

We note that while in general the cardinality of $W$ in the
characterization of Wyner's common information is known only to
satisfy the upper bound $\card{\W}\le \cardX\times\cardY$, we obtain
that cardinality $\cardW=2(K-1)$, which is much smaller when $\X$ and/or
$\Y$ is large, suffices to achieve $\eps$-common information as
$\eps\to0$.

Given data $(x^m,y^m)$ from the extended model 
\begin{subequations} 
\begin{equation}
X^m \markov W \markov Y^m,
\end{equation}
with
\begin{align} 
P_{X^m|W}(x^m|w) &= \prod_{i=1}^m P_{X|W}(x_i|w) \\
P_{Y^m|W}(y^m|w) &= \prod_{i=1}^m P_{Y|W}(y_i|w) \\
P_{X^m,Y^m}(x^m,y^m) &= \prod_{i=1}^m P_{X,Y}(x_i,y_i),
\end{align}%
\label{eq:full-markov-common}%
\end{subequations}
it further follows from \propref{prop:common} that 
\begin{equation}
R^{K-1}_* \defeq S^{K-1}_*+T^{K-1}_*
\label{eq:R*-def}
\end{equation}
with $S^{K-1}_*$ and $T^{K-1}_*$ as defined via \eqref{eq:stmk*-def},
is, as $\eps\to0$, a sufficient statistic for inferences about the
$\eps$-common information variable $W$, i.e., we have the Markov chain
\begin{equation}
W \markov R^{K-1}_* \markov \bigl(S_*^{K-1},T_*^{K-1}\bigr) \markov
(X^m,Y^m),\quad \eps\to0. 
\end{equation}
In particular, we have the following result, a proof of which is
provided in \appref{app:suffstat-common}.
\begin{corollary}
\label{corol:suffstat-common}
In the solution to the optimization in \propref{prop:common} for the
extended model \eqref{eq:full-markov-common},
\begin{align} 
&P_{W|X^m,Y^m}(w|x^m,y^m) \notag\\
&\quad\qquad= \frac{\sigma_\abs{w}}{2\nuclear{\dtmt}} \left( 1 +
  m\,\sgn(w)\, \nuclear{\dtmt}^{1/2}\, r_\abs{w}^* \right) + o(\sqrt{\eps}),
\label{eq:suffstat-common}
\end{align}
where, consistent with \eqref{eq:R*-def},
\begin{equation}
r_i^* = s_i^* + t_i^*,
\label{eq:ri*-def}
\end{equation}
and $s_i^*$ and $t_i^*$ are as defined in \eqref{eq:sti*-def}.
\end{corollary}

The sufficiency relation of \corolref{corol:suffstat-common} can be
equivalently expressed in the form
\begin{equation*}
\lim_{\eps\to0} \frac{I(W;X^m,Y^m)}{I(W;R^{K-1}_*)} = 1.
\end{equation*}
In essence, \corolref{corol:suffstat-common} shows that in making
inferences about the $\eps$-common information variable $W$ from
high-cardinality data $(X^m,Y^m)$, it is sufficient to extract a
low-dimensional real-valued sufficient statistic $R^{K-1}$.  And we
emphasize that a consequence of the way common information is defined
is that the sufficient statistic $R_*^{K-1}$ involves \emph{separate}
processing of $X^m$ and $Y^m$.

\subsection{Relating Common Information to Dominant Structure}
\label{sec:relate}

The $\eps$-common information variable $W$ can be related to the
dominant structure sequence pair $(U^{K-1},V^{K-1})$.  To develop
this, let us equivalently express $W$ as
\begin{subequations} 
\begin{equation}
W \defeq W^{K-1} = (W_1,\dots,W_{K-1}),
\end{equation}
where each $W_i$ is a variable defined over alphabet
\begin{equation}
\W_\circ\defeq\{-1,0,+1\}
\label{eq:W0-def}
\end{equation}
according to
\begin{equation}
W_i \defeq \begin{cases} 
+1 & W = i \\
-1 & W = -i \\
 0 & \text{otherwise}.
\end{cases}
\end{equation}
\label{eq:Wi-def}%
\end{subequations}

We can interpret $W_i$ as effectively capturing the $\eps$-common
information in $(U_i,V_i)$, which is defined, consistent with
\defref{def:eci}, as
\begin{equation} 
C_\eps(U_i,V_i) = \min_{P_{\Wt_i|U_i,V_i}\in\cPt_\eps} I(\Wt_i;U_i,V_i),
\label{eq:eps-ci-def-uv}
\end{equation}
where
\begin{align}
&\cPt_\eps \defeq \notag\\
& \Bigl\{\, P_{\Wt_i|U_i,V_i}\in\cP^{\Wt_i},\ \text{some $\cWt_i$} \colon U_i\markov \Wt_i\markov V_i \ \text{and} \notag\\
&\quad
P_{U_i|\Wt_i}(\cdot|\wt_i)\!\in\!\smash{\nbhd_{\sqrt{\smash[b]{\delta(\eps)}}}^{\U_i}(P_{U_i})},\
P_{V_i|\Wt_i}(\cdot|\wt_i)\!\in\!\smash{\nbhd_{\sqrt{\smash[b]{\delta(\eps)}}}^{\V_i}(P_{V_i})},\notag\\
&\, \quad  \text{for all $\wt_i\!\in\!\cWt_i$ and $\delta(\cdot)>0$ such that
  $\lim_{\eps\to0}\delta(\eps)\!\to\!0$.} \ \Bigr\}. 
\label{eq:cPe-def-uv}
\end{align}

In particular, we have the following result, a proof of which is
provided in \appref{app:common-relate}.
\begin{corollary}
\label{corol:common-relate}
Given $P_{X,Y}\in\nbhdk_\eps^{\X\times\Y}(P_XP_Y)$ for $\eps>0$, and
let $W^{K-1}$ be the representation \eqref{eq:Wi-def} of the
optimizing $\eps$-common information variable $W$ in
\propref{prop:common}.  Then
\begin{equation}
C_\eps(X,Y) = I(W;X,Y) = \sum_{i=1}^{K-1} I(W_i;X,Y) + o(\eps),
\label{eq:C-decomp}
\end{equation}
where 
\begin{equation} 
I(W_i;X,Y) = \sigma_i + o(\eps),\quad i=1,\dots,K-1.
\label{eq:C-comp}
\end{equation}
Moreover, if $(U^{K-1},V^{K-1})$ are the optimizing
$\epst$-multi-attributes in \propref{prop:ib-double} for some
$\epst>0$, then 
\begin{equation}
C_\eps(U_i,V_i) = \epst^2\, I(W_i;X,Y) + o(\epst^2\eps),\quad i=1,\dots,K-1.
\label{eq:CUV}
\end{equation}
\end{corollary}

The associated data processing implications follow from the extended
Markov structure \eqref{eq:full-markov-common}.  In particular, an
(asymptotically) sufficient statistic for making decisions about $W_i$
from $(X^m,Y^m)$ is 
\begin{equation} 
R_i^*=S_i^*+T_i^*,
\label{eq:common-ss-i}
\end{equation}
i.e., we have the Markov chain
\begin{equation*}
W_i \markov R_i^* \markov (S_i^*,T_i^*) \markov (X^m,Y^m),\quad\eps\to0.
\end{equation*}
In particular, we have the following result, a proof of which is
provided in \appref{app:Wi-markov}.
\begin{corollary}
\label{corol:Wi-markov}
For $W_i$ as defined in \eqref{eq:Wi-def}, 
\begin{align} 
&P_{W_i|X^m,Y^m}(w_i|x^m,y^m) \notag\\
&\quad = \begin{cases}
\ds\frac{\sigma_i}{2\,\nuclear{\dtmt}} \bigl(1 + m\,w_i\,\nuclear{\dtmt}^{1/2}\,r_i^* \bigr) +
o(\sqrt{\eps}) & w_i=\pm1 \\
\ds \vphantom{\Biggm|^M}\left(1-\frac{\sigma_i}{\nuclear{\dtmt}} \right) + o(\sqrt{\eps}) & w_i=0,
\end{cases}
\label{eq:Wi-markov}
\end{align}
whose dominant term depends on $x^m,y^m$ only through $r_i^*$.
\end{corollary}

By comparison, $\Wt_i$ satisfies the asymptotic Markov structure
\begin{equation*}
\Wt_i \markov Z_i \markov R_i^* \markov (X^m,Y^m),\quad \epst,\eps\to0,
\end{equation*}
where
\begin{equation}
Z_i\defeq U_i+V_i.
\label{eq:Zi-def}
\end{equation}
In particular, we have the following result, a proof of which is
provided in \appref{app:Wti-markov}.
\begin{corollary}
\label{corol:Wti-markov}
For $\Wt_i$ as defined in \eqref{eq:eps-ci-def-uv},
\begin{subequations}
\begin{align}
&P_{\Wt_i|Z_i,X^m,Y^m}(\wt_i|z_i,x^m,y^m) \notag\\
&\quad= o(\epst\sqrt{\eps}) + 
\begin{cases}
\bigl( 1 + \sgn(\wt_i\,z_i)\,\sqrt{\sigma_i} \bigr)/2 & z_i=\pm2 \\
0 & z_i=0,
\end{cases}
\label{eq:markov-2}
\end{align}
whose dominant term depends on $x^m,y^m$ (and thus $r_i^*$) only through $z_i$, and
\begin{align} 
&P_{Z_i|X^m,Y^m}(z_i|x^m,y^m) \notag\\
&\quad= o(\epst) + 
\begin{cases} 
\bigl( 1 + \epst\, m\, \sgn(z_i)\, r_i^*\bigr)/4 & z_i=\pm2 \\
1/2 & z_i=0,
\end{cases}
\label{eq:markov-1}
\end{align}
whose dominant term depends on $x^m,y^m$ only through $r_i^*$.
\label{eq:markov-pair}%
\end{subequations}
\end{corollary}

\section{Efficient Learning of Modal Decompositions}
\label{sec:ace}

Since the universal features developed in
\secref{sec:unifeature-char} are naturally expressed in terms
of the modal decomposition \eqref{eq:modal} of the joint distribution
$P_{X,Y}$, efficient learning of this decomposition from data is key
to the practical applicability of these features.  We 
discuss some of the main aspects of such computational issues in this
section.

We begin by more fully describing the scenario of interest.  To this
point in our development, for pedagogical reasons our model has
been that $P_{X,Y}$ is known, by which we mean practically that it has
been reliably estimated from some training phase.  When this is the
case, the problem of efficiently computing the modal decomposition is
simply one of efficient computation of the SVD of the associated
matrix $\dtmt$.

However, in practice, of course, learning $P_{X,Y}$ is an important
aspect of the overall feature selection process in the inference
pipeline.  Accordingly, this section describes how this learning
can be accomplished efficiently from samples as part of the
computation of the modal decomposition.

It will be convenient in our exposition to first consider the modal
computation from known $P_{X,Y}$, then use the resulting foundation to
address the problem of learning the modal decomposition from samples.

\subsection{Computing the Modal Decomposition}

Given $P_{X,Y}$, computation of the SVD of $\dtmt$ is a
straightforward exercise in numerical linear algebra.  In particular,
from $P_{X,Y}$ we compute marginals $P_X$ and $P_Y$, then construct
$\dtmt$ via \eqref{eq:dtmt-def}, then apply any of many
well-established numerical methods for computing the SVD of a
matrix---see, e.g., \cite{gvl12,ys11,tb97}.  However, in this section
we further develop an interpretation of the resulting computation in
the context of probabilistic analysis that will be particularly useful
in the sequel.  We emphasize at the outset that the results of this
subsection are largely not new, but rather establish the viewpoints
and interpretations we need for our subsequent development.

\subsubsection{Orthogonal Iteration}
\label{sec:orthog-iter}

Among the oldest and simplest approaches to the computation of the
principal singular value and vector of a matrix is referred to as
\emph{power iteration} or the \emph{power method}
\cite{gvl12,ys11,tb97}.  Moreover, power iteration can be used in a
sequential manner to recover any number of the largest singular values
and their corresponding singular vectors.  However, when the first
$1<k\le K-1$ dominant modes are desired, it is more efficient to compute
them in parallel via a generalization of power iteration.  The most
direct generalization is referred to as \emph{orthogonal iteration}
\cite[Section~7.3.2]{gvl12}.\footnote{A refinement of orthogonal
  iteration referred to as \emph{QR iteration}
  \cite[Section~7.3.3]{gvl12} forms the basis of most practical
  implementations, and can be used in conjunction with various
  acceleration techniques.}  This algorithm has a corresponding
relation to the generalized variational characterizations of the SVD
we have used throughout our analysis.

To implement orthogonal iteration, we start with some $\cardX\times k$
matrix $\bPsib^{X,0}_{(k)}$, which is typically chosen at random, and then
execute the following iterative procedure:
\begin{enumerate}
\item Set $l=0$.
\item Orthogonalize $\bPsib^{X,l}_{(k)}$ 
to obtain $\bPsih^{X,l}_{(k)}$
via the (thin or reduced) QR decomposition \cite{gvl12}
\begin{equation}
\bPsib^{X,l}_{(k)} = \bPsih^{X,l}_{(k)}\, \cholX{l}_{(k)},
\label{eq:qr-x}
\end{equation}
in which $\cholX{l}_{(k)}$ is a $k\times k$ upper triangular matrix.
\item Compute
\begin{equation}
\bPsib^{Y,l}_{(k)} = \dtmt\, \bPsih^{X,l}_{(k)},
\label{eq:dtmt-map}
\end{equation}
then orthogonalize to obtain $\bPsih^{Y,l}_{(k)}$ via
the (thin or reduced) QR decomposition
\begin{equation}
\bPsib^{Y,l}_{(k)} = \bPsih^{Y,l}_{(k)} \, \cholY{l}_{(k)},
\label{eq:qr-y}
\end{equation}
in which $\cholY{l}_{(k)}$ is a $k\times k$ upper triangular matrix.
\item Produce the update 
\begin{equation}
\bPsib^{X,l+1}_{(k)} = \dtmt^\T \bPsih^{Y,l}_{(k)}.
\label{eq:dtmt-adj-map}
\end{equation}
\item Increment $l$ and return to Step 2.
\end{enumerate}
The QR decompostions employed in the orthogonalizations
in this algorithm can be directly implemented using, e.g., the Gram-Schmidt
procedure.  However, numerical stability can be improved in practice
through the use of Householder transformations \cite{gvl12}.

The convergence of this procedure depends on the properties of 
\begin{equation}
\bA = \bigl(\bPsi^X_{(k)}\bigr)^\T\bPsib^{X,0}_{(k)},
\label{eq:bPsib0-decomp}
\end{equation}
where $\bPsi^X_{(k)}$ is the matrix of dominant singular vectors
defined in \eqref{eq:bPsi-k-def}.  In particular, 
using $a_{i,j}$ to denote
the $(i,j)$\/th entry of $\bA$, 
when
$\sigma_1,\dots,\sigma_{k+1}$ are unique and there exist distinct
$j_1,\dots,j_k$ such that $a_{i,j_i}\neq0$ for each $i$, then we
obtain, as $l\to\infty$, \cite[Theorem~7.3.1]{gvl12}
\begin{align*}
\bPsih^{X,l}_{(k)} &\to \bPsi^X_{(k)}\\
\bPsih^{Y,l}_{(k)} &\to \bPsi^Y_{(k)}\\
\bigl(\bPsih^{Y,l}_{(k)}\bigr)^\T \dtmt\,\bPsih^{X,l}_{(k)}  &\to \bSi_{(k)},
\end{align*}
where $\bSi_{(k)}$ is a diagonal matrix with diagonal entries
$\sigma_1,\dots,\sigma_k$.  Moreover, convergence is exponentially
fast.\footnote{However, it is worth emphasizing that the closer a pair
  of dominant singular values are to each other, the poorer the
  convergence rate.}  When $\sigma_1,\dots,\sigma_k$ are not distinct
(but still $\sigma_k>\sigma_{k+1}$),
natural generalizations of these results are obtained
\cite[Theorem~7.3.1]{gvl12}. 

\subsubsection{Statistical Interpretation as the ACE Algorithm}
\label{sec:ace-interp}

The use of orthogonal iteration to compute the dominant modes in
\eqref{eq:modal} has a direct statistical interpretation that
corresponds to what is referred to as the alternating conditional
expectations (ACE) algorithm \cite{bf85,wm04}.

In particular, choosing the correspondences
\begin{align*} 
\fh_i(x) \defeq
\frac{\psih^{X,l}_i(x)}{\sqrt{P_X(x)}},
&\qquad
\fb_i(x) \defeq
\frac{\psib^{X,l}_i(x)}{\sqrt{P_X(x)}},
\\ 
\gh_i(y) \defeq \frac{\psih^{Y,l}_i(y)}{\sqrt{P_Y(y)}},
&\qquad
\gb_i(y) \defeq \frac{\psib^{Y,l}_i(y)}{\sqrt{P_Y(y)}}, 
\end{align*}%
for $i=1,\dots,k$, with
\begin{align*} 
\bPsih^{X,l}_{(k)} &= 
\begin{bmatrix} \bpsih^{X,l}_1 & \cdots & \bpsih^{X,l}_k \end{bmatrix} \\
\bPsih^{Y,l}_{(k)} &= 
\begin{bmatrix} \bpsih^{Y,l}_1\, &\cdots& \bpsih^{Y,l}_k\, \end{bmatrix}
\end{align*}
and
\begin{align*} 
\bPsib^{X,l}_{(k)} &= 
\begin{bmatrix} \bpsib^{X,l}_1 & \cdots & \bpsib^{X,l}_k \end{bmatrix} \\
\bPsib^{Y,l}_{(k)} &= 
\begin{bmatrix} \bpsib^{Y,l}_1\, &\cdots& \bpsib^{Y,l}_k\, \end{bmatrix},
\end{align*}
we can rewrite the procedure of \secref{sec:orthog-iter}
in the form of \algoref{alg:ace-comp-k}, which iteratively
computes both the dominant $k$ features $(f^k_*,g^k_*)$ and
\begin{equation*}
\sigma^{(k)}=\sum_{i=1}^k \sigma_i,
\end{equation*}
the Ky Fan $k$-norm of $\dtmt$.  As such, the convergence behavior
follows immediately from the corresponding analysis of orthogonal
iteration.\footnote{We emphasize that the Cholesky decompositions in
  \algoref{alg:ace-comp-k} are unique when the associated covariance
  matrices are full rank, which is the case when: 1) the singular
  values are distinct; and 2) the covariance matrix of the
  initialization is positive definite, with components correlated with
  each of the dominant features in the modal decomposition.}

To obtain \algoref{alg:ace-comp-k}, we use that \eqref{eq:qr-x} and
\eqref{eq:qr-y} can be equivalently expressed in the form\footnote{For
  convenience, in this derivation, we drop the dependence on iteration
  (superscript $l$) from our the notation, consistent with the
  notation in \algoref{alg:ace-comp-k}.}
\begin{align*}
\fb^k(x) &= \bigl(\cholXg_{(k)}\bigr)^\T \fh^k(x),\quad x\in\X \\
\gb^k(y) &= \bigl(\cholYg_{(k)}\bigr)^\T \gh^k(y),\quad y\in\Y,
\end{align*}
via which we also obtain
\begin{align*} 
\sum_{x\in\X} P_X(x)\, \fb^k(x)\,\fb^k(x)^\T &= 
\bigl(\bPsib^X_{(k)})^\T \bPsib^X_{(k)} = 
\bigl(\cholXg_{(k)})^\T \cholXg_{(k)} \\
\sum_{y\in\Y} P_Y(y)\, \gb^k(y)\,\gb^k(y)^\T &= 
\bigl(\bPsib^Y_{(k)})^\T \bPsib^Y_{(k)} =
\bigl(\cholYg_{(k)})^\T \cholYg_{(k)},
\end{align*}
since
\begin{equation*}
\bigl(\bPsih^X_{(k)})^\T \bPsih^X_{(k)} = 
\bigl(\bPsih^Y_{(k)})^\T \bPsih^Y_{(k)} = \bI.
\end{equation*}
Additionally, we use that, via
\eqref{eq:dtmts-def}, 
\eqref{eq:dtmt-map} and \eqref{eq:dtmt-adj-map} 
can be equivalently expressed
in the form 
\begin{align*}
\gb^k(y) 
&= \frac1{\sqrt{P_Y(y)}}\, \sum_{x\in\X} \dtmts(y,x)\,
  \sqrt{P_X(x)}\,\fh^k(x) \\
&= \bE{\fh^k(X)\mid Y=y} - \bE{\fh^k(X)} \\
\fb^k(x) 
&= \frac1{\sqrt{P_X(x)}}\, \sum_{x\in\X} \dtmts(y,x)\,
  \sqrt{P_Y(y)}\,\gh^k(y) \\
&= \bE{\gh^k(Y)\mid X=x} - \bE{\gh^k(Y)}.
\end{align*}

\begin{algorithm}[tbp]
\caption{ACE Algorithm for Multiple Mode Computation}
\label{alg:ace-comp-k}
\begin{algorithmic}[]
\Require{Joint distribution $P_{X,Y}$, number of modes $k$} \\ 
1. Initialization: randomly choose $\fb^k(x), \ \forall x\in\X$ 
\Repeat\\
	\quad 2a. Center: $\fb^k(x) \gets \fb^k(x) - \bE{\fb^k(X)},
        \quad \forall x\in\X$\\
        \quad 2b. Cholesky factor:\\
        \quad\qquad
        $\bE{\fb^k(X)\,\fb^k(X)^\T}=\bigl(\cholXg_{(k)}\bigr)^\T
        \cholXg_{(k)}$\\ 
	\quad 2c. Whiten: \\ 
        \quad\qquad $\fh^k(x) \gets \bigl(\cholXg_{(k)}\bigr)^{-\T} \, \fb^k(x),
        \ \forall x\in\X$ \\ 
	\quad 2d. $\gb^k(y) \gets \bE{\fh^k(X)\mid Y=y}, \ \forall y\in\Y$\\
	\quad 2e. Center: $\gb^k(y) \gets \gb^k(y) - \bE{\gb^k(Y)},
        \quad \forall y\in\Y$\\  
        \quad 2f. Cholesky factor:\\
        \quad\qquad$\bE{\gb^k(Y)\,\gb^k(Y)^\T}=\bigl(\cholYg_{(k)}\bigr)^\T \cholYg_{(k)}$\\
	\quad 2g. Whiten: \\
        \quad\qquad $\gh^k(y) \gets \bigl(\cholYg_{(k)}\bigr)^{-\T} \,
        \gb^k(y), \ \forall y \in \Y$ \\ 
	\quad 2h. $\fb^k(x) \gets \bE{\gh^k(Y)\mid X=x}, \ \forall x\in\X$\\
        \quad 2i. $\sigmah^{(k)} \gets \bE{\fb^k(X)^\T \gh^k(Y)}$
\Until $\sigmah^{(k)}$ stops increasing.
\end{algorithmic}
\end{algorithm}

Note, too, that via Bayesian estimation theory, the conditional expectations
in \algoref{alg:ace-comp-k} can be interpreted as minimum mean-square
error (MMSE) estimates, and thus we can equivalently write the
corresponding steps 2h and 2d as optimizations; specifically, we have,
respectively, the variational characterizations
\begin{subequations} 
\begin{align}
\fb^k(\cdot) &\gets \argmin_{f^k(\cdot)} \E{\bnorm{f^k(X)-\gh^k(Y)}^2} \\
\gb^k(\cdot) &\gets \argmin_{g^k(\cdot)} \E{\bnorm{\fh^k(X)-g^k(Y)}^2},
\end{align}
\label{eq:var-ce}%
\end{subequations}
where we note that these optimizations can, themselves, be carried out
by an iterative procedure.  Such implementations can be attractive
when there are contraints on $f^k_*(\cdot)$ and $g^k_*(\cdot)$ based on,
e.g., domain knowledge and/or other considerations.

\subsection{Estimating the Modal Decomposition from Data}
\label{sec:modal-approx}

When the joint distribution $P_{X,Y}$ is unknown, as is common in
applications, but we have (labeled) training data 
\begin{equation}
\cT \defeq \{(x_1,y_1),\dots,(x_n,y_n)\},
\label{eq:cT-def}
\end{equation}
drawn i.i.d.\ from $P_{X,Y}$, we can replace $P_{X,Y}$ in 
\algoref{alg:ace-comp-k} with the empirical distribution
\begin{equation}
\Ph_{X,Y}(x,y) \defeq \frac1n \sum_{i=1}^n \kron_{x=x_i}\,\kron_{y=y_i}
\label{eq:Phn-def}
\end{equation}
to generate an estimate of the modal decomposition.  In this case, the
expectations in \algoref{alg:ace-comp-k} are all with respect to the
corresponding empirical distributions.
In particular, those in steps
2a-b and 2e-f are with respect to, respectively,
\begin{equation*}
\Ph_X(x) \defeq \sum_y \Ph_{X,Y}(x,y)\quad\text{and}\quad
\Ph_Y(y) \defeq \sum_x \Ph_{X,Y}(x,y),
\end{equation*}
while those in steps 2d and 2h are with respect to, respectively,
\begin{equation*}
\Ph_{X|Y}(x|y) \defeq \frac{\Ph_{X,Y}(x,y)}{\Ph_Y(y)}
\quad\text{and}\quad
\Ph_{Y|X}(y|x) \defeq \frac{\Ph_{X,Y}(x,y)}{\Ph_X(x)},
\end{equation*}
and that in step 2i is with respect to $\Ph_{X,Y}(x,y)$.

Evidently, in this version of \algoref{alg:ace-comp-k}, the
computational complexity scales with the number of training samples
$n$.  There are a variety of ways to reduce this complexity in
practice.  For example, among other possibilities, in each basic
iteration we can choose to operate on only a (comparatively small)
randomly chosen subset of the training data and exploit bootstrapping
techniques.

It is also worth emphasizing that in some scenarios we may have both
\emph{labeled} and \emph{unlabeled} training data available, the
latter of which correspond to samples $x_1,\dots,x_{n'}$ and
$y_1,\dots,y_{n'}$, drawn i.i.d.\ from $P_X$ and $P_Y$, respectively.
While labeled data is typically expensive to obtain, since the
labeling process often involves a significant amount of manual labor,
unlabeled data is comparatively inexpensive to obtain, since no
correspondences need be identified.  As such, it is often possible to
accurately estimate $P_X$ and $P_Y$.  In such scenarios, the
corresponding version of \algoref{alg:ace-comp-k} replaces $P_{X,Y}$
with an estimate based the labeled training data subject the marginal
constraints $P_X$ and $P_Y$, which can be constructed in a variety of
ways; see, e.g., \cite{dvg73,rwmw74,mh85}.

In the sequel, we quantify the accuracy of the modal decomposition
when estimated from data.  In light of
the preceding discussion, for this analysis, $\fh_i^*$, $\gh_i^*$, and
$\sigmah_i$ for $i=1,\dots,K$ are defined via the modal
decomposition
\begin{equation} 
\Ph_{X,Y}(x,y) = P_X(x)\,P_Y(y)
\left[ 1 + \sum_{i=1}^K \sigmah_i \, \fh_i^*(x)\,\gh_i^*(y) \right],
\label{eq:modal-emp}
\end{equation}
where $\sigmah_1\ge \cdots\ge \sigmah_K\ge0$ and
$\bE{\fh_i^*(X)\,\fh_j^*(X)} =
\bE{\gh_i^*(Y)\,\gh_j^*(Y)}=\kron_{i=j}$ for $i,j\in\{1,\dots,K\}$.
We emphasize that in \eqref{eq:modal-emp} we only have $\sigmah_K=0$
and $\bE{\fh_i(X)}=\bE{\gh_i(Y)}=0$ for $i\in\{1,\dots,K\}$, when
$\Ph_X=P_X$ and $\Ph_Y=P_Y$.\footnote{As such these properties
  effectively hold when $n\gg\max\{\cardX,\cardY\}$.}  Accordingly, in
analysis, we will frequently find it convenient to use the 
equivalent zero-mean features
\begin{subequations} 
\begin{align}
\fc_i^*(x) &\defeq \fh_i^*(x) - \bE{\fh_i^*(X)} \label{eq:fc-def}\\
\gc_i^*(y) &\defeq \gh_i^*(y) - \bE{\gh_i^*(Y)} \label{eq:gc-def}
\end{align}
\label{eq:fgc-def}%
\end{subequations}
for $i=1,\dots,K$.

The decomposition \eqref{eq:modal-emp} corresponds to the singular
value decomposition of the quasi-CDM $\dtmh$ whose $(y,x)$\/th entry
is
\begin{equation}
\dtmhs(x,y) \defeq
\frac{\Ph_{X,Y}(x,y)-P_X(x)\,P_Y(y)}{\sqrt{P_X(x)}\sqrt{P_Y(y)}}, 
\label{eq:dtmhs-def}
\end{equation}
i.e.,
\begin{align} 
\dtmh &\defeq
\left[\sqrt{\bP_Y}\right]^{-1} \left[ \bPh_{Y,X} - \bP_Y\,\bP_X \right]
\left[\sqrt{\bP_X}\right]^{-1} \label{eq:dtmh-def}\\
&= \sum_{i=1}^K \sigmah_i \bpsih^Y_i \bigl(\bpsih^X_i\bigr)^\T,
\label{eq:dtmh-svd}
\end{align}
where the singular vectors in \eqref{eq:dtmh-svd} have elements
\begin{subequations} 
\begin{align} 
\psih^X_i(x) &\defeq \sqrt{P_X(x)}\, \fh_i^*(x),\qquad
x\in\X, \label{eq:psihX-def}\\ 
\psih^Y_i(y) &\defeq \sqrt{P_Y(y)}\, \gh_i^*(y),\qquad
y\in\Y, \label{eq:psihY-def} 
\end{align}
\label{eq:psih-def}%
\end{subequations}
for $i=1,\dots,K$. 

There are several aspects of the modal decomposition estimation whose
sample complexity is of interest, which we now address.

\subsubsection{Sample Complexity of Maximal Correlation Estimates}
\label{sec:sampcomp-maxcorr}

In this section, we determine the number of samples required to obtain
accurate estimates $\sigmah_1,\dots,\sigmah_k$ of
$\sigma_1,\dots,\sigma_k$, for $k\in\{1,\dots,K\}$.\footnote{Although
  the case $k=K$ is typically less of interest (since $\sigma_K=0$),
  we include it for completeness.  For this case, $f_K^*(X)$ and
  $g_K^*(Y)$ can be chosen freely subject to the constraint that they
  are uncorrelated with $f^{K-1}_*(X)$ and $g^{K-1}_*(Y)$,
  respectively.}  Specifically, we focus on the measure
\begin{equation}
\mu_1^k(P_{X,Y},\Ph_{X,Y}) \defeq \sum_{i=1}^k \abs{\sigmah_i-\sigma_i},
\label{eq:sigma-meas}
\end{equation}
and related quantities.

We begin with the following tail probability bound, 
a proof of which is provided in \appref{app:sval-est-tail}.
\begin{proposition}
\label{prop:sval-est-tail}
For $P_{X,Y}\in\simpXY$, let $p_0>0$ be such that 
\begin{equation}
P_X(x)\ge p_0\quad\text{and}\quad P_Y(y)\ge p_0,\qquad\text{all
  $x\in\X$, $y\in\Y$}.
\label{eq:marg-bnd}
\end{equation}
Then for  $k\in\{1,\dots,K\}$ and
  $0\le\delta\le \sqrt{k/2}/p_0$,
\begin{equation}
\prob{\sum_{i=1}^k \babs{\sigmah_i - \sigma_i} \ge \delta} 
\le \expop{\frac14 - \frac{p_0^2\,\delta^2 n}{8k}},
\label{eq:sval-est-tail}
\end{equation}
where $\sigmah_i$ for $i=1,\dots,K$ are defined via 
\eqref{eq:modal-emp} with
$\Ph_{X,Y}$ denoting the
empirical distribution based on $n$ training samples.  
\end{proposition}

A key consequence of \propref{prop:sval-est-tail} is the following
corollary.
\begin{corollary}
\label{corol:sval-est-mse}
Suppose $P_{X,Y}\in\simpXY$ is such that \eqref{eq:marg-bnd} is satisfied
for some $p_0>0$.  Then for $k\in\{1,\dots,K\}$ and $n$
sufficiently large that $n\ge 16\ln(4kn)$,
\begin{equation} 
\E{\left(\sum_{i=1}^k \bigl|\sigmah_i-\sigma_i\bigr|\right)^2} \le
\frac{6k+8k\ln(nk)}{p_0^2n},
\label{eq:sval-est-mse}
\end{equation}
where $\sigmah_i$ for $i=1,\dots,K$ are defined via 
\eqref{eq:modal-emp} with
$\Ph_{X,Y}$ denoting the
empirical distribution based on $n$ training samples.  
\end{corollary}

The proof of \corolref{corol:sval-est-mse}, provided in
\appref{app:sval-est-mse}, makes use of the following simple lemma,
which is a straightforward exercise in calculus.
\begin{lemma}
\label{lem:varphi-ab}
Given $a,b>0$, the convex function
$\varphi_{a,b}\colon \reals\mapsto\reals$ defined via 
\begin{equation} 
\varphi_{a,b}(\omega) \defeq \omega + a \e^{-b\omega},
\label{eq:varphi-ab-def}
\end{equation}
has its minimum at
\begin{equation}
\omega_* \defeq \argmin_\omega \varphi_{a,b}(\omega) = \frac1b \ln(ab),
\label{eq:varphi-ab-argmin}
\end{equation}
where it takes value
\begin{equation}
\min_{\omega} \varphi_{a,b}(\omega) = \varphi_{a,b}(\omega_*)=
\frac{1+\ln(ab)}{b}.
\label{eq:varphi-ab-min}
\end{equation}
\end{lemma}

Additional consequences of
\propref{prop:sval-est-tail} and \corolref{corol:sval-est-mse} are
that
\begin{equation*} 
\prob{\bbabs{\sum_{i=1}^k \bigl(\sigmah_i-\sigma_i\bigr)} \ge \delta}
\le \expop{\frac14 - \frac{p_0^2\,\delta^2 n}{8k}}
\end{equation*}
and
\begin{equation*} 
\E{\left(\sum_{i=1}^k \bigl(\sigmah_i - \sigma_i\bigr)\right)^2}
\le
\frac{6k+8k\ln(nk)}{p_0^2n},
\end{equation*}
respectively, which follow from 
the triangle inequality; specifically,
\begin{align} 
\bbabs{\kyfank{\bA_1}-\kyfank{\bA_2}} 
&= \bbabs{\sum_{i=1}^k 
  \bigl(\sigma_i(\bA_1)-\sigma_i(\bA_2)\bigr)} \notag\\
&\le  
 \sum_{i=1}^k \babs{\sigma_i(\bA_1)-\sigma_i(\bA_2)},
\label{eq:kyfan-bnd}
\end{align}
for any $\bA_1,\bA_2\in\reals^{k_1\times k_2}$ and
$k\in\bigl\{1,\min\{k_1,k_2\}\bigr\}$,
with  $\sigma_1(\cdot)\ge \cdots \ge
\sigma_{\min\{k_1,k_2\}}(\cdot)$ denoting
the ordered singular values of its (matrix) argument.\footnote{Note that
  \eqref{eq:kyfan-bnd}, in turn, means that, more generally,
  \lemref{lem:kyfan-stab} in \appref{app:sval-est-tail} also
  quantifies the stability of Ky Fan $k$-norms.}

And still further consequences of 
\propref{prop:sval-est-tail} and \corolref{corol:sval-est-mse} are
that
\begin{equation*}
\prob{\sum_{i=1}^k \bigl(\sigmah_i - \sigma_i\bigr)^2
 \ge \delta^2} 
\le \expop{\frac14 - \frac{p_0^2\,\delta^2 n}{8k}}
\end{equation*}
and
\begin{equation*}
\E{\sum_{i=1}^k \bigl(\sigmah_i-\sigma_i\bigr)^2} \le
\frac{6k+8k\ln(nk)}{p_0^2n},
\end{equation*}
respectively, which follow from the standard norm inequality
\begin{equation}
\bnorm{a^k} \le \bnorm{a^k}_1\defeq\sum_{i=1} |a_i|,\qquad\text{any $k$
  and $a^k$}.
\label{eq:l2-l1-bnd}
\end{equation}

Finally, for $\eps$-dependent $X$ and $Y$, variables
$X^{(k)},Y^{(k)}$ defined via \eqref{eq:Pxyk-exp} have mutual
information $I(X^{(k)};Y^{(k)})$ given by
\eqref{eq:PIk-def}.   Accordingly, a natural estimate of this mutual
information is
\begin{equation*}
\Ih(X^{(k)};Y^{(k)}) \defeq \frac12 \sum_{i=1}^k \sigmah_i^2,
\end{equation*}
for which the error is
\begin{equation}
\Ih(X^{(k)};Y^{(k)}) - I(X^{(k)};Y^{(k)})
= \frac12 \sum_{i=1}^k \bigl(\sigmah_i^2 - \sigmah_i^2\bigr) + o(\eps^2).
\label{eq:I-est-error}
\end{equation}
The results of \propref{prop:sval-est-tail} and
\corolref{corol:sval-est-mse} can be used to bound the error
\eqref{eq:I-est-error}; specifically, we have the following corollary,
whose proof is provided in \appref{app:I-est-err}.
\begin{corollary}
\label{corol:I-est-err}
Suppose $P_{X,Y}\in\simpXY$ is such that \eqref{eq:marg-bnd} is satisfied
for some $p_0>0$, and let $\sigmah_i$ for $i=1,\dots,K$ be 
defined via \eqref{eq:modal-emp} with $\Ph_{X,Y}$ denoting the
empirical distribution based on $n$ training samples.  Then 
for any $k\in\{1,\dots,K\}$  and $0\le\delta\le
\sqrt{k/2}/p_0^2$,
\begin{equation}
\prob{\bbabs{\frac12\sum_{i=1}^k \bigl(\sigmah_i^2 - \sigma_i^2\bigr)} \ge \delta} 
\le \expop{\frac14 - \frac{p_0^4\,\delta^2 n}{8k}},
\label{eq:I-est-tail}
\end{equation}
and, for $n$ such that $n\ge 16\ln(4kn)$,
\begin{equation}
\E{\bbabs{\frac12 \sum_{i=1}^k \bigl(\sigmah_i^2-\sigma_i^2\bigr)}^2} \le
\frac{6k+8k\ln(nk)}{p_0^4n}.
\label{eq:I-est-mse}
\end{equation}
\end{corollary}

More generally, it should be emphasized that, as the proofs of
\propref{prop:sval-est-tail}, \corolref{corol:sval-est-mse}, and
\corolref{corol:I-est-err} reveal,
their results hold not only for the $k$ dominant singular values
$\sigmah_i$ and $\sigma_i$, but for arbitrary (corresponding) subsets
of $k$ singular values.

\subsubsection{Sample Complexity of Feature Estimates}
\label{sec:sampcomp-features}

In this section, we determine the number of samples required to obtain
accurate estimates 
\begin{equation*}
\fc^k_* = \bigl(\fc_1^*,\dots,\fc_k^*\bigr) \quad\text{and}\quad
\gc^k_* = \bigl(\gc_1^*,\dots,\gc_k^*\bigr)
\end{equation*}
of the features $f^k_*$ and $g^k_*$, respectively, for
$k\in\{1,\dots,K-1\}$.  Our development focuses on measuring the
accuracy of these estimates by the extent to which they preserve as
much of the mutual information between $X$ and $Y$ as possible, in the
local sense, corresponding to $\sigma_1^2+\cdots+\sigma_k^2$.
Specifically, we measure this via\footnote{To avoid unnecessarily
  cumbersome expressions, we have left implicit the conditioning on
  $\smash{\fc^k_*(\cdot)}$ in the expectation in \eqref{eq:fq-meas}.}
\begin{align} 
&\mu_2^k(P_{X,Y},\Ph_{X,Y}) \notag\\
&\qquad\defeq \Ed{P_Y}{\bnorm{\bEd{P_{X|Y}}{f^k_*(X)}}^2 -
    \bnorm{\bEd{P_{X|Y}}{\fc^k_*(X)}}^2}.
\label{eq:fq-meas}
\end{align}
To facilitate interpretation of the measure \eqref{eq:fq-meas}, note that since
\begin{equation}
\bfrob{\dtmt\,\bfVX}^2 = \Ed{P_Y}{\bnorm{\bEd{P_{X|Y}}{f^k(X)}}^2}
\label{eq:ce-frob-rel}
\end{equation}
with $\bfVX$ denoting the collection of feature vectors associated with
$f^k$ as defined in \eqref{eq:bfVX-k-def}, we have
\begin{align*} 
&\max_{f^k\in\cF_k}
\Ed{P_Y}{\bnorm{\bEd{P_{X|Y}}{f^k(X)}}^2} \notag\\
&\qquad\qquad\qquad= 
\Ed{P_Y}{\bnorm{\bEd{P_{X|Y}}{f^k_*(X)}}^2} = \sum_{i=1}^k \sigma_i^2,
\end{align*}
where $\cF_k$ is as defined in \eqref{eq:cF-def}.

We begin with the following tail probability bound, a
proof of which is  provided in
\appref{app:svec-est-tail}.
\begin{proposition}
\label{prop:svec-est-tail}
Let $P_{X,Y}\in\simpXY$ be such that \eqref{eq:marg-bnd} is satisfied
for some $p_0>0$.  Then for $k\in\{1,\dots,K\}$ and $0\le\delta\le
4k$,\footnote{With slight abuse of notation, we use
  $\smash{\probd{\smash{\fc^k_*}}{\cdot}}$ 
to denote probability with respect to the
  distribution governing the random map \smash{$\fc^k_*(\cdot)$}.}
\begin{align} 
&\probd{\fc^k_*}{\Ed{P_Y}{\bnorm{\bEd{P_{X|Y}}{f^k_*(X)}}^2 - \bnorm{\bEd{P_{X|Y}}{\fc^k_*(X)}}^2}  \ge \delta}
  \notag\\
&\ \quad\qquad\qquad\qquad\qquad\qquad 
  \le \bigl(\cardX+\cardY\bigr) \expop{- \frac{p_0\,
  \delta^2 n}{64\,k^2}},
\label{eq:svec-est-tail}
\end{align}
where $\fc_i^*$ for $i=1,\dots,K-1$ are defined via \eqref{eq:fc-def}
and \eqref{eq:modal-emp}, with $\Ph_{X,Y}$ denoting the empirical
distribution based on $n$ training samples.
\end{proposition}

Note that by symmetry, it also follows immediately from
\propref{prop:svec-est-tail} that
\begin{align} 
&\probd{\gc^k_*}{\Ed{P_X}{\bnorm{\bEd{P_{Y|X}}{g^k_*(Y)}}^2 - \bnorm{\bEd{P_{Y|X}}{\gc^k_*(Y)}}^2}  \ge \delta}
  \notag\\
&\ \quad\qquad\qquad\qquad\qquad\qquad 
  \le \bigl(\cardX+\cardY\bigr) \expop{- \frac{p_0\,
  \delta^2 n}{64\,k^2}},
\end{align}
where, analogously, $\gc_i^*$ for $i=1,\dots,K-1$ are defined via
\eqref{eq:gc-def} and \eqref{eq:modal-emp}.

A key consequence of \propref{prop:svec-est-tail} is the following
corollary, whose proof, provided in \appref{app:svec-est-mse}, makes
use of \lemref{lem:varphi-ab}.
\begin{corollary}
\label{corol:svec-est-mse}
Let $P_{X,Y}\in\simpXY$ be such that \eqref{eq:marg-bnd} is satisfied
for some $p_0>0$.  Then for $k\in\{1,\dots,K\}$ and $n$ sufficiently
large that 
\begin{subequations} 
\begin{align} 
\frac{p_0 n}{64} &\ge \frac1{\bigl(\cardX+\cardY\bigr)} \label{eq:ss-a}\\
\intertext{and}
\frac{p_0 n}{4} &\ge \ln\left(\frac{p_0 n}{64}
\bigl(\cardX+\cardY\bigr)\right), \label{eq:ss-b}
\end{align}
\label{eq:ss-constraint}%
\end{subequations}
we have
\begin{align} 
&\Ed{\fc^k_*}{\left(\Ed{P_Y}{\bnorm{\bEd{P_{X|Y}}{f^k_*(X)}}^2 -
      \bnorm{\bEd{P_{X|Y}}{\fc^k_*(X)}}^2}\right)^2} \notag\\
&\qquad\qquad\le \frac{64 k^2 \Bigl( \ln \bigl[ p_0 n
    \bigl(\cardX+\cardY\bigr)\bigr] - 3 \Bigr)}{p_0 n},
\label{eq:svec-est-mse}
\end{align}
where $\fc_i^*$ for $i=1,\dots,K-1$ are defined  via \eqref{eq:fc-def} and
\eqref{eq:modal-emp}, with $\Ph_{X,Y}$ denoting the empirical
distribution based on $n$ training samples, and
where (with slight abuse of notation) we use $\bEd{\fc^k_*}{\cdot}$ to
denote expectation with respect to the distribution governing the
random map \smash{$\fc^k_*$}.
\end{corollary}

\subsubsection{Complementary Sample Complexity Bounds}
\label{sec:sampcomp-compl}

In \propref{prop:sval-est-tail}, we bound the sample complexity of
maximal correlation estimates via a Frobenius norm bound, while in
\propref{prop:svec-est-tail}, we bound the sample complexity of
feature estimates via a spectral norm bound.  However, we may
interchange these analyses, using spectral norm bounds to analyze
sample complexity of maximal correlation estimates, and Frobenius norm
bounds to analyze the sample complexity of feature estimates.  

In this way, we obtain complementary results.  In particular, we have
the following alternative bound on the sample complexity of maximal
correlation estimates, a proof of which
is provided in \appref{app:sval-est-tail-alt}.
\begin{proposition}
\label{prop:sval-est-tail-alt}
For $P_{X,Y}\in\simpXY$, let $p_0>0$ be such that \eqref{eq:marg-bnd}
is satisfied.  Then for $k\in\{1,\dots,K\}$ and $0\le\delta\le
k$,
\begin{equation}
\prob{\sum_{i=1}^k \babs{\sigmah_i - \sigma_i} \ge \delta} 
\le \bigl(\cardX+\cardY\bigr) \expop{-\frac{p_0\, \delta^2
  n}{4 k^2}},
\label{eq:sval-est-tail-alt}
\end{equation}
where $\sigmah_i$ for $i=1,\dots,K$ are defined via
\eqref{eq:modal-emp} with $\Ph_{X,Y}$ denoting the empirical
distribution based on $n$ training samples.
\end{proposition}
Moreover, \propref{prop:sval-est-tail-alt} can be used to obtain an
alternative version of \corolref{corol:sval-est-mse}.

Likewise, we have the following alternative bound on the sample
complexity of feature estimates, a proof of which is provided in
\appref{app:svec-est-tail-alt}. 
\begin{proposition}
\label{prop:svec-est-tail-alt}
Let $P_{X,Y}\in\simpXY$ be such that \eqref{eq:marg-bnd} is satisfied
for some $p_0>0$.  Then for  $k\in\{1,\dots,K\}$ and $0\le\delta\le
(4/p_0)/\sqrt{k/2}$, 
\begin{align} 
&\probd{\fc^k_*}{\Ed{P_Y}{\bnorm{\bEd{P_{X|Y}}{f^k_*(X)}}^2 - \bnorm{\bEd{P_{X|Y}}{\fc^k_*(X)}}^2}  \ge \delta}
  \notag\\
&\ \quad\qquad\qquad\qquad\qquad\qquad 
  \le \expop{\frac14 - \frac{p_0^2\, \delta^2 n}{128k}},
\label{eq:svec-est-tail-alt}
\end{align}
where $\fc_i^*$ for $i=1,\dots,K-1$ are defined via
\eqref{eq:fc-def} and \eqref{eq:modal-emp}, with $\Ph_{X,Y}$ denoting
the empirical distribution based on $n$ training samples.
\end{proposition}
\propref{prop:svec-est-tail-alt} can similarly be used to obtain an
alternative version of \corolref{corol:svec-est-mse}.

Comparing \eqref{eq:sval-est-tail-alt} to \eqref{eq:sval-est-tail} for
maximum correlation estimates, and \eqref{eq:svec-est-tail-alt} to
\eqref{eq:svec-est-tail} for feature estimates, we see that the
alternative bounds depend on the parameters in different ways, and
apply in different regimes.  As such, each may be better than the
other in different regimes.

\subsubsection{A Related Measure of Feature Quality}
\label{sec:fq-alt}

A natural measure of feature quality closely related to that defined
in \eqref{eq:fq-meas} is
\begin{align} 
&\mu_{2'}^k(P_{X,Y},\Ph_{X,Y}) \notag\\
&\qquad\defeq
\bfrob{\bE{f^k_*(X)\,g^k_*(Y)^\T}-\bE{\fc^k_*(X)\,\gc^k_*(Y)^\T}}.
\label{eq:fq-meas-alt}
\end{align}
See \appref{app:fq-alt} for an analysis of \eqref{eq:fq-meas-alt}, which
establishes that sample complexity bounds very similar to those
for \eqref{eq:fq-meas} apply.

\subsubsection{Sample Complexity Error Exponent Analysis}
\label{sec:errexp}

Further sample complexity results can be obtained in the limit
$n\to\infty$ via large deviations analysis, complementing the results
of Sections~\ref{sec:sampcomp-maxcorr}--\ref{sec:sampcomp-compl}.

In this analysis, for a given $\Ph_{X,Y}$ we focus on the empirical
DTM $\dtmh$ whose $(y,x)$\/th entry is\footnote{By contrast, in the
  preceding sections the analysis focused on the quasi-CDM defined via
  \eqref{eq:dtmhs-def}, which uses true instead of empirical marginals
  and removes the zeroth mode, resulting in the decomposition
  \eqref{eq:modal-emp}.  However, the re-use of notation is
  convenient.  Also, more generally 
  $\dtmhs(x,y)\defeq0$ for all $(x,y)\in\X\times\Y$ such that $\Ph_X(x)=0$
  or $\Ph_Y(y)=0$, consistent with
  \eqref{eq:dtms-extend}.}
\begin{equation}
\dtmhs(x,y) \defeq 
\frac{\Ph_{X,Y}(x,y)}{\sqrt{\Ph_X(x)}\sqrt{\Ph_Y(y)}},
\label{eq:dtmhs-alt-def}
\end{equation}
for which $\fh_i^*$, $\gh_i^*$, and $\sigmah_i$ for
$i=1,\dots,K$ are defined via the modal decomposition
\begin{equation} 
\Ph_{X,Y}(x,y) = \Ph_X(x)\,\Ph_Y(y)
\left[ 1 + \sum_{i=1}^{K-1} \sigmah_i \, \fh_i^*(x)\,\gh_i^*(y) \right],
\label{eq:modal-emp-alt}
\end{equation}
where $\sigmah_1\ge \cdots\ge \sigmah_{K-1}\ge0$ and
$\bEd{\Ph_X}{\fh_i^*(X)\,\fh_j^*(X)} =
\bEd{\Ph_Y}{\gh_i^*(Y)\,\gh_j^*(Y)}=\kron_{i=j}$ for $i,j\in\{1,\dots,K-1\}$.
Eq.~\eqref{eq:modal-emp-alt}
corresponds to the singular value decomposition 
\begin{align} 
\dtmh &= \sum_{i=0}^{K-1} \sigmah_i \bpsih^Y_i \bigl(\bpsih^X_i\bigr)^\T,
\label{eq:dtmh-svd-alt}
\end{align}
where for $i=1,\dots,K-1$ the
singular vectors in \eqref{eq:dtmh-svd-alt} are related to the feature
vectors in \eqref{eq:modal-emp-alt} as in \eqref{eq:psih-def}, and 
where [cf.\ \eqref{eq:sv0-def}] 
\begin{equation*}
\sigmah_0=1,\qquad \bpsih^X_0(x)=\sqrt{\Ph_X(x)},\qquad
\bpsih^Y_0(y)=\sqrt{\Ph_Y(y)}.
\end{equation*}

Motivated by the analyses in the previous sections, 
we consider the following neighborhoods of
$P_{X,Y}$:
\begin{subequations} 
\begin{align}
&\nbhdsf_\delta(P_{X,Y}) \defeq \left\{\Ph_{X,Y} \in \simpXY \colon
\bfrob{\dtmh - \dtm} \le \delta\right\} \label{eq:nbhdsf-def}\\ 
&\nbhdss_\delta(P_{X,Y}) \defeq \left\{\Ph_{X,Y} \in \simpXY \colon
\bspectral{\dtmh - \dtm} \le \delta\right\} \label{eq:nbhdss-def}\\ 
&\begin{aligned}
\nbhds^k_\delta(P_{X,Y}) \defeq &\Bigl\{\Ph_{X,Y} \in \simpXY \colon\\
&\qquad\qquad\bbabs{\bfrob{\dtm \bPsi^X_{(k)}}^2-\bfrob{\dtm
    \bPsih^X_{(k)}}^2} \le \delta\Bigr\}, \\
&\qquad\qquad\qquad\qquad\qquad\qquad1\le k \le K-1, 
\end{aligned}%
\label{eq:nbhdsk-def}%
\end{align}%
\label{eq:nbhds-defs}%
\end{subequations}
for any $\delta > 0$, where $\dtmh$ is the DTM corresponding to
$\Ph_{X,Y}$.  We denote the
$k$ dominant singular vectors using $\bpsih^X_i$, and $\bpsih^Y_i$,
and define
\begin{equation} 
\bPsih^X_{(k)} \defeq \begin{bmatrix} \bpsih^X_1 & \cdots
  & \bpsih^X_k \end{bmatrix} \ \ \text{and}\ \ 
\bPsih^Y_{(k)} \defeq \begin{bmatrix} \bpsih^Y_1 & \cdots
  & \bpsih^Y_k \end{bmatrix}.
\label{eq:bPsih-def}
\end{equation}

Our main result relates the error exponents for \eqref{eq:nbhdsk-def}
to those of \eqref{eq:nbhdsf-def} and \eqref{eq:nbhdss-def}; a proof
is provided in \appref{app:eststab}.
\begin{proposition}
\label{prop:eststab}
For any $P_{X,Y}\in\relint(\simpXY)$, any 
$0<\delta<\bmin(P_{X,Y})$ with
\begin{equation}
\bmin(P_{X,Y}) \defeq \min_{x\in\X, y\in\Y} \dtms(x,y)>0,
\label{eq:bmin-def}
\end{equation}
and
$\nbhdsf_\delta(P_{X,Y})$, $\nbhdss_\delta(P_{X,Y})$, and
$\nbhds^k_\delta(P_{X,Y})$ as defined 
in \eqref{eq:nbhds-defs},
\begin{subequations} 
\begin{gather}
E\bigl(\nbhds^k_{4 \delta\sqrt{k} }(P_{X,Y})\bigr) 
\ge \uE\bigl(\nbhdsf_{\delta}(P_{X,Y})\bigr) = E_*\bigl(\nbhdsf_{\delta}(P_{X,Y})\bigr) \label{eq:eststab-F}\\
\begin{aligned}
E\bigl(\nbhds^k_{4 \delta k}(P_{X,Y})\bigr) 
\ge \uE\bigl(\nbhdss_{\delta}(P_{X,Y})\bigr) 
&= E_*\bigl(\nbhdss_{\delta}(P_{X,Y})\bigr) \\
&\ge
E_*\bigl(\nbhdsf_{\delta}(P_{X,Y})\bigr), 
\end{aligned}
\label{eq:eststab-s}
\end{gather}
\label{eq:eststab}%
\end{subequations}
for $k\in\{1,\dots,K-1\}$, where for $\nbhds\subseteq\simpXY$,
\begin{align} 
E(\nbhds) &\defeq
-\limsup_{n\to\infty} \frac1n\log\prob{\Ph_{X,Y}
  \in \simpXY \backslash \nbhds} \label{eq:errexp-def}\\
E_*(\nbhds) &\defeq
-\lim_{n\to\infty} \frac1n\log\prob{\Ph_{X,Y}
  \in \simpXY \backslash \nbhds}\\
\uE(\nbhds) &\defeq \inf_{\Ph_{X,Y} \in
  \simpXY\backslash\nbhds} D(\Ph_{X,Y}\|P_{X,Y}).
\label{eq:uE-def}
\end{align}
\end{proposition}
\propref{prop:eststab} quantifies, e.g., the relative difficulties of
achieving small
\begin{equation*}
\sum_{x\in\X,y\in\Y} \bigl(\dtmhs(x,y)-\dtms(x,y)\bigr)^2,
\end{equation*}
which corresponds to \eqref{eq:nbhdsf-def}, small\footnote{This is the
  special ($k=1$) case of expressing $\kyfank{\dtm-\dtmh}$ in the form
\begin{equation*}
\max_{(f^k,g^k)\in\cF_k\times\cG_k} \Bigl[
\bE{f^k(X)^\T g^k(Y)}-\bEd{\Ph_{X,Y}}{f^k(X)^\T g^k(Y)} \Bigr].
\end{equation*}
}
\begin{equation*}
\max_{(f,g)\in\cF_1\times\cG_1} \Bigl(
\bE{f(X)\,g(Y)}-\bEd{\Ph_{X,Y}}{f(X)\,g(Y)} \Bigr),
\end{equation*}
which corresponds to \eqref{eq:nbhdss-def} when $\Ph_X=P_X$ and
$\Ph_Y=P_Y$ (and which is a good approximation for moderately large $n$),
and small
\begin{equation*}
\bbabs{\Ed{P_Y}{\bnorm{\bEd{P_{X|Y}}{f^k_*(X)}}^2 - \bnorm{\bEd{P_{X|Y}}{\fh^k_*(X)}}^2}}
\end{equation*}
which corresponds to \eqref{eq:nbhdsk-def}---and which is closely
related to achieving small
\begin{equation*}
\bbfrob{\bE{f^k_*(X)\,g^k_*(Y)^\T}-\bE{\fh^k_*(X)\,\gh^k_*(Y)^\T}},
\end{equation*}
according to the discussion of \secref{sec:fq-alt}.

Finally, the following lemma, whose proof is provided in
\appref{app:chernoff-local}, provides a local characterization of
related Chernoff exponents.\footnote{For related local exponent
  analysis, see, e.g., \cite{hx19}.}
\begin{lemma}
\label{lem:chernoff-local}
For $P_Z\in\relint(\simpZ)$ and every $h \colon \Z \rightarrow \reals$
such that $\E{h(Z)} \ne 0$ and $\varop{h(Z)} > 0$, and with $\Ph_Z$ 
denoting the empirical distribution formed from $n$
i.i.d.\ samples of $P_Z$,
we have
\begin{multline}
\lim_{\gamma \to 0^+} \lim_{\vphantom{0^+} n \rightarrow \infty}
\frac{2}{\gamma^2 n} \log
\prob{\left|\frac{\bEd{\Ph_Z}{h(Z)}}{\E{h(Z)}} - 1\right| \ge
\gamma } \\
= -\frac{\bigl(\E{h(Z)}\bigr)^2}{\varop{h(Z)}}.
\label{eq:chernoff-local}
\end{multline}
\end{lemma}

We can apply \lemref{lem:chernoff-local} to $h(Z)=f(X)\,g(Y)$ with
$Z=(X,Y)$ for different choices of $f$ and $g$. In particular, it
follows immediately that for any $P_{X,Y}\in\relint(\simpXY)$, and any
$f$ and $g$ such that
\begin{align*} 
\bE{f(X)^2} = \bE{g(Y)^2} &= 1 \\
\bE{f(X)\,g(Y)} &\neq 0\\
 \bvarop{f(X)\,g(Y)} &> 0,
\end{align*}
we have
\begin{align} 
&-\lim_{\Delta\to0^+} \frac1{\Delta^2} \lim_{n\to\infty} \frac1n
\notag\\
&\qquad\qquad\qquad{}\cdot 
\log
\left[ \prob{\bbabs{\frac{\bEd{\Ph_{X,Y}}{f(X)\,g(Y)}}{\bE{f(X)\,g(Y)}} - 1} \ge
    \Delta} \right]\notag\\
&\qquad= \frac12 \, \frac{\bE{f(X)\,g(Y)}^2}{\bvarop{f(X)\,g(Y)}}.
\label{eq:chernoff-local-fg}
\end{align}

As one instance of \eqref{eq:chernoff-local-fg}, we can choose
$f=f_1^*$ and $g=g_1^*$ to quantify the sample complexity of
estimating $\sigma_1$.  As
  another, we can choose
\begin{equation*}
f(x)=\frac{\kron_{x=x_0}}{\sqrt{P_X(x)}}\quad\text{and}\quad
g(y)=\frac{\kron_{y=y_0}}{\sqrt{P_Y(y)}},
\end{equation*}
for some $(x_0,y_0)$, to quantify the sample
complexity of estimating the $(x_0,y_0)$\/th entry of $\dtm$, i.e.,
\begin{equation*}
\bE{f(X)\,g(Y)} =
\frac{P_{X,Y}(x_0,y_0)}{\sqrt{P_X(x_0)}\,\sqrt{P_Y(y_0)}} = \dtms(x_0,y_0).
\end{equation*}

\section{Collaborative Filtering and Matrix Factorization}
\label{sec:cf}

A variety of high-dimensional learning and inference problems can be
addressed within the preceding framework of analysis.  One example is
the design of recommender systems \cite{ms17} based on collaborative
filtering \cite{gnot92}.  These systems aim to predict the preferences
of individual users for various items from (limited) knowledge of some
of their and other users' item preferences, such as may be obtained
from ratings data and/or records of prior choices.  Among the most
successful forms of collaborative filtering to date have been matrix
factorization methods based on latent factor models and involving
low-rank approximation techniques \cite{srebro04,kbv09,rt11}, a subset of
which are formulated as matrix completion problems
\cite{cr09,kmo10,ct10,cp10} or variations thereof
\cite{gwlcm10,VaswaniBJN18}.  

In applying such methods, the system designer must (implicitly or
otherwise) choose: 1) how to model the available data expressing user
preferences; 2) what matrix representation to factor; and 3) a
criterion for evaluating the quality of candidate factorizations.  The
large literature in this area reflects the many choices available.  In
this section, we formulate the collaborative filtering problem as one
of Bayesian decision making involving multi-attributes, and show that
its solution corresponds to a matrix factorization method that differs
in some significant respects from existing ones.

\subsection{Bayesian Attribute Matching}
\label{sec:bama}

As a convenient context popularized in \cite{bl07}, consider the
content-provider problem of recommending movies to subscribers.  Let
$\X$ be the collection of subscribers, and let $\Y$ be the collection
of available movies.  In turn, $(X,Y)=(x,y)\in\X\times\Y$ denotes the
event that the next instant a movie is watched, it will be subscriber
$x$ watching movie $y$, and $P_{X,Y}(x,y)$ denotes the probability of
this event.

With this notation, the associated conditional $P_{Y|X}(y|x)$ denotes
the probability that if $x$ is the next subscriber to watch a movie,
he/she will select movie $y$.  From this perspective, the
recommendation problem can be interpreted as identifying values of $y$
for which this conditional probability is high for the given $x$, or
more generally sampling from $P_{Y|X}(\cdot|x)$.  Alternatively, if
one seeks to avoid biasing the recommendation according to $P_Y$ and
replace it with a uniform distribution, we sample from the distribution
proportional to $P_{X|Y}(x|\cdot)$
instead.

In practice, we must estimate $P_{X,Y}$ from data $(x_1,y_1),\dots,
(x_n,y_n)$, where $(x_j,y_j)$ is a record of subscriber $X=x_j$ having
selected movie $Y=y_j$ to watch at some point in the past.  In
particular, we treat these $n$ records as i.i.d.\ samples from
$P_{X,Y}$.  In the regime of interest, there are comparatively few
training samples $n$ relative to the joint alphabet size $\X \times
\Y$, so to obtain meaningful results the procedure for estimating
$P_{X,Y}$ must take this into account.  In the sequel, this is
accomplished by exploiting attribute variables, as we now develop.

In developing the key concepts, it is convenient to initially treat
$P_{X,Y}$ as known, then return to the scenario of interest in which
only the empirical distribution $\Ph_{X,Y}$ is available.

To begin, we view the multi-attribute variables $U^k$ and $V^k$ in the
Markov chain $U^k\markov X \markov Y \markov V^k$ obtained in
\secref{sec:uni-coop} (and \secref{sec:uni-ib}) as the dominant
attributes of subscribers and movies, respectively.  In turn the
corresponding $S^k_*$ and $T^k_*$, as defined in \eqref{eq:stmk*-def},
represent sufficient statistics for the detection of these attibutes.

Conceptually, for each movie $y$, there is an associated movie
multi-attribute $V^k(y)$ generated randomly from $y$ according to
$P_{V^k|Y}(\cdot|y)$, as defined via \eqref{eq:CYk-opt}, that
expresses its dominant characteristic.  Likewise, for the target
subscriber $x$, there is an associated movie multi-attribute
$V_\circ^k(x)$ generated randomly from $x$ according to
$P_{V^k|X}(\cdot|x)$, as defined via \eqref{eq:PXVi-opt}, that
expresses his/her preferred movie characteristic.  Having defined
these multi-attributes, we can express the recommendation problem as
one of Bayesian decision-making among multiple (not mutually
exclusive) hypotheses.  Specifically, $\cE_y(x)$ denotes the event
that there is an attribute match with movie $y$ for subscriber $x$,
i.e.,
\begin{equation}
\cE_y(x) \defeq \bigl\{ V^k(y)=V_\circ^k(x) \bigr\}.  
\label{eq:cEy-def}
\end{equation}

The following result characteries the movie recommendation rule that results
from maximizing the expected number of matches.  A proof is
provided in \appref{app:attribute-match}. 
\begin{proposition}
\label{prop:attribute-match}
Given $k\in\{1,\dots,K-1\}$, $l\in\{1,\dots,\cardY\}$,
$P_{X,Y}\in\relint(\simpXY)$, and a collection $\cYh(x)$ of $l$
(distinct)  movies for subscriber $x\in\X$, let the number
that are a match be
\begin{equation}
M \defeq \sum_{y\in\cYh(x)} \kron_{\cE_y(x)},
\label{eq:M-def}
\end{equation}
where $\cE_y(x)$ as defined in \eqref{eq:cEy-def}, with
the movie multi-attributes $V^k(y)$ (for
movie $y\in\Y$) and $V_\circ^k(x)$ (for user $x\in\X$) being independent and
distributed according to, respectively, $P_{V^k|Y}(\cdot|y)$ and
$P_{V^k|X}(\cdot|x)$ as defined via \propref{prop:mpe-coop} and
\corolref{corol:mpe-coop}, i.e., according to\footnote{Note that
  \eqref{eq:VVo-dist} implies, e.g., 1)
  $V_\circ^k\markov (X,Y)\markov V^k$; 2) $X\markov Y\markov V^k$; and
  3) $V_\circ^k\markov X\markov Y$.  These, in turn, imply, e.g., 
  $V_\circ^k\markov X\markov V^k$ and $V_\circ^k\markov X\markov V^k$.} 
\begin{align} 
&P_{V_\circ^k,V^k,X,Y}(v_\circ^k,v^k,x,y) \notag\\
&\qquad\qquad= P_{V^K|X}(v_\circ^k|x)\, P_{V^K|Y}(v^k|y)\, P_{X,Y}(x,y).
\label{eq:VVo-dist}
\end{align}
Then 
\begin{equation} 
\E{M} 
\le \frac1{2^k} \sum_{y\in\cYh^*(x)} 
\left( 1 + \eps^2 \sum_{i=1}^k \sigma_i \, f_i^*(x)\,
g_i^*(y) \right) + o(\eps^2),
\label{eq:exp-match-bnd}
\end{equation}
where 
\begin{equation}
\cYh^*(x) \defeq \bigl\{ y_1^*(x),\dots,y_l^*(x) \bigr\}
\end{equation}
is constructed sequentially according to 
\begin{subequations} 
\begin{align} 
y_1^*(x) &\defeq \argmax_{y\in\Y} 
\sum_{i=1}^k \sigma_i \,
f_i^*(x)\, g_i^*(y) \label{eq:ystar-disc-final-1}\\
y_j^*(x) &\defeq \!\!\smash[b]{\argmax_{y\in\Y\setminus\{y_1^*(x),\dots,y_{j-1}^*(x)\}}
\sum_{i=1}^k} \sigma_i \, f_i^*(x)\, g_i^*(y), \\
&\quad\qquad\qquad\qquad\qquad\qquad\qquad\qquad j=2,\dots,l. \notag
\end{align}%
\label{eq:ystar-disc-final}%
\end{subequations}
Moreover, the inequality in \eqref{eq:exp-match-bnd} holds with equality when we
choose $\cYh(x)=\cYh^*(x)$.
\end{proposition}

Note that in the case $l=1$, the criterion in
\propref{prop:attribute-match} specializes to the probability of a
decision error, which our result establishes is minimized by the use
of the \emph{maximum a posteriori} (MAP) decision rule, generalizing
the familiar result for Bayesian hypothesis testing.  More generally,
\propref{prop:attribute-match} establishes that we maximize the
expected number of matches in our list by an MAP \emph{list} decision
rule: we recommend to subscriber $x$ the $l$ movies having the $l$
highest probabilities of an attribute match.  Note, too, that using
\eqref{eq:gi-condexp} we can write the $k$-dimensional (weighted)
inner product that is the core computation in
\eqref{eq:ystar-disc-final} in the form
\begin{equation*}
\sum_{i=1}^k \sigma_i \, f_i^*(x)\, g_i^*(y)
= \sum_{i=1}^k g_i^*(y)\, \E{g_i^*(Y)\mid X=x},
\end{equation*}
which provides additional interpretation of the maximum inner product
decision rule.  

\subsection{Interpretation as Matrix Factorization}

To interpret Bayesian attribute matching as a form of matrix
factorization, we have the following result establishing the
decision rule as a maximum likelihood one based on a rank-reduced
approximation to $P_{X,Y}$.
\begin{corollary}
\label{corol:matrix-completion}
Given $k\in\{1,\dots,K-1\}$, $l\in\{1,\dots,\cardY\}$, and
$P_{X,Y}\in\relint(\simpXY)$,
the optimum recommendation list in \propref{prop:attribute-match} can
be expressed in the form
\begin{subequations} 
\begin{align} 
y^*_1(x) &= \argmax_{y\in\Y} P_{X|Y}^{(k)}(x|y) \\
y_j^*(x) &= \!\!\smash[b]{\argmax_{y\in\Y\setminus\{y_1^*(x),\dots,y_{j-1}^*(x)\}}}
P_{X|Y}^{(k)}(x|y), \\
&\quad\qquad\qquad\qquad\qquad\qquad\qquad\qquad j=2,\dots,l. \notag
\end{align}%
\end{subequations}
where
\begin{equation*}
P_{X|Y}^{(k)}(x|y) \defeq \frac{P_{X,Y}^{(k)}(x,y)}{P_Y(y)}.
\end{equation*}
with $P_{X,Y}^{(k)}$ as defined in \eqref{eq:Pxyk-exp}.
\end{corollary}

This corollary is an immediate consequence of the fact that the
objective function in \eqref{eq:ystar-disc-final} can be equivalently
written in terms of $P^{(k)}_{X,Y}$;
specifically,
\begin{equation*}
\sum_{i=1}^k \sigma_i \, f_i^*(x)\, g_i^*(y) =
\frac{P^{(k)}_{X,Y}(x,y)}{P_X(x)\,P_Y(y)} -1 .
\end{equation*}

Moreover, $P^{(k)}_{X,Y}$ is the distribution corresponding to
$\dtmt^{(k)}$, which is the rank-$k$ approximation to $\dtmt$ obtained
by retaining the dominant $k$ modes in the SVD of $\dtmt$.
Equivalently, $\dtmt^{(k)}$ is the rank-constrained approximation to
$\dtmt$ obtained by minimizing $\bfrob{\dtmt-\dtmt^{(k)}}$, which
follows from the well-known matrix approximation theorem
\cite{ey36}
\cite[Corollary~7.4.1.3(a) and Section~7.4.2]{hj12}
\cite[Theorem~2.4.8]{gvl12}:\footnote{As
  discussed in \cite{gws93}, the original version of this
  approximation theorem was actually due to Schmidt \cite{es07}.} 
\begin{lemma}[Eckart-Young] 
\label{lem:ey}
If $\bA$ and $\bAt$ are $k_1\times k_2$ matrices such that $\bA$ has
singular values 
$\sigma_1(\bA)\ge \cdots \ge \sigma_{\min\{k_1,k_2\}}(\bA)$ and
$\rank(\bAt)\le k$, then
\begin{equation} 
\bfrob{\bA-\bAt}^2 \ge \sum_{i=k+1}^{\min\{k_1,k_2\}} \sigma_i^2.
\end{equation}
\end{lemma}

\subsection{Collaborative Filtering Based on Attribute Matching}

The preceding results lead directly to a straightforward collaborative
filtering procedure.  In particular, given a history \eqref{eq:cT-def}
of $n$ prior movie selections by users, modeled as drawn i.i.d.\ from
$P_{X,Y}$, we form the empirical distribution $\Ph_{X,Y}$, and use
this distribution in \propref{prop:attribute-match} and
\corolref{corol:matrix-completion}.  Consistent with the discussion in
\secref{sec:modal-approx}, we focus on the case in
which $P_X$ and $P_Y$ can be accurately estimated, but
$P_{X,Y}$ cannot.

As such, we effectively obtain the dominant $k$ modes from the modal
decomposition \eqref{eq:modal-emp} using \algoref{alg:ace-comp-k} with
the empirical distribution,  then use the resulting $\sigmah_i$, $\fh^*_i$, and
$\gh^*_i$, for $i=1,\dots,k$, to form the score function
\begin{equation*}
\sum_{i=1}^k \sigmah_i\,\fh^*_i(x)\,\gh^*_i(y),
\end{equation*}
whose maxima over $y$ for a given subscriber $x$ produce the movie
recommendations.  As our analysis in
\secref{sec:modal-approx} reflects, we can expect the estimated modes to
be accurate provided $k$ is sufficiently small relative to
$n$. The complete procedure takes the form
of \algoref{alg:collab-filter}.

\begin{algorithm}[tbp]
\caption{Collaborative Filtering by Attribute-Matching}
\label{alg:collab-filter}
\begin{algorithmic}[]
\Require{Subscriber list $\X$, movie list $\Y$, selections history
  $\cT$, i.e., $\Ph_{X,Y}$, dimension 
  $k$, recommendation list size $l$, target subscriber $x$}\\ 
   1. Estimate $k$ modes of $P_{X,Y}$ from $\Ph_{X,Y}$ via ACE:\\
   \qquad\qquad $\sigmah_i$, $\fh_i(\cdot)$, $\gh_i(\cdot)$,\quad for $i=1,\dots,k$\\
   2. Initialize recommendation list: $\cYh^*=\varnothing$\\
   3. Initialize candidates list: $\cYb=\Y$
\For{$j=1,\dots,l$}\\
\quad 4a. $\ds y^* \gets \argmax_{y\in\cYb} \sum_{i=1}^k \sigmah_i\,
\fh_i^*(x)\,\gh_i^*(y)$ \\
\quad 4b. Update recommendation list: $\cYh^* \gets \cYh^* \cup \{y^*\}$\\
\quad 4c. Update candidates list: $\cYb \gets \cYb \setminus \{y^*\}$
\EndFor
\end{algorithmic}
\end{algorithm}

It is worth emphasizing that the attribute matching approach to
collaborative filtering dictates factoring $\dtmh$ as defined in
\eqref{eq:dtmh-def}, which differs from other approaches used in the
literature.  For example, popular alternatives 
 include factoring the
matrix representation $\bPh_{Y,X}$ of $\Ph_{X,Y}$ itself
\cite{kbv09}, and factoring the matrix representation for pointwise
mutual information \cite{ch90} (information
density \cite{hv93}), i.e., the matrix whose $(y,x)$\/th entry is
\begin{equation}
\log\frac{\Ph_{X,Y}(x,y)}{P_X(x)\,P_Y(y)},
\label{eq:pmi-def}
\end{equation}
as arises in natural language processing \cite{MikolovCCD13,lg14}.

\subsection{Extensions}

Finally, a natural alternative to the procedure of \corolref{corol:matrix-completion}
are the recommendations
\begin{subequations} 
\begin{align} 
y^*_1(x) &= \argmax_{y\in\Y} P_{Y|X}^{(k)}(y|x) \\
y_j^*(x) &= \!\!\smash[b]{\argmax_{y\in\Y\setminus\{y_1^*(x),\dots,y_{j-1}^*(x)\}}}
P_{Y|X}^{(k)}(y|x), \\
&\quad\qquad\qquad\qquad\qquad\qquad\qquad\qquad j=2,\dots,l. \notag
\end{align}%
\end{subequations}
where
\begin{equation*}
P_{Y|X}^{(k)}(y|x) \defeq \frac{P_{X,Y}^{(k)}(x,y)}{P_X(x)}.
\end{equation*}
Unlike that of \corolref{corol:matrix-completion}, this procedure
includes the effect of $P_Y$ in its recommendations.  It is obtained
by replacing \eqref{eq:M-def} with
\begin{equation*}
M \defeq \sum_{y\in\cYh(x)} P_Y(y)\,\kron_{\cE_y(x)},
\end{equation*}
and  analytically extending the local
analysis to $\eps=1$, i.e., relaxing the weak dependence constraint.

\section{Softmax Regression}
\label{sec:nn}

As we develop in this section, a further characterization of the
universal features in \propref{prop:modal} is as the optimizing
parameters in softmax regression (i.e., multinomial logistic
regression) in the weak-dependence regime.  Softmax regression
\cite{cmb06}, which originated with the introduction of logistic
regression by Cox \cite{drc58}, has proven to be an extraordinarily
useful classfication architecture in a wide range of practical
applications, and has well known approximation properties---see, e.g.,
\cite{gc89,hsw89,arb93}.  As such, viewing our results from this
perspective yields additional interpretations and insights.  More
generally, this form of regression can be expressed as an elementary
form of neural network, and thus its analysis is useful in relating
the preceding results to aspects of contemporary neural networks.

In softmax regression, for a class index $Y$ and $k$-dimensional
real-valued data $S$ we fit a posterior of the form
\begin{equation*} 
\Pt_{Y|S}^{g,\beta}(y|s) 
= P_Y(y)\, \e^{s^\T g(y) + \beta(y) - \alpha(s)},
\end{equation*}
to some $P_{S,Y}$ by choosing parameters $g$ and $\beta$.  Note that
$g$ is defined by\footnote{In this section, we omit the supercript
  notation that we previously used to explicity indicate the dimension of
  a multi-dimensional variable.}
$\nncardY$ parameter vectors $g(1),\dots,g(\nncardY)$, each of
dimension $k$.  Likewise, $\beta$ is defined by $\nncardY$ scalar
parameters $\beta(1),\dots,\beta(\nncardY)$.  

We characterize the optimizing softmax parameters in the weak-dependence
regime as follows; a proof is provided in \appref{app:softmax}.
\begin{proposition}
\label{prop:softmax}
Given  $P_{X,Y}\in\simpXY$ such that
$X,Y$ are $\eps$-dependent for some $\eps>0$, a
dimension $k$,  and $s=f(x)$ for some $f\colon\X\mapsto\reals^k$, let
\begin{equation*}
P_{S,Y}(s,y) = \sum_{\{x\colon f(x)=s \}} P_{X,Y}(x,y)
\end{equation*}
be the induced distribution, and let $P_S$ denote the
associated induced marginal, with $\bmu_S$ its mean, and $\bLa_S$ its
covariance, which we assume to be nonsingular.\footnote{It is
  comparatively straightforward to verify that when $\bLa_S$ is
  singular, the proposition holds provided we replace the inverse $\bLa_S^{-1}$
  wherever it appears with the Moore-Penrose pseudoinverse
  $\bLa_S^\mppi$ and add to $g_{*,S}(y)$ any
  $g_\varnothing(y)\in\nullspace(\bLa_S)$ such that
  $\bE{g_\varnothing(Y)}=\bzero$.}  Let 
\begin{align}
\cPt^\Y_s(P_Y) \defeq \Bigl\{ &P\in\simpY\colon \notag\\
&\smash[t]{P =\Pt_{Y|S}^{g,\beta}(y|s) 
\defeq P_Y(y)\, \e^{s^\T g(y) + \beta(y) - \alpha(s)}}\vphantom{P} \notag\\
&\text{for some $\beta\colon\Y\mapsto\reals$ and
  $g\colon\Y\mapsto\reals^k$}
\Bigr\}
\label{eq:discrim}
\end{align}
denote the exponential family with natural statistic $g(y)$ and
natural parameter $s\in\cS$,  where $\cS\defeq 
f(\X)$.  Then 
\begin{align} 
&\min_{\Pt_{Y|S}(\cdot|s)\in\cPt^\Y_s(P_Y)}
\sum_{s\in\cS} P_S(s)\, D\bigl(P_{Y|S}(\cdot|s)\bigm\|\Pt_{Y|S}(\cdot|s)\bigr)  \notag\\
&\qquad = I(Y;S) - \frac12\, \E{\bnorm{\bLa_S^{-1/2}\bigl(\bmu_{S|Y}(Y)-\bmu_S\bigr)}^2} + o(\eps^2)
\label{eq:D*-softmax}
\end{align}
with
\begin{equation}
\bmu_{S|Y}(y) \defeq \bE{S|Y=y},
\label{eq:bmu-S|Y-def}
\end{equation}
and is achieved by the parameters 
\begin{subequations} 
\begin{align}
g(y) &= g_{*,S}(y) \defeq \bLa_S^{-1} \left( \bmu_{S|Y}(y) - \bmu_S \right) + o(\eps) \label{eq:g*-softmax}\\
\beta(y) &= \beta_{*,S}(y) \defeq -\bmu_S^\T \, g_{*,S}(y) + o(\eps), \label{eq:bt*-softmax}
\end{align}
i.e., 
\begin{align} 
&\Pt_{Y|S}^*(y|s) \notag\\
&\ \propto P_Y(y) \, \expop{(s\!-\!\bmu_S)^\T \bLa_S^{-1}
  (\bmu_{S|Y}(y)\!-\!\bmu_S) } \bigl(1+o(1)\bigr).
\label{eq:Pt*-softmax}
\end{align}
\label{eq:*-softmax}%
\end{subequations}
\end{proposition}

\subsection{Relation to Gaussian Mixture Analysis}
\label{sec:mixture}

It is useful to note that the optimum posterior \eqref{eq:Pt*-softmax}
in \propref{prop:softmax} matches that for a Gaussian mixture in which
the components depend weakly on the class index, despite the fact that
there are no Gaussian assumptions in the proposition.  In particular,
suppose that $P_{S|Y}(\cdot|y)=\gauss(\bmu_{S|Y}(y),\bLa_{S|Y})$,
where $\bLa_{S|Y}$ is positive definite and, as the notation reflects,
does not depend on $y$, and where $\bmu_{S|Y}(y) \defeq \bmu_S +
\eps\,\bfe(y)$ with $\bE{\bfe(Y)}=\bzero$ and $\eps>0$.  Then
\begin{align} 
&P_{Y|S}(y|s) \notag\\
&\ \propto P_Y(y)\,P_{S|Y}(s|y) \notag\\
&\ \propto P_Y(y)\, \expop{-\frac12 (s-\bmu_{S|Y}(y))^\T \bLa_{S|Y}^{-1}
  (s-\bmu_{S|Y})} \notag\\
&\ = P_Y(y)\, \exp \biggl\{-\frac12 \Bigl[(s-\bmu_S)^\T\bLa_{S|Y}^{-1}
  (s-\bmu_S) \notag\\
&\qquad\qquad\qquad +2\,(s-\bmu_{S|Y}(y))^\T\bLa_{S|Y}^{-1}
    (\bmu_S-\bmu_{S|Y}(y)) 
  \notag\\
&\qquad\qquad\qquad 
  + (\bmu_S-\bmu_{S|Y}(y))^\T \bLa_{S|Y}^{-1}
  (\bmu_S-\bmu_{S|Y}(y))\Bigr]\biggr\} \label{eq:exp-three}\\
&\ \propto P_Y(y)\, \expop{
  (s-\bmu_{S|Y}(y))^\T\bLa_{S|Y}^{-1} (\bmu_{S|Y}(y)-\bmu_S)}\notag\\
  &\qquad\qquad\qquad\qquad\qquad\qquad\qquad\qquad\qquad\qquad{}\cdot(1+o(1))
\label{eq:mixture-weak} \\
&\ = P_Y(y)\, \expop{
  (s\!-\!\bmu_S)^\T\bLa_S^{-1} (\bmu_{S|Y}(y)\!-\!\bmu_S)}(1+o(1)),
\label{eq:mixture-weak-b}
\end{align}
as $\eps\to0$,
where to obtain \eqref{eq:exp-three} we have used the simple expansion
\begin{equation*}
s-\bmu_{S|Y}(y) = (s-\bmu_S) + (\bmu_S-\bmu_{S|Y}(y)),
\end{equation*}
and to obtain \eqref{eq:mixture-weak} we have used that in the
exponent in \eqref{eq:exp-three} the first term does not depend on
$y$, the second term is $O(\eps)$, and the third term is $o(\eps)$.
To obtain \eqref{eq:mixture-weak-b} we have used that
$\bmu_{S|Y}-\bmu_S$ and $\bLa_S^{-1}-\bLa_{S|Y}^{-1}$ are both $o(1)$.
Hence, \eqref{eq:mixture-weak-b} and \eqref{eq:Pt*-softmax} match, to
first order.

\subsection{Optimum Feature Design}

\propref{prop:softmax} describes the optimizing softmax weights $g$
and biases $\beta$ for a given choice of $f$.  When we further
optimize over the choice of $f$, we obtain a direct relation to the
modal decomposition of \propref{prop:modal} and the universal posterior
\eqref{eq:Y|X-expfam}.  In particular, we have the following result, a
proof of which is provided in \appref{app:softmax-corol}.
\begin{corollary}
\label{corol:softmax}
Given dimension $k\in\{1,\dots,K-1\}$, if $P_{X,Y}$ is such that $f^k_*$ as defined in
\eqref{eq:fgk*-def} is injective (i.e., a one-to-one function), then
\begin{align} 
\smash[b]{\min_{\substack{\{\text{injective $f$}\colon\\ \qquad
      s=f(x)\}}}} \ 
\sum_{s\in\cS} P_S(s)\, 
  &D\bigl(P_{Y|S}(\cdot|s)\bigm\|\Pt_{Y|S}^*(\cdot|s)\bigr) \notag\\
&\quad\qquad = \frac12 \sum_{i=k+1}^K \sigma_i^2 + o(\eps^2),
\label{eq:D**-softmax}
\end{align}
and is achieved by 
\begin{equation}
f_i(x)=f_i^*(x),\quad x\in\X,\quad i=1,\dots,k,
\end{equation}
with $f_i^*$ as defined in \propref{prop:modal} .
Moreover, for this choice of $f$, the
parameters 
\begin{equation*} 
g_{*,S}(y)=\bigl(g_1^{*,S}(y),\dots,g_k^{*,S}(y)\bigr)
\quad\text{and}\quad \beta_{*,S}(y)
\end{equation*}
in \propref{prop:softmax} take the form
\begin{subequations} 
\begin{align} 
g_i^{*,S}(y) &= \sigma_i \, g_i^*(y),\quad y\in\Y,\quad i=1,\dots,k,
\label{eq:g**-softmax}\\
\beta_{*,S} &= 0 \label{eq:bt**-softmax}
\end{align}
\end{subequations}
where $g_i^*(y)$ and $\sigma_i$ are as defined in
\propref{prop:modal}, and thus
\begin{align} 
&\Pt^*_{Y|S_*}(y|f^k_*(x)) \notag\\
&\qquad\qquad\propto P_Y(y) \, \expop{\sum_{i=1}^k \sigma_i
  f^*_i(x)\, g^*_i(y)}\bigl(1 + o(1)\bigr).
\end{align}
\end{corollary}

\subsection{Neural Network Insights}

Since softmax regression can be interpreted as a simple neural network
classifier with a single hidden layer \cite{mp69,gbc17}, our results
can be equivalently expressed in terms of the optimization of such
networks.  Moreover, with the interpretation of universal features as
a solution to a local information bottleneck as developed in
\secref{sec:uni-ib}, these results shed insight into recent analyses
of deep learning based on such bottlenecks
\cite{tz15a,tz15b,afdm16,szt17,as17} as well as related
information-theoretic analyses \cite{mrrg96}.

\begin{figure*}[tbp]
\centering
\includegraphics[width=0.75\textwidth]{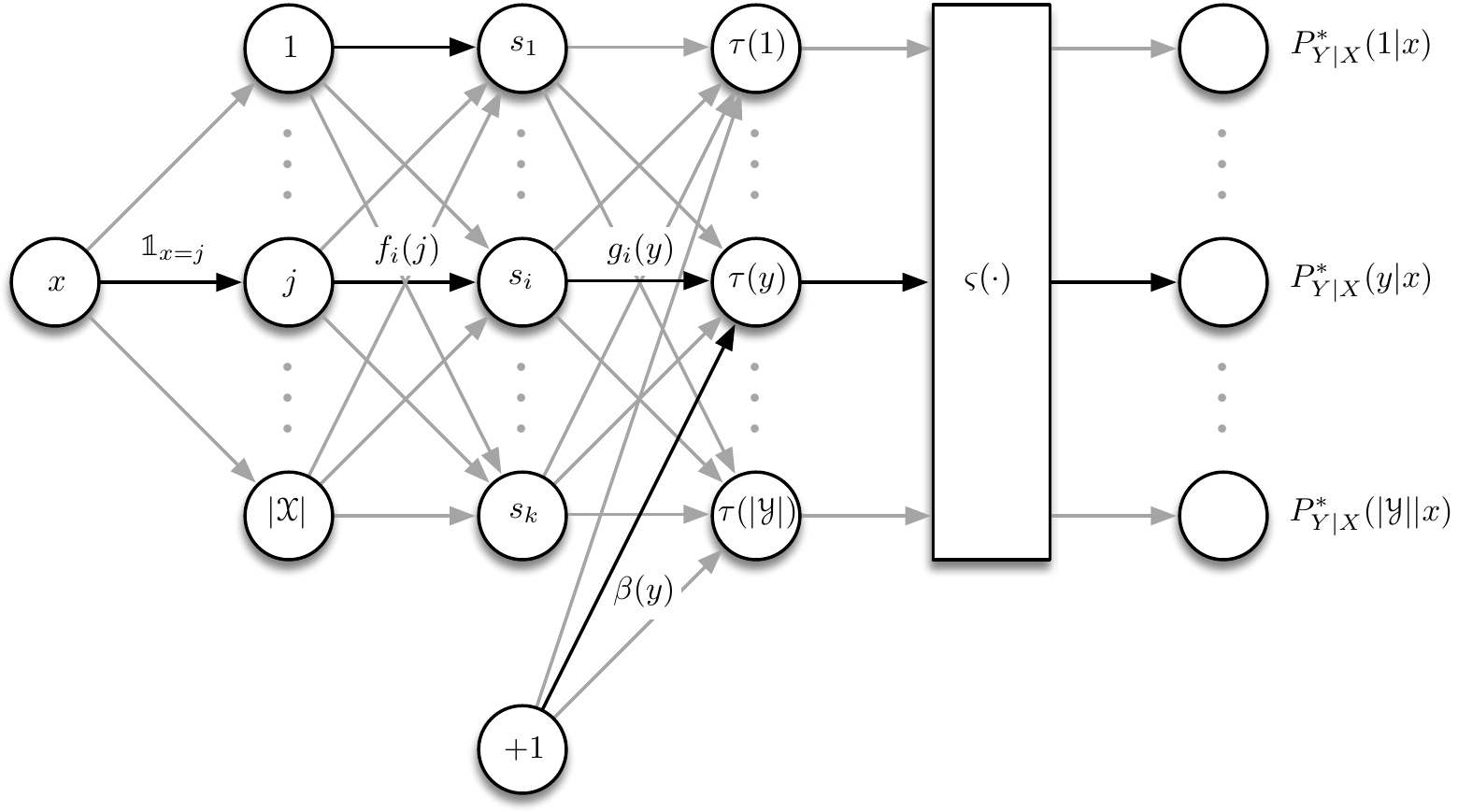}
\caption{A neural network representation of our softmax regression
  framework.  In this network, we use a one-hot representation of the
  input $x$, corresponding to the kronecker weights $\kron_{x=j}$.
  The hidden layer is characterized by the feature weights $f_i(j)$
  for $i\in\{1,\dots,k\},\ j\in\{1,\dots,\nncardX\}$, and the output
  layer is parameterized by the weights $g_i(y)$ and biases
  $\beta(y)$, for $i\in\{1,\dots,k\},\ y\in\{1,\dots,\nncardY\}$.  The
  the softmax processing, as defined in \eqref{eq:softmax}, is
  represented by the sigmoid function $\sigmoid(\cdot)$ and operates
  on the unnormalized log-posterior $\tau(y)$.
\label{fig:nn}}
\end{figure*}

The neural network architecture associated with softmax analysis is,
in our notation, as depicted in \figref{fig:nn}.  The input layer
uses a so-called ``one-hot'' representation of the input $x$,
corresponding to weights $\kron_{x=j}$.  Next, in the hidden layer,
features $s_i=f_i(x)$ of the input $x$ are generated using weights
$f_i(j)$.  Finally, in the output layer, the (unnormalized) log-posterior
$\tau(y)$ is constructed according to
\begin{equation*} 
\tau(y) = \sum_{i=1}^k s_i\, g_i(y) + \beta(y), 
\qquad y=1,\dots,\nncardY
\end{equation*}
using output layer weights $g_i(y)$ and biases $\beta(y)$.  The $\tau(y)$ are
then combined and normalized to produce the posterior via the softmax
processing 
\begin{subequations} 
\begin{equation} 
P^*_{Y|X}(y|x) 
= \frac{\e^{\tau(y)}}{\ds\sum_{y'=1}^{\nncardY}
  \e^{\tau(y')}}
= \sigmoid\left( -\ln \sum_{y'\ne y} \e^{\tau(y')-\tau(y)} \right),
\end{equation}
where 
\begin{equation}
\sigmoid(\omega) \defeq \frac{1}{1+\e^{-\omega}}
\label{eq:sigmoid-def}
\end{equation}
\label{eq:softmax}%
\end{subequations}
is the sigmoid function \cite{gbc17}.

For such networks and their multi-layer generalizations, the
optimization of the weights and biases is typically carried out using
stochastic gradient descent (SGD) \cite{gbc17}.  As such,
\propref{prop:softmax} implies that SGD is effectively computing
empirical conditional expectations, and as such corresponds to an
approximation to one step of the ACE algorithm.  More generally,
\corolref{corol:softmax} establishes that jointly optimizing the
softmax parameters and data embeddings (features) can be accomplished
iteratively via the full ACE algorithm.

To see this, let $\Ph_{S,Y}$ denote the empirical distribution for
induced training data
\begin{equation*} 
\cT_f=\{(s_1,y_1),\dots,(s_n,y_n)\}
\end{equation*}
generated from $P_{S,Y}$, and note that
\begin{align*} 
&\sum_{s\in\cS} \Ph_S(s)\, D\bigl(\Ph_{Y|S}(\cdot|s) \bigm\|
\Pt_{Y|S}^{g,\beta}(\cdot|s)\bigr) \notag\\
&\qquad\qquad\qquad= \Hh(Y|S) - \underbrace{\bEd{\Ph_{S,Y}}{\log
  \Pt_{Y|S}^{g,\beta}(Y|S)}}_{\defeq\ell(g,\beta)}, 
\end{align*}
where $\ell(g,\beta)$ is the likelihood function, and $\Hh(Y|S)$
denotes the empirical conditional entropy.  Then if the number of
training samples is sufficiently large that,
effectively,\footnote{There are only $\cardY-1$ degrees of freedom in
  $P_Y$, so that when $\cardS$ is large, as we assume, $P_Y$ can be
  more accurately estimated from a given number of samples than
  $P_{S,Y}$, since the latter is described by $\cardS\cardY-1$ degrees
  of freedom.}
\begin{equation*}
\Ph_Y(y)=\sum_{s\in\cS} \Ph_{S,Y}(s,y) = P_Y(y),\quad y\in\Y,
\end{equation*}
the maximum-likelihood parameters are, via \propref{prop:softmax}
\begin{subequations} 
\begin{align}
\gh_{*,S}(y) &= \bLah_S^{-1} \left( \bmuh_{S|Y}(y) - \bmuh_S \right) +
o(\eps) \label{eq:gh*-softmax}\\ 
\betah_{*,S}(y) &= -\bmuh_S^\T \, \gh_{*,S}(y) + o(\eps), 
\label{eq:bth*-softmax}
\end{align}
\end{subequations}
where
\begin{align*}
\bmuh_S &= \bEd{\Ph_S}{S} \\
\bmuh_{S|Y}(y) &= \bEd{\Ph_{S|Y}(\cdot|y)}{S} \\
\bLah_S &= \Ed{\Ph_S}{\bigl(S-\bmuh_S\bigr)\bigl(S-\bmuh_S\bigr)^\T},
\end{align*}
with 
\begin{align*} 
\Ph_S(s) &= \sum_{y\in\Y} \Ph_{S,Y}(s,y),\quad s\in\cS\\
\Ph_{S|Y}(s|y) &= \frac{\Ph_{S,Y}(s,y)}{P_Y(y)},\quad s\in\cS,\ y\in\Y.
\end{align*}

Likewise, further optimizing the likelihood for the training data
\eqref{eq:cT-def} with respect to $f$ such that $s=f(x)$ under the
condition that $\Ph_S(s)=P_S(s)$ for $s\in\cS$ yields that
\begin{equation*} 
\fh_i(x) = \fh_i^*(x) \ \text{and}\ 
\gh_i(y) = \sigmah_i \, \gh_i^*(y),\quad i\in\{1,\dots,k\}
\end{equation*}
for $i\in\{1,\dots,k\}$, where $\fh_i$, $\gh_i$, and $\sigmah_i$ are
as defined in the analysis of \secref{sec:modal-approx} and can
(effectively) be computed via \algoref{alg:ace-comp-k}; specifically,
they characterize the modal decomposition of the empirical
distribution as expressed by \eqref{eq:modal-emp}.

Such analysis suggests the potential for alternatives to SGD that more
directly approximate empirical conditional expectation, and for
interpretations of the iterative matrix factorizations inherent in,
e.g., \cite{VegasZ04,DziugaiteR15}.  Moreover, our analysis provides
an upper bound \eqref{eq:D**-softmax} on performance against which the
performance of various weight optimization strategies can be measured.

Finally---and perhaps as importantly---we can view existing neural
network implementations as a tool for efficiently computing
conditional expectations.  Indeed, direct computation of empirical
conditional expectations can be prohibitively expensive in practice
for typical alphabet sizes, which the use of SGD can circumvent.

Ultimately, these insights clarify both the importance of nonlinear
activation functions in neural network, and the value of multilayer
architectures, the details of which will be more fully described and
developed in a subsequent paper.

\section{Gaussian Distributions and Linear Features}
\label{sec:gauss}

There is a natural counterpart to the development of the paper to this
point for the case when $X,Y$ are jointly Gaussian.
As we will describe, the resulting features in this case
are linear, and closely related to canonical correlation analysis
(CCA) \cite{hh36,hsst04,hs12} and principal component analysis (PCA)
\cite{kp01,hh33,itj02}.  More generally, the associated framework provides an
analysis for the case of arbitrary distributions of continuous-valued
variables subject to linear processing constraints.

\subsection{Gaussian Variables}
\label{sec:gauss-vars}

We begin with some convenient terminology, notation, and conventions.
Our development focuses on Gaussian variables that take the form of
(column) vectors.  We use $\gauss(\mu_Z,\bLa_Z)$ to denote the
corresponding distribution of such a variable\footnote{In our Gaussian
  development, to avoid certain notational conflicts we drop the use
  of boldface characters for random vectors, but retain them for
  nonrandom ones, and to further simplify notation, we also forgo the
  use of superscripts to indicate the dimension of a variable, as in
  \secref{sec:nn}.} $Z\in\Zg$, where $\mu_Z$ and $\bLa_Z$ denote the
associated mean vector and covariance matrix, respectively,
parameterizing the distribution, i.e.,
\begin{equation}
P_Z(z) =\frac{\detb{\bLa_Z}^{-1/2}}{(2\uppi)^{\dimZ/2}}
\expop{-\frac12 \,(z-\mu_Z)^\T 
  \bLa_Z^{-1}\, (z-\mu_Z)},
\label{eq:Pz-gauss}
\end{equation}
with $\detb{\cdot}$ denoting the determinant of its argument.  Without
loss of generality, we restrict our attention to variables $Z$ such
that $\bLa_Z$ is (strictly) positive definite since otherwise we may
eliminate the associated redundancy by reducing the dimensionality of
$Z$ until the covariance matrix is positive definite.  Also, for
simplicity of exposition we restrict our attention to zero-mean
variables whenever possible, while recognizing that nonzero means are
unavoidable when conditioning on other such variables.  The extension
to the more general case is straightforward.

It will frequently be convenient to work with the following equivalent
representation of a random variable.
\begin{definition}[Normalized Variable]
\label{def:norm-var}
For a variable $Z\in\reals^{K_Z}$  with mean $\mu_Z$ and
covariance $\bLa_Z$, the
corresponding normalized variable is 
\begin{equation}
\Zt\defeq\bLa_Z^{-1/2} (Z-\mu_Z)
\label{eq:normvar-def}
\end{equation}
and has mean $0$ and covariance $\bI$.
\end{definition}
In the sequel, we will generally use $\tilde{\ }$ notation to indicate
variables normalized according to \defref{def:norm-var}.

Next, consider an arbitrary pair of Gaussian variables, $Z\in\Zg$ and
$W\in\Wg$, which are jointly represented by
\begin{subequations} 
\begin{equation} 
C=\begin{bmatrix} Z \\ W
\end{bmatrix}
\sim\gauss(\bzero,\bLa_C),
\label{eq:C-def}
\end{equation}
where
\begin{equation}
  \bLa_C = \E{C C^\T} = \begin{bmatrix} \bLa_Z & \bLa_{ZW}
  \\ \bLa_{WZ} & \bLa_W
\end{bmatrix},
\end{equation}
\label{eq:C-form}%
\end{subequations}
so $Z\sim\gauss(\bzero,\bLa_Z)$, $W\sim\gauss(\bzero,\bLa_W)$,
$\bLa_{WZ}=\bE{W Z^\T}$, and $\bLa_{ZW}=\bLa_{WZ}^\T$.

It will frequently be convenient to express the relationship between
such variables in the familiar innovations form, the notation for
which we summarize as follows.
\begin{lemma}[Innovations Form and MMSE Estimation]
\label{lem:innov-form}
For any zero-mean jointly Gaussian variables $Z,W$ characterized by 
$\bLa_Z$, $\bLa_W$, and $\bLa_{ZW}$,
we have
\begin{equation}
Z = \bGa_{Z|W} \, W + \nu_{W\to Z},
\end{equation}
with gain matrix
\begin{equation}
\bGa_{Z|W}\defeq \bLa_{ZW} \bLa_W^{-1},
\label{eq:gain-matrix}
\end{equation}
and where $\nu_{W\to Z}\sim\gauss(\bzero,\bLa_{Z|W})$ is independent of
$W$ and thus
\begin{equation}
\bLa_{Z|W} = \bE{\nu_{W\to Z}\, \nu_{W\to Z}^\T} 
= \bLa_Z - \bLa_{ZW} \bLa_W^{-1} \bLa_{WZ}.
\label{eq:nu-cov}
\end{equation}
Moreover, the MMSE estimate of $Z$ based on $W$ follows immediately as
\begin{equation}
\Zh(W) = \bE{Z|W} = \bGa_{Z|W}\, W,
\label{eq:mmse-est}
\end{equation}
for which the  mean-square error (MSE) in the 
resulting estimation error $\nu_{W\to Z}$ is, from \eqref{eq:nu-cov},
\begin{equation}
\mse^{Z|W} \defeq \E{\bnorm{\nu_{W\to Z}}^2} = \tr\bigl(\bLa_{Z|W}\bigr).
\label{eq:mmse}
\end{equation}
\end{lemma}

\subsection{The Modal Decomposition of Covariance}
\label{sec:svd-cov}

For the model of interest involving zero-mean jointly Gaussian
$X\in\Xg$ and $Y\in\Yg$ with covariances $\bLa_X$ and $\bLa_Y$,
respectively, and cross-covariance $\bLa_{XY}$, it follows that the
correlation structure among the equivalent normalized variables 
\begin{equation}
\At \defeq \begin{bmatrix} \Xt \\ \Yt 
\end{bmatrix}
\label{eq:At-def}
\end{equation}
is
\begin{equation}
  \bLa_\At = \begin{bmatrix} \bI & \dtmg^\T \\ \dtmg & \bI 
\end{bmatrix},
\label{eq:bLa-At-def}
\end{equation}
where
\begin{equation} 
\dtmg \defeq  \bLa_Y^{-1/2} \, \bLa_{YX} \, \bLa_X^{-1/2}
= \bLa_Y^{-1/2} \ \bGa_{Y|X} \, \bLa_X^{1/2}
\label{eq:dtmg-def}
\end{equation}
plays the role in Gaussian analysis corresponding to $\dtmt$ in the
discrete case \cite{hwz18}.  We recognize $\dtmg$ as, of course, the
central quantity in CCA, and refer to it as the \emph{canonical
  correlation matrix (CCM)}.

The SVD of $\dtmg$ takes the form
\begin{equation}
\dtmg = \bPsi^Y \bSi \, \bigl( \bPsi^X\bigr)^\T = \sum_{i=1}^\mindim \sigma_i \,
\bpsi^Y_i \bigl( \bpsi^X_i\bigr)^\T, 
\label{eq:dtmg-svd}
\end{equation}
with
\begin{equation}
\mindim \defeq \min\{\dimX,\dimY\},
\label{eq:Kg-def}
\end{equation}
where $\bSi$ is an $\dimY\times\dimX$ diagonal matrix whose $\mindim$
diagonal entries are $\sigma_1,\dots,\sigma_\mindim$, where
\begin{subequations} 
\begin{align} 
\bPsi^X &= \begin{bmatrix} \bpsi^X_1 & \cdots & \bpsi^X_\dimX 
\end{bmatrix} \\
\bPsi^Y &= \begin{bmatrix} \bpsi^Y_1 & \cdots & \bpsi^Y_\dimY
\end{bmatrix}
\end{align}
\end{subequations}
are $\dimX\times\dimX$ and $\dimY\times\dimY$ orthogonal matrices,
respectively, and where, as before, we order the singular values
according to $\sigma_1\ge \dots\ge \sigma_\mindim$.  

Analogous to the case of finite alphabets, $\dtmg$ is a contractive operator representing conditional
expectation, i.e., $\sigma_i\le 1$ for $i=1,\dots,\mindim$, as is the
case for $\dtmt$ in the finite-alphabet case.  In particular, this
follows from the following standard result, a derivation of which we
provide for convenience in \appref{app:sv-ccm}.
\begin{fact}
\label{fact:sv-ccm}
Let $\bM$ be a matrix such that 
\begin{equation*}
\bLa = \begin{bmatrix} \bI & \bM \\ \bM^\T & \bI \end{bmatrix}
\end{equation*}
is symmetric, and let $\sigma_i(\bM)$ denote the $i$\/th singular
value of $\bM$.  Then $\bLa$ is positive semidefinite if and only if
$\sigma_i(\bM)\le1$ for all $i$.  More specifically, the $i$\/th pair
of eigenvalues of $\bLa$ are $1\pm\sigma_i(\bM)$ and the remaining
eigenvalues are all unity.
\end{fact}

In turn, the SVD \eqref{eq:dtmg-svd} yields the following modal
decomposition of 
the covariance $\bLa_{YX}$.  
\begin{proposition}
\label{prop:modal-gauss}
Let $X\in\Xg$, $Y\in\Yg$ be zero-mean jointly Gaussian variables
characterized by $\bLa_X$, $\bLa_Y$, and $\bLa_{XY}$, and let
\eqref{eq:dtmg-svd} denote the SVD of its CCM \eqref{eq:dtmg-def}.
Then there exist invertible linear mappings (coordinate
transformation)
\begin{subequations} 
\begin{align}
S^* \defeq \bff^*(X) = \begin{bmatrix} f_1^*(X) & \cdots & f_\dimX^*(X) 
\end{bmatrix}^\T &\defeq \bigl(\bF^*\bigr)^\T X \\
T^* \defeq \bg^*(Y) = \begin{bmatrix} g_1^*(Y) & \cdots & g_\dimY^*(Y) 
\end{bmatrix}^\T &\defeq \bigl(\bG^*\bigr)^\T Y
\end{align}
\label{eq:ST*-gauss}%
\end{subequations}
satisfying
\begin{subequations} 
\begin{align}
\bE{\bff^*(X)\,\bff^*(X)^\T} = \bigl(\bF^*\bigr)^\T \bLa_X \, \bF^*
&= \bI \label{eq:F-uncorr}\\
\bE{\bg^*(Y)\,\bg^*(Y)^\T} = \bigl(\bG^*\bigr)^\T \bLa_Y \, \bG^*  &=
\bI,\label{eq:G-uncorr} 
\end{align}
\label{eq:FG-uncorr}%
\end{subequations}
such that 
\begin{equation}
\bE{\bg^*(Y)\,\bff^*(X)^\T} = \bSi,
\label{eq:bLa-ST*}
\end{equation}
i.e.,
\begin{equation}
\bLa_{YX} = \bigl(\bG^*\bigr)^{-\T} \bSi\, \bigl(\bF^*\bigr)^{-1}
= \bLa_Y\,\bG^* \,\bSi\, \bigl(\bF^*\bigr)^\T \bLa_X.
\label{eq:cov-modal}
\end{equation}
\end{proposition}

\begin{IEEEproof}
Let
\begin{subequations} 
\begin{align}
\bF^* &\defeq \bLa_X^{-1/2} \, \bPsi^X  \\
\bG^* &\defeq \bLa_Y^{-1/2}\, \bPsi^Y ,
\end{align}
\label{eq:FG*-gauss}%
\end{subequations}
which we note satisfy \eqref{eq:FG-uncorr}
\begin{align*}
\bigl(\bF^*\bigr)^\T \bLa_X \, \bF^*  = 
\bigl(\bPsi^X\bigr)^\T \bLa_X^{-1/2} \bLa_X \bLa_X^{-1/2} \bPsi^X 
&= \bI \\
\bigl(\bG^*\bigr)^\T \bLa_Y \, \bG^*  =
\bigl(\bPsi^Y\bigr)^\T \bLa_Y^{-1/2} \bLa_Y 
\bLa_Y^{-1/2} \bPsi^Y &= \bI.
\end{align*}
Moreover, since
\begin{align*}
\bigl(\bF^*\bigr)^{-1} &= \bigl(\bPsi^X\bigr)^\T \bLa_X^{1/2}  \\
\bigl(\bG^*\bigr)^{-1} &= \bigl(\bPsi^Y \bigr)^\T \bLa_Y^{1/2},
\end{align*}
it follows that \eqref{eq:cov-modal} is satisfied, i.e., 
\begin{equation*}
\bigl(\bG^*\bigr)^{-\T} \bSi\, \bigl(\bF^*\bigr)^{-1}
= \bLa_Y^{1/2}\, \bPsi^Y \bSi\, \bigl(\bPsi^X\bigr)^\T \bLa_X^{1/2}
=\bLa_{YX},  
\end{equation*}
where to obtain the last equality we have used \eqref{eq:dtmg-svd}.
\end{IEEEproof}

One consequence of \propref{prop:modal-gauss} are the following conditional
expectation relations, which are derived in \appref{app:ce-gauss}
\begin{corollary}
\label{corol:ce-gauss}
The features $\bff^*$ and $\bg^*$ defined via \eqref{eq:FG*-gauss} satisfy
\begin{subequations} 
\begin{align}
\bSi\, \bff^*(X) &= \bE{\bg^*(Y)\mid X} \\
\bSi\, \bg^*(Y) &= \bE{\bff^*(X)\mid Y}.
\end{align}
\end{subequations}
\end{corollary}

Finally, note that $\dtmg$ has the interpretation of the gain matrix in
estimates of $\Yt$ based on $\Xt$ (and vice-versa).  In particular, 
from \lemref{lem:innov-form} it is readily verified that we have the
innovations form
\begin{equation}
\Yt = \dtmg \Xt + \tilde{\nu},
\label{eq:innov-form-model}
\end{equation}
i.e., $\bGa_{\Yt|\Xt} = \dtmg$, with
\begin{equation*} 
\bE{\tilde{\nu} \tilde{\nu}^\T} = \bLa_{\Yt|\Xt} = \bI - \dtmg \dtmg^\T,
\label{eq:NormalizedGaussianEstimationError}
\end{equation*}
so the resulting MSE in the MMSE estimate
\begin{equation}
\hat{\Yt}(\Xt) = \bE{\Yt|\Xt} = \dtmg \Xt
\label{eq:Yt-est}
\end{equation}
is 
\begin{equation}
\mset \defeq \bE{\norm{\tilde{\nu}}^2} = \tr\bigl(\bI -\dtmg\dtmg^\T\bigr) = \dimY - \frob{\dtmg}^2.
\label{eq:mset-exp}
\end{equation}
As such, the SVD \eqref{eq:dtmg-svd} has the further interpretation as
a modal decomposition of
the MMSE estimator \eqref{eq:Yt-est}.

\subsection{Variational Characterization of the Modal Decomposition}

As in the discrete case, the linear features
$\bigl(\bff^*,\bg^*\bigr)$ in \propref{prop:modal-gauss} can be
equivalently obtained from a variational characterization, via which
we obtain the key connection to CCA \cite{hh36,hsst04,hs12}.
\begin{proposition}
\label{prop:var-modal-gauss}
For any $k\in\{1,\dots,\mindim\}$, let 
$\bF^*_{(k)}$ and $\bG^*_{(k)}$ denote the first $k$ columns of $\bF^*$
and $\bG^*$, respectively, in \propref{prop:modal-gauss},
 i.e., 
\begin{subequations} 
\begin{align} 
S^*_{(k)} &\defeq \bigl(\bF^*_{(k)}\bigr)^\T X
\defeq \begin{bmatrix} f^*_1(X) & \cdots & f^*_k(X) 
\end{bmatrix}^\T \label{eq:S*-gauss} \\
T^*_{(k)} &\defeq \bigl(\bG^*_{(k)}\bigr)^\T Y \defeq \begin{bmatrix} g^*_1(Y) & \cdots & g^*_k(Y) 
\end{bmatrix}^\T.
\label{eq:T*-gauss} 
\end{align}
\label{eq:ST*-k-gauss}%
\end{subequations}
Then
\begin{align}
\bigl(\bF^*_{(k)},\bG^*_{(k)}\bigr) 
&= \argmin_{(\bF_{(k)},\bG_{(k)})\in\cL}
\E{\bnorm{\bF_{(k)}^\T X - \bG_{(k)}^\T Y}^2} \notag\\
&=
\argmax_{(\bF_{(k)},\bG_{(k)})\in\cL}{\sigma(\bF_{(k)},\bG_{(k)})} ,
\end{align}
where
\begin{align} 
\sigma(\bF_{(k)},\bG_{(k)}) 
&\defeq \E{\bigl(\bF_{(k)}^\T X\bigr)^\T
  \bG_{(k)}^\T Y} \notag\\
&= \tr\bigl(\bG_{(k)}^\T\bLa_{YX}\bF_{(k)}\bigr) 
\end{align}
and
\begin{equation} 
\cL \defeq \Bigl\{ (\bF_{(k)},\bG_{(k)}) \colon
 \bF_{(k)}^\T \bLa_X\,\bF_{(k)} = \bG_{(k)}^\T \bLa_Y\, \bG_{(k)} = \bI
\Bigr\}.
\label{eq:cL-def}
\end{equation}
Moreover, the resulting maximal correlation (generalized Pearson
correlation coefficient) is
\begin{equation}
\sigma(\bF^*_{(k)},\bG^*_{(k)}) =
\tr\Bigl(\bigl(\bG^*_{(k)}\bigr)^\T\bLa_{YX}\bF^*_{(k)}\Bigr) 
= \sum_{i=1}^k \sigma_i,
\label{eq:maxcorr}
\end{equation}
the Ky Fan $k$-norm of $\dtmg$.
\end{proposition}

\begin{IEEEproof}
Without loss of generality, we reparameterize $\bF_{(k)}$ and
$\bG_{(k)}$ in terms of new matrices\footnote{We refer to $\bfWX$ and
  $\bfWY$ as the \emph{feature weights} associated with the linear
  features $S$ and $T$, and note that they play the role in Gaussian
  analysis corresponding to that played by the feature vectors $\bfVX$
  and $\bfVY$ in the discrete case.}  $\bfWX$ and $\bfWY$ according to
\begin{subequations} 
\begin{align}
\bF_{(k)} &= \bLa_X^{-1/2} \, \bfWX  \\
\bG_{(k)} &= \bLa_Y^{-1/2} \, \bfWY ,
\end{align}
\label{eq:FG-def}%
\end{subequations}
in which case 
\begin{equation} 
\sigma(\bF_{(k)},\bG_{(k)}) 
= \tr\bigl(\bG_{(k)}^\T\bLa_{YX}\bF_{(k)}\bigr)
= \tr\Bigl(\bigl(\bfWY\bigr)^\T \dtmg \, \bfWX\Bigr), 
\label{eq:veccor}
\end{equation}
and \eqref{eq:cL-def} dictates that
\begin{equation} 
\bigl(\bfWX\bigr)^\T \bfWX = \bigl(\bfWY\bigr)^\T \bfWY = \bI.
\label{eq:orthog-gauss}
\end{equation}
From \lemref{lem:svd-k}, it follows immediately that \eqref{eq:veccor}
is maximized subject to 
\eqref{eq:orthog-gauss} when
we choose 
\begin{subequations} 
\begin{align} 
\bfWX &= \bPsi^X_{(k)} \label{eq:bfWX-opt} \\
\bfWY &= \bPsi^Y_{(k)}, \label{eq:bfWY-opt}
\end{align}
\label{eq:bfWXY-opt}%
\end{subequations}
where
\begin{subequations} 
\begin{align} 
\bPsi^X_{(k)} &\defeq 
\begin{bmatrix} \bpsi^X_1 & \cdots  & \bpsi^X_k \end{bmatrix} \label{eq:bPsiX-k-gauss-def}\\
\bPsi^Y_{(k)} &\defeq 
\begin{bmatrix} \bpsi^Y_1 & \cdots  & \bpsi^Y_k \end{bmatrix},
\label{eq:bPsiY-k-gauss-def}
\end{align}%
\label{eq:bPsi-k-gauss-def}%
\end{subequations}
and the resulting maximal correlation is \eqref{eq:maxcorr}, i.e., 
\begin{subequations} 
\begin{align}
\bF^*_{(k)} &= \bLa_X^{-1/2} \bPsi^X_{(k)} \label{eq:F*-k-def}\\
\bG^*_{(k)} &= \bLa_Y^{-1/2} \bPsi^Y_{(k)} \label{eq:G*-k-def},
\end{align}
\label{eq:FG*-k-def}%
\end{subequations}
as claimed.
\end{IEEEproof}

\subsection{Local Gaussian Information Geometry}
\label{sec:local-gauss}

It will sometimes be useful to define a local analysis for Gaussian
variables.  For such analysis, there is a natural counterpart of the
$\chi^2$-divergence \eqref{eq:chi2-def} used in the finite-alphabet
case.  In particular, we will make use of the following notion of
neighborhood.
\begin{definition}[Gaussian $\eps$-Neighborhood]
\label{def:nbhdg}
For a given $\eps>0$, the $\eps$-neighborhood of a $K_0$-dimensional
Gaussian distribution $P_0=\gauss(\mu_0,\bLa_0)$ with positive
definite $\bLa_0$ is the set of
Gaussian distributions in the following generalized
divergence 
ball of radius $\eps^2$ about $P_0$, i.e., 
\begin{subequations} 
\begin{align} 
&\nbhdg_\eps^{K_0}(P_0) \defeq \bigl\{ P'=\gauss(\mu,\bLa) \colon
\Db(P'\|P_0) \le \eps^2 K_0 \bigr\},
\end{align}
where for $P=\gauss(\mu_P,\bLa_P)$ and $Q=\gauss(\mu_Q,\bLa_Q)$ with
positive definite $\bLa_Q$, 
\begin{align} 
&\Db(P\|Q) \defeq (\mu_P-\mu_Q)^\T\bLa_Q^{-1}(\mu_P-\mu_Q) \notag\\
&{}\qquad\qquad\qquad+ \frac12
\bfrob{\bLa_Q^{-1/2}\bigl(\bLa_P-\bLa_Q\bigr)\bLa_Q^{-1/2}}^2.
\label{eq:Db-def}
\end{align} 
\label{eq:nbhdg-def}%
\end{subequations}
\end{definition}

Just as $D(\cdot\|\cdot)$, is invariant to a change of coordinates,
$\Db(\cdot\|\cdot)$ is invariant to invertible linear transformation
of variables, i.e., mappings of the form $Z'=\bA\,Z + \bc$ with
nonsingular $\bA$.  In particular, we have the following result.
\begin{lemma}
\label{lem:Db-invariance}
Let $\gauss(\mu_P,\bLa_P)$ and $\gauss(\mu_Q,\bLa_Q)$ be 
$K_0$-dimensional Gaussian distributions with nonsingular $\bLa_Q$.  Then
for any nonsingular matrix $\bA$ vector $\bc$ of compatible
dimensions, 
\begin{align} 
&\Db\bigl(\gauss(\mu_P,\bLa_P)\bigm\|\gauss(\mu_Q,\bLa_Q)\bigr) \notag\\
&\qquad=\Db\bigl(\gauss\bigl(\bA\mu_P+\bc,\bA\bLa_P\bA^\T\bigr)\bigm\|\gauss\bigl(\bA\mu_Q+\bc,\bA\bLa_Q\bA^\T\bigr)\bigr).
\label{eq:Db-invariance}
\end{align}
\end{lemma}

A proof of this invariance is provided in \appref{app:Db-invariance},
and makes use of the following simple identity.
\begin{lemma}
\label{lem:frob-ident}
For any symmetric matrices $\bM_1$ and $\bM_2$ of equal dimension,
\begin{equation}
\bfrob{\bM_1^{1/2}\bM_2\bM_1^{1/2}}^2 = \tr\bigl(\bM_1\bM_2\bM_1\bM_2\bigr).
\end{equation}
\end{lemma}

\subsection{Weakly Correlated Variables}
\label{sec:eps-corr}

An instance of the local analysis of \secref{sec:local-gauss}
corresponds to weak correlation between variables, a concept we
formally define as follows. 
\begin{definition}[$\eps$-Correlation]
\label{def:eps-corr}
Let $Z$ and $W$ be zero-mean jointly Gaussian with dimensions $\dimZ$
and $\dimW$, respectively.  Then $Z$ and $W$ are $\eps$-correlated
when 
\begin{equation}
P_{Z,W}\in\nbhdg_\eps^{\dimZ+\dimW}(P_ZP_W),
\end{equation}
where $P_Z$ and $P_W$
are the marginal distributions associated with $P_{Z,W}$.
\end{definition}

The following lemma, a proof of which is provided in
\appref{app:nbhdg-frob}, is useful in further characterizing
$\eps$-correlated variables.
\begin{lemma}
\label{lem:nbhdg-frob}
For any $\eps>0$ and zero-mean, $\eps$-correlated jointly Gaussian
variables $Z,W$ characterized by $\bLa_Z$, $\bLa_W$, and $\bLa_{ZW}$,
\begin{equation} 
\Db(P_{Z,W}\|P_ZP_W) = \eps^2 \bfrob{\bPhi^{Z|W}}^2,
\label{eq:nbhdg-frob}
\end{equation}
where 
\begin{equation}
\bPhi^{Z|W} \defeq \frac1\eps\, \bLa_Z^{-1/2} \bLa_{ZW} \bLa_W^{-1/2},
\label{eq:innov-matrix}
\end{equation}
which we refer to as the \emph{innovation matrix}.
\end{lemma}
In particular, it follows immediately from \lemref{lem:nbhdg-frob} that $Z,W$ being
$\eps$-correlated is equivalent to the condition 
\begin{equation}
\bfrob{\bPhi^{Z|W}}^2 \le \dimZ+\dimW.
\label{eq:eps-corr-char}
\end{equation}

It also follows that $Z,W$ are $\eps$-correlated when, on
average, $P_{Z|W}(\cdot|w)\in\nbhdg_\eps^{\dimZ+\dimW}(P_Z)$.  The following
lemma is useful in establishing this result; a proof is provided in
\appref{app:Db-cond-local}. 
\begin{lemma}
\label{lem:Db-cond-local}
For any $\eps>0$ and zero-mean, $\eps$-correlated jointly Gaussian
variables $Z,W$ characterized by $\bLa_Z$, $\bLa_W$, and $\bLa_{ZW}$,
\begin{equation} 
\Db(P_{Z|W}(\cdot|w)\|P_Z) =
\eps^2\bnorm{\bPhi^{Z|W}w}^2 + o(\eps^2),
\label{eq:Db-cond-local}
\end{equation}
where $\bPhi^{Z|W}$ is as defined in \eqref{eq:innov-matrix}.
\end{lemma}

Our further equivalent condition for $\eps$-correlation is then a consequence
of the following result, a proof of which is provided in
\appref{app:Ib-identity}. 
\begin{lemma}
\label{lem:Ib-identity}
For any $\eps>0$ and zero-mean, $\eps$-correlated jointly Gaussian
variables $Z,W$ characterized by $\bLa_Z$, $\bLa_W$, and $\bLa_{ZW}$,
we have\footnote{Note, too, that by symmetry we have
\begin{equation}
\bEd{P_Z}{\Db(P_{W|Z}(\cdot|Z)\|P_W)} = \bEd{P_W}{\Db(P_{Z|W}(\cdot|W)\|P_Z)} .
\end{equation}
Indeed, $\bPhi^{W|Z}=\bigl(\bPhi^{Z|W}\bigr)^\T$.}
\begin{equation}
\bEd{P_W}{\Db(P_{Z|W}(\cdot|W)\|P_Z)} = 
\Db(P_{Z,W}\|P_ZP_W)\bigl(1+o(1)\bigr).
\end{equation}
\end{lemma}

Finally, yet another such equivalent notion of $\eps$-correlation is
\begin{equation}
I(Z;W) = D(P_{Z,W}\|P_ZP_W) \le \eps^2 (\dimZ+\dimW)
\label{eq:I-gauss}
\end{equation}
where for Gaussian distributions, KL divergence takes the familiar form
\begin{align} 
&D\bigl(\gauss(\mu_P,\bLa_P)\|\gauss(\mu_Q,\bLa_Q)\bigr) \notag\\
&\qquad= \frac12 \Bigl[ (\mu_P-\mu_Q)^\T\bLa_Q^{-1}(\mu_P-\mu_Q) \notag\\
&\qquad\qquad\qquad{} +
\tr\bigl(\bLa_Q^{-1}\bLa_P - \bI\bigr) - \ln\bdetb{\bLa_P\bLa_Q^{-1}} \Bigr].
\label{eq:KL-gauss}
\end{align}
To establish \eqref{eq:I-gauss}, we make use of the following simple
fact, whose proof is provided in \appref{app:logdet}.
\begin{fact}
\label{fact:logdet}
Let $\delta$ be an arbitrary positive constant and $\bA$ an arbitrary
matrix.  Then
\begin{equation*} 
\ln\bdetb{\bI - \delta \bA\,\bA^\T} = -\delta \frob{\bA}^2 + o(\delta).
\end{equation*}
\end{fact}

As a first step, we have the following result, a proof of 
which is provided in \appref{app:D-local}. 
\begin{lemma}
\label{lem:D-local}
For any $\eps>0$ and zero-mean, $\eps$-correlated jointly Gaussian
variables $Z,W$ 
characterized by $\bLa_Z$, $\bLa_W$, and
$\bLa_{ZW}$, 
\begin{equation} 
D\bigl(P_{Z|W}(\cdot|w)\bigm\|P_Z\bigr) =
\frac12 \, \Db\bigl(P_{Z|W}(\cdot|w)\bigm\|P_Z\bigr) + o(\eps^2).
\end{equation}
\end{lemma}
The equivalence \eqref{eq:I-gauss} is then an immediate consequence of the
following corollary, whose proof is provided in \appref{app:I-local}.
\begin{corollary}
\label{corol:I-local}
For any $\eps>0$ and zero-mean, $\eps$-correlated jointly Gaussian
variables $Z,W$ characterized by $\bLa_Z$, $\bLa_W$, and $\bLa_{ZW}$,
\begin{equation} 
I(Z;W)= \frac12 \Db(P_{Z,W}\|P_ZP_W)\bigl(1+o(1)\bigr).
\label{eq:I-local}
\end{equation}
\end{corollary}

Finally, the following lemma is a useful generalization; a proof is
provided in \appref{app:KL-gauss-frob}.
\begin{lemma}
\label{lem:KL-gauss-frob}
Let $P_{X,Y}$ and $Q_{X,Y}$ denote candidate jointly Gaussian
distributions for $\eps$-correlated variables $X,Y$ with given
covariances $\bLa_X$, $\bLa_Y$, and $\eps>0$, where $\bLa_{XY}^P$ and
$\bLa_{XY}^Q$ denote the respective cross-covariances.  Then
\begin{equation}
D(P_{X,Y}\|Q_{X,Y}) = \frac12\, \frob{\dtmg_P-\dtmg_Q}^2 + o(\eps^2),
\label{eq:KL-gauss-frob}
\end{equation}
where [cf.\ \eqref{eq:dtmg-def}]
\begin{subequations} 
\begin{align} 
\dtmg_P &= \bLa_Y^{-1/2}\,\bLa_{YX}^P \bLa_X^{-1/2} \\
\dtmg_Q &= \bLa_Y^{-1/2}\,\bLa_{YX}^Q \bLa_X^{-1/2}.
\end{align}
\label{eq:DPPk}%
\end{subequations}
\end{lemma}

\subsection{Modal Decomposition of Jointly Gaussian Distributions}
\label{sec:svd-mi-gauss}

\secref{sec:svd-cov} describes how the SVD of $\dtmt$ provides a modal
decomposition of covariance for the jointly Gaussian $X,Y$ model.  
As related analysis, this section describes how in the weak
correlation regime, this SVD 
also provides a modal decomposition of mutual information and, more
generally, the joint distribution $P_{X,Y}$.

First, since
\begin{equation}
\bPhi^{Y|X} = \frac1\eps\, \dtmg,
\end{equation}
specializing \lemref{lem:nbhdg-frob}, we obtain that $X,Y$ are
$\eps$-correlated when 
\begin{equation}
\bfrob{\dtmg}^2 = \sum_{i=1}^\mindim \sigma_i^2 \le \eps^2 (\dimX+\dimY).
\label{eq:nbhdg-est-ball}
\end{equation}
In turn, when $X,Y$
are $\eps$-correlated we have,
specializing \corolref{corol:I-local},
\begin{equation} 
I(X;Y) 
= \frac12 \sum_{i=1}^\mindim \sigma_i^2 + o(\eps^2).
\label{eq:IXY-gauss}
\end{equation}

An interpretation of \eqref{eq:IXY-gauss} is obtained in terms of the
modal decomposition of $P_{X,Y}$, as we now describe.  In this Gaussian
scenario, in contrast to the finite alphabet case, the SVD is both a
logarithmic-domain one and asymptotic.  In particular, observe that as
$\eps\to0$, with $\bLa_\At$ as given by \eqref{eq:bLa-At-def} for
$\At$ as defined in \eqref{eq:At-def}, we have
\begin{align} 
\bLa_\At^{-1} 
&= \begin{bmatrix}
(\bI-\dtmg^\T\dtmg)^{-1} & -\dtmg^\T(\bI-\dtmg\dtmg^\T)^{-1} \\
-\dtmg(\bI-\dtmg^\T\dtmg)^{-1} & (\bI-\dtmg\dtmg^\T)^{-1}
\end{bmatrix} \notag\\
&= \begin{bmatrix} \bI +\dtmg^\T\dtmg& -\dtmg^\T \\ -\dtmg & \bI +
  \dtmg\dtmg^\T 
\end{bmatrix} + o(\eps^2) \label{eq:bLa-Zt-inv-2}\\
&= \begin{bmatrix} \bI & -\dtmg^\T \\ -\dtmg & \bI
\end{bmatrix} + o(\eps),
\label{eq:bLa-Zt-inv}
\end{align}
whence
\begin{align*}
&P_{\Xt,\Yt}(\xt,\yt) \notag\\
&\ = \expop{-\frac{\dimA}{2} \ln(2\uppi) -\frac12 \xt^\T\xt - \frac12
  \yt^\T\yt + \yt^\T\dtmg\,\xt + o(\eps)}\\
&\ = P_\Xt(\xt) \, P_\Yt(\yt) \, \expop{\sum_{i=1}^\mindim \sigma_i\,
  \xt^\T\bpsi^X_i (\bpsi^Y_i)^\T\yt + o(\eps)} \\
&\ = P_\Xt(\xt) \, P_\Yt(\yt) \, \prod_{i=1}^\mindim \expop{ \sigma_i \,
  \xt^\T\bpsi^X_i (\bpsi^Y_i)^\T\yt} \bigl(1 + o(1)\bigr).
\end{align*}
with $\dimA=\dimX+\dimY$.

As a result, we have 
\begin{align}
&P_{X,Y}(x,y) \notag\\
&\qquad=\detb{\bLa_X}^{-1/2}\detb{\bLa_Y}^{-1/2}\,
  P_{\Xt,\Yt}(\bLa_X^{-1/2} x,\bLa_Y^{-1/2} y) \notag\\ 
&\qquad= P_X(x) \,P_Y(y) \, \left(\prod_{i=1}^\mindim \e^{\sigma_i\,
f_i^*(x)\, g_i^*(y)}\right)\bigl(1+o(1)\bigr),
\label{eq:svd-gauss-form}
\end{align}
where $f_i^*$ and $g_i^*$ are the linear functions
\eqref{eq:ST*-gauss} determined in \propref{prop:modal-gauss}.

Furthermore, meaningful approximations to this joint distribution
arise by considering, for $k<\mindim$,
\begin{subequations} 
\begin{equation} 
\dtmg^{(k)}_* \defeq
\bPsi^Y_{(k)}\, \bSi_{(k)} \bigl(\bPsi^X_{(k)}\bigr)^\T,
\end{equation}
where
\begin{equation}
\bSi_{(k)} \defeq \bigl( \bPsi^Y_{(k)} \bigr)^\T \dtmg\, \bPsi^X_{(k)}
\label{eq:bSi-k-def}
\end{equation}
\label{eq:Bk*-def}%
\end{subequations}
is a diagonal matrix whose diagonal elements are $\sigma_1,\dots,\sigma_k$.
In particular, if we let  $X^{(k)}$ and $Y^{(k)}$ denote
zero-mean jointly Gaussian variables with the same marginals as $X$
and $Y$, respectively, but covariance\footnote{Note that
  $\bLa_{YX}^{(k)*}$ so-defined is a valid cross covariance matrix, i.e.,
  \begin{equation*} 
\begin{bmatrix} \bLa_Y & \bLa_{YX}^{(k)*}
    \\ \bigl(\bLa_{YX}^{(k)*}\bigr)^\T & \bLa_X 
  \end{bmatrix}
  \end{equation*}
is positive definite, as can be verified using \factref{fact:sv-ccm}.}
 [cf.\ \eqref{eq:cov-modal}]
\begin{align} 
\bLa_{YX}^{(k)*}\defeq \bLa_{Y^{(k)} X^{(k)}} 
&= \bLa_Y^{1/2} \, \dtmg^{(k)}_*\, \bLa_X^{1/2} \notag\\
&= \bigl(\bG^*_{(k)}\bigr)^{\mppi\T} \bSi_{(k)}\,
\bigl(\bF^*_{(k)}\bigr)^\mppi \notag\\
&= \bLa_Y\,\bG^*_{(k)} \,\bSi_{(k)}\, \bigl(\bF^*_{(k)}\bigr)^\T \bLa_X,
\label{eq:bLak-def}
\end{align}
where $\bF^*_{(k)}$ and $\bG^*_{(k)}$ are as defined in
\eqref{eq:FG*-k-def} and $\bSi_{(k)}$ is as defined in
\eqref{eq:bSi-k-def}, 
then it follows that the joint distribution of these new variables
takes the form
\begin{align*} 
&P_{X,Y}^{(k)*}(x,y) \defeq P_{X^{(k)},Y^{(k)}}(x,y)\notag\\
&\qquad= P_X(x) \,P_Y(y) \, \left(\prod_{i=1}^k \e^{\sigma_i\,
f_i^*(x)\, g_i^*(y)}\right)\bigl(1+o(1)\bigr),
\end{align*}
and 
\begin{equation*}
I(X^{(k)};Y^{(k)}) = \frac12 \sum_{i=1}^k \sigma_i^2 + o(\eps^2).
\label{eq:mi-approx}
\end{equation*}

\subsubsection*{Nonlinear Features and nonGaussian Distributions}

In concluding this section, we briefly discuss connections to
nonlinear features and nonGaussian variables.  First, with respect to
the former, it is worth noting that if we seek a modal decomposition
of the form \eqref{eq:modal} for the Gaussian case, an infinite number
of terms must be involved: the modal decomposition
takes the form
\begin{subequations} 
\begin{equation}
P_{X,Y}(x,y) 
= P_X(x) \,P_Y(y) \, \left( 1+ \sum_{i=1}^\infty \sigmat_i\,
\ft_i^*(x)\, \gt_i^*(y) \right),
\end{equation}
with 
\begin{equation} 
\begin{gathered} 
\bE{\ft_i^*(\Xt)} = \bE{\gt_i^*(\Yt)} = 0,\ i\!=\!1,2,\dots \\
\bE{\ft_i^*(\Xt)\,\ft^*_j(\Xt)} = 
\bE{\gt_i^*(\Yt)\,\gt^*_j(\Yt)} = 
\kron_{i=j},\ i,j\!=\!1,2,\dots,
\label{eq:svd-gauss-alt-constr}
\end{gathered}
\end{equation}%
\label{eq:svd-gauss-alt}%
\end{subequations}
And in such an expansion, it is important
to emphasize that the terms involving linear features need not
dominate; a brief discussion is provided in \appref{app:nonlinear}.

Second, with respect to nonGaussian distributions, we first emphasize
that in \propref{prop:modal-gauss} (and, in turn,
\propref{prop:var-modal-gauss}), only the second-moment properties of
the joint distribution $P_{X,Y}$ are required to derive the optimizing
linear features.  As such, those results obviously apply more broadly.
Additionally, when the nonGaussian variables are defined on finite
(but real-valued) alphabets, we can equivalently interpret the CCA
optimization problem as that of HGR maximal correlation with the
features constrained to be linear, i.e., maximizing the vector
correlation \eqref{eq:hgr-k-veccor} over linear $f^k$ and $g^k$.  Such
constraints may be practically motivated, for example.  In such cases,
we can relate the CCM $\dtmg$ from our Gaussian analysis to the
associated CDM $\dtm$ from our discrete analysis; the details are
provided in \appref{app:dtm-dtmg}.

Finally, modal decompositions of the form \eqref{eq:svd-gauss-alt} for
classes of nonGaussian distributions, generalizing aspects of the
results of \appref{app:nonlinear}, are developed in, e.g.,
\cite{az12,mz16,mz17}; see also the references therein, including the
early work of Lancaster \cite{hol58,hol69}.

\subsection{Latent Gaussian Attributes and Statistical Model}
\label{sec:latent-model-gauss}

In this section, we describe useful interpretations of the modal
decomposition for Gaussian variables in terms of latent variable
analysis, in a manner analogous to that of \secref{sec:latent-model}
for distributions over finite-alphabets.  In this case, our
development is more directly related to its roots in factor analysis
\cite{twa03,ab94} as introduced by Spearman \cite{cs1904}.

We begin with the introduction of latent Gaussian attribute
variables.   Although our definition includes a correlation
constraint, in our Gaussian case analysis we do not limit our
attention to the vanishing correlation regime.
\begin{definition}[Gaussian $\eps$-Attribute]
\label{def:gauss-attribute}
For [cf.\ \eqref{eq:eps-corr-char}]
\begin{equation}
0 < \eps \le \sqrt{\frac{\dimW}{\dimZ+\dimW}},
\label{eq:eps-range}
\end{equation}
the variable $W\in\Wg$ is a Gaussian $\eps$-attribute of $Z\in\Zg$ if:
1) $\dimW\le\dimZ$ and $\bLa_W$ is nonsingular; 2) $W,Z$ are jointly
Gaussian; 3) $W,Z$ are $\eps$-correlated but $\bLa_{WZ}\ne\bzero$; and 4)
$W$ conditionally independent of all other variables in the model
given $Z$.
\end{definition}

\begin{definition}[Gaussian $\eps$-Attribute Configuration]
\label{def:config-gauss}
Given a zero-mean Gaussian variable $Z\in\Zg$ with covariance
$\bLa_Z$, then for $\eps$ satisfying \eqref{eq:eps-range}, an
$\eps$-attribute $W$ of $Z$ is characterized by its configuration
[cf.\ \eqref{eq:eps-corr-char}]
\begin{equation}
\Cg_\eps^\dimZ(\bLa_Z)
= \Bigl\{ \dimW,\, \bLa_W,\, \bPhi^{Z|W} \colon
\bfrob{\bPhi^{Z|W}}^2\le\dimZ+\dimW \Bigr\} ,
\label{eq:CgZ-def}
\end{equation}
with $\bPhi^{Z|W}$ as defined in \eqref{eq:innov-matrix}.
\end{definition}

As in the case of discrete variables, the notion of a multi-attribute
is also useful in the Gaussian case.
\begin{definition}[Gaussian $\eps$-Multi-Attribute]
\label{def:gauss-multi-attribute}
A Gaussian $\eps$-multi-attribute is a Gaussian $\eps$-attribute
satisfying the additional property that 
\begin{equation}
\bspectral{\bPhi^{Z|W}}^2\le \frac{\dimZ+\dimW}{\dimW},
\label{eq:gauss-multi-attrib-cond}
\end{equation}
with $\bPhi^{Z|W}$ as defined in \eqref{eq:innov-matrix}.
\end{definition}
Note that \eqref{eq:gauss-multi-attrib-cond} is a stronger version of the
$\eps$ correlation property, since
$\bspectral{\bPhi^{Z|W}}\le\bfrob{\bPhi^{Z|W}}\le
\dimW\,\bspectral{\bPhi^{Z|W}}$. 

\begin{definition}[Gaussian $\eps$-Multi-Attribute Configuration]
\label{def:config-gauss-multi}
Given a zero-mean Gaussian variable $Z\in\Zg$ with covariance
$\bLa_Z$, then for $\eps$ satisfying \eqref{eq:eps-range}, an
$\eps$-multi-attribute $W$ of $Z$ is characterized by its configuration
[cf.\ \eqref{eq:gauss-multi-attrib-cond}] 
\begin{align} 
&\Cgm_\eps^\dimZ(\bLa_Z) \notag\\
&\qquad= \left\{ \dimW,\, \bLa_W,\, \bPhi^{Z|W} \colon
\bspectral{\bPhi^{Z|W}}^2\le\frac{\dimZ+\dimW}{\dimW} \right\} ,
\end{align}
with $\bPhi^{Z|W}$ as defined in \eqref{eq:innov-matrix}.
\end{definition}

For inferences about an attribute $W$, we consider features of the
form
\begin{equation*}
h(Z) = \bH^\T Z = \bigl(\bfWgen\bigr)^\T \Zt,
\end{equation*}
where $\bfWgen$ is the associated feature weight matrix.  Without loss
of generality, we restrict our attention to normalized (zero-mean)
$h(Z)$, so that\footnote{Indeed, if they were not, so long as the
  columns of $\bfWgen$ are linearly independent, so
  $\bigl(\bfWgen\bigr)^\T\bfWgen$ is invertible, we can transform
  $\bH$ into $\bHt$ via
\begin{equation*}
\bHt = \bH \Bigl( \bigl(\bfWgen\bigr)^\T \bfWgen \Bigr)^{-1/2}
 = 
\underbrace{\bfWgen\, \Bigl( \bigl(\bfWgen\bigr)^\T \bfWgen
  \Bigr)^{-1/2}}_{\defeq\bfWgent},  
\end{equation*}
where we note $\bigl(\bfWgent\bigr)^\T \bfWgent = \bI$.}
\begin{equation*}
\bE{h(Z)\,h(Z)^\T} = \bH^\T \bLa_Z \bH = \bigl(\bfWgen\bigr)^\T\bfWgen = \bI.
\end{equation*}

In the context of a given model $P_{X,Y}$, the Gaussian
$\eps$-attribute variables $U$ and $V$ for $X$ and $Y$, respectively,
are characterized by the (now Gauss-) Markov structure
\begin{equation}
U \markov X \markov Y \markov V,
\label{eq:markov-gauss}
\end{equation}
where $U\in\Ug$ and $V\in\Vg$ for some dimensions $\dimU$ and $\dimV$.

The following familiar fact will be useful, whose short proof we provide for
convenience in \appref{app:gm-cov-prod}.
\begin{fact}
\label{fact:gm-cov-prod}
Normalized zero-mean Gaussian variables $\Zt_1,\Zt_2,\Zt_3$ 
form a Markov chain $\Zt_1\markov \Zt_2\markov \Zt_3$ if and only if
\begin{equation}
\bLa_{\Zt_1\Zt_3} = \bLa_{\Zt_1\Zt_2}\,\bLa_{\Zt_2\Zt_3}.
\label{eq:gm-cov-prod}
\end{equation}
\end{fact}

As an application of \factref{fact:gm-cov-prod}, we have, for example,
\begin{subequations} 
\begin{align} 
\bPhi^{Y|U} &= \dtmg\,\bPhi^{X|U} \label{eq:bPhiXU2YU}\\
\bPhi^{X|V} &= \dtmg^\T\bPhi^{Y|V}. \label{eq:bPhiYV2XV}
\end{align}
\label{eq:bPhiYX2UV}%
\end{subequations}

For inferences about attributes $U$ and $V$, we will generally
consider statistics of the form
\begin{subequations} 
\begin{align} 
S_{(k)} &\defeq \bigl(\bF_{(k)}\bigr)^\T X
\defeq \begin{bmatrix} f_1(X) & \cdots & f_k(X) 
\end{bmatrix}^\T \label{eq:S-gauss} \\
T_{(k)} &\defeq \bigl(\bG_{(k)}\bigr)^\T Y \defeq \begin{bmatrix}
  g_1(Y) & \cdots & g_k(Y)  
\end{bmatrix}^\T
\label{eq:T-gauss} 
\end{align}
\label{eq:ST-gauss}%
\end{subequations}
for some dimension $k\in\{1,\dots,\mindim\}$ and feature matrices
$\bF_{(k)}\in\reals^{\dimX\times k}$ and
$\bG_{(k)}\in\reals^{\dimY\times k}$.  Without loss of generality we
restrict our attention to normalized features,
i.e. $(\bF_{(k)},\bG_{(k)})\in\cL$ with $\cL$ as defined in
\eqref{eq:cL-def}.  As we will develop, the particular choices
$S^*_{(k)},T^*_{(k)}$ defined in \eqref{eq:ST*-k-gauss} play a special
role.

For arbitrary jointly Gaussian $W$ and $Z$, we use $\mse^{W|Z}$ to
denote the MSE in the MMSE estimate of $W$ based on $Z$, so with
respect to our specific variables of interest,
$\mse^{U|S}(\bF_{(k)})$, $\mse^{V|S}(\bF_{(k)})$,
$\mse^{U|T}(\bG_{(k)})$, and $\mse^{V|T}(\bG_{(k)})$ denote the
associated MSEs, with their dependencies on $\bF_{(k)}$ and
$\bG_{(k)}$ made explicit.

\subsection{MMSE Universal Features}
\label{sec:uni-rieg}

In this formulation, we seek to determine optimum $k$-dimensional
features for estimating a pair of unknown Gaussian
attributes $(U,V)$ for $(X,Y)$ in the Gauss-Markov model
\eqref{eq:markov-gauss}, where $k\in\{1,\dots,\mindim\}$.

As in the finite-alphabet setting, we view the configurations of attributes
$U$ and $V$ as randomly drawn by nature from a RIE.  In this case,
this ensemble is also defined via the spherical symmetry of
\defref{def:sphere-sym}.

\begin{definition}[Gaussian Rotation Invariant Ensemble]
\label{def:rie-gauss} 
Given $\eps$ satisfying \eqref{eq:eps-range}, the Gaussian rotationally
invariant ensemble (RIE) for an attribute $W$ of a Gaussian variable
$Z$ is the collection of all jointly Gaussian attribute configurations
of the form \eqref{eq:CgZ-def}
together with a probability measure over the collection such that
$\bPhi^{Z|W}$ is spherically symmetric.
\end{definition}

Let $\Cg^\dimX_{\epsX}(\bLa_X)$ and $\Cg^\dimY_{\epsY}(\bLa_Y)$ denote
configurations for attributes $U$ and $V$, respectively, in the sense
of \defref{def:config-gauss}, i.e., 
\begin{subequations}
\begin{align}
\Cg_{\epsX}^\dimX(\bLa_X) &= \Bigl\{ \dimU,\, \bLa_U,\, \bPhi^{X|U} \colon
\bfrob{\bPhi^{X|U}}^2\le\dimU+\dimX \Bigr\} \\
\Cg_{\epsY}^\dimY(\bLa_Y) &= \Bigl\{ \dimV,\, \bLa_V,\, \bPhi^{Y|V} \colon
\bfrob{\bPhi^{Y|V}}^2\le\dimV+\dimY \Bigr\}.
\end{align}
\label{eq:UV-configs-gauss}%
\end{subequations}
where [cf.\ \eqref{eq:eps-range}]
\begin{equation}
0< \epsX^2 \le \frac{\dimU}{\dimU+\dimX} \quad\text{and}\quad
0< \epsY^2 \le \frac{\dimV}{\dimV+\dimY}.
\label{eq:eps-range-XY}
\end{equation}

In what follows, we denote the MSE in the MMSE estimates $U$ and $V$ based
on $S_{(k)}$ as defined in \eqref{eq:S-gauss}, respectively, via
\begin{subequations} 
\begin{align}
\mse^{U|S}\bigl(\Cg^\dimX_{\epsX}(\bLa_X),\bF_{(k)}\bigr)\quad\text{and}\quad
\mse^{V|S}\bigl(\Cg^\dimY_{\epsY}(\bLa_Y),\bF_{(k)}\bigr),
\end{align}
and those for the MMSE estimates based on $T_{(k)}$ as defined in
\eqref{eq:T-gauss} via, respectively,
\begin{align}
\mse^{U|T}\bigl(\Cg^\dimX_{\epsX}(\bLa_X),\bG_{(k)}\bigr)\quad\text{and}\quad
\mse^{V|T}\bigl(\Cg^\dimY_{\epsY}(\bLa_Y),\bG_{(k)}\bigr).
\end{align}
\label{eq:mse-def}%
\end{subequations}
In turn, we let
\begin{subequations} 
\begin{align} 
\mseb^{U|S}\bigl(\bF_{(k)}\bigr)
&\defeq
\bEd{\mathrm{RIE}}{\mse^{U|S}\bigl(\Cg^\dimX_{\epsX}(\bLa_X),\bF_{(k)}\bigr)}\\
\mseb^{V|S}\bigl(\bF_{(k)}\bigr)
&\defeq
\bEd{\mathrm{RIE}}{\mse^{V|S}\bigl(\Cg^\dimY_{\epsY}(\bLa_Y),\bF_{(k)}\bigr)}\\
\mseb^{U|T}\bigl(\bG_{(k)}\bigr)
&\defeq
\bEd{\mathrm{RIE}}{\mse^{U|T}\bigl(\Cg^\dimX_{\epsX}(\bLa_X),\bG_{(k)}\bigr)}\\
\mseb^{V|T}\bigl(\bG_{(k)}\bigr)
&\defeq
\bEd{\mathrm{RIE}}{\mse^{V|T}\bigl(\Cg^\dimY_{\epsY}(\bLa_Y),\bG_{(k)}\bigr)},
\end{align}
\label{eq:mseb-def}%
\end{subequations}
where $\bEd{\mathrm{RIE}}{\cdot}$ denotes expectation with respect to
the Gaussian RIEs for $\Cg^\dimX_{\epsX}(\bLa_X)$ and
$\Cg^\dimY_{\epsY}(\bLa_Y)$.  

For this scenario, we have following result, a proof of which is provided in
\appref{app:mmse-rie}. 
\begin{proposition}
\label{prop:mmse-rie}
Given zero-mean jointly Gaussian $X\in\Xg$, $Y\in\Yg$ characterized
by $\bLa_X$, 
$\bLa_Y$, and $\bLa_{XY}$, and attributes $U$ and $V$ of $X$ and $Y$,
respectively, each drawn from a Gaussian RIE for some $\epsX$ and $\epsY$,
respectively, satisfying \eqref{eq:eps-range-XY}, then for any
dimension $k\in\{1,\dots,\mindim\}$, the multi-objective minimization
\begin{align} 
\min_{(\bF_{(k)},\bG_{(k)})\in\cL} &\Bigl(
  \mseb^{U|S}(\bF_{(k)}),\mseb^{V|S}(\bF_{(k)}), \notag\\ 
&\qquad\qquad\mseb^{U|T}(\bG_{(k)}),\mseb^{V|T}(\bG_{(k)})\Bigr) 
\label{eq:mse-rie-pareto}
\end{align}
has a unique Pareto optimal solution, which is achieved by
$\bigl(\bF_{(k)}^*,\bG_{(k)}^*\bigr)$ as defined in
\eqref{eq:FG*-gauss}.  Moreover,
\begin{subequations} 
\begin{align}
\mseb^{U|S}(\bF_{(k)}^*) &= 
\tr(\bLa_U)\left[1 - \epsX^2 \bar{E}^{X|U}_0\,k
 \right]\\
\mseb^{V|S}(\bF_{(k)}^*) &= \tr(\bLa_V)\left[1 - \epsY^2 \bar{E}^{Y|V}_0
  \sum_{i=1}^k \sigma_i^2 \right] \\ 
\mseb^{U|T}(\bG_{(k)}^*) &=
\tr(\bLa_U)\left[1 - \epsX^2 \bar{E}^{X|U}_0 \sum_{i=1}^k \sigma_i^2 \right]
\\
\mseb^{V|T}(\bG_{(k)}^*)\bigr) &=
\tr(\bLa_V)\left[1 - \epsY^2\bar{E}^{Y|V}_0\, k \right],
\end{align}
\label{eq:mseb-opt}%
\end{subequations}
where $\bar{E}^{X|U}_0$ and $\bar{E}^{Y|V}_0$ are nonnegative
constants that do not depend on $\epsX$, $\epsY$, $k$, or $P_{X,Y}$.
\end{proposition}
We emphasize that \propref{prop:mmse-rie} is not asymptotic: we do not
require $\epsX,\epsY\to0$.

\subsection{MMSE Cooperative Game}
\label{sec:uni-coopg}

A characterization of the associated cooperative game for MSE
minimization, in which nature chooses the attribute that can be most
accurately estimated, is given by the following.  A proof is provided
in \appref{app:mmse-coop}.
\begin{proposition}
\label{prop:mmse-coop}
Given zero-mean jointly Gaussian $X\in\Xg$, $Y\in\Yg$ characterized by
$\bLa_X$, 
$\bLa_Y$, and $\bLa_{XY}$, parameters $\epsX,\epsY$ of
multi-attributes $U$ and $V$, respectively, satisfying
\eqref{eq:eps-range-XY}, and a dimension $k\in\{1,\dots,\mindim\}$,
then the multi-objective minimization
\begin{align} 
\smash[b]{\min_{\substack{(\Cg_{\epsX}^\dimX(\bLa_X),\Cg_{\epsY}^\dimY(\bLa_Y))\in\C_{(k)},\\
(\bF_{(k)},\bG_{(k)})\in\cL}}}
\Bigl( &\mse^{U|S}(\Cg_{\epsX}^\dimX(\bLa_X),\bF_{(k)}), \notag\\
&\quad\mse^{V|S}(\Cg_{\epsY}^\dimY(\bLa_Y),\bF_{(k)}), \notag\\
&\quad\quad\mse^{U|T}(\Cg_{\epsX}^\dimX(\bLa_X),\bG_{(k)}), \notag\\
&\quad\quad\quad\mse^{V|T}(\Cg_{\epsY}^\dimY(\bLa_Y),\bG_{(k)})\Bigr),
\label{eq:mse-coop-pareto}
\end{align}
where 
\begin{align}
\C_{(k)} \defeq \Bigl\{
&\bigl(\Cg_{\epsX}^\dimX(\bLa_X),\Cg_{\epsY}^\dimY(\bLa_Y)\bigr)
\colon \notag\\
&\qquad \dimU=\dimV=k, \notag\\ 
&\qquad \bspectral{\bLa_U^{-1}} \le1,\quad 
\bspectral{\bLa_V^{-1}} \le1, 
\Bigr\},
\label{eq:gauss-config-constr}
\end{align}
has a unique Pareto optimal solution, which is achieved by
$\bigl(\bF_{(k)}^*,\bG_{(k)}^*\bigr)$ as
defined in \eqref{eq:FG*-gauss}, and
$\bigl(\Cgm_{\epsX,*}^\dimX(\bLa_X),\Cgm_{\epsY,*}^\dimY(\bLa_Y)\bigr)$
characterized by
\begin{subequations} 
\begin{equation} 
\bLa_U = \bLa_V =\bI
\end{equation}
and
\begin{align} 
\bLa_{XU} &= \epsX\,\sqrt{\frac{K_X+k}{k}}\, \bLa_X\,
\bF^*_{(k)} \label{eq:bLaXU-opt}\\
\bLa_{YV} &= \epsY\,\sqrt{\frac{K_Y+k}{k}}\, \bLa_Y\, \bG^*_{(k)}.
\label{eq:bLaYV-opt}
\end{align}
\label{eq:opt-config-gauss}%
\end{subequations}
Moreover, 
\begin{subequations} 
\begin{align} 
\mse^{U|S}\!\bigl(\Cgm^\dimX_{\epsX,*}(\bLa_X),\bF^*_{(k)},\bG^*_{(k)}\bigr) 
&=k-\epsX^2(\dimX+k) \\
\mse^{V|S}\!\bigl(\Cgm^\dimY_{\epsY,*}(\bLa_Y),\bF^*_{(k)},\bG^*_{(k)}\bigr) 
&=k-\epsY^2\!\left(\frac{\dimY\!+\!k}{k}\right) \sum_{i=1}^k\sigma_i^2  \\
\mse^{U|T}\!\bigl(\Cgm^\dimX_{\epsX,*}(\bLa_X),\bF^*_{(k)},\bG^*_{(k)}\bigr) 
&=k-\epsX^2\!\left(\frac{\dimX\!+\!k}{k}\right) \sum_{i=1}^k\sigma_i^2  \\
\mse^{V|T}\!\bigl(\Cgm^\dimY_{\epsY,*}(\bLa_Y),\bF^*_{(k)},\bG^*_{(k)}\bigr) 
&=k-\epsY^2(\dimY+k).
\end{align}
\end{subequations}
\end{proposition}

Note that \propref{prop:mmse-coop} is also not asymptotic: it does not
require $\epsX,\epsY\to0$.  Note, too, that via
\propref{prop:mmse-coop} we obtain the multi-attributes $U$ and $V$
for which the features $\bigl(\bF_{(k)}^*,\bG_{(k)}^*\bigr)$ are
sufficient statistics.

\subsection{The Local Gaussian Information Bottleneck}
\label{sec:ib-gauss}

The following result establishes the optimum attributes in the MMSE
cooperative game of \secref{sec:uni-coopg} coincide 
with those of a Gaussian version of the local information
double bottleneck problem.  A proof is provided in
\appref{app:ib-double-gauss}. 
\begin{proposition}
\label{prop:ib-double-gauss}
Let $X\in\Xg$, $Y\in\Yg$ be zero-mean jointly Gaussian variables characterized by
$\bLa_X$, $\bLa_Y$, and $\bLa_{XY}$, and given $\epsX,\epsY>0$, 
let $U$ and $V$ be Gaussian $\epsX$- and
$\epsY$-multi-attributes of $X$ and $Y$, respectively, with
$\dimU=\dimV=k$.  Then
\begin{equation}
I(U;V) \le \frac{\epsX^2\epsY^2}{2}
\left(\frac{\dimX+k}{k}\right)\left(\frac{\dimY+k}{k}\right)
\sum_{i=1}^k \sigma_i^2 + o(\epsX^2\epsY^2),
\label{eq:IUV-opt}
\end{equation}
where the inequality holds with equality when the configurations of
$U$ and $V$ are given by \eqref{eq:opt-config-gauss}, in which case
\begin{equation}
\bLa_{UV} =
\epsX\epsY\,\sqrt{\frac{\dimX+k}{k}}\,\sqrt{\frac{\dimY+k}{k}}\,\bSi_{(k)}, 
\end{equation}
where $\bSi_{(k)}$ is as defined in \eqref{eq:bSi-k-def}.
\end{proposition}

\propref{prop:ib-double-gauss} can be equivalently expressed in the
form of a solution to a symmetric version of the Gaussian information
bottleneck problem \cite{cgtw05} in the weak dependence regime.  In
particular, we have the following corollary, a proof of which is
provided in \appref{app:ib-double-gauss-corol}.
\begin{corollary}
\label{corol:ib-double-gauss}
Let $X,Y$ be zero-mean jointly Gaussian variables characterized by
$\bLa_X$, $\bLa_Y$, and $\bLa_{XY}$, and let $U$ and $V$ be variables
in the Gauss-Markov chain \eqref{eq:markov-gauss} such that we satisfy
the independence relations $\bLa_U=\bLa_V=\bI$, the conditional
independence relations that\footnote{The elements of $U$ are
  conditionally independent given $X$ when
  $\bLa_{U|X}=\bLa_U-\bLa_{XU}^\T\bLa_X^{-1}\bLa_{XU}$ is diagonal,
  and similarly for the $V,Y$ relation.}
  $\bLa_{XU}^\T\bLa_X^{-1}\bLa_{XU}$ and
  $\bLa_{YV}^\T\bLa_Y^{-1}\bLa_{YV}$ are diagonal, and the dependence
  constraints $\max\bigl\{I(U_i;X),I(V_i;Y)\bigr\}\le\eps^2/2$ for
  $i=1,\dots,k$.  Then
\begin{equation}
\max_{U,V} I(U;V) = \frac{\eps^4}{2} \sum_{i=1}^k \sigma_i^2 + o(\eps^4).
\label{eq:IUV-opt-eps}
\end{equation}
Moreover, the maximum is achieved by the configurations
[cf.\ \eqref{eq:bLaXU-opt}--\eqref{eq:bLaYV-opt}]
\begin{subequations} 
\begin{align} 
\bLa_{XU} &= \eps\, \bLa_X\,
\bF^*_{(k)} \label{eq:bLaXU-opt-var}\\
\bLa_{YV} &= \eps\, \bLa_Y\, \bG^*_{(k)},
\label{eq:bLaYV-opt-var}
\end{align}
\label{eq:opt-config-gauss-var}%
\end{subequations}
in which case
\begin{equation}
\bLa_{UV} = \eps^2\,\bSi_{(k)}.
\label{eq:UV-opt-gauss}
\end{equation}
with $\bSi_{(k)}$ as defined in \eqref{eq:bSi-k-def}.
\end{corollary}

It further follows that $(S_{(k)}^*,T_{(k)}^*)$ is a sufficient
statistic for inferences about the optimizing $(U,V)$, i.e., for any
dimension $k\in\{1,\dots,\mindim\}$ we have
the Markov chains
\begin{equation}
(U,V) \markov (S_{(k)}^*,T_{(k)}^*) \markov (X,Y)
\end{equation}
and
\begin{equation}
U \markov S_{(k)}^* \markov T_{(k)}^* \markov V.
\end{equation}
In particular, we have the following result; a proof is provided in
\appref{app:suffstat-gauss}. 
\begin{corollary}
\label{corol:suffstat-gauss}
In the solution to the optimization in \propref{prop:ib-double-gauss},
\begin{equation}
p_{U,V|X,Y}(u,v|x,y) = p_{U|X}(u|x)\, p_{V|Y}(v|y) 
\label{eq:use-attrib}
\end{equation}
with 
\begin{subequations} 
\begin{align} 
p_{U|X}(\cdot|x) &=
\gauss\bigl( \eps\, s^*_{(k)}\,,\, (1 - \eps^2)\, \bI
\bigr) \label{eq:PUS-form} \\
p_{V|Y}(\cdot|y) &=
\gauss\bigl( \eps\, t^*_{(k)} \,,\, (1 - \eps^2)\, \bI
\bigr) \label{eq:PVT-form} 
\end{align}
\label{eq:suffstat-gauss}%
\end{subequations}
where [cf.\ \eqref{eq:ST*-k-gauss}] $s_{(k)}^*=\bF_{(k)}^*\, x$ and
$t_{(k)}^*=\bG_{(k)}^*\,y$, where we note that
\eqref{eq:suffstat-gauss} depend on $(x,y)$ only through
$(s_{(k)}^*,t_{(k)}^*)$.  Moreover, 
\begin{subequations} 
\begin{align}
p_{U|S_{(k)}^*,T_{(k)}^*,V}\bigl(u|s_{(k)}^*,t_{(k)}^*,v\bigr) 
&= p_{U|S_{(k)}^*}\bigl(u|s_{(k)}^*\bigr) \label{eq:US-markov}\\
p_{V|S_{(k)}^*,T_{(k)}^*,U}\bigl(v|s_{(k)}^*,t_{(k)}^*,u\bigr) 
&= p_{V|T_{(k)}^*}\bigl(v|t_{(k)}^*\bigr),
\label{eq:VT-markov}
\end{align}
\end{subequations}
and 
\begin{subequations}
\begin{align} 
p_{V|X}(\cdot|x) &= \gauss\bigl(\eps\,\bSi_{(k)}\,
s_{(k)}^*\,,\,\bI-\eps^2\,\bSi_{(k)}^2\bigr) \label{eq:PVS-form}\\
p_{U|Y}(\cdot|y) &= \gauss\bigl(\eps\,\bSi_{(k)}\,
t_{(k)}^*\,,\,\bI-\eps^2\,\bSi_{(k)}^2\bigr). \label{eq:PUT-form}
\end{align}
\label{eq:suffstat-condexp-gauss}%
\end{subequations}
\end{corollary}
We emphasize that the sufficient statistic pair
$(S_{(k)}^*,T_{(k)}^*)$ involves separate processing of $X$ and $Y$.
We also emphasize that \corolref{corol:suffstat-gauss} is not an
asymptotic result---it holds for finite $\eps$.

The more classical one-sided Gaussian information bottleneck problem
\cite{cgtw05} can also be analyzed in the weak-dependence regime.  For
example, we have the following result, a proof of which is provided in
\appref{app:ib-single-gauss}. 
\begin{proposition}
\label{prop:ib-single-gauss}
Let $X\in\Xg$, $Y\in\Yg$ be jointly Gaussian variables characterized by
$\bLa_X$, $\bLa_Y$, and $\bLa_{XY}$, and given $\eps>0$, let $U$ and
$V$ be variables in the Gauss-Markov chain \eqref{eq:markov-gauss} such
that we satisfy the independence relations $\bLa_U=\bLa_V=\bI$, the
conditional independence relations that
$\bLa_{XU}^\T\bLa_X^{-1}\bLa_{XU}$ and
$\bLa_{YV}^\T\bLa_Y^{-1}\bLa_{YV}$ are diagonal, and the dependence
constraints $\max\bigl\{I(U_i;X),I(V_i;Y)\bigr\}\le\eps^2/2$ for
$i=1,\dots,k$.  Then
\begin{equation}
\max_{U} I(U;Y) = \max_{V} I(V;X) = \frac{\eps^2}{2} \sum_{i=1}^k
\sigma_i^2 + o(\eps^2). 
\end{equation}
Moreover, the maximum is achieved by the configurations
\eqref{eq:opt-config-gauss-var}. 
\end{proposition}

Note, finally, that $(S,T)$ given by \eqref{eq:S*-gauss} and
\eqref{eq:T*-gauss} are sufficient statistics for inferences about the
resulting $(U,V)$, which we emphasize are obtained by \emph{separate}
processing of $X$ and $Y$.

\subsection{Gaussian Common Information}
\label{sec:common-gauss}

We now develop the relationship between the
optimizing Gaussian multi-attributes $(U,V)$ in \secref{sec:ib-gauss} (and
\secref{sec:uni-coopg}), and the common information associated with the
pair $(X,Y)$ characterized by a given joint distribution $P_{X,Y}$.

In the Gaussian case, common information can be readily evaluated,
without the local restriction of the finite-alphabet case, and takes the
following form, as shown in \cite[Corollary~1]{sc15}.  
For convenience, the proof is
provided in \appref{app:common-gauss}.
\begin{proposition}
\label{prop:common-gauss}
Let $X\in\Xg$, $Y\in\Yg$ be zero-mean jointly Gaussian variables characterized by
$\bLa_X$, $\bLa_Y$, and $\bLa_{XY}$.   Then
\begin{equation} 
C(X,Y) = \min_{\substack{P_{W|X,Y}\colon\\ X\markov W\markov Y}} I(W;X,Y) = 
\frac12 \sum_{i=1}^\mindim
\log\left(\frac{1+\sigma_i}{1-\sigma_i}\right). 
\label{eq:wci-gauss}
\end{equation}
Moreover, an optimizing $P_{W|X,Y}$ is Gaussian with
\begin{subequations} 
\begin{align} 
\bLa_{XW} &=
\bLa_X\,\bF^*_{(\mindim)}\,\bSi_{(\mindim)}^{1/2} \\
\bLa_{YW} &= 
\bLa_Y\,\bG^*_{(\mindim)}\,\bSi_{(\mindim)}^{1/2}.
\end{align}
\label{eq:Wg-opt}%
\end{subequations}
\end{proposition}

Note that since for $0<\omega<1$,
\begin{equation*}
\frac12 \ln \frac{1+\omega}{1-\omega} \ge \omega,
\end{equation*}
we have
\begin{equation*}
C(X;Y) \ge  \sum_{i=1}^\mindim \sigma_i = \nuclear{\dtmg},
\end{equation*}
where the bound is tight in the limit of weak correlation, i.e., 
\begin{equation*}
\frac{C(X;Y)}{\nuclear{\dtmg}}\to1 \qquad\text{as $\nuclear{\dtmg}\to0$}.
\end{equation*}

We further have 
\begin{equation} 
W\markov R^*_{(\mindim)}\markov (S^*_{(\mindim)},T^*_{(\mindim)})\markov(X,Y).
\end{equation}
where 
\begin{equation}
R^*_{(\mindim)} \defeq S^*_{(\mindim)} + T^*_{(\mindim)},
\label{eq:R*-gauss-def}
\end{equation}
with $S^*_{(\mindim)}$ and $T^*_{(\mindim)}$ as defined in
\eqref{eq:ST*-k-gauss}.   In particular, we have the following result, a
proof of which is provided in \appref{app:suffstat-common-gauss}.
\begin{corollary}
\label{corol:suffstat-common-gauss}
In the solution to the optimization in \propref{prop:common-gauss}
\begin{subequations} 
\begin{align} 
\bE{W|X,Y} &= \bSi_{(\mindim)}^{1/2} \bigl(\bI+\bSi_{(\mindim)}\bigr)^{-1}
 R^*_{(\mindim)} \\
\bLa_{W|X,Y} &= \bigl(\bI -
\bSi_{(\mindim)}\bigr)\bigl(\bI+\bSi_{(\mindim)}\bigr)^{-1}.
\end{align}
\label{eq:suffstat-common-gauss}%
\end{subequations}
\end{corollary}

\subsection{Relating Common Information to Dominant Structure}
\label{sec:relate-gauss}

The common information auxiliary variable $W$ of
\propref{prop:common-gauss} is naturally related to the unrestricted
dominant (i.e., optimizing) multi-attributes $(U,V)$ of
\secref{sec:uni-coopg} (whose restricted form arises in
\secref{sec:ib-gauss}).  By restricted, we mean
\begin{equation*}
\epsX = \sqrt{\frac{k}{\dimX+k}}\quad\text{and}\quad
\epsY = \sqrt{\frac{k}{\dimY+k}}
\end{equation*}
in \propref{prop:mmse-coop}, or, equivalently, $\eps=1$ in
\propref{prop:ib-double-gauss}.  In particular, the following result,
a proof of which is provided in \appref{app:Wc-opt}, establishes that
common information can be equivalently characterized by
\begin{equation} 
C(X,Y) = \min_{\substack{P_{W|X,Y}\colon\\ X\markov W\markov
    Y\\W\markov(U,V)\markov(X,Y)}} I(W;X,Y) . 
\label{eq:wci-gauss-var}
\end{equation}
so that the optimizing $W$ satisfies
\begin{equation}
W\markov (U,V)\markov \bigl(S^*_{(\mindim)},T^*_{(\mindim)}\bigr)\markov (X,Y).
\end{equation}
\begin{corollary}
\label{corol:Wc-opt}
Let $X,Y$ be zero-mean jointly Gaussian variables characterized by
$\bLa_X$, $\bLa_Y$, and $\bLa_{XY}$, and let $(U,V)$ be the
unrestricted dominant $\mindim$-dimensional multi-attributes.  If
$\Wc$ is chosen so that $\Wc\markov (U,V)\markov (X,Y)$ is a
Gauss-Markov chain with 
\begin{equation}
\bLa_{\Wc U} = \bLa_{\Wc V} = \bSi^{1/2}_{(\mindim)},
\label{eq:Wc-opt-gauss}
\end{equation}
and $\bLa_{\Wc}=\bI$, then
\begin{equation}
I(\Wc;X,Y) = C(X,Y),
\end{equation}
where $C(X,Y)$ is as given in \propref{prop:common-gauss}.
\end{corollary}

When $\Wc$ is constructed according to \corolref{corol:Wc-opt}, we have
the additional Markov structure
\begin{equation}
\Wc\markov (U+V)\markov R^*_{(\mindim)}\markov (X,Y).
\end{equation}
Specifically, we have the following readily verified result.
\begin{corollary}
With $\Wc$ as constructed in 
\corolref{corol:Wc-opt}, we have
\begin{align}
\bE{\Wc|U,V} &= \bSi_{(\mindim)}^{1/2}
\bigl(\bI+\bSi_{(\mindim)}\bigr)^{-1} (U+V) \\
\bLa_{\Wc|U,V} &=
\bigl(\bI-\bSi_{(\mindim)}\bigr) \bigl(\bI+\bSi_{(\mindim)}\bigr)^{-1}.
\end{align}
\end{corollary}

\subsection{An Interpretation of PCA}
\label{sec:pca}

PCA \cite{kp01,hh33,itj02} can be interpreted as a special case of the
preceding results.  Specifically, in some important instances, the
form of dimensionality reduction realized by PCA corresponds to the
optimum $k$-dimensional statistics $S_*=\bff^*(Y)$ and $T_*=\bg^*(X)$
as defined in \eqref{eq:S*-gauss} and \eqref{eq:T*-gauss},
respectively, for the universal estimation of the unknown
$k$-dimensional attributes $U$ and $V$ under any of our formulations.

\begin{example}
\label{ex:pca}
As an illustration, suppose we have the innovations form
\begin{equation*}
Y = X + \nu_{X\to Y},
\end{equation*}
where $X$ and $Y$ are $\mindim$-dimensional, and where
$\bLa_\nu=\sigma_\nu^2\,\bI$ but $\bLa_X$ is arbitrary.  Moreover, let
\begin{equation*} 
\bLa_X = \bUps \, \bLa \, \bUps^\T
\end{equation*}
denote the diagonalization of $\bLa_X$, so the columns of
\begin{equation}
\bUps = \begin{bmatrix} \bups_1 & \cdots & \bups_\mindim \end{bmatrix},
\label{eq:bUps-columns}
\end{equation}
are orthonormal, and $\bLa$ is diagonal with entries $\la_1 \ge \la_2
\ge \dots \ge \la_\mindim$.  Then it is immediate that $\bLa_Y$ has
diagonalization
\begin{equation*}
\bLa_Y = \bUps \, \bigl(\bLa+\sigma_\nu^2\,\bI\bigr) \,\bUps^\T.
\end{equation*}

In this case, it follows immediately that $\dtmg$ has SVD
\begin{equation} 
\dtmg 
= \bLa_Y^{-1/2} \bLa_X^{1/2}
= \underbrace{\bUps}_{=\bPsi^Y}\, \underbrace{\bigl( \bI +
    \sigma_\nu^2\, \bLa^{-1}\bigr)^{-1/2}}_{=\bSi}
\underbrace{\bUps^\T}_{=\bigl(\bPsi^X\bigr)^\T}. 
\label{eq:dtmg-pca}
\end{equation}

As a result, we have, for a given $1\le k\le \mindim$,
that \eqref{eq:S*-gauss} specializes to
\begin{align}
\bF^*_{(k)}
&= \bigl(\bUps \, \bLa^{-1/2} \bUps^\T\bigr)\, \bUps_{(k)}  \notag\\
&= \bUps_{(k)} \, \bLa_{(k)}^{-1/2} ,
\label{eq:f-gen-pca}
\end{align}
where
$\bUps_{(k)}$
denotes the $\mindim\times k$ matrix consisting of the first $k$ columns of
$\bUps$, i.e., 
\begin{equation*}
\bUps_{(k)} = \begin{bmatrix} \bups_1 & \cdots & \bups_k 
\end{bmatrix},
\end{equation*}
and where $\bLa_{(k)}$ denotes the $k\times k$ upper left submatrix of
$\bLa$, i.e., the matrix whose diagonal entries are
$\la_1,\dots,\la_k$.  
Likewise, \eqref{eq:T*-gauss} specializes to
\begin{align}
\bG^*_{(k)}
&= \Bigl(\bUps\, \bigl(\bLa+\sigma_\nu^2\,\bI\bigr)^{-1/2}
\bUps^\T\Bigr)\, \bUps_{(k)}  
\notag\\  
&=  \bUps_{(k)}\, \bigl(\bLa_{(k)}+\sigma_\nu^2\,\bI\bigr)^{-1/2}.
\label{eq:g-gen-pca}
\end{align}

In turn, it follows from \eqref{eq:f-gen-pca} and \eqref{eq:g-gen-pca}
that the $k$-dimensional PCA vector
\begin{subequations} 
\begin{align} 
S^\mathrm{PCA} &= \bff^\mathrm{PCA}(X) \defeq \bUps_{(k)}^\T \, X
\label{eq:S-pca} \\
\intertext{is a sufficient statistic for inferences about the unknown
  $U$ and $V$ based on $X$, and}
T^\mathrm{PCA} &= \bg^\mathrm{PCA}(Y) \defeq \bUps_{(k)}^\T \, Y
\label{eq:T-pca}
\end{align}
\label{eq:ST-pca}%
\end{subequations}
is a sufficient statistic for such inferences based on $Y$, i.e., we
have the Markov structure
\begin{subequations} 
\begin{align}
&(U,V) \markov S^\mathrm{PCA} \markov X\\
&(U,V) \markov T^\mathrm{PCA} \markov Y.
\end{align}
\end{subequations}

\end{example}

Beyond this illustrative example, for a general jointly Gaussian pair
$(X,Y)$, the statistics 
\begin{equation*}
S_{(k)}=\bigl(\bF^*_{(k)}\bigr)^\T X \quad\text{and}\quad
T_{(k)}=\bigl(\bG^*_{(k)}\bigr)^\T Y
\end{equation*}
specialize to 
(invertible transformations of) the PCA statistics \eqref{eq:ST-pca}
whenever $\dimX=\dimY=\mindim$ and $\bLa_X$ 
and $\bLa_Y$ are simultaneously diagonalizable, i.e., when they share
the same set of eigenvectors \eqref{eq:bUps-columns}, which is
equivalent to the condition that $\bLa_X$ and $\bLa_Y$ commute (see,
e.g., \cite[Theorem~1.3.12]{hj12}).  In fact, if $\bLa_X$ has distinct
eigenvalues and commutes with $\bLa_Y$, then there is a polynomial
$\pi(\cdot)$ of degree at most $\mindim-1$ such that $\bLa_Y=\pi(\bLa_X)$, which
follows from the Cayley-Hamilton theorem (see, e.g.,
\cite[Theorem~2.4.3.2 and Problem~1.3.P4]{hj12}).

\subsection{Efficient Learning of Covariance Modal Decompositions}
\label{sec:ace-g}

The linear features $\bff(x)$ and $\bg(y)$ in the modal decomposition
of covariance $\bLa_{XY}$ are readily constructed via an iterative
procedure.  In particular, a natural approach corresponds to applying
orthogonal iteration to $\dtmg$ to generate the dominant modes of its
SVD.  The resulting procedure has a statistical interpretation as an
ACE algorithm.  

To obtain the Gaussian version of the ACE algorithm
\algoref{alg:ace-comp-k}, it suffices to note that the conditional
expectations in this case are all linear---specifically,
\begin{subequations} 
\begin{align}
\bE{\bF_{(k)}^\T X \mid Y=y} 
&= \bF_{(k)}^\T \E{X\mid Y=y} \notag\\
&=  \bF_{(k)}^\T \bLa_{XY} \,\bLa_Y^{-1}\, y
\end{align}
and
\begin{align}
\bE{\bG_{(k)}^\T Y \mid X=x} 
&= \bG_{(k)}^\T \E{Y\mid X=x} \notag\\
&=  \bG_{(k)}^\T \bLa_{XY}^\T\bLa_X^{-1}\, x,
\end{align}
and that [cf.\ \eqref{eq:bLa-ST*}]
\begin{equation}
\E{\bF_{(k)}^\T X\,\bigl(\bG_{(k)}^\T Y\bigr)^\T} = \bF_{(k)}^\T
\bLa_{XY} \, \bG_{(k)}. 
\end{equation}
\label{eq:gauss-expectations}%
\end{subequations}
The resulting procedure then takes the form of
\algoref{alg:aceg-comp-k}.
Computational complexity behavior is
analogous to the corresponding algorithm for discrete data.
As in our discussion of \secref{sec:ace-interp}, 
steps 2f and 2c can be equivalently
expressed in their respective variational forms [cf.\ \eqref{eq:var-ce}] 
\begin{subequations} 
\begin{align}
\bFb_{(k)} &\gets \argmin_{\bF} \E{\bnorm{\bF_{(k)}^\T X-\bGh_{(k)}^\T Y}^2} \\
\bGb_{(k)} &\gets \argmin_{\bG} \E{\bnorm{\bFh_{(k)}^\T X-\bG_{(k)}^\T Y}^2},
\end{align}
\label{eq:var-ce-g}%
\end{subequations}
and evaluated iteratively or otherwise.  

\begin{algorithm}[tbp]
\caption{Gaussian ACE, Multiple Mode Computation}
\label{alg:aceg-comp-k}
\begin{algorithmic}[]
\Require{Covariance matrices $\bLa_{XY}$, $\bLa_X$, and $\bLa_Y$;
  dimension $k$}
\\ 1. Initialization: randomly choose $\bFb_{(k)}$
\Repeat\\
        \quad 2a. Cholesky factor:\\ 
        \quad\qquad $\bFb_{(k)}^\T\bLa_X\bFb_{(k)} =
        \bigl(\cholXg_{(k)}\bigr)^\T \cholXg_{(k)}$\\
        \quad 2b. Whiten:\\
        \quad\qquad $\bFh_{(k)} = \bFb_{(k)}\,\bigl(\cholXg_{(k)}\bigr)^{-1}$\\
	\quad 2c. $\bGb_{(k)} \gets \bLa_Y^{-1}\,\bLa_{YX}\,\bFh_{(k)}$\\
	\quad 2d. Cholesky factor:\\
        \quad\qquad $\bGb_{(k)}^\T\bLa_Y\bGb_{(k)} =
        \bigl(\cholYg_{(k)}\bigr)^\T \cholYg_{(k)}$\\
        \quad 2e. Whiten:\\
        \quad\qquad $\bGh_{(k)} = \bGb_{(k)}\,\bigl(\cholYg_{(k)}\bigr)^{-1}$\\
	\quad 2f. $\bFb_{(k)} \gets \bLa_X^{-1}\,\bLa_{YX}^\T\bGh_{(k)}$\\
        \quad 2g. $\sigmah^{(k)} \gets
        \tr\bigl(\bGh_{(k)}^\T\bLa_{YX}\,\bFb_{(k)}\bigr)$ 
\Until $\sigmah^{(k)}$ stops increasing.
\end{algorithmic}
\end{algorithm}

When the covariance structure $\bLa_X$, $\bLa_Y$, and $\bLa_{XY}$ is
unknown, but we have training data
\begin{equation}
\cT \defeq \{(x_1,y_1),\dots,(x_n,y_n)\},
\label{eq:cT-def-gauss}
\end{equation}
drawn i.i.d.\ from the associated Gaussian distribution, we can use
sample covariance matrices in place of the true ones in
\algoref{alg:aceg-comp-k}, viz.,
\begin{align*} 
\bLah_X &= \frac1n \sum_{i=1}^n (x_i-\bmuh_X)\,(x_i-\bmuh_X)^\T \\
\bLah_Y &= \frac1n \sum_{i=1}^n (y_i-\bmuh_Y)\,(y_i-\bmuh_Y)^\T \\
\bLah_{XY} &= \frac1n \sum_{i=1}^n (x_i-\bmuh_X)\,(y_i-\bmuh_Y)^\T,
\end{align*}
for example, 
where
\begin{equation*} 
\bmuh_X = \frac1n \sum_{i=1}^n x_i \quad\text{and}\quad
\bmuh_Y = \frac1n \sum_{i=1}^n y_i.
\end{equation*}
Sample complexity analysis of modal estimation in this
Gaussian case can be carried out in a manner analogous to that
described in \secref{sec:modal-approx}.

\subsection{Gaussian Attribute Matching}
\label{sec:gam}

The preceding analysis can be used to develop the natural Gaussian
counterpart of the Bayesian attribute matching formulation of
collaborative filtering in \secref{sec:cf}.  Our analysis can be
interpreted as a formulation of a problem of high-dimensional linear
estimation.  As such, they can be viewed in a broader context that
includes related results on shrinkage-based methods (see, e.g.,
\cite{lw04} and the references therein) that generalize James-Stein
estimators \cite{js61}.  Recent work more directly analogous to the
analyses of matrix factorization in, e.g., collaborative filtering as
discussed in \secref{sec:cf} include, e.g., \cite{rfp10,nw11}.  As
such, this section provides an additional interpretation of such
relationships.

For the purposes of illustration, consider a simple problem of
low-level computer vision.  Let $Y$ denote a vector representing a
(e.g., rasterized) $\dimY$-pixel target image of some scene of
interest, and let $X$ denote a vector representing a (linearly)
distorted $\dimX$-pixel source image of the scene.  Such distortions
could include, e.g., complex geometric transformations, nonuniform
sampling, spatially-varying filtering, and noise.  Then
$P_{Y|X}(\cdot|x)$ denotes the probability density for the target
image associated with a given source image $x$.

Given a choice for $k\in\{1,\dots,\mindim\}$, the $k$-dimensional
variables $U$ and $V$ in the Gaussian Markov chain
\eqref{eq:markov-gauss} correspond to the dominant attributes of
source and target images, respectively, and where $S_{(k)}^*$ and
$T_{(k)}^*$ represent sufficient statistics for the estimation of
these attributes.  

Conceptually, for each target image $y$, there is an associated target
attribute $V(y)$ generated randomly from $y$ according to
$P_{V|Y}(\cdot|y)$ that expresses the dominant attribute of the target image.
Likewise, for the source image $x$, there is an associated target
attribute $V_\circ(x)$ generated randomly from $x$ according to
$P_{V|X}(\cdot|x)$.  

Next, let $\Delta_y(x)$ denote how close the target attribute of
target image $y$ is to the target attribute of the source image $x$,
i.e.,
\begin{equation*}
\Delta_y(x) \defeq V(y) - V_\circ(x),
\end{equation*}
and define the  set
\begin{equation}
\cYh(x) \defeq \argmin_{y\in\Yg} \E{\bnorm{\Delta_y(x)}^2}.
\label{eq:cYh-def-gauss}
\end{equation}

The following characterization of $\cYh(x)$---the collection of target
images whose attributes match that of the source image most
closely---is useful in our development.  A proof is provided in
\appref{app:match-set-gauss}.
\begin{lemma}
\label{lem:match-set-gauss}
Given $k\in\{1,\dots,\mindim\}$ and zero-mean jointly Gaussian $X,Y$
characterized by $\bLa_X$, $\bLa_Y$, and $\bLa_{XY}$, define
$k$-dimensional Gaussian multi-attributes in the Gauss-Markov structure
\eqref{eq:markov-gauss} according to
\corolref{corol:ib-double-gauss} for some $\eps>0$.   Then for a given $x\in\Xg$ and $\cYh(x)$ as defined
\eqref{eq:cYh-def-gauss}, it follows $y\in\cYh$ if and only if
the associated (linear) features are related according to
\begin{equation}
g_i^*(y) = \sigma_i\, f_i^*(x),\qquad i=1,\dots,k.
\label{eq:match-set-gauss-rel}
\end{equation}
\end{lemma}

Among the target images $y$ for which the attribute match with $x$ is
closest, we seek the most likely, which we denote using $y^*(x)$.  We
have the following characterization of $y^*(x)$.  A proof is 
provided in \appref{app:attribute-match-gauss}.
\begin{proposition}
\label{prop:attribute-match-gauss}
Given $k\in\{1,\dots,\mindim\}$ and zero-mean jointly Gaussian
$X\in\Xg$, $Y\in\Yg$ 
characterized by $\bLa_X$, $\bLa_Y$, and $\bLa_{XY}$, define
$k$-dimensional Gaussian multi-attributes in the Gauss-Markov chain
\eqref{eq:markov-gauss} according to 
\corolref{corol:ib-double-gauss} for some $\eps>0$.
Then for a given $x\in\Xg$ and $\cYh(x)$ as defined
\eqref{eq:cYh-def-gauss}, we have that
\begin{equation}
y^*(x) \defeq \argmax_{y\in\cYh(x)} p_Y(y),
\label{eq:attribute-match-obj}
\end{equation}
with $p_Y=\gauss(\bzero,\bLa_Y)$ denoting the marginal for $Y$,
satisfies
\begin{subequations} 
\begin{equation}
y^*(x) = \bigl(\bG^*_{(k)}\bigr)^{\mppi\T}\, \bSi_{(k)}\,
\bigl(\bF^*_{(k)}\bigr)^\T x, 
\end{equation}
where the Moore-Penrose pseudoinverse of $\bG_{(k)}$ takes the form
\begin{equation}
\bigl(\bG^*_{(k)}\bigr)^\mppi = \bigl(\bG^*_{(k)}\bigr)^\T \bLa_Y.
\end{equation}
\label{eq:y*-gauss}%
\end{subequations}
\end{proposition}

The optimizing $y^*(x)$ in \propref{prop:attribute-match-gauss} has
the interpretation as an MMSE estimate based not on $\bLa_{XY}$ but on
the approximation $\bLa_{XY}^{(k)*}$ of rank $k$ defined in
\eqref{eq:bLak-def}.  In particular, the following corollary is an
immediate consequence of \eqref{eq:bLak-def}.
\begin{corollary}
An equivalent characterization of \eqref{eq:y*-gauss} in
\propref{prop:attribute-match-gauss} is
\begin{equation} 
y^*(x) = \bLa_{YX}^{(k)*}\, \bLa_X^{-1} \, x,
\end{equation}
where $\bLa_{YX}^{(k)*}$ is as defined in \eqref{eq:bLak-def}.
\end{corollary}

\subsubsection*{Connections to PCA}

As we now illustrate, PCA naturally arises in special cases of the
preceding matching framework.  In particular, returning to the
scenario of the example in \secref{sec:pca}, we first interchange the roles of $U$ and
$V$, and $X$ and $Y$, obtaining that the optimum target image $x^*(y)$
for a given source image $y$ based on Gaussian attribute matching is
\begin{subequations} 
\begin{equation}
x^*(y) = \bigl(\bF^*_{(k)}\bigr)^{\mppi\T}\, \bSi_{(k)}\,
\bigl(\bG^*_{(k)}\bigr)^\T y
\end{equation}
where
\begin{equation}
\bigl(\bF^*_{(k)}\bigr)^\mppi = \bigl(\bF^*_{(k)}\bigr)^\T \bLa_X.
\end{equation}
\label{eq:x*-gauss}%
\end{subequations}

Gaussian attribute matching in this scenario takes a familiar form.
In particular, specializing \eqref{eq:x*-gauss} using \eqref{eq:dtmg-pca},
\eqref{eq:f-gen-pca}, and \eqref{eq:g-gen-pca}, we obtain
\begin{align*} 
&x^*(y) \\
&\ = \bUps_{(k)} \bLa_{(k)}^{1/2}
\bigl(\bI+\sigma_\nu^2\,\bLa_{(k)}^{-1}\bigr)^{-1/2}\,\bigl(\bLa_{(k)}+\sigma_\nu^2\,\bI)\bigr)^{-1/2}\, \bUps_{(k)}^\T \, y\\ 
&\ = \bUps_{(k)}
\bLa_{(k)}\,\bigl(\bLa_{(k)}+\sigma_\nu^2\,\bI)\bigr)^{-1}\,
\bUps_{(k)}^\T \, y.
\end{align*}

The result is, of course, a standard approach to simple (linear)
denoising, whereby a signal of interest is expanded in the basis
prescribed by PCA, only the dominant modes are retained, and the
associated coefficients are appropriately attenuated.  As such, our
analysis provides an additional interpretation of such processing,
further insights into which also arise in the next section.

\subsection{Rank-Constrained Linear Regression}
\label{sec:nn-gauss}

In this section, we develop the counterpart to our softmax regression
analysis for jointly Gaussian variables, which is a form of
rank-constrained linear regression.  Such regression problems have a
long history.  Indeed, Young \cite{gy40} recognized the relationship
between early factor analysis and low-rank approximation.  Subsequent
results on the topic appear in, e.g., \cite{crr65}, and, later, in
\cite[Theorem~10.2.1]{drb75} \cite{aji75}.  Later still,
interpretations of the special case of PCA in terms of neural networks
appeared in, e.g., \cite{Oja82,BaldiH89,Oja92}, and the general case
of CCA in \cite{jk92}, in which an alternative to the ACE algorithm of
\secref{sec:ace-g} is involved in its implementation.  The results
of this section provide some complementary perspectives.

To begin, the counterpart of \propref{prop:softmax} is immediate in
the Gaussian case.  In particular, the following simple
result expresses that the particular exponential family form is
not restrictive.
\begin{proposition}
\label{prop:softmax-gauss}
Let $X\in\Xg$, $Y\in\Yg$ be zero-mean jointly Gaussian variables
characterized by $\bLa_X$, $\bLa_Y$, and $\bLa_{XY}$.  Furthermore,
given a dimension $k\in\{1,\dots,\mindim\}$, let $S=\bF^\T X$ for some
$\dimX\times k$ matrix $\bF$, so $\bLa_{YS} = \bLa_{YX}\, \bF$ and
$\bLa_S=\bF^\T\bLa_X\bF$ are the induced covariances.  Then the joint
probability density for $S,Y$ takes the form
\begin{align} 
P_{S,Y}(s,y) 
&= P_Y(y)\, P_{S|Y}(s|y) \notag\\
&= \gauss(y;\bzero,\bLa_Y)\,\gauss\bigl(s;\bG^\T y,
\bLa_S-\bG^\T\bLa_Y\,\bG\bigr), 
\end{align}
where
\begin{equation}
\bG^\T Y \defeq \bE{S|Y}.
\label{eq:bG-ce-rel}
\end{equation}
\end{proposition}

\begin{IEEEproof}
It suffices to exploit that since \eqref{eq:bG-ce-rel} is the MMSE
estimate of $S$ given $Y$, we have
\begin{equation*}
S = \bG^\T Y + \nu,
\end{equation*}
where the error $\nu$ is independent of $Y$.  Moreover,
\begin{equation*} 
\bLa_{S|Y} 
= \bLa_S - \bLa_{SY} \bLa_Y^{-1} \bLa_{YS}
= \bLa_S - \bG^\T \bLa_Y\, \bG ,
\end{equation*}
where to obtain the last equality we have used that 
\begin{equation*}
\bG^\T = \bLa_{SY}\,\bLa_Y^{-1}.
\end{equation*}
since $S,Y$ are jointly Gaussian.
\end{IEEEproof}

In turn, the counterpart to \corolref{corol:softmax} is the following
result optimizing $\bF$, whose proof is provided in
\appref{app:softmax-gauss-corol}.
\begin{proposition}
\label{prop:softmax-gauss-corol}
Let $X\in\Xg$, $Y\in\Yg$ be $\eps$-correlated zero-mean jointly
Gaussian variables whose joint density $P_{X,Y}$ is characterized by
$\bLa_X$, $\bLa_Y$, and $\bLa_{XY}$.  Furthermore, given a dimension
$k\in\{1,\dots,\mindim\}$, let
\begin{align}
&\cPt^{\dimX,\dimY}_k(\bLa_X,\bLa_Y) \defeq \notag\\
&\qquad\qquad\biggl\{ 
P\colon 
P =\gauss\left(\bzero,\begin{bmatrix} \bLa_X & \bLat_{XY} \\ \bLat_{XY}^\T
  & \bLa_Y  \end{bmatrix} \right),\notag\\
&\ \qquad\qquad\qquad\text{some $\bLat_{XY}$\! with $\rank(\bLat_{XY})\le k$} \biggr\}
\end{align}
denote the collection of zero-mean jointly Gaussian distributions with
rank-constrained cross-covariance.  Then for
$\Pt_{X,Y}\in\cPt^{\dimX,\dimY}_k(\bLa_X,\bLa_Y)$,
\begin{equation*}
D(P_{X,Y}\| \Pt_{X,Y}) 
\ge \sum_{i=k+1}^\mindim \sigma_i^2 + o(\eps^2),
\end{equation*}
where the inequality holds with equality when $\Pt_{X,Y}$ has
cross-covariance $\bLat_{YX} = \bLa_{YX}^{(k)*}$, with
$\bLa_{YX}^{(k)*}$ as given by \eqref{eq:bLak-def}.
\end{proposition}
This result expresses that among all $k$-dimensional linear
restrictions $S=\bF^\T X$ of the data, that corresponding to
$\bF=\bF_{(k)}^*$ is optimum.  

In turn, given the choice $S=\smash[b]{\bigl(\bF_{(k)}^*\bigr)^\T} X$, the
matrix $\bG^{(k)}_*$ defines the 
weights in the associated estimate of $Y$; specifically,
[cf.\ \eqref{eq:y*-gauss}]
\begin{equation}
\Yh^* = \bigl(\bLa_{XY}^{(k)*}\bigr)^\T \bLa_X^{-1} X = 
\bLa_Y\, \bG^*_{(k)} \, \bSi_{(k)}\, S.
\label{eq:Yh-KL-gauss}
\end{equation}
We emphasize that the estimate
\eqref{eq:Yh-KL-gauss}, in which $\bLa_{XY}^{(k)*}$ can be
equivalently expressed in the form 
\begin{align}
\bLa_{XY}^{(k)*} &=
\bigl(\bLa_X^{1/2}\,\bPsi^X_{(k)}\bigr)
\bigl(\bLa_X^{1/2}\,\bPsi^X_{(k)}\bigr)^\mppi \, \bLa_{XY},
\end{align}
is generally different from the MMSE estimator limited to rank $k$,
which can be expressed in the following form \cite{aji75}, a
derivation of which is provided in
\appref{app:mmse-low-rank}.
\begin{proposition}
\label{prop:mmse-low-rank}
For zero-mean, jointly Gaussian $X\in\Xg$ and $Y\in\Yg$ characterized by
covariance $\bLa_X$, $\bLa_Y$, and $\bLa_{XY}$, then given 
$k\in\{1,\dots,\mindim\}$,
\begin{subequations} 
\begin{equation} 
\Yh^\circ 
= \argmin_{\substack{\{\Yh\colon \Yh = \bGat_{Y|X}
    X,\\ \rank(\bGat_{Y|X})\le k\}}} \bE{\norm{Y-\Yh}^2} 
= \bigl(\bLa_{XY}^{(k)\circ}\bigr)^\T \bLa_X^{-1} X ,
\label{eq:mmse-opt}
\end{equation}
where
\begin{equation}
\bLa_{XY}^{(k)\circ} = 
\bigl( \bLa_X^{1/2}\, \bPsit^X_{(k)} \bigr) 
\bigl( \bLa_X^{1/2}\, \bPsit^X_{(k)} \bigr)^\mppi \,
\bLa_{XY} ,
\end{equation}
with $\bPsit^X_{(k)}$ denoting the first (dominant) $k$ columns of
$\bPsit^X$ in the (alternative) SVD [cf.\ \eqref{eq:dtmg-svd}]
\begin{equation}
\bLa_{YX}\, \bLa_X^{-1/2} = \bLa_Y^{1/2}\, \dtmg = \bPsit^Y \bSit
\,\bigl(\bPsit^X\bigr)^\T
\label{eq:mmse-svd}
\end{equation}
\label{eq:mmse-est-k}%
\end{subequations}
in which $\bPsit^X$ and $\bPsit^Y$ are orthogonal matrices and $\bSit$
is a diagonal matrix.
\end{proposition}
We note, in particular, that the generally different estimators
\eqref{eq:mmse-est-k} and \eqref{eq:Yh-KL-gauss} 
coincide when $\bLa_Y=\bI$, since the SVDs
\eqref{eq:dtmg-svd} and \eqref{eq:mmse-svd} are identical in this case.

Finally, the implied rank-constrained linear regression procedure
is as follows.   First, we assume that sufficient unlabeled training
data is available that $\bLa_X$ and $\bLa_Y$ are accurately
recovered.  Second, from the labeled training data we obtain the
empirical covariance $\bLah_{XY}$.   We then let $\Ph_{X,Y}$
denote the distribution of zero-mean jointly Gaussian variables
characterized by $\bLa_X$, $\bLa_Y$, and $\bLah_{XY}$ and
apply \propref{prop:softmax-gauss-corol} with $P_{X,Y}=\Ph_{X,Y}$
to obtain that the (locally) divergence-minimizing (cross-entropy maximizing)
regression parameters are given by 
\begin{equation}
\bLah_{YX}^{(k)*} \defeq 
\bLa_Y\,\bGh^*_{(k)} \,\bSih_{(k)}\, \bigl(\bFh^*_{(k)}\bigr)^\T \bLa_X,
\end{equation}
where $\bFh_{(k)}$, $\bGh_{(k)}$, and $\bSih_{(k)}$ correspond to the
$k$ dominant modes in the modal decomposition of the empirical
cross-covariance, viz., [cf.\ \eqref{eq:cov-modal}]
\begin{equation} 
\bLah_{YX} = \bLa_Y\,\bGh^* \,\bSih\, \bigl(\bFh^*\bigr)^\T \bLa_X.
\label{eq:cov-modal-est}
\end{equation}
In turn, the quality of the model fit is given by
\begin{equation}
D\bigl(\Ph_{X,Y}\bigm\| \Ph^{(k)*}_{X,Y}\bigr) 
= \sum_{i=k+1}^\mindim \sigmah_i^2 + o(\eps^2),
\end{equation}
where $\Ph^{(k)*}_{X,Y}$ denotes the (optimized) distribution of
zero-mean jointly Gaussian variables characterized by $\bLa_X$,
$\bLa_Y$, and $\bLah_{XY}^{(k)*}$, and
$\sigmah_1,\dots,\sigmah_\mindim$ are the diagonal entries of $\bSih$.

\subsection{Comments on Nonlinear CCA and PCA}  

A variety of nonlinear generalizations of CCA and PCA have been
developed.  Examples include versions of nonlinear CCA as developed in
\cite{Akaho01,AndrewABL13,mwl16,BentonKGRZA17}, as well as nonlinear
PCA as developed in \cite{Kramer91}, both of which are expressed via
layered neural network architectures.  As such,
the relationships between the results of Sections~\ref{sec:nn} and
\ref{sec:nn-gauss}---and between Sections~\ref{sec:modal} and
\ref{sec:svd-cov}---may also facilitate interpretating such forms of
CCA and PCA and relating them to their linear counterparts in ways
that extend existing analyses, examples of which include
\cite{itj02,Licciardi12}, and particularly the recent work
\cite{pft18}, the analysis of which is perhaps closest in spirit to that
of this paper.  Additionally, such nonlinear generalizations play a
role in the analysis of independent component analysis (ICA), as
described in, e.g., \cite{KarhunenOWVJ97,eo97}, and here, too, there
is the potential to complement existing information-theoretic
analyses, such as those in \cite{LeeGBS00,prf15,prf16,ap16}.

\section{Semi-Supervised Learning}
\label{sec:semi}

There are a wide variety of problems that deviate from the standard
supervised learning model on which we have focused by using labeled
data in more limited ways, instead relying more on 
unlabeled data in their training.  These are
typically referred to as \emph{semi-supervised} learning problems, and
there is a rich taxonomy and literature; see, e.g.,
\cite{csz06} and the many references therein, including the early work
\cite{hs65}.  While a broader development on the topic is beyond the
scope of the present paper, in this section we briefly discuss some of
the most immediate implications of our analysis to some such problems.

\subsection{Indirect Learning}
\label{sec:indirect}

A problem of significant interest is that of unsupervised learning, in
which only unlabeled data is available to train the system.  These
correspond to clustering problems, and there are a number of classical
approaches, originating with the work of Pearson \cite{kp1894}; see,
e.g., \cite{dhs00} and the references therein for a summary.

In practice, there can be many valid clusterings of data, some more
useful than others for a given target application.  For instance, in
the case of movies, one could cluster by any number of attributes,
including time period, genre, etc.  One can view these alternatives as
capturing different measures of proximity in carrying out the
clustering.  But if one is interested in clustering movies according
to the way people select movies to watch, then the measure of
proximity is less straightforward to quantify.  

In such cases, auxiliary labeled data can be used to effectively
capture the right notion of distance for such problems, and express
them in terms of universal features.  To develop this notion of
``indirect'' learning, which has similarities in spirit to methods
such as those described in \cite{bbbm06}, let
\begin{equation*}
X\markov Y\markov Z
\end{equation*}
denote a Markov chain of discrete variables in which $Y\in\Y$
represents the data we seek to cluster (e.g., movies), $Z\in\Z$
represents the class index, and $X\in\X$ represents auxiliary data
(e.g., people).  We assume that in general $\X$ and $\Y$ are large
alphabets, but that $\Z$ may be comparatively small, and that we have
an empirical distribution $\Ph_{X,Y}$ obtained from i.i.d.\ training
data from $P_{X,Y}$ (e.g., the Neflix database), but no training
samples of $Z$ from which to directly estimate $P_{Z|Y}$, or
even $P_Z$.

For this scenario, our universal analysis suggests the following
procedure.  First, for some suitably small $k\in\{1,\dots,K-1\}$, we
extract the $k$ dominant modes in the decomposition \eqref{eq:modal}
from $\Ph_{X,Y}$ (via, e.g., the ACE algorithm), then use the
resulting estimate of $g$ to define a new variable
$T=g(Y)\in\reals^k$.  In turn, our softmax analysis reveals that the
following model for the latent variable $Z$
[cf.\ \eqref{eq:*-softmax}]
\begin{align} 
&\Pt_{Z|T}^*(z|t) \notag\\
&\ \propto P_Z(z) \, \expop{(t\!-\!\bmu_T)^\T \bLa_T^{-1}
  (\bmu_{T|Z}(z)\!-\!\bmu_T) } \bigl(1+o(1)\bigr)
\label{eq:PZ|T}
\end{align}
is locally universal, and unsupervised learning corresponds to fitting
this model to the (induced) samples of $T$.

Our softmax analysis further implies a rather natural model fitting
procedure.  In particular, as discussed in \secref{sec:mixture}, the
resulting distribution $P_Y$, matches, to first order, that of a
Gaussian mixture, where $P_{T|Z}(\cdot|z)$ for $z\in\Z$ are the
Gaussian components.  Hence, this suggests that carrying out Gaussian
mixture modeling on the estimate of $P_T$ obtained from the training
data---e.g., via the Expectation-Maximization (EM) algorithm
\cite{dlr77}---to learn the parameters $\bmu_{T|Z}(t)$, $\bLa_{T|Z}$,
and $P_Z$, and (soft) clustering according to the resulting
$P_{Z|T}(\cdot|t)$ is locally optimal.  Of course, such soft
clustering can be replaced by any of a number of hard-decision
alternatives if desired, such as that based on the Lloyd algorithm
\cite{gg91}, which correspond to so-called $k$-means\footnote{Note that
  $k$ refers a different quantity (specifically, $\cardZ$) in this
  nomenclature than it does in our use.} clustering on the induced
samples of $T=g(Y)$.

In practice, this procedure is straightforward to apply and effective.
For example, applying it to, e.g., the Netflix database yields
meaningful movie clusterings.  For related developments and additional
insights, see, e.g., \cite{qmz18}.

\subsection{Partially-Supervised Learning}
\label{sec:MNIST}

Another class of learning system architectures is one in which labeled
data is used to design a classifier of interest, but the design of the
features themselves for such a classifier is based on unlabeled data.
These can be viewed as partially-supervised learning systems, and can
provide performance close to that of fully supervised architectures
while requiring significantly less labeled training data.  In such
cases, the feature extraction step corresponds to unsupervised
dimensionality reduction, for which there are a variety of well
established methods, both linear and nonlinear; see, e.g.,
\cite{swshl06}.  

The characterization of common information in terms of universal
features, as described in \secref{sec:common}, suggests a natural
framework for nonlinear dimensionality reduction, and, in turn,
constructing such partially-supervised learning systems, which we
illustrate through an example involving handwritten digit recognition,
using the MNIST database \cite{MNIST}.

The MNIST database consists of a set of $n=60\,000$ training images
$x^{(1)},\dots,x^{(n)}$ and a set of $n' = 10\,000$ test images, each
depicting a single handwritten digit from the set
$\Z\defeq\{\mathtt{0},\dots,\mathtt{9}\}$.  We let $z^{(i)}\in\Z$
denote the label corresponding to training image $x^{(i)}$, and let
$z\in\Z$ denote that for a given test image $x$, which are all
provided in the database.  Each training and test image is a
black-and-white, $28 \times 28$ pixels size, and quantized to 8-bits
per pixel (corresponding to intensity levels $\{0,\dots,255\}$), so
$\cardX=28\cdot28\cdot256=200\,704$ is the image alphabet size.

Using the labeled data
$\bigl(x^{(1)},z^{(1)}\bigr),\dots,\bigl(x^{(n)},z^{(n)}\bigr)$, we
seek to train a classifier based on our framework to predict the label
$z$ of a test image $x$ as accurately as possible.

\subsubsection{Classification Architecture}
\label{sec:mnist-arch}

The architecture we develop for this application involves three stages
in a manner corresponding to a two-layer neural network. The first
stage is a preprocessing step that converts the test image $x$ to a
representation $y=q(x)$ from a smaller alphabet $\Y$.  In the second
stage, we extract a low-dimensional real-valued feature $r=h(y)$ from
the image representation $y$.  Finally, in the third stage we classify
the image based on this low-dimensional feature using a predictor
$\pred(\cdot)$, generating label $\zh=\pred(h(q(x)))$.  

We restrict our attention to designs based on semi-supervised
learning.  Specifically, $q$ and $h$ are designed from the unlabeled
data $x^{(1)},\dots,x^{(n)}$ in an unsupervised manner, while $\pred$
is designed in a supervised manner from the reduced labeled data
\begin{subequations} 
\begin{equation}
\bigl(r^{(1)},z^{(1)}\bigr),\dots,\bigl(r^{(n)},z^{(n)}\bigr),
\end{equation}
with
\begin{equation} 
r^{(d)}=h(y^{(d)}),\qquad d=1,\dots,n.
\end{equation}
\label{eq:reduced-labeled-data}%
\end{subequations}

The details of our classifier design are as follows

\paragraph*{Stage 1 (Preprocessing)}  

\begin{figure}[tbp]
\centering
\includegraphics[width=.75\columnwidth]{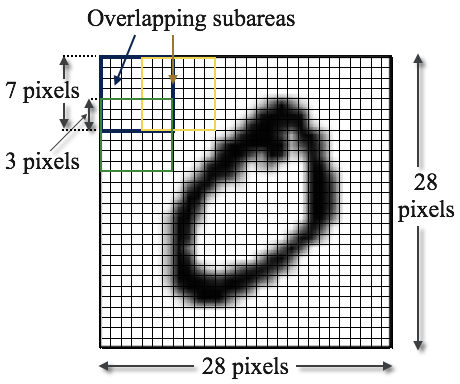}
\caption{Image representation for the preprocessing stage of the
  semi-supervised handwritten digit classifier.  Each $28\times28$
  MNIST database image is decomposed into an array of $6 \times 6 = 36$
  subimages, each of size $7\times7$ pixels, and each
  overlapping with its immediate neighbors by 3 pixels, horizontally
  and/or vertically.
\label{fig:MNIST}}
\end{figure}

As depicted in \figref{fig:MNIST}, we first decompose each MNIST
database image into an array of $6 \times 6 = 36$ overlapping
subimages, each of size $7 \times 7$ pixels, with immediately
neighboring subimages overlapping by $3$ pixels, horizontally and/or
vertically.  We denote the $(i,j)$\/th subimage by $\yt_{i,j}$, for
$i,j=1,\dots,6$, which takes value in an alphabet of size
$\card{\Yt}=7\cdot7\cdot256=12\,544$.

Second, quantize each subimage in a lossy manner to reduce the size of
the alphabet $\Yt$.  For this purpose, for each $(i,j)$, we cluster
all the subimages $\yt_{i,j}^{(1)},\dots,\yt_{i,j}^{(n)}$ extracted
from the training data using the ``balanced iterative reducing and
clustering using hierarchies'' (BIRCH) algorithm \cite{zrl96}, which
is simple and has computationally efficient (linear complexity).  In
particular, we use the implementation \cite{kgn17} with threshold
parameter to $256 \sqrt{3}$ and branching factor $1000$.  Each
subimage $\yt_{i,j}$ is then represented by the cluster to which it
maps, which we denote using $y_{i,j}$.\footnote{The resulting
  alphabets $\Y_{i,j}$ differ in size, ranging from roughly 10 for
  subimages at the perimeter of the image, to roughly 500 for
  subimages in the middle.}  We further use $y$ to denote the
resulting composite image presentation, i.e.,
\begin{equation*} 
y= \begin{bmatrix} y_{1,1} & \cdots & y_{1,6} \\ \vdots & \ddots &
  \vdots \\ y_{6,1} & \cdots & y_{6,6} 
\end{bmatrix}.
\end{equation*}

\paragraph*{Stage 2: Feature Extraction}

We generate a $k$-dimensional feature from the unlabeled training data
that captures as much of the common information among the subimages as
possible; in our experiment we choose $k=500$.  In particular, for
each of the $m=\binom{36}{2}=630$ pairs
$\bigl(y_{i,j},y_{i',j'}\bigr)$ of preprocessed subimages, we
determine the $k'$ dominant modes of the empirical pairwise
distribution $\Ph_{Y_{i,j},Y_{i',j'}}$ generated from the reduced
unlabeled training data
\begin{equation*}
y^{(1)},\dots,y^{(n)}.
\end{equation*}
In our experiment we choose $k'=16$, and use \algoref{alg:ace-comp-k}
to obtain these modes.  We then order this aggregate list of
$k'\,m=10\,080$ modes by singular value, and construct our feature set
from the subset corresponding to the overall $k$ largest singular
values, which we denote using $\sigmah_1,\dots,\sigmah_k$.

Specifically, with $\bigl\{(i_l,j_l),(i'_l,j'_l)\bigr\}$ denoting the
indices of the subimage pair whose $m_l$\/th mode has singular value
$\sigmah_l$, and with $\fh_{(i_l,j_l),(i'_l,j'_l),m_l}^*$ and
$\gh_{(i_l,j_l),(i'_l,j'_l),m_l}^*$ denoting the corresponding feature
functions in the decomposition of $\Ph_{Y_{i_l,j_l},Y_{i'_l,j'_l}}$,
for a test image with representation $y$, 
we choose as our $k$-dimensional feature
\begin{subequations} 
\begin{equation} 
r = h(y) = \bigl(h_1(y),\dots,h_k(y)\bigr)
\end{equation}
with
\begin{equation} 
h_l(y) = \fh_{(i_l,j_l),(i'_l,j'_l),m_l}^*(y_{i_l,j_l}) + \gh_{(i_l,j_l),(i'_l,j'_l),m_l}^*(y_{i'_l, j'_l}).
\end{equation}
\label{eq:mnist-feature}%
\end{subequations}
We emphasize that, in accordance with our development in
\secref{sec:common} that leads to \eqref{eq:common-ss-i}, the elements
of \eqref{eq:mnist-feature} are sufficient statistics for the relevant
components of the common information between the associated subimage
pairs.

\paragraph*{Stage 3 (Feature Classification)}

The final stage implements a low-dimensional feature classifier
generated from the reduced labeled training data
\eqref{eq:reduced-labeled-data}, with $h$ as defined in
\eqref{eq:mnist-feature}.  In particular, we choose a linear support
vector machine (SVM) \cite{cv95} for this purpose.

\subsubsection{Performance Evaluation}

When we evaluate the performance of the classifier of
\secref{sec:mnist-arch} on the full the set of $10\,000$ MNIST test
images, we achieve an digit recognition error probability of 3.02\%.

Our classifier, which is characterized by its $k=500$ scalar features,
is naturally compared to alternatives with similar numbers of
features.  For example, one such alternative classifier would omit the
preprocessing and feature extraction stages, and apply a linear SVM
directly to the original representation of image data, corresponding
to $28 \cdot 28 = 784$ scalar features.  This involves training
parameters of the linear classifier in fully-supervised manner, yet
only achieves an error probability of 8.17\% based on our experimental
analysis.  This reflects the importance of
nonlinearities inherent in the feature formation stages of our
architecture.

As another alternative, we can compare our achitecture to a deep
neural network (DNN) with two hidden layers and using sigmoidal
activation functions \cite{gbc17}, and trained in a fully supervised
manner.  For instance, using 300 units in the first layer and 100
units in the second corresponds to a total of $300+100=400$ scalar
features, and yields an error probability of 3.05\%
\cite{MNIST,lbbh98}, which is comparable to that of our classifer,
which is effectively a network with a \emph{single} hidden layer.  As
such, this reflects the effectiveness of the universal features
extracted via our methodology, which we further emphasize are designed
in an unsupervised manner---without taking into account the inference
task.

As a final evaluation, in Stage 3 of our architecture, we reduced the
amount of (labeled) data used to train the classifier from $n=60\,000$
to $n/2=30\,000$, while still using all $n$ \emph{unlabeled} training
samples for Stages 1 and 2.  In this case, we obtain an only mildly
degraded error probability of 3.4\%, which is a reflection of the
efficiency with which our architecture uses labeled training data, by
restricting its use to the final stage.

\section{Concluding Remarks}
\label{sec:conc}

In recent years, there has been rapid growth in activity in the area
of statistical inference and machine learning, largely motivated by
the increasingly abundant computational resources available for
implementing the associated methods, which in turn has led to an
ever-expanding set of rich application domains.  In turn, the
literature on these topics has expanded commensurately, both in depth
and breadth.  As such, beyond its specific contributions, this paper
can be viewed as a constrained effort to identify meaningful
connections and relationships between lines of research---and the
corresponding results---that have often been somewhat distinct, using
a convenient local information-theoretic analysis.  From this perpective,
there are abundant opportunities for further research, and the broader
challenge of a unifying treatment remains.

\appendices

\addtocounter{section}{1}

\section{Appendices for \protect\secref{sec:modal}}
\label{app:modal}

\subsection{Proof of \protect\propref{prop:contractive}}
\label{app:contractive}

It suffices to show that the maximum eigenvalue of $\dtm\dtm^\T$ is at
most unity.  To this end, note that via \eqref{eq:dtm-alt} and
\eqref{eq:dtma-alt} we have
\begin{align}
\dtm\dtm^\T 
= \left[\sqrt{\bP_Y}\right]^{-1} \bP_{Y|X}
\bP_{X|Y} \sqrt{\bP_Y}. \label{eq:use-PX|Y-def}
\end{align}
Now $\bP_{Y|X}$ and $\bP_{X|Y}$ are both column-stochastic matrices,
so their product $\bP_{Y|X}\,\bP_{X|Y}$ is as well.  As such, this
product has maximum eigenvalue of unity, which follows from, e.g.,
\cite[Theorem~8.3.4]{hj12} and the fact that by definition a matrix
$\bA$ is column stochastic if $\bone^\T \bA = \bone^\T$.  Finally, since
\eqref{eq:use-PX|Y-def} is a similarity transformation of
$\bP_{Y|X}\,\bP_{X|Y}$, it has the same eigenvalues.

Finally, \eqref{eq:svec0-def} can be verified by direct calculation
using \eqref{eq:dtms-def}:
\begin{align*} 
\sum_{x\in\X} \dtms(x,y)\,\sqrt{P_X(x)} 
&= 
\frac1{\sqrt{P_Y(y)}} \sum_{x\in\X} P_{X,Y}(x,y) = \sqrt{P_Y(y)} \\
\sum_{y\in\Y} \dtmas(y,x)\,\sqrt{P_Y(y)} 
&= 
\frac1{\sqrt{P_X(x)}} \sum_{y\in\Y} P_{X,Y}(x,y) = \sqrt{P_X(x)}.
\end{align*}
\hfill\IEEEQED

\subsection{DTM Characterization}
\label{app:dtm-char}

As notation, let $\dtm(P_{X,Y})$ denote the $\Y\times\X$ dimensional
DTM associated with joint distribution $P_{X,Y}$.  Moreover, let
$\dtmset$ denote the set of all DTMs, i.e., 
\begin{equation} 
\dtmset=\dtm(\simpXY),
\label{eq:dtmset-def}
\end{equation}
and let $\dtmsett$ denote the set of all DTMs corresponding to
distributions with strictly positive probabilities, i.e.,
\begin{equation}
\dtmsett=\dtm(\relint(\simpXY)).
\label{eq:dtmsett-def}
\end{equation}

As further notation, for a matrix $\bA$, we use $\bA>\bzero$ to denote
that every entry of $\bA$ is positive, and, likewise $\bA\ge\bzero$
when all entries are nonnegative.

The following proposition characterizes $\dtmsett$ in \eqref{eq:dtmsett-def}.
\begin{proposition}
\label{prop:dtm-1}
A matrix $\bM$ is a DTM corresponding to a joint distribution in
$\relint(\simpXY)$ if and only if $\bM > \bzero$ and $\spectral{\bM} =
1$, i.e.,
\begin{equation}
\dtmsett = \{\bM \in \reals^{\cardY \times \cardX} \colon \bM > \bzero \text{ and }
\spectral{\bM} = 1\},
\label{eq:dtmsett-form}
\end{equation}
where $\spectral{\cdot}$ denotes the spectral norm of its argument.
\end{proposition}

\begin{IEEEproof} 
The ``only if'' part of the claim is immediate.  Indeed, since
$\bM=\dtm(P_{X,Y})$ for some strictly positive $P_{X,Y}$, it follows
that $\bM >\bzero$.  Moreover, as developed in \secref{sec:modal},
$\spectral{\dtm(P_{X,Y})} = 1$.

For the ``if'' part of the claim, consider any matrix $\bM \in
\reals^{\cardY \times \cardX}$ satisfying $\bM > \bzero$ and
$\spectral{\bM} = 1$.  We can contruct a $P_{X,Y}$ for which $\bM$ is
its DTM.  To see this, first note that $\bM^\T \bM > \bzero$, $\bM
\bM^\T > \bzero$, and $\lambda(\bM^\T \bM) = \spectral{\bM}^2 = 1$,
where $\lambda(\cdot)$ denotes the largest eigenvalue of its argument.
Then, applying the Perron-Frobenius theorem \cite[Theorem
  8.2.2]{hj12}, it follows that there exist unit-norm vectors
$\bpsi^X\!(\bM)$ and $\bpsi^Y\!(\bM)$ with strictly positive elements such
that
\begin{equation*}
\bM^\T \bM \,\bpsi^X\!(\bM) = \bpsi^X\!(\bM) \quad \text{and} \quad
\bM \bM^\T \bpsi^Y\!(\bM) = \bpsi^Y\!(\bM) .
\end{equation*}
In turn, this implies that $\bpsi^X\!(\bM)$ and $\bpsi^Y\!(\bM)$ are
the right and left singular vectors corresponding to the unit
principal singular value of $\bM$, respectively, i.e.,
\begin{equation}
\label{Eq:SVD Relations}
\bM \, \bpsi^X\!(\bM) = \bpsi^Y\!(\bM) \quad \text{and} \quad \bM^\T \bpsi^Y\!(\bM) = \bpsi^X\!(\bM) . 
\end{equation}  

We now define a $P_{X,Y}$ lying on the simplex, and show that its DTM
is $\bM$.   In particular, we let 
\begin{equation*}
P_{X,Y}(x,y) \defeq \begin{bmatrix} \bP_{Y,X}
\end{bmatrix}_{y,x},\quad x\in\X,\ y\in\Y,
\end{equation*}
where
\begin{equation} 
\bP_{Y,X} \defeq \diag\bigl(\bpsi^Y\!(\bM)\bigr) \, \bM \diag\bigl(\bpsi^X\!(\bM)\bigr),
\label{eq:PYX-Mform}
\end{equation}
with $\diag(\cdot)$ denoting a diagonal matrix with diagonal entries specified
by its (vector) argument.

That $P_{X,Y}$ is strictly postive follows by construction, since the
quantities forming $\bP_{Y,X}$ are all strictly positive.  To verify that it
sums to unity, observe that
\begin{align*} 
\smash{\sum_{x,y}} P_{X,Y}(x,y) 
&= \bone^\T \bP \, \bone \notag\\
&= \bone^\T \diag\bigl(\bpsi^Y\!(\bM)\bigr) \, \bM\, \diag\bigl(\bpsi^X\!(\bM)\bigr) \,
\bone \notag\\ 
&= \bpsi^Y\!(\bM)^\T \bM \, \bpsi^X\!(\bM) \notag\\
&= \bpsi^Y\!(\bM)^\T \bpsi^Y\!(\bM) \notag\\
&= 1.
\end{align*}

Moreover, applying \eqref{Eq:SVD Relations}, we obtain that the
marginals take the form
\begin{subequations} 
\begin{align} 
P_Y(y) 
&= \bP \, \bone \notag\\
&= \diag\bigl(\bpsi^Y\!(\bM)\bigr) \, \bM \, \diag\bigl(\bpsi^X\!(\bM)\bigr) \, \bone \notag\\
&= \diag\bigl(\bpsi^Y\!(\bM)\bigr) \, \bM \, \bpsi^X\!(\bM) \notag\\
&= \diag\bigl(\bpsi^Y\!(\bM)\bigr) \, \bpsi^Y\!(\bM) \notag\\
&= \bpsi^Y\!(\bM)^2  
\label{Eq:Y Marginal}
\end{align}
and
\begin{align} 
P_X(x) &= \bP^\T \bone \notag\\
&= \diag\bigl(\bpsi^X\!(\bM)\bigr) \, \bM^\T \diag\bigl(\bpsi^Y\!(\bM)\bigr) \, \bone \notag\\
&= \diag\bigl(\bpsi^X\!(\bM)\bigr) \, \bM^\T \bpsi^Y\!(\bM) \notag\\
&= \diag\bigl(\bpsi^X\!(\bM)\bigr) \, \bpsi^X\!(\bM) \notag\\
&= \bpsi^X\!(\bM)^2,   
\label{Eq:X Marginal}
\end{align}%
\label{eq:XY-marginals}%
\end{subequations}
where $\bpsi^X\!(\bM)^2$ and $\bpsi^Y\!(\bM)^2$ are vectors whose
elements are the squares of the elements of $\bpsi^X\!(\bM)$ and
$\bpsi^Y\!(\bM)$, respectively.  

Hence, using \eqref{eq:XY-marginals} in \eqref{eq:PYX-Mform}
we obtain
\begin{align} 
\bM &= \left[ \sqrt{\diag\bigl(\bpsi^Y\!(\bM)\bigr)} \right]^{-1} \bP_{Y,X}
\left[\sqrt{\diag\bigl(\bpsi^X\!(\bM)\bigr)} \right]^{-1} \notag\\
&= \left[ \sqrt{\bP_Y} \right]^{-1} \bP_{Y,X} \left[\sqrt{\bP_X}
  \right]^{-1},
\end{align}
where $\bP_X$ and $\bP_Y$ are diagonal matrices whose diagonal
elements are the elements of $P_X$ and $P_Y$, respectively, which are
all strictly positive.   Hence, $\bM$ is the DTM corresponding to the
$P_{X,Y}$ we have constructed, i.e., $\bM=\dtm(P_{X,Y})$.
\end{IEEEproof}

\medskip 
The following generalization of \propref{prop:dtm-1} characterizes
$\dtmset$ in \eqref{eq:dtmset-def}.
\begin{proposition}
\label{prop:dtm-2}
A matrix $\bM$ is a DTM corresponding to a joint distribution in
$\simpXY$ if and only if $\bM\ge \bzero$, $\spectral{\bM}=1$, and 
each of $\bM^\T \bM$ and $\bM \bM^\T$
have a strictly positive eigenvector corresponding to their unit
eigenvalue, i.e., 
\begin{align*}
\dtmset = \Bigl\{ &\bM \in \reals^{\cardY \times \cardX} \colon 
\bM \geq \bzero, \ \spectral{\bM} = 1, \\ 
& \exists\, \bpsi^X\!(\bM)>\bzero \text{ s.t. } \bM^\T \bM \,
\bpsi^X\!(\bM) = \bpsi^X\!(\bM), \\ 
& \exists\, \bpsi^Y\!(\bM)>\bzero \text{ s.t. } \bM \bM^\T
\bpsi^Y\!(\bM) = \bpsi^Y\!(\bM) \Bigr\} .  
\end{align*}
\end{proposition}

\begin{IEEEproof}
For the ``only if'' part, since $\bM=\dtm(P_{X,Y})$ for some $P_{X,Y}
\in \simpXY$, it follows by construction that $\bM\ge\bzero$, and as
developed in \secref{sec:modal}, $\spectral{\dtm} = 1$ with
corresponding right and left principal singular vectors $\bpsi^X(\bM)$
and $\bpsi^Y(\bM)$ whose elements are $\{\sqrt{P_X},\ x\in\X\}$ and
$\{\sqrt{P_Y},\ y\in\Y\}$, respectively, which are strictly positive
according to our assumption at the outset of this appendix.  As such,
these strictly positive $\bpsi^X(\bM)$ and $\bpsi^Y(\bM)$ must be
eigenvectors of $\dtm^\T \dtm$ and $\dtm \dtm^\T$ corresponding to the
unit eigenvalue.

The ``if'' part follows from the same proof as that for
\propref{prop:dtm-1} mutatis mutandis.  However, we must be careful
when applying the Perron-Frobenius theorem
\cite[Theorem~8.3.1]{hj12} to $\bM \ge \bzero$ as it
only guarantees that the eigenvectors $\bpsi^X\!(\bM)$ and
$\bpsi^Y\!(\bM)$ are entrywise nonnegative.  If an entry of
$\bpsi^X\!(\bM)$ or $\bpsi^Y\!(\bM)$ were zero, then the corresponding
column or row of 
\begin{equation*} 
\bP_{Y,X} =
\diag(\bpsi^Y\!(\bM))\,\bM\,\diag(\bpsi^X\!(\bM)),
\end{equation*}
which defines $P_{X,Y}$, would be zero.  In turn, this would imply that
$P_X(x)=0$ for some $x\in\X$ or $P_Y(y)=0$ for some $y\in\Y$, which
would mean that $P_{X,Y} \not\in \simpXY$, so that $\bM$ could not be
a DTM.   Accordingly, we add the $\bpsi^X\!(\bM) > \bzero$ and
$\bpsi^Y\!(\bM) > \bzero$ conditions in the statement of the
proposition.
\end{IEEEproof}

\begin{remark}
It is worth noting that a nonnegative
square matrix $\bA \ge \bzero$ has strictly positive left and right
eigenvectors corresponding to its Perron-Frobenius eigenvalue (or
spectral radius) $\rho(\bA)$ if and only if the triangular block form of
$\bA$ is a direct sum of irreducible nonnegative square matrices whose
spectral radii are also $\rho(\bA)$---see Theorem~3.14 and the preceding
discussion in \cite[Chapter~2, Section~3]{NonnegativeMatricesBP}. This
means that $\bM^\T \bM$ and $\bM \bM^\T$ have strictly positive
eigenvectors corresponding to their spectral radius of unity if and
only if they have the aforementioned direct form structure after
suitable similarity transformations using permutation matrices.
\end{remark}

Finally, we establish the following.
\begin{proposition}
\label{prop:dtm-3}
The DTM function $\dtm\colon \simpXY \mapsto \dtmset$ is bijective and continuous. 
\end{proposition}

\begin{IEEEproof}
The DTM function $\dtm\colon \simpXY \mapsto \dtmset$ is bijective
because: 1) its range is defined to be $\dtmset$; and 2) the proof of
\propref{prop:dtm-1} (and, in turn, its extension
\propref{prop:dtm-2}) delineates the inverse function.

To prove that $\dtm\colon \simpXY \mapsto \dtmset$ is continuous,
consider any sequence of distributions $\{P_{X,Y}^n \in \simpXY,\ 
n=1,2,\dots\}$ such that for all $(x,y)\in\X\times\Y$
\begin{equation*}
\lim_{n \rightarrow  \infty}{P_{X,Y}^n(x,y)} = P_{X,Y}(x,y) . 
\end{equation*}
By the triangle inequality, we have, for all $x \in \X$,
\begin{align*}
\bigl|P_{X}^n(x) - P_{X}(x)\bigr| & =  \Biggl|\sum_{y \in \Y}{P_{X,Y}^n(x,y) - P_{X,Y}(x,y)}\Biggr| \\
& \leq \sum_{y \in \Y}{\left|P_{X,Y}^n(x,y) - P_{X,Y}(x,y)\right|} ,
\end{align*}
which implies that $P_{X}^n(x) \to P_{X}(x)$ as $n \rightarrow \infty$
for all $x \in \X$.  Likewise, $P_{Y}^n(y)\to P_{Y}(y)$ as
$n\rightarrow \infty$ for all $y \in \Y$.  Hence, we have
\begin{equation*}
\lim_{n \rightarrow
   \infty}{\frac{P_{X,Y}^n(x,y)}{\sqrt{P_{X}^n(x)P_{Y}^n(y)}}} =
 \frac{P_{X,Y}(x,y)}{\sqrt{P_{X}(x)P_{Y}(y)}} 
\end{equation*}
for all $(x,y)\in\X\times\Y$, which means that the elements of
$\dtm\left(P_{X,Y}^n\right)$ converge to the elements of
$\dtm\left(P_{X,Y}\right)$, and where we note that the denominator terms
  are strictly positive according to our assumption at the outset of
  this appendix.  Therefore, the DTM function is continuous.
\end{IEEEproof}

\subsection{Conditional Expectation Operator Representations}
\label{app:ce-rep}

It is reasonable to ask why it is natural to focus on the SVD of the
CDM $\dtmt$ corresponding to $\dtmts$, as opposed to other commonly
used representations of the conditional expectation operator
$P_{X|Y}$, such as simply
\begin{equation*}
\dtms_0(x,y) \defeq P_{X,Y}(x,y),
\end{equation*}
or 
\begin{equation*} 
\dtms_1(x,y) \defeq \frac{P_{X,Y}(x,y)}{P_X(x)\,P_Y(y)},
\end{equation*}
whose logarithm is the pointwise mutual information \cite{ch90} (also
referred to as the information density \cite{hv93}). While fulling
addressing this question is beyond the scope of the present
development, we can show that $\dtmts$ generates inner product spaces
with the ``right'' properties, and that it does so uniquely over a
reasonable class of candidates.

Our characterization takes the form of the following proposition, in which
$\bP_{X|Y}$ is the representation of the
conditional expectation operator $\bE{\cdot|Y=y}$ defined in
\eqref{eq:dtma-alt}.  In particular, expressing a function $f$ as a
length-$\cardX$ (column) vector $\bff$, 
\begin{equation*}
\bE{f(X)|Y=y} = \bP_{X|Y}^\T\bff.
\end{equation*}
\begin{proposition}
\label{prop:dtm-opt}
Define an inner product on $\reals^\cardX$ using a distribution
$Q_X$
\begin{equation*}
\ip{\bff_1}{\bff_2}_{Q_X} \defeq \sum_{x\in\X} Q_X(x)\, f_1(x)\, f_2(x),
\end{equation*}
yielding $\ell^2(\X,Q_X)$, and similarly use $P_Y$ to convert
$\reals^\cardY$ into $\ell^2(\Y,P_Y)$, i.e., 
\begin{equation*}
\ip{\bg_1}{\bg_2}_{P_Y} \defeq \sum_{y\in\Y} P_Y(y)\, g_1(y)\, g_2(y).
\end{equation*}
Then
\begin{equation*}
\min_{Q_X} \max_{\bff\in\ell^2(\X,Q_X)}
\frac{\bnorm{\bP_{X|Y}^\T\bff}_{P_Y}}{\bnorm{\bff}_{Q_X}} = 1,
\end{equation*}
and, moreover,
\begin{equation*}
Q_X^* = P_X = P_Y\,P_{X|Y}
\end{equation*}
is the unique minimizer. 
\end{proposition}
\begin{IEEEproof}
Note that for all $Q_X$ we have $\bone\in\ell^2(\X,Q_X)$ with
$\norm{\bone}_{Q_X}=1$.  Also,
$\bnorm{\bP_{X|Y}^\T\bone}_{P_Y} = \norm{\bone}_{P_Y} = 1$. Hence,
\begin{equation*}
\max_{\bff\in\ell^2(\X,Q_X)}
\frac{\bnorm{\bP_{X|Y}^\T \bff}_{P_Y}}{\bnorm{\bff}_{Q_X}} \ge
  1,\quad\text{for all $P_Y$ and $Q_X$}.
\end{equation*}
But we know that $Q_X=P_X$ achieves the lower bound, which proves the
minimum.   In particular, via Jensen's inequality we have
\begin{equation*} 
\E{\E{f(X) | Y}^2} \le  \E{\E{f(X)^2|Y}} = \E{f(X)^2}.
\end{equation*}

To prove that $P_X$ is the unique minimizer, suppose we use $Q_X\neq
P_X$ for the inner product.  Then consider the adjoint operator, which
for $\bff\in\ell^2(\X,Q_X)$ and $\bg\in\ell^2(\Y,P_Y)$ is defined by
\begin{align*} 
\bEd{P_Y}{\bE{f(X)|Y}\,g(Y)} 
&= \bEd{P_{X,Y}}{f(X)\,g(Y)} \\
&= \bEd{P_X}{f(X)\,\bE{g(Y)|X}} \\
&= \Ed{Q_X}{f(X)\,\bE{g(Y)|X}\, \frac{P_X(X)}{Q_X(X)}}.
\end{align*}
So the adjoint operator is
\begin{equation*}
(P_{X|Y}^*\,g)(x) = \frac{P_X(x)}{Q_X(x)}\, \bEd{P_{Y|X}}{g(Y)|X=x}.
\end{equation*}
Now observe that 
\begin{equation*}
(P_{X|Y}^*\, 1)(x) = \frac{P_X(x)}{Q_X(x)},
\end{equation*}
so $\norm{\bone}_{P_Y}=1$ and
\begin{equation*}
\bnorm{P_{X|Y}^*\,1}_{Q_X}^2 = \sum_{x\in\X} Q_X(x)
\frac{P_X(x)^2}{Q_X(x)^2} = 1\!+\!D_{\chi^2}(Q\|P) > 1,
\end{equation*}
where the last inequality follows because $P_X\neq Q_X$.  Hence the
largest singular value of $P_{X|Y}^*$ is strictly greater than unity.
Hence, $Q_X^*=P_X$ is the unique minimizer.
\end{IEEEproof}

Note that \propref{prop:dtm-opt} shows that given $P_{X,Y}$, the
\emph{only} choice of inner products that make $P_{Y|X}$ and $P_{X|Y}$
adjoints and contractive operators (so that the data processing
inequality is satisfied locally) are those with respect to $P_X$ and
$P_Y$.  It also establishes that if we are given only $P_{X|Y}$, we
are free to choose $P_Y$, but we must choose the corresponding $P_X$
for the other inner product to obtain the required contraction
property.

We comment that the restriction in \propref{prop:dtm-opt} to inner
products corresponding to weighting by distribution is natural.  In
general, each inner product corresponds to a positive semidefinite
matrix $\bA$, i.e., $\ip{\bff}{\bg}_{\bA} = \bff^\T \bA \bg$.  For
simplicity, we neglect the orthogonal matrices in the spectral
decomposition of $\bA$, and only consider diagonal matrices $\bA$ with
strictly positive diagonal entries, which correspond to weighted inner
products.  Moreover, we restrict the diagonal entries to sum to unity
to have a ``well-defined'' problem (indeed, allowing arbitrary scaling
would make the infimum in our proposition zero).

\addtocounter{section}{1}
\section{Appendices for \protect\secref{sec:geom-simplex}}
\label{app:geom-simplex}

\subsection{Proof of \protect\propref{prop:h-P-equiv}}
\label{app:h-P-equiv}

The first part of the proposition is immediate: from
\eqref{eq:P-to-h} we obtain both
\begin{equation*}
\bEd{\refgen}{h(Z)} 
= \sum_{z\in\Z} \refgen(z)\, h(z) 
= \frac1{\eps} \sum_{z\in\Z} \bigl( P(z) - \refgen(z) \bigr) = 0,
\end{equation*}
and 
\begin{equation}
\fvgen(z) = \sqrt{P_0(z)}\, h(z) =
\frac{P(z)-\refgen(z)}{\eps\sqrt{P_0(z)}} = \ivgen(z),
\end{equation}
where we have further used \eqref{eq:fvgen-def}, and where to obtain
the last equality we have used \eqref{eq:ivgen-def}.  The second part of
the proposition is trivially true when $h\equiv0$.  When $h\not\equiv0$, it
suffices to note that $P$ in \eqref{eq:h-to-P} satisfies
\begin{equation*}
\sum_{z\in\Z} P(z) = \sum_{z\in\Z} \refgen(z) + \eps\, \bEd{\refgen}{h(Z)} = 1,
\end{equation*}
and that $P(z)\in[0,1]$ whenever
\begin{equation*}
\eps \le \min\left\{ \frac{-1}{\ds\min_{z\in\Z} h(z)} \, ,\,
\frac{\ds1-\max_{z\in\Z} \refgen(z)}{\ds\max_{z\in\Z} h(z)\, \max_{z\in\Z}
  \refgen(z)}\right\}, 
\end{equation*}
where we have used that since $h\not\equiv0$,
\begin{equation*}
\min_{z\in\Z} h(z) <0\quad\text{and}\quad \max_{z\in\Z} h(z)>0,
\end{equation*}
and that since $\refgen\in\relint(\simpZ)$, 
\begin{equation*}
\max_{z\in\Z} \refgen(z) < 1.
\end{equation*}
Finally, using, in turn, \eqref{eq:ivgen-def}, \eqref{eq:h-to-P}, and
\eqref{eq:fvgen-def},  we obtain
\begin{equation*}
\ivgen(z) = 
\frac{P(z) - \refgen(z)}{\eps\sqrt{\refgen(z)}} =
\sqrt{\refgen(z)}\, h(z)= \fvgen(z).
\end{equation*}
\hfill\IEEEQED

\subsection{Proof of \protect\corolref{corol:ll-equiv}}
\label{app:ll-equiv}

To obtain the first part of the corollary, we have that given $P$ there exists
$h$ such that 
\begin{equation} 
\frac1{\eps}\, \log\frac{P(z)}{\refgen(z)} 
= \frac1{\eps} \log\bigl(1+\eps h(z)\bigr)
= h(z) + h_\eps(z),
\label{eq:LL-h-rel}
\end{equation}
with $h_\eps(z)$ denoting an $o(1)$ term, where to obtain the first
equality we have used \eqref{eq:P-to-h}, and to obtain the second
equality we have used the first-order Taylor series approximation
$\log(1+\omega)=\omega+o(\omega)$.  In turn, using
\eqref{eq:fvgen-0mean} it follows that
\begin{equation} 
h_\mathrm{LL}(z) = h(z) + \hti_\eps(z)
\label{eq:hLL-h-rel}
\end{equation}
where $\hti_\eps$ is a function such that 
\begin{equation} 
\hti_\eps(z)=o(1),\ \eps\to0,\ z\in\Z
\quad\text{and}\quad \Ed{\refgen}{\hti_\eps(Z)}=0.
\label{eq:ht-char}
\end{equation}
Multiplying both sides of \eqref{eq:hLL-h-rel} by
$\sqrt{\refgen(z)}$ yields the \eqref{eq:LL-phi}.

To obtain the second part of the corollary, we use from
\eqref{eq:h-to-P} that given $h$ satisfying \eqref{eq:fvgen-0mean}
there exists $P$ such that \eqref{eq:LL-h-rel} holds for sufficiently
small $\eps$. Subtracting the mean with respect to $\refgen$ from
\eqref{eq:LL-h-rel} then yields \eqref{eq:h-LL}. \hfill\IEEEQED

\subsection{Proof of \protect\lemref{lem:Eh-char}}
\label{app:Eh-char}

We have
\begin{align} 
\Ed{\distgen}{h(Z)}
&= \sum_{z\in\Z} \distgen(z) \, h(z) \notag\\
&= \sum_{z\in\Z} \bigl(\refgen(z) + \eps \sqrt{\refgen(z)} \,
\, \ivgen(z)\bigr) \, \frac{\fvgen(z)}{\sqrt{\refgen(z)}} \label{eq:use-ivgen-def-first}\\ 
&= \sum_{z\in\Z} \sqrt{\refgen(z)}\,\fvgen(z) + 
\eps \sum_{z\in\Z} \ivgen(z)\, \fvgen(z) \label{eq:two-terms}\\ 
&= \eps \bip{\ivgen}{\fvgen}, \label{eq:use-fvgen-0mean}
\end{align}
where to obtain \eqref{eq:use-ivgen-def-first} we have used
\eqref{eq:ivgen-def} and \eqref{eq:fvgen-def}, and to obtain
\eqref{eq:use-fvgen-0mean} we have used that \eqref{eq:fvgen-0mean},
i.e., $\fvgen\in\ivspacegen(\refgen)$, 
implies that the first term in \eqref{eq:two-terms} is zero in
accordance with \eqref{eq:ivspace-def}. \hfill\IEEEQED

\subsection{Proof of Lemma~\ref{lem:D-char}}
\label{app:div-deriv}

With the feature functions
\begin{equation*}
L_i(z) \defeq \frac1{\eps} \left( \frac{P_i(z)}{\refgen(z)}-1 \right)
\end{equation*}
we have, for $i=1,2$,
\begin{align} 
\log \frac{P_i(z)}{\refgen(z)}
&= \log \bigl( 1+ \eps L_i(z) \bigr) \notag\\
&= \eps L_i(z) - \frac12 \eps^2 L_i(z)^2 + o(\eps^2) \label{eq:use-taylor-2}\\
&= \eps \frac{\phi_i(z)}{\sqrt{\refgen(z)}} - \frac12 \eps^2
  \frac{\phi_i(z)^2}{\refgen(z)} + o(\eps^2),
\label{eq:use-prop-first}
\end{align}
where to obtain \eqref{eq:use-taylor-2} we have used
the second-order Taylor series approximation
\begin{equation*}
\log (1+\omega) = \omega - \frac12 \omega^2 + o(\omega^2),\quad\text{as $\omega\to0$},
\end{equation*}
and where to obtain \eqref{eq:use-prop-first} we have used
\eqref{eq:phi-to-fvgen} from the first part of
\propref{prop:h-P-equiv}.
Hence, 
\begin{align*} 
\log \frac{P_1(z)}{P_2(z)}
&= \log\frac{P_1(z)}{\refgen(z)} - \log\frac{P_2(z)}{\refgen(z)} \\
&= \eps \frac{\phi_1(z)-\phi_2(z)}{\sqrt{\refgen(z)}} - \frac12 \eps^2
  \frac{\phi_1(z)^2-\phi_2(z)^2}{\refgen(z)} + o(\eps^2),
\end{align*}
and, in turn,
\begin{align*} 
&D(P_1\|P_2) \\
&\ = \sum_{z\in\Z} P_1(z)  \log \frac{P_1(z)}{P_2(z)} \\
&\ = \sum_{z\in\Z} \refgen(z)  \log \frac{P_1(z)}{P_2(z)} + 
\sum_{z\in\Z} \bigl(P_1(z)-\refgen(z)\bigr)  \log \frac{P_1(z)}{P_2(z)} \\
&\ = \sum_{z\in\Z} \refgen(z) \, \eps \,
\frac{\phi_1(z)-\phi_2(z)}{\sqrt{\refgen(z)}} \\
&\qquad {} - \sum_{z\in\Z} \refgen(z)\, \frac12\eps^2 \,
\frac{\phi_1(z)^2-\phi_2(z)^2}{\refgen(z)} \\ 
&\qquad\qquad {} + \sum_{z\in\Z} \eps \sqrt{\refgen(z)} \, \phi_1(z) \, \eps
  \frac{\phi_1(z)-\phi_2(z)}{\sqrt{\refgen(z)}} \\
&\qquad\qquad\qquad {} - \sum_{z\in\Z} \eps \sqrt{\refgen(z)} \,
  \phi_1(z)\, \frac12\eps^2 \,
  \frac{\phi_1(z)^2-\phi_2(z)^2}{\refgen(z)}  \\
&\qquad\qquad\qquad\qquad{} + o(\eps^2) \\
&\ = 0-\frac12\eps^2 \sum_{z\in\Z} \bigl( \phi_1(z)^2 -\phi_2(z)^2
  \bigr) \\
&\qquad\qquad{}+ \eps^2\sum_{z\in\Z} \phi_1(z) \bigl( \phi_1(z) -
  \phi_2(z) \bigr) + o(\eps^2) \\ 
&\ = \frac12\eps^2 \bigl[ \|\phi_2\|^2 -\|\phi_1\|^2 + 2\|\phi_1\|^2 -
    2\ip{\phi_1}{\phi_2} \bigr] + o(\eps^2) \\
&\ = \frac12 \eps^2 \|\phi_1-\phi_2\|^2 + o(\eps^2).
\end{align*}
\hfill\IEEEQED

\subsection{Proof of \protect\lemref{lem:refdist-invar}}
\label{app:refdist-invar}

To obtain \eqref{eq:refdist-invar}, it suffices to note that since
\begin{equation*}
\Pt_0(z) = P_0(z) + \eps\,\sqrt{P_0(z)}\, \phit_0(z)
\end{equation*}
for some $\phit_0(z)$ such that $\norm{\phit_0}\le 1$, we have, for $i=1,2$,
\begin{align*}
\phit_i(z) 
&= \frac{P_i(z) - (P_0(z) +
  \eps\,\sqrt{P_0(z)}\,\phit_0(z))}{\eps\,\sqrt{P_0(z) 
   + \eps\,\sqrt{P_0(z)}\,\phit_0(z)}} \\
&= \frac{\bigl(P_i(z)-P_0(z)\bigr)-\eps\,\sqrt{P_0(z)}\,\phit_0(z)}{\eps\,\sqrt{P_0(z)}\,
\sqrt{1 + \eps\,\frac{\phit_0(z)}{\sqrt{P_0(z)}}}}\\
&= \bigl(\phi_i(z)-\phit_0(z)\bigr)\,\bigl(1+o(1)\bigr),
\end{align*}
where to obtain the last equality we have used that $(1+\omega)^{-1/2} =
1 + o(1)$ as $\omega\to0$.  \hfill\IEEEQED

\subsection{Proof of \protect\lemref{lem:eps-dep-equiv}}
\label{app:eps-dep-equiv}

It suffices to note that since
\begin{align*}
D_{\chi^2}\bigl(P_{Z,W} \bigm\| P_{Z}P_{W}\bigr)
&= \Ed{P_Z}{D_{\chi^2}\bigl(P_{W|Z}(\cdot|Z) \bigm\| P_{W}\bigr)} \notag\\
&= \Ed{P_W}{D_{\chi^2}\bigl(P_{Z|W}(\cdot|W)\bigm\| P_{Z} \bigr)},
\end{align*}
we have
for all $z\in\Z$,
\begin{align*} 
&D_{\chi^2}\bigl(P_{W|Z}(\cdot|z) \bigm\| P_{W}\bigr) \,
\min_{z'\in\Z} P_{Z}(z') \notag\\
&\qquad\qquad\le D_{\chi^2}\bigl(P_{Z,W} \bigm\| P_{Z}P_{W}\bigr) \notag\\
&\qquad\qquad\qquad\qquad\le 
\max_{z'\in\Z} D_{\chi^2}\bigl(P_{W|Z}(\cdot|z') \bigm\| P_{W}\bigr),\\
\intertext{and, similarly, for all $w\in\W$,}
&D_{\chi^2}\bigl(P_{Z|W}(\cdot|w)\bigm\| P_{Z} \bigr) \,
\min_{w'\in\W} P_{W}(w') \notag\\
&\qquad\qquad\le D_{\chi^2}\bigl(P_{Z,W} \bigm\| P_{Z}P_{W}\bigr) \notag\\
&\qquad\qquad\qquad\qquad\le 
\max_{w'\in\W} D_{\chi^2}\bigr(P_{Z|W}(\cdot|w') \bigm\| P_{Z} \bigr),
\end{align*}
where both the constituent minima are finite and nonzero as $\eps\to0$
due to \eqref{eq:PZ-PW-const}.
\hfill\IEEEQED

\subsection{Proof of \protect\lemref{lem:weak-dep-KL}}
\label{app:weak-dep-KL}

The ``if'' part of the lemma follows from using 
$O(\eps)$-dependence between $Z$
and $W$ in the form 
\eqref{eq:Oeps-dep-Z|W} with \lemref{lem:D-char} to obtain
\begin{equation} 
D(P_{Z|W}(\cdot|w)\|P_Z) 
= \frac{\eps^2}{2}\bnorm{\bphi^{Z|W}_w}^2 +o(\eps^2),
\quad\text{$w\in\W$},
\label{eq:D-dep}
\end{equation}
where for each $w\in\W$,
\begin{equation*}
\ivgen^{Z|W}_w(z) \defeq \frac{P_{Z|W}(z|w)-P_Z(z)}{\eps\,\sqrt{P_Z(z)}},\quad
z\in\Z
\end{equation*}
is the information vector associated with $P_{Z|W}(\cdot|w)$, and for
which $\bnorm{\bivgen^{Z|W}_w}\le1$.  Alternatively, the inequality 
\begin{equation*} 
D(p\|q)\le
\ln\bigl(1+D_{\chi^2}(p\|q)\bigr)\le D_{\chi^2}(p\|q),
\end{equation*}
valid for all finite $\Z$ and $p,q\in\simpZ$,
which is derived in, e.g., \cite[Theorem~5]{gs02}, is sufficient to
obtain this part of the lemma, using $p=P_{Z,W}$ and $q=P_ZP_W$.

To obtain the ``only if'' part of the lemma, note that for any finite
$\Z$ and $p,q\in\simpZ$, we have, using Pinsker's inequality
\cite{ck11},
\begin{align} 
D_{\chi^2}(p\|q) 
&\le \frac1{\min_{z\in\Z} q(z)} \sum_{z\in\Z} \bigl( p(z)-q(z)\bigr)^2
\notag\\
&\le \frac1{\min_{z\in\Z} q(z)} \left( \sum_{z\in\Z}
\babs{p(z)-q(z)}\right)^2 \notag\\
&\le \frac{2 D(p\|q)}{\min_{z\in\Z} q(z)}.
\label{eq:chi2-ub}
\end{align}
The result then follows setting, again, $p=P_{Z,W}$ and $q=P_ZP_W$, since
the minimum in \eqref{eq:chi2-ub} is finite and nonzero due to 
\eqref{eq:PZ-PW-const}.
\hfill\IEEEQED

\subsection{Proof of \protect\lemref{lem:KL-eps-dep-equiv}}
\label{app:KL-eps-dep-equiv}

It suffices to note that since
\begin{align*}
I(Z;W)
&\defeq D\bigl(P_{Z,W} \bigm\| P_{Z}P_{W}\bigr) \\
&= \Ed{P_Z}{D\bigl(P_{W|Z}(\cdot|Z) \bigm\| P_{W}\bigr)} \notag\\
&= \Ed{P_W}{D\bigl(P_{Z|W}(\cdot|W)\bigm\| P_{Z} \bigr)},
\end{align*}
we have
for all $z\in\Z$,
\begin{align*} 
&D\bigl(P_{W|Z}(\cdot|z) \bigm\| P_{W}\bigr) \,
\min_{z'\in\Z} P_{Z}(z') \notag\\
&\qquad\qquad\le I(Z;W)\le 
\max_{z'\in\Z} D\bigl(P_{W|Z}(\cdot|z') \bigm\| P_{W}\bigr),\\
\intertext{and, similarly, for all $w\in\W$,}
&D\bigl(P_{Z|W}(\cdot|w)\bigm\| P_{Z} \bigr) \,
\min_{w'\in\W} P_{W}(w') \notag\\
&\qquad\qquad\le I(Z;W) \le 
\max_{w'\in\W} D\bigr(P_{Z|W}(\cdot|w') \bigm\| P_{Z} \bigr),
\end{align*}
where both the constituent minima are finite and nonzero as $\eps\to0$
due to \eqref{eq:PZ-PW-const}.
\hfill\IEEEQED

\subsection{Proof of \lemref{lem:mismatch-k}}
\label{app:mismatch-k}

Since the rule is to decide based on comparing the projection 
\begin{equation*}
\sum_{i=1}^k \ell_i \, \Bigl( \bEd{P_1}{h_i(Z)}-\bEd{P_2}{h_i(Z)} \Bigr)
\end{equation*}
to a threshold, via Cram\'er's Theorem \cite{dz98} the error exponent
under $P_i$ is
\begin{equation}
E_i(\la) = \min_{P\in\cS(\la)} D(P\|P_i),
\label{eq:Ei-def}
\end{equation}
where 
\begin{align} 
\cS(\la) &\defeq \bigl\{ P\in\simpZ \colon \notag \\
&\Ed{P}{h^k(Z)} = 
\la \, \Ed{P_1}{h^k(Z)} + (1-\la) \, \Ed{P_2}{h^k(Z)} \bigr\}.
\label{eq:cSk-def}
\end{align}

Now since \eqref{eq:h-0mean} holds, 
from \lemref{lem:Eh-char} we obtain
\begin{equation*}
\Ed{P_i}{h_l(Z)} = \eps\, \ip{\phi_i}{\fvgen_l}, \quad \text{ $i=1,2$ and
  $l =1,\dots,k$},  
\end{equation*}
which we express compactly as 
\begin{equation*}
\Ed{P_i}{h^k(Z)} = \eps \, \ip{\phi_i}{\fvgen^k},\quad i=1,2.
\end{equation*}
Hence, the constraint \eqref{eq:cSk-def} is expressed in information
space as
\begin{equation*} 
\ip{\phi}{\fvgen_l} 
= \ip{\la \, \phi_1 + (1-\la) \, \phi_2}{\fvgen_l}, \quad l = 1,
\dots, k, 
\end{equation*}
i.e.,
\begin{equation} 
\ip{\phi}{\fvgen^k} 
= \ip{\la \, \phi_1 + (1-\la) \, \phi_2}{\fvgen^k}.
\label{eq:cSk-interp}
\end{equation}
In turn, the optimizing $P$ in \eqref{eq:Ei-def}, which we denote by
$P^*$, lies in the exponential family through $P_i$ with natural
statistic $h^k(z)$, i.e., the $k$-dimensional family whose members are
of the form
\begin{equation*}
\log \Pt_{\theta^k}(z) = \sum_{l=1}^k \theta_l \, h_l(z) + \log P_i(z)
- \alpha(\theta^k),
\end{equation*}
for which the associated information vector is
\begin{equation*} 
\eps \, \phit_{\theta^k}(z) = \sum_{l=1}^k \theta_l \, \fvgen_l(z) +
\eps\, \phi_i(z) 
-\alpha(\theta^k) \sqrt{\refgen(z)} + o(\eps),
\end{equation*}
so
\begin{equation*}
\eps \, \ip{\phit_{\theta^k}}{\fvgen_l} = \theta_l + \eps \,
\ip{\phi_i}{\fvgen_l} + o(\eps),
\end{equation*}
where we have used \eqref{eq:h-uncorr}.
Hence, via \eqref{eq:cSk-interp} we obtain
that the intersection with the linear family \eqref{eq:cSk-def} is at
$P^*=P_{{\theta^k}^*}$ with 
\begin{equation*}
\theta^*_l = \eps \, \ip{\la\, \phi_1 + (1-\la)\,\phi_2 - \phi_i}{\fvgen_l} + o(\eps),
\end{equation*}
and thus 
\begin{align} 
E_i(\la)
&= D(P^*\|P_i) \notag\\
&= \frac12 \biggl\|\sum_{l=1}^k \theta_k^* \, \fvgen_l\biggr\|^2 + \frac12
\alpha({\theta^k}^*)^2 + o(\eps^2) \notag\\ 
&= \frac12 \sum_{l=1}^k \bigl(\theta_k^*\bigr)^2 + \frac12
\alpha({\theta^k}^*)^2 + o(\eps^2) \notag\\ 
&= \frac{\eps^2}{2} \sum_{l=1}^k \ip{\la\,\phi_1 + (1-\la)\,\phi_2 -
\phi_i}{\fvgen_l}^2 + o(\eps^2),
\end{align}
where to obtain the second equality we have again exploited
\eqref{eq:h-uncorr}, and
where to obtain the last equality we have used that 
\begin{equation*}
\alpha\bigl({\theta^k}^*\bigr) = o(\eps^2)
\end{equation*}
since ${\theta^k}^*=\bigO(\eps)$ and
\begin{equation*}
\alpha(0)=0,\quad\text{and}\quad
\nabla\alpha(0) =
\bEd{P_i}{h^k(Z)} = \eps\,\ip{\phi_i}{\fvgen^k} = \bigO(\eps).
\end{equation*}
Finally, $E_1(\la)=E_2(\la)$ when $\la=1/2$, so the overall error probability
has exponent \eqref{eq:mm-exp-k}.  \hfill\IEEEQED

\section{Appendices for \protect\secref{sec:unifeature-char}}
\label{app:unifeature-char}

\subsection{Proof of \protect\lemref{lem:orthog}}
\label{app:orthog}

Via \eqref{eq:phiZWi-def}, we have
\begin{align*} 
&\sum_{z\in\Z} \phi^{Z|W_i}_{w_i}(z)\, \phi^{Z|W_j}_{w_j}(z) \notag\\
&\qquad\qquad = \frac1{\eps^2} \biggl[ \sum_{z\in\Z}
\frac{P_{Z|W_i}(z|w_i)\,P_{Z|W_j}(z|w_j)}{P_Z(z)} \notag\\
&\qquad\qquad\qquad\qquad{} - \sum_{z\in\Z}
\bigl( P_{Z|W_i}(z|w_i)  + P_{Z|W_j}(z|w_j)\bigr) \notag\\
&\qquad\qquad\qquad\qquad\qquad{} + \sum_{z\in\Z} P_Z(z) \biggr] = 0,
\end{align*}
where the first sum within the brackets is 1 since, using the pairwise
marginal and conditional independencies,
\begin{align*}
&\frac{P_{Z|W_i}(z|w_i)\,P_{Z|W_j}(z|w_j)}{P_Z(z)}\\
&\qquad\qquad\qquad=
\frac{P_{W_i|Z}(w_i|z)\,P_{W_j|Z}(w_j|z)\,P_Z(z)}{P_{W_i}(w_i)\,P_{W_j}(w_j)}
\\
&\qquad\qquad\qquad= \frac{P_{W_i,W_j|Z}(w_i,w_j|z)\,P_Z(z)}{P_{W_i,W_j}(w_i,w_j)}\\
&\qquad\qquad\qquad= P_{Z|W_i,W_j}(z|w_i,w_j).
\end{align*}
\hfill\IEEEQED

\subsection{Proof of \protect\lemref{lem:ivma-sum}}
\label{app:ivma-sum}

Due to the conditional independence among the $W^k$,
\begin{align}
P_{Z|W^k}(z|w^k) &= \frac{P_Z(z)}{P_{W^k}(w^k)} \, \prod_{i=1}^k
P_{W_i|Z}(w_i|z) \notag\\
&= \frac{P_Z(z)}{\pi\bigl(w^k\bigr)} \, \prod_{i=1}^k
\frac{P_{Z|W_i}(z|w_i)}{P_Z(z)},  
\label{eq:bayes-feature-set}
\end{align}
with
\begin{equation*} 
\pi\bigl(w^k\bigr) = \frac{P_{W^k}(w^k)}{\prod_{i=1}^k P_{W_i}(w_i)}
= \sum_{z'} P_Z(z') \prod_{i=1}^k
\frac{P_{Z|W_i}(z'|w_i)}{P_Z(z')}.
\end{equation*}
Moreover, 
\begin{align} 
&P_Z(z)\prod_{i=1}^k
\frac{P_{Z|W_i}(z|w_i)}{P_Z(z)} \notag\\
&\quad= P_Z(z) \prod_{i=1}^k \left( 1 + \frac{\eps}{\sqrt{P_Z(z)}}\,
  \phi^{Z|W_i}_{w_i} \right) \notag\\
&\quad= P_Z(z) + \eps\, \sqrt{P_Z(z)}\, \sum_{i=1}^k
  \phi^{Z|W_i}_{w_i} + o(\eps),
\label{eq:presum}
\end{align}
where to obtain \eqref{eq:presum} we have used \factref{fact:prod-sum}.
In turn, summing \eqref{eq:presum} over $z$ we obtain
\begin{equation}
\pi\bigl(w^k\bigr) = 1 + o(\eps),
\label{eq:Zuk-exp}
\end{equation}
where we have used that since $\phi^{Z|W_i}_{w_i}\in\ivspacegen$,
\begin{equation*}
\sum_z \sqrt{P_Z(z)}\, \phi^{Z|W_i}_{w_i}(z) = 0.
\end{equation*}

Hence, using \eqref{eq:presum} and \eqref{eq:Zuk-exp}
with \eqref{eq:pxuk-def} 
in \eqref{eq:bayes-feature-set}, we obtain \eqref{eq:ivma-sum}.
\hfill\IEEEQED

\subsection{Proof of \protect\lemref{lem:rie}}
\label{app:rie}

First, note that the $(i,j)$\/th entry of $\bA_1^\T\bZ\,\bA_2$ is
$\ba_{1,i}^\T \bZ \ba_{2,j}$, 
where $\ba_{1,i}$ and $\ba_{2,j}$ denote the $i$\/th and $j$\/th
columns of $\bA_1$ and $\bA_2$, respectively.  Hence,
\begin{align} 
\E{\bfrob{\bA_1^\T\bZ\,\bA_2}^2} 
&= \ev\biggl[\sum_{i,j} \bigl( \ba_{1,i}^\T \,\bZ \,\ba_{2,j}
  \bigr)^2\biggr] \notag\\ 
&= \sum_{i,j} \E{\bigl(\ba_{1,i}^\T \bZ\, \ba_{2,j}\bigr)^2}.
\label{eq:frob-exp}
\end{align}
Next, with $Z_{ij}$ denoting the $(i,j)$\/th element of $\bZ$, note
that 
\begin{align} 
\E{ \bigl(\ba_{1,i}^\T\bZ\,\ba_{2,j}\bigr)^2}
&= \ev\Bigl[ \bigl(\underbrace{\ba_{1,i}^\T\bQ_{1,i}^\T}_{\defeq
      \tilde{\ba}_{1,i}^\T}\,\bZ\,\underbrace{\bQ_{2,j}\,\ba_{2,j}}_{\defeq
      \tilde{\ba}_{2,j}^{\vphantom{\T}}}\bigr)^2 \Bigr] \label{eq:quad-xfm}\\
&= \bnorm{\ba_{1,i}}^2\, \bnorm{\ba_{2,j}}^2\,
\ev\Bigl[\bigl(\underbrace{\bfe_1^\T\bZ\,\bfe_1}_{=Z_{11}}\bigr)^2\Bigr] 
\label{eq:quad-val}
\end{align}
where to obtain \eqref{eq:quad-xfm} we have used \defref{def:sphere-sym} with
orthogonal matrices $\bQ_{1,i}$ and $\bQ_{2,j}$, and to obtain
\eqref{eq:quad-val} we have chosen $\bQ_{1,i}$ and $\bQ_{2,j}$ so that 
\begin{equation*}
\tilde{\ba}_{1,i} = \bnorm{\ba_{1,i}}\, \bfe_1 \qquad\text{and}\qquad
\tilde{\ba}_{2,j} = \bnorm{\ba_{2,j}}\, \bfe_1.
\end{equation*}
In turn, substituting \eqref{eq:quad-val} into \eqref{eq:frob-exp} yields
\begin{align} 
\E{\bfrob{\bA_1^\T\bZ\,\bA_2}^2} 
&= \E{Z_{11}^2} \, \sum_{i,j} \bnorm{\ba_{1,i}}^2\, \bnorm{\ba_{2,j}}^2 \notag\\
&= \E{Z_{11}^2} \, \sum_i \bnorm{\ba_{1,i}}^2 \, \sum_j \bnorm{\ba_{2,j}}^2 \notag\\
&= \E{Z_{11}^2} \, \bfrob{\bA_1}^2\, \bfrob{\bA_2}^2.
\label{eq:frob-form}
\end{align}
Finally, with $\tilde{\bQ}_l$ denoting the permutation matrix that
interchanges the first and $l$\/th columns of the identity matrix, it
follows from \defref{def:sphere-sym} with $\bQ_1=\tilde{\bQ}_i$ and
$\bQ_2=\tilde{\bQ}_j$ that $Z_{ij}\smash{\eqd}\vphantom= Z_{11}$, and thus
\begin{equation*}
\E{\bfrob{\bZ}^2} = k_1 k_2 \, \E{Z_{11}^2},
\end{equation*}
which when used in conjunction with \eqref{eq:frob-form} yields
\eqref{eq:rie-prop}.
\hfill\IEEEQED

\subsection{Proof of \protect\propref{prop:mpe-rie}}
\label{app:mpe-rie}

Without loss of generality, as discussed in \secref{sec:geom-inf} we
assume that $f^k$ and $g^k$ are normalized according to
\eqref{eq:cF-def} and \eqref{eq:cG-def}, so the
associated feature vectors $\bfVX$ and $\bfVY$, respectively, satisfy
\eqref{eq:fvXY-orthon}.

We first analyze the error probability in decisions about the value of
$V$ based on $S^k$.  To begin, we have
\begin{align}
\Eb^{V|S}(f^k) 
&=\lim_{m\to\infty}\frac{-\Ed{\mathrm{RIE}}{\log\pe^{V|S}(\C^Y_\eps(P_Y),f^k)}}{m} \notag\\
&= \Ed{\mathrm{RIE}}{\lim_{m\to\infty}\frac{-\log
      \pe^{V|S}(\C^Y_\eps(P_Y),f^k,v_*,v_*')}{m}} \label{eq:use-pairwise}\\
&= \bEd{\mathrm{RIE}}{E^{V|S}(\C^Y_\eps(P_Y),f^k,v_*,v_*')},
\label{eq:useEVS-pairwise-notation}
\end{align}
where to obtain the
\eqref{eq:use-pairwise} we have used standard pairwise exponent analysis.
Specifically, (with slight abuse of notation)
$\pe^{V|S}(\C^Y_\eps(P_Y),f^k,v,v')$ denotes  
the pairwise error probability distinguishing
distinct $v$ and $v'$ in $\V$ based on $s^k$, and
\begin{equation}
(v_*,v_*') = \argmin_{\{v,v'\in\V\colon v\neq v'\}}
  \pe^{V|S}(\C^Y_\eps(P_Y),f^k,v,v'),
\end{equation}
whose dependence on $f^k$ and $\C^X_\eps(P_X)$ we leave implicit in
our notation.  Finally, in \eqref{eq:useEVS-pairwise-notation} we have used
the notation
\begin{align*} 
&E^{V|S}(\C^Y_\eps(P_Y),f^k,v,v') \notag\\
&\qquad\qquad\qquad\defeq
\lim_{m\to\infty}\frac{-\log
      \pe^{V|S}(\C^Y_\eps(P_Y),f^k,v,v')}{m}
\end{align*}
for any distinct $v$ and $v'$.

Now
\begin{align} 
&E^{V|S}(\C^Y_\eps(P_Y),f^k,v,v') \notag\\
&\qquad= \frac{\eps^2}{8} \sum_{i=1}^k \left( \bigl(
\bphi^{X|V}_{v}-\bphi^{X|V}_{v'} \bigr)^\T \bfvX_i \right)^2 
+ o(\eps^2) \label{eq:use-lem-mismatch-k} \\  
&\qquad= \frac{\eps^2}{8} \bnorm{\bigl(\bfVX\bigr)^\T\dtm^\T
  \bigl(\bphi^{Y|V}_{v}-\bphi^{Y|V}_{v'}\bigr)}^2 +
o(\eps^2) \label{eq:use-Y|V-to-X|V} \\
&\qquad= \frac{\eps^2}{8} \bnorm{\bigl(\bfVX\bigr)^\T\dtm^\T
  \bPhi^{Y|V}\bigl(\bfe_{v}-\bfe_{v'}\bigr)}^2 + o(\eps^2),
\label{eq:Efg-char-k}
\end{align}
where to obtain \eqref{eq:use-lem-mismatch-k} we have used
\lemref{lem:mismatch-k} with $\phi^{Y|V}_v$ and $\phi^{Y|V}_{v'}$ as
defined in \eqref{eq:phiYV-def}, to obtain \eqref{eq:use-Y|V-to-X|V}
we have used \eqref{eq:Y|V-to-X|V}, and in \eqref{eq:Efg-char-k} we
have exploited elementary vector notation (with an abuse of notation
as discussed in footnote \ref{fn:abuse}).
Moreover, for fixed $v$ and $v'$,
\begin{align} 
&\bEd{\mathrm{RIE}}{ E^{V|S}(\C^Y_\eps(P_Y),f^k,v,v') }  \notag\\
&\ = \Ed{\mathrm{RIE}}{ 
\frac{\eps^2}{8}
  \bnorm{\bigl(\bfVX\bigr)^\T\dtm^\T\bPhi^{Y|V}\bigl(\bfe_v-\bfe_{v'}\bigr)}^2
  }  + o(\eps^2) \notag\\ 
&\ =  \frac{\eps^2\,\Ed{\mathrm{RIE}}{\bfrob{\bPhi^{Y|V}}^2}}{4\,\cardY\,\cardV}
  \bfrob{\dtm\,\bfVX}^2 + o(\eps^2) \label{eq:use-lem-rie-first}
\end{align}
where to obtain \eqref{eq:use-lem-rie-first} we have used \lemref{lem:rie}.

Then since \eqref{eq:use-lem-rie-first} does not depend on $v$ or $v'$,
it follows from the law of total expectation that
\eqref{eq:useEVS-pairwise-notation} satisfies
\begin{align} 
&\bEd{\mathrm{RIE}}{ E^{V|S}(\C^Y_\eps(P_Y),f^k,v_*,v_*') }  \notag\\
&\ =
  \eps^2\,\underbrace{\frac{\Ed{\mathrm{RIE}}{\bfrob{\bPhi^{Y|V}}^2}}{4\,\cardY\,\cardV}}_{\defeq
    \Eb_0^{Y|V}} 
  \bfrob{\dtm\,\bfVX}^2 + o(\eps^2) \\
&\ \le \Eb_0^{Y|V} \, \eps^2 \sum_{i=1}^k \sigma_i^2 + o(\eps^2),
\label{eq:EEfg-k}
\end{align}
where to obtain \eqref{eq:EEfg-k} we have used \lemref{lem:eig-k} with
the relevant constraint in \eqref{eq:fvXY-orthon} induced by our choice of
normalization \eqref{eq:cF-def}.  Moreover, the
inequality in \eqref{eq:EEfg-k} holds with equality when we choose
$\bfVX$ according to \eqref{eq:bfVX-opt}, i.e., the optimal features
are $f^k=f^k_*$.

We analyze the error probability in decisions about $V$ based on $T^k$
similarly.   In particular, we obtain 
\begin{align} 
\Eb^{V|T}(g^k) 
&=\lim_{m\to\infty}\frac{-\Ed{\mathrm{RIE}}{\log\pe^{V|T}(\C^Y_\eps(P_Y),g^k)}}{m}
\notag\\  
&=  \frac{\eps^2\,\Ed{\mathrm{RIE}}{\bfrob{\bPhi^{Y|V}}^2}}{4\,\cardY\,\cardV}
  \bfrob{\bfVY}^2 + o(\eps^2) \\
&= \Eb_0^{Y|V} \, \eps^2\, k + o(\eps^2),
\end{align}
for any $\bfVY$ satisfying \eqref{eq:fvXY-orthon}, i.e., any choice of
(normalized) $g^k$.

We obtain the error probability in decisions about the value of
$U$ from symmetry considerations.  In particular, it suffices to interchange
the roles of $U$ and $V$, and $X$ and $Y$---so
$\dtms$ is replaced with its adjoint---in the preceding analysis,
which yields that 
\begin{equation*}
\Eb^{U|T}(g^k) \le
\underbrace{\frac{\Ed{\mathrm{RIE}}{\bfrob{\bPhi^{X|U}}^2}}{4\,\cardX\,\cardU}}_{\defeq
    \Eb_0^{X|U}} \, \eps^2 \sum_{i=1}^k \sigma_i^2 + o(\eps^2), 
\end{equation*}
with equality when $g^k=g^k_*$, and 
\begin{equation*}
\Eb^{U|S}(f^k) = \Eb_0^{X|U} \, \eps^2\, k + o(\eps^2),
\end{equation*}
for any choice of (normalized) $f^k$.

It follows that the inequalities in \eqref{eq:pe-rie-pareto}
simultaneously all hold with equality for 
the choices $f^k=f^k_*$ and $g^k=g^k_*$.\hfill\IEEEQED

\subsection{Proof of \protect\propref{prop:mpe-coop}}
\label{app:mpe-coop}

Without loss of generality, as discussed in \secref{sec:geom-inf} we
assume that $f^k$ and $g^k$ are normalized according to
\eqref{eq:cF-def} and \eqref{eq:cG-def}, so the associated feature
vectors $\bfVX$ and $\bfVY$, respectively, satisfy
\eqref{eq:fvXY-orthon}.

We first analyze the error probability in decisions about the value of
$V^k$ based on $S^k$.  To begin, the error exponent in decisions
about $V_i$ satisfies
\begin{align} 
&E^{V_i|S}(\C^{\Y,k}_\eps(P_Y),f^k) \notag\\
&\quad \le
  E^{V_i|S}(\C^{\Y,k}_\eps(P_Y),f^k,v_i,v_i') \label{eq:use-pairwise-again}\\ 
&\quad = \frac{\eps^2}8 \, \bnorm{
  \bigl(\bfVX\bigr)^\T\dtm^\T
  \bigl(\bphi^{Y|V_i}_{v_i}-\bphi^{Y|V_i}_{v_i'}\bigr)}^2  +
o(\eps^2) \label{use-Efg-char-k}\\ 
&\quad \le \frac{\eps^2}2 \, \max_{v_i\in\V_i} \bnorm{
  \bigl(\bfVX\bigr)^\T\dtm^\T  \bphi^{Y|V_i}_{v_i}}^2 
  + o(\eps^2) \label{eq:use-triangle-first}\\
&\quad \le \frac{\eps^2}2 \, \bspectral{\bfVX}^2\, \max_{v_i\in\V_i}
\bnorm{ \dtm^\T \bphi^{Y|V_i}_{v_i}}^2 +
o(\eps^2) \label{eq:use-spectral-first}\\ 
&\quad \le \frac{\eps^2}2 \max_{v_i\in\V_i} \bnorm{ \dtm^\T
  \bphi^{Y|V_i}_{v_i}}^2 + o(\eps^2) \label{eq:use-spec-bnd} \\
&\quad = \frac{\eps^2}2 \max_{v_i\in\V_i} \bnorm{ \dtm^\T
  \bphit^{Y|V_i}_{v_i}}^2\bigl(\delta^{Y|V_i}_{v_i}\bigr)^2 + o(\eps^2) \label{eq:use-delta} \\
&\quad \le \frac{\eps^2}2 \max_{v_i\in\V_i} \bnorm{ \dtm^\T
  \bphit^{Y|V_i}_{v_i}}^2 + o(\eps^2) \label{eq:use-delta-bnd} \\
&\quad = \frac{\eps^2}2 \bnorm{ \dtm^\T
  \bphit^{Y|V_i}_*}^2 + o(\eps^2), 
\label{eq:EViS-bnd}
\end{align}
where \eqref{eq:use-pairwise-again} follows from standard pairwise
error analysis with $E^{V_i|S}(\C^{\Y,k}_\eps(P_Y),f^k,v_i,v_i')$
denoting the pairwise error exponent in distinguising distinct $v_i$
and $v_i'$ in $\V_i$, to obtain \eqref{use-Efg-char-k} we have adapted
\eqref{eq:use-Y|V-to-X|V} in the proof of \propref{prop:mpe-rie}, to
obtain \eqref{eq:use-triangle-first} we have used the triangle
inequality, to obtain \eqref{eq:use-spectral-first} we have used
\factref{fact:frob-submult}, to obtain \eqref{eq:use-spec-bnd} we have
used that $\bspectral{\bfVX}=\bspectral{\bfVY}=1$, and to obtain
\eqref{eq:use-delta} and \eqref{eq:use-delta-bnd} we have used the
decomposition
\begin{equation*}
\bphi^{Y|V_i}_{v_i} = \bphit^{Y|V_i}_{v_i} \, \delta^{Y|V_i}_{v_i}
\end{equation*}
where $\bnorm{\bphit^{Y|V_i}_{v_i}}=1$ and $\babs{\delta^{Y|V_i}_{v_i}}\le1$.
In \eqref{eq:EViS-bnd}, we have introduced the notation
\begin{equation*} 
\bphit^{Y|V_i}_* \defeq \argmax_{\left\{ \bphit^{Y|V_i}_{v_i} \colon
  v_i\in\V_i\right\}} \bnorm{ \bB^\T \bphit^{Y|V_i}_{v_i}}^2,
\end{equation*}
and note that
since $V^k$ is a multi-attribute, by \lemref{lem:orthog} the matrix
\begin{subequations} 
\begin{equation}
\bPhit^{Y|V^k}_*
\defeq \begin{bmatrix} \bphit^{Y|V_1}_* & \cdots &
  \bphit^{Y|V_k}_*  \end{bmatrix} 
\label{eq:bPhitYVk-coop-def}
\end{equation}
has orthogonal columns, so
\begin{equation}
\bigl(\bPhit^{Y|V^k}_*\bigr)^\T \bPhit^{Y|V^k}_* = \bI.
\label{eq:bPhitYVk-coop-orthog}
\end{equation}
\label{eq:bPhitYVk-coop-char}%
\end{subequations}

Hence, for each $i\in\{1,\dots,k\}$,
\begin{align}
&\max_{\C^{\Y,k}_\eps(P_Y),f^k} \ \min_{j\le i} \ 
  E^{V_j|S}(\C^{\Y,k}_\eps(P_Y),f^k) \notag\\
&\qquad \le \max_{\C^{\Y,k}_\eps(P_Y),f^k} \ \min_{j\le i} \ 
\frac{\eps^2}2 \bnorm{ \dtm^\T
  \bphit^{Y|V_j}_\dtm}^2 + o(\eps^2) \label{eq:use-maxmin-obj-ub}\\
&\qquad = \smash[b]{\max_{\substack{\bPhit^{Y|V^i}_\dtm \colon\\
  (\bPhit^{Y|V^i}_\dtm)^\T \bPhit^{Y|V^i}_\dtm = \bI}}} \vphantom{\max_{\bPhit^{Y|V^j}_\dtm}}
\ \min_{j\le i}  \ 
\frac{\eps^2}2 \bnorm{ \dtm^\T \bphit^{Y|V_j}_\dtm}^2 +
o(\eps^2) \label{eq:use-bPhitYVk-coop} \\
&\qquad = \max_{\cS\subset\reals^\dimY\colon \dim(\cS)=i} \ 
\min_{\bphit\in\cS\colon \norm{\bphit}=1} \ 
\frac{\eps^2}2 \bnorm{ \dtm^\T \bphit}^2 + o(\eps^2) \label{eq:use-subspace-def}\\
&\qquad = \frac{\eps^2}2\, \sigma_i^2 + o(\eps^2),
\label{eq:max-min-ub}
\end{align}
where to obtain \eqref{eq:use-maxmin-obj-ub} we have used
\eqref{eq:EViS-bnd}, to obtain \eqref{eq:use-bPhitYVk-coop} we have used
\eqref{eq:bPhitYVk-coop-char}, to obtain \eqref{eq:use-subspace-def} we have
used the definition of a subspace, and to obtain \eqref{eq:max-min-ub}
we have used \lemref{lem:maxmin}.

We further note that the inequalities leading to the right-hand side
of \eqref{eq:max-min-ub} hold with equality for all
$i\in\{1,\dots,k\}$ when we choose
\begin{equation*} 
\cV_i = \{ +1,-1\} \quad\text{and}\quad
\bphi^{Y|V_i}_{+1} = - \bphi^{Y|V_i}_{-1} = \bpsi^Y_i,
\end{equation*}
for $i=1,\dots,k$ (so $p_{V_i}(+1)=p_{V_i}(-1)=1/2$), and
\begin{equation*} 
\bfVX = \bPsi^X_{(k)},
\end{equation*}
with $\bPsi^X_{(k)}$ as defined in
\eqref{eq:bPsi-k-def}.  In particular, the equalities leading to
\eqref{eq:EViS-bnd} all hold with equality with these choices so
\eqref{eq:use-maxmin-obj-ub} holds with equality, and, via
\lemref{lem:maxmin}, \eqref{eq:max-min-ub} holds when $\cS$ is the
space spanned by the columns of $\bPsi^Y_{(i)}$ and
$\bphit=\bpsi^Y_i$.  

We similarly analyze the error probability in decisions about 
$U^k$ based on $S^k$.  In particular, 
the error exponent in decisions
about $U_i$ satisfies
\begin{equation} 
E^{U_i|S}(\C^{\X,k}_\eps(P_X),f^k) 
\le \frac{\eps^2}2 \bnorm{ \bphit^{X|U_i}_*}^2 + o(\eps^2), 
\label{eq:EUiS-bnd}
\end{equation}
where
we have used the
decomposition
\begin{equation*}
\bphi^{X|U_i}_{u_i} = \bphit^{X|U_i}_{u_i} \, \delta^{X|U_i}_{u_i}
\end{equation*}
with $\bnorm{\bphit^{X|U_i}_{u_i}}=1$ and
$\babs{\delta^{X|U_i}_{u_i}}\le1$, and
where
\begin{equation*} 
\bphit^{X|U_i}_* \defeq \argmax_{\left\{ \bphit^{X|U_i}_{u_i} \colon
  u_i\in\U_i\right\}} \bnorm{ \bphit^{X|U_i}_{u_i}}^2.
\end{equation*}
Analgously, we note that
since $U^k$ is a multi-attribute, by \lemref{lem:orthog} the matrix
\begin{subequations} 
\begin{equation}
\bPhit^{X|U^k}_*
\defeq \begin{bmatrix} \bphit^{X|U_1}_* & \cdots &
  \bphit^{X|U_k}_*  \end{bmatrix} 
\label{eq:bPhitXUk-coop-def}
\end{equation}
has orthogonal columns, so
\begin{equation}
\bigl(\bPhit^{X|U^k}_*\bigr)^\T \bPhit^{X|U^k}_* = \bI.
\label{eq:bPhitXUk-coop-orthog}
\end{equation}
\label{eq:bPhitXUk-coop-char}%
\end{subequations}

Hence, for each $i\in\{1,\dots,k\}$,
\begin{align}
&\max_{\C^{\X,k}_\eps(P_X),f^k} \ \min_{j\le i} \ 
  E^{U_j|S}(\C^{\X,k}_\eps(P_X),f^k) \notag\\
&\qquad \le \max_{\cS\subset\reals^\dimY\colon \dim(\cS)=i} \ 
\min_{\bphit\in\cS\colon \norm{\bphit}=1} \ 
\frac{\eps^2}2 \bnorm{ \bphit}^2 + o(\eps^2) 
\\
&\qquad = \frac{\eps^2}2 + o(\eps^2).
\label{eq:max-min-ub-alt}
\end{align}
In this case, it is
straightforward to verify that the corresponding inequalities leading
to \eqref{eq:max-min-ub-alt}---and so to \eqref{eq:EUiS-bnd} as
well---all hold with equality when
\begin{align*}
\cU_i=\{+1,-1\}\quad\text{and}\quad
\bphi^{X|U_i}_{+1} = - \bphi^{X|U_i}_{-1} = \bphit^{X|U_i}_*,
\end{align*}
for any $\bPhit^{X|U^k}_*$ satisfying
\eqref{eq:bPhitXUk-coop-orthog}, and when $\bfVX=\bPhit^{X|U^k}_*$.

The associated error probabilities for decisions about $U^k$ and $V^k$
based on $T^k$ follow from symmetry considerations.  
In particular, it suffices to interchange
the roles of $U$ and $V$, and $X$ and $Y$---so
$\dtms$ is replaced with its adjoint---in the preceding analysis. This
immediately yields that for $i\in\{1,\dots,k\}$,
\begin{equation} 
\max_{\C^{\X,k}_\eps(P_X),g^k} \ \min_{j\le i} \ 
  E^{U_j|T}(\C^{\X,k}_\eps(P_X),g^k) 
 = \frac{\eps^2}2\, \sigma_i^2 + o(\eps^2),
\label{eq:max-min-sym}
\end{equation}
which is achieved by the choices
\begin{equation*} 
\cU_i = \{ +1,-1\} \quad\text{and}\quad
\bphi^{X|U_i}_{+1} = - \bphi^{X|U_i}_{-1} = \bpsi^X_i,
\end{equation*}
for $i=1,\dots,k$ (so $p_{U_i}(+1)=p_{U_i}(-1)=1/2$), and
\begin{equation*} 
\bfVY = \bPsi^Y_{(k)},
\end{equation*}
with $\bPsi^Y_{(k)}$ as defined in
\eqref{eq:bPsi-k-def}.

And it likewise yields, also for $i\in\{1,\dots,k\}$, that
\begin{equation} 
\max_{\C^{\Y,k}_\eps(P_Y),g^k} \ \min_{j\le i} \ 
  E^{V_j|T}(\C^{\Y,k}_\eps(P_Y),g^k) 
= \frac{\eps^2}2 + o(\eps^2),
\label{eq:max-min-sym-alt}
\end{equation}
which is achieved by the choices
\begin{align*}
\cV_i=\{+1,-1\}\quad\text{and}\quad
\bphi^{Y|V_i}_{+1} = - \bphi^{Y|V_i}_{-1} = \bphit^{Y|V_i}_*,
\end{align*}
for any $\bPhit^{Y|V^k}_*$ satisfying
\eqref{eq:bPhitYVk-coop-orthog}, and when $\bfVY=\bPhit^{Y|V^k}_*$.

It follows that the inequalities in \eqref{eq:pe-coop-pareto}
simultaneously all hold with equality for the choices $f^k=f^k_*$ and
$g^k=g^k_*$. and \eqref{eq:CXYk-opt}.\hfill\IEEEQED

\subsection{Proof of \protect\corolref{corol:suffstat}}
\label{app:suffstat}

First, note that
\begin{align} 
&P_{U^k|X^m,Y^m}(u^k|x^m,y^m) \notag \\
&\qquad\qquad = P_{U^k|X^m}(u^k|x^m) \label{eq:use-full-markov-uxyv}\\
&\qquad\qquad = \prod_{i=1}^k P_{U_i|X^m}(u_i|x^m) \label{eq:use-full-orthog}\\
&\qquad\qquad = \prod_{i=1}^k P_{U_i}(u_i)\,
  \frac{P_{X^m|U_i}(x^m|u_i)}{P_{X^m}(x^m)} \notag\\ 
&\qquad\qquad = \prod_{i=1}^k P_{U_i}(u_i)\, 
   \prod_{j=1}^m
   \frac{P_{X|U_i}(x_j|u_i)}{P_X(x_j)} \label{eq:use-full-markov-xu}\\  
&\qquad\qquad = \left(\frac12\right)^k \prod_{i=1}^k \prod_{j=1}^m
   \bigl( 1 + \eps\, u_i\, f_i^*(x_j) \bigr) \label{eq:use-CXk-opt}\\
&\qquad\qquad = \left(\frac12\right)^k \left(
1 + \eps \sum_{i=1}^k u_i\, \sum_{j=1}^m f_i^*(x_j) \right) + o(\eps)
\label{eq:use-prod-sum}\\
&\qquad\qquad = \left(\frac12\right)^k \left( 1 + \eps\, m
\sum_{i=1}^k u_i\, s_i* \right) + o(\eps),
\label{eq:Pu|xy-form}
\end{align}
where to obtain \eqref{eq:use-full-markov-uxyv} we have used the
Markov structure \eqref{eq:full-markov-uxyv}, to obtain
\eqref{eq:use-full-orthog} we have used that $U^k$ is a
multi-attribute of $X^m$,
to obtain \eqref{eq:use-full-markov-xu} we have used
\eqref{eq:full-markov-xu} and \eqref{eq:full-markov-xy} with \eqref{eq:Xm-subprod}, to obtain
\eqref{eq:use-CXk-opt} we have used \eqref{eq:CXk-opt}, and to obtain
\eqref{eq:use-prod-sum} we have used \factref{fact:prod-sum}.

Next, from symmetry considerations, we obtain the analogous result
\begin{equation} 
P_{V^k|X^m,Y^m}(v^k|x^m,y^m) 
= \left(\frac12\right)^k \left( 1 + \eps\, m \sum_{i=1}^k v_i\, t_i^*
\right) + o(\eps),
\label{eq:Pv|xy-form}
\end{equation}

We then obtain \eqref{eq:Puv|xy-form} by substituting
\eqref{eq:Pu|xy-form} and \eqref{eq:Pv|xy-form} into
\begin{multline*}
P_{U^k,V^k|X^m,Y^m}(u^k,v^k|x^m,y^m) \\
=P_{U^k|X^m,Y^m}(u^k|x^m,y^m) \,
  P_{V^k|X^m,Y^m}(v^k|x^m,y^m),
\end{multline*}
which is a consequence of the
Markov structure \eqref{eq:full-markov-uxyv}.

Finally, using the preceding results we have
\begin{align*}
&P_{U^k|S_*^k,T_*^k,V^k}(u^k|s_*^k,t_*^k,v^k) \notag\\
&\qquad= \frac{P_{U^k,V^k|S_*^k,T_*^k}(u^k,v^k|s_*^k,t_*^k)}{P_{V^k|S_*^k,T_*^k}(v^k|s_*^k,t_*^k)} \\
&\qquad= \frac{P_{U^k,V^k|X^m,Y^m}(u^k,v^k|x^m,y^m) 
+ o(\eps)}{P_{V^k|Y^m}(v^k|y^m)+o(\eps)}
\\ 
&\qquad= \frac{P_{U^k|X^m}(u^k|x^m)\,P_{V^k|Y^m}(v^k|y^m) +
  o(\eps)}{P_{V^k|Y^m}(v^k|y^m)+o(\eps)} \\ 
&\qquad= P_{U^k|X^m}(u^k|x^m) + o(\eps),
\end{align*}
which combined with \eqref{eq:Pu|xy-form} verifies \eqref{eq:U|STV}, and
\eqref{eq:V|STU} follows from symmetry considerations. \hfill\IEEEQED

\subsection{Proof of \protect\propref{prop:ib-double}}
\label{app:ib-double}

The following lemma is useful in our proof.
\begin{lemma}
\label{lem:Iuv}
Given $\eps>0$ and configurations $\C^\X_\eps(P_X)$ and
$\C^\Y_\eps(P_Y)$ for $\eps$-attributes $U$ and $V$, respectively, we
have
\begin{equation}
\frac{P_{U,V}(u,v)}{P_U(u)\,P_V(v)} 
= 1 + \eps^2 \, \sigmat(u,v),
\label{eq:Puv-form}
\end{equation}
and
\begin{equation}
I(U;V) = \frac{\eps^4}2\, \sum_{u\in\U,v\in\V} P_U(u)\,P_V(v) \, \sigmat(u,v)^2
+ o(\eps^4),
\label{eq:Iuv-form}
\end{equation}
where
\begin{equation}
\sigmat(u,v) \defeq 
\bigl(\bphi^{Y|V}_v\bigr)^\T \dtm \,
  \bphi^{X|U}_u.
\label{eq:sigmat-def}
\end{equation}
\end{lemma}

\begin{IEEEproof}[Proof of \lemref{lem:Iuv}]
First, we obtain \eqref{eq:Puv-form} via
\begin{align}
&\frac{P_{U,V}(u,v)}{P_U(u)\,P_V(v)} \notag\\
&\ = \sum_{x\in\X,y\in\Y}
\frac{P_{Y|V}(y|v)\,P_{Y|X}(y|x)\,P_{X|U}(x|u)}{P_Y(y)} \notag\\
&\ = \sum_{x\in\X,y\in\Y} \!\!
P_{Y,X|U}(y,x|u) + P_{X,Y|V}(x,y|v) - P_{X,Y}(x,y) \notag\\
&\ \qquad\qquad\qquad{} + \
\frac{P_{Y|V}(y|v)-P_Y(y)}{\sqrt{P_Y(y)}} \notag\\
&\ \qquad\qquad\qquad\qquad\qquad{} \cdot\frac{P_{X,Y}(x,y)}{\sqrt{P_X(x)}\,\sqrt{P_Y(y)}} \notag\\
&\ \qquad\qquad\qquad\qquad\qquad\qquad{} \cdot \frac{P_{X|U}(x|u)-P_X(x)}{\sqrt{P_X(x)}} \notag\\
&\ = 1 + \eps^2 \, \sigmat(u,v),
\end{align}
with $\sigmat(u,v)$ as defined in \eqref{eq:sigmat-def}.  In turn, 
we obtain \eqref{eq:Iuv-form} via
\begin{align}
&I(U;V) \notag\\
&\quad = D(P_{U,V}\|P_U P_V) \notag\\
&\quad = \sum_{u\in\U,v\in\V} P_{U,V}(u,v) \log
\frac{P_{U,V}(u,v)}{P_U(u)\,P_V(v)} \notag\\ 
&\quad = \sum_{u\in\U,v\in\V} P_{U,V}(u,v) 
\biggl[ \eps^2 \sigmat(u,v) - \frac{\eps^4}2
  \sigmat(u,v)^2 + o(\eps^4) \biggr] \label{eq:use-log-exp2}\\ 
&\quad = \smash{\sum_{u\in\U,v\in\V}}\vphantom{\sum} P_U(u)\,P_V(v) \,
\bigl[1+ \eps^2 \sigmat(u,v) \bigr] \notag\\
&\qquad \qquad\qquad {} \cdot
\biggl[ \eps^2 \sigmat(u,v) - \frac{\eps^4}2 \sigmat(u,v)^2 +
  o(\eps^4) \biggr]  \label{eq:use-Puv-form}\\
&\quad = \frac{\eps^4}2\,
\sum_{u\in\U,v\in\V} P_U(u)\,P_V(v) \, \sigmat(u,v)^2 +
o(\eps^4), \label{eq:use-zero-cond}
\end{align}
where to obtain \eqref{eq:use-log-exp2}  we have used 
\eqref{eq:Puv-form} and the Taylor
series expansion
$\log(1+\omega) = \omega - \omega^2/2 + o(\omega^2)$, where to obtain
\eqref{eq:use-Puv-form} we  
have again used \eqref{eq:Puv-form}, and where to obtain
\eqref{eq:use-zero-cond} we have used that 
\begin{equation*}
\sum_{u\in\U,v\in\V} P_U(u)\,P_V(v) \, \sigmat(u,v) =0
\end{equation*}
as a consequence of \eqref{eq:pxyuv-constr}, since $\sigmat(u,v)$
takes the form \eqref{eq:sigmat-def}.
\end{IEEEproof}

Proceeding to the proof of \propref{prop:ib-double}, we have
\begin{align} 
&I(U^k;V^k) \notag\\
&\quad= \frac{\eps^4}{2} \sum_{u^k,v^k} P_{U^k}(u^k)\, P_{V^k}(v^k) \,
\sigmat(u^k,v^k)^2 + o(\eps^4) \label{eq:use-Iuv-lem}\\
&\quad\le \frac{\eps^4}{2} \, \max_{u^k,v^k} \sigmat(u^k,v^k)^2 + o(\eps^4)\notag\\
&\quad= \frac{\eps^4}{2} \Bigl(\bigl(\bphi^{Y|V^k}\bigr)^\T \dtm \,
\bphi^{X|U^k}\Bigr)^2 + o(\eps^4) \label{eq:use-max-notation}\\ 
&\quad= \frac{\eps^4}{2} \Bigl( \sum_{i=1}^k \sum_{j=1}^k
\bigl(\bphi^{Y|V_i}\bigr)^\T \dtm \,  \bphi^{X|U_j} \Bigr)^2
+ o(\eps^4) \label{eq:use-k-decomp}\\
&\quad= \frac{\eps^4}{2} \, \bfrob{\bigl(\bPhi^{Y|V^k}\bigr)^\T \dtm \,
  \bPhi^{X|U^k}}^2 + o(\eps^4) \label{eq:use-bPhi}\\
&\quad= \frac{\eps^4}{2} \, \bfrob{\bigl(\bPhit^{Y|V^k}\bDel^{Y|V^k}\bigr)^\T \dtm \,
  \bPhit^{X|U^k}\bDel^{X|U^k}}^2 + o(\eps^4) \label{eq:use-bDels}\\
&\quad\le \frac{\eps^4}{2} \, \left( \max_i \bnorm{\bphi^{Y|V_i}}^2
\right) \left( \max_i \bnorm{\bphi^{X|U_i}}^2
\right) \notag\\
&\qquad\qquad\qquad{} \cdot
\bspectral{\bPhit^{Y|V^k}}^2 \, \bfrob{\dtm \,
  \bPhit^{X|U^k}}^2 + o(\eps^4) \label{eq:use-fact}\\
&\le \frac{\eps^4}{2} \, \bfrob{\dtm \, \bPhit^{X|U^k}}^2 +
o(\eps^4) \label{eq:use-phit-orthog} \\
&\quad\le \frac{\eps^4}{2} \, \sum_{i=1}^k \sigma_i^2 + o(\eps^4),
\label{eq:IUV-k}
\end{align}
where to obtain \eqref{eq:use-Iuv-lem} we have used
\eqref{eq:Iuv-form} of \lemref{lem:Iuv} with $U=U^k$ and $V=V^k$ so
\begin{equation}
\sigmat(u^k,v^k) \defeq 
\bigl(\bphi^{Y|V^k}_{v^k}\bigr)^\T \dtm \, \bphi^{X|U^k}_{u^k},
\label{eq:sigmat-def-k}
\end{equation}
in \eqref{eq:use-max-notation} we have introduced the notation
\begin{equation*} 
\bphi^{X|U^k} \defeq \bphi^{X|U^k}_{u^k_{\max}}\quad\text{and}\quad
\bphi^{Y|V^k} \defeq \bphi^{Y|V^k}_{v^k_{\max}}
\end{equation*}
where
\begin{equation*}
(u^k_{\max},v^k_{\max})  \defeq \argmax_{u^k,v^k} \sigmat(u^k,v^k)^2.
\end{equation*}
To obtain \eqref{eq:use-k-decomp} we have used
\lemref{lem:ivma-sum} together with the notation
\begin{equation*} 
\bphi^{X|U_i} \defeq \bphi^{X|U_i}_{u_i^{\max}}\quad\text{and}\quad
\bphi^{Y|V_i} \defeq \bphi^{Y|V_i}_{v_i^{\max}}
\end{equation*}
with
\begin{align*} 
u^k_{\max} &= (u_1^{\max},\dots,u_k^{\max}) \\
v^k_{\max} &= (v_1^{\max},\dots,v_k^{\max}),
\end{align*}
and in \eqref{eq:use-bPhi} we have introduced the notation
\begin{align*}
\bPhi^{X|U^k} & \defeq
\begin{bmatrix} \bphi^{X|U_1} & \cdots & \bphi^{X|U_k} \end{bmatrix} \\
\bPhi^{Y|V^k} & \defeq 
\begin{bmatrix} \bphi^{Y|V_1} & \cdots & \bphi^{Y|V_k} \end{bmatrix}.
\end{align*}
To obtain \eqref{eq:use-bDels} have used 
the factorizations
\begin{align*}
\bPhi^{X|U^k} &= \bPhit^{X|U^k} \bDel^{X|U^k} \\
\bPhi^{Y|V^k} &= \bPhit^{Y|V^k} \bDel^{Y|V^k}
\end{align*}
where, due to \lemref{lem:orthog},
\begin{equation*}
\bigl(\bPhit^{X|U^k}\bigr)^\T \bPhit^{X|U^k} =
\bigl(\bPhit^{Y|V^k}\bigr)^\T \bPhit^{Y|V^k} = \bI
\end{equation*}
and where $\bDel^{X|U^k}$ and $\bDel^{Y|V^k}$ are diagonal matrices,
to obtain \eqref{eq:use-fact} we have repeatedly used
\factref{fact:frob-submult}, to obtain \eqref{eq:use-phit-orthog} we
have used the properties of the factorization, and to obtain
\eqref{eq:IUV-k} we have used \lemref{lem:eig-k}.  Finally, it is
straightforward to verify that the inequalities leading to
\eqref{eq:IUV-k} all hold with equality when we choose the particular
configurations \eqref{eq:CXYk-opt}, i.e., when
\begin{subequations} 
\begin{align}
\bphi^{X|U_i}_{u_i} &= u_i\, \bpsi^X_i,\quad u_i\in\{+1,-1\}\\
\bphi^{Y|V_i}_{v_i} &= v_i\, \bpsi^Y_i,\quad v_i\in\{+1,-1\}.
\end{align}
\label{eq:opt-uv-configs}%
\end{subequations}

To obtain \eqref{eq:PUV-k}, 
first note that, starting from \eqref{eq:sigmat-def-k},
\begin{align} 
\sigmat(u^k,v^k) 
&= \bigl(\bphi^{Y|V^k}_{v^k}\bigr)^\T \dtm\, \bphi^{X|U^k}_{u^k} \notag\\
&= \left(\sum_{i=1}^k \bphi^{Y|V_i}_{v_i}\right)^\T \dtm\, \left(
\sum_{j=1}^k \bphi^{X|U_j}_{u_j} \right) +
o(1) \label{eq:use-lem-ivma-sum}\\  
&= \left(\sum_{i=1}^k v_i\, \bpsi^Y_i \right)^\T \dtm\, \left(
\sum_{j=1}^k u_j\, \bpsi^X_j \right) + o(1) \label{eq:use-opt-uv-configs}
\\ 
&= \sum_{i=1}^k \sum_{j=1}^k u_j\, v_i\, \bigl(\bpsi^Y_i\bigr)^\T \dtm\, 
 \bpsi^X_j + o(1) \notag\\ 
&= \sum_{i=1}^k u_i\, v_i\, \sigma_i + o(1), \label{eq:sigmat-opt}
\end{align}
where to obtain \eqref{eq:use-lem-ivma-sum} we have used \lemref{lem:ivma-sum}
to obtain \eqref{eq:use-opt-uv-configs} we have used
\eqref{eq:opt-uv-configs}, and to obtain 
\eqref{eq:sigmat-opt} we have used \eqref{eq:dtm-svd-form}.

Hence, using \eqref{eq:Puv-form} in
\lemref{lem:Iuv} with $U=U^k$ and $V=V^k$ and 
substituting \eqref{eq:sigmat-opt}, we obtain
\begin{align*}
P_{U^k,V^k}(u^k,v^k)
&= P_{U^k}(u^k)\,P_{V^k}(v^k)\, \bigl(1 + \eps^2\, \sigmat(u^k,v^k)\bigr) \\
&= \frac1{4^k} \left( 1 + \eps^2 \sum_{i=1}^k u_i\, v_i\, \sigma_i
\right) + o(\eps^2),
\end{align*}
viz., \eqref{eq:PUV-k}.
\hfill\IEEEQED

\subsection{Proof of \protect\propref{prop:ib-single}}
\label{app:ib-single}

First, we note that, in accordance with the discussion of
\secref{sec:eps-dep},  the conditions of the proposition imply that
$U^k$ has a configuration of the form $\C^{\X,}_{\eps(1+o(1))}$.  Next,
we have
\begin{align}
I(U^k;Y) 
&= \frac1{2^k} \sum_{u^k}
D(P_{Y|U^k}(\cdot|u^k)\|P_Y) \label{eq:use-equiprob}\\ 
&= \frac{\eps^2}{2^{k+1}} \sum_{u^k} \bnorm{\bphi^{Y|U^k}_{u^k}}^2 +
o(\eps^2) \label{eq:use-div-lemma} \\
&= \frac{\eps^2}{2^{k+1}} \sum_{u^k} \bnorm{\dtm\, \bphi^{X|U^k}_{u^k}}^2 +
o(\eps^2) \label{eq:use-ivYXUk-map} \\
&= \frac{\eps^2}{2^{k+1}} \sum_{u^k} \bbnorm{\dtm\, \sum_{i=1}^k \bphi^{X|U_i}_{u_i}}^2 +
o(\eps^2) \label{eq:use-ivma-sum}\\ 
&= \frac{\eps^2}{2} \sum_{i=1}^k \bnorm{\dtm\, \bphi^{X|U_i}}^2 +
o(\eps^2)  \label{eq:use-phiXUi-sym}\\
&= \frac{\eps^2}{2}\bfrob{\dtm\, \bPhi^{X|U^k}}^2 + o(\eps^2) \notag\\
&= \frac{\eps^2}{2}\bfrob{\dtm\, \bPhit^{X|U^k}\, \bDel^{X|U^k}}^2 +
o(\eps^2) \label{eq:use-bPhiXUk-def} \\
&\le \frac{\eps^2}{2}\bfrob{\dtm\, \bPhit^{X|U^k}}^2\,
\bspectral{\bDel^{X|U^k}}^2 + o(\eps^2) \label{eq:use-frob-submult-disc}\\
&\le \frac{\eps^2}{2}\bfrob{\dtm\, \bPhit^{X|U^k}}^2 + o(\eps^2) \notag\\
&\le \frac{\eps^2}{2} \sum_{i=1}^k \sigma_i^2 + o(\eps^2),
\label{eq:IUkY-bnd}
\end{align}
where to obtain \eqref{eq:use-equiprob} we have used that all
configurations $u^k$ are equiprobable due to condition 1, to obtain
\eqref{eq:use-div-lemma} we have used \lemref{lem:D-char}, to obtain
\eqref{eq:use-ivYXUk-map} we have used \eqref{eq:X|U-to-Y|U} with
$U=U^k$, to obtain \eqref{eq:use-ivma-sum} we have used
\lemref{lem:ivma-sum}, and to obtain \eqref{eq:use-phiXUi-sym} we have
used that constraint 2 implies that
\begin{equation}
\bphi^{X|U_i}_{+1} = -\bphi^{X|U_i}_{-1} \defeq \bphi^{X|U_i},
\label{eq:phiXUi-sym}
\end{equation}
for $i=1,\dots,K-1$, since 
\begin{equation*}
\sum_x P_{U_i}(u)\, \phi^{X|U_i}_u = 0.
\end{equation*}
To obtain \eqref{eq:use-bPhiXUk-def} we have used the notation 
\begin{equation*}
\bPhi^{X|U^k} \defeq \begin{bmatrix} \bphi^{X|U_1} & \cdots & \bphi^{X|U_k} 
\end{bmatrix}
\end{equation*}
with factorization
\begin{equation*}
\bPhi^{X|U^k} = \bPhit^{X|U^k} \bDel^{X|U^k}
\end{equation*}
where, due to \lemref{lem:orthog},
\begin{equation}
\bigl(\bPhit^{X|U^k}\bigr)^\T \bPhit^{X|U^k} = \bI
\label{eq:bPhitXUk-orthog}%
\end{equation}
and $\bDel^{X|U^k}$ is a diagonal matrix whose $i$\/th diagonal entry is
$\|\bphi^{X|U_i}\| \le 1 + o(1)$.  To obtain
\eqref{eq:use-frob-submult-disc} we have used
\factref{fact:frob-submult}, and to obtain \eqref{eq:IUkY-bnd} we have
used \lemref{lem:eig-k} with the constraint
\eqref{eq:bPhitXUk-orthog}.  Furthermore, the inequalities leading to
\eqref{eq:IUkY-bnd} hold with equality when we choose
\begin{equation*}
\bPhi^{X|U^k} = \bPsi^X_{(k)},
\end{equation*}
with $\bPsi^X_{(k)}$ as defined in \eqref{eq:bPsi-k-def}, so the optimum
configuration is \eqref{eq:CXk-opt}.

The corresponding result, including \eqref{eq:CYk-opt}, for the
maximization of $I(V^k;X)$ subject to $I(V_i;Y)\le \eps^2/2$ and the
other corresponding constraints follows immediately from symmetry
considerations.  \hfill\IEEEQED

\subsection{Proof of \protect\propref{prop:common}}
\label{app:common}

First, without loss of generality let us choose $\delta(\cdot)$ such that
\begin{equation}
\delta(\eps)\ge\eps,
\label{eq:delta-choice}
\end{equation}
in which case, we have, for all $\eps$ sufficiently small,
\begin{align} 
P_{X,Y} &\in\nbhdk_\eps^{\X\times\Y}(P_XP_Y) \label{eq:nbhdk-given}\\
&\subset
\nbhd_\eps^{\X\times\Y}(P_XP_Y) \label{eq:use-sub-eps-nbhd}\\
&\subset
\nbhd_{\delta(\eps)}^{\X\times\Y}(P_XP_Y) \label{eq:use-delta-choice}\\
&\subset
\nbhd_{\sqrt{\smash[b]{2\delta(\eps)}}}^{\X\times\Y}(P_XP_Y).
\label{eq:XY-near-indep}
\end{align}
where \eqref{eq:nbhdk-given} is given, where
\eqref{eq:use-sub-eps-nbhd} follows from  
\eqref{eq:sub-eps-nbhd}, where \eqref{eq:use-delta-choice} follows
from \eqref{eq:delta-choice}, and where \eqref{eq:XY-near-indep} holds
when $\eps\le 2$.

Next, from \eqref{eq:ivXY|W} and \eqref{eq:phiXWYW-ivspace} it follows
that for $w\in\W$, 
\begin{equation}
\phit^{X,Y|W}_w(x,y) = \phic^{X,Y|W}_w(x,y)  + o(1),\quad x\in\X,\ y\in\Y, 
\label{eq:phit-phic}
\end{equation}
wherein
\begin{align}
&\bnorm{\bphic^{X,Y|W}_w}^2 \notag\\
&\quad= \frac12 \sum_{x,y} \Bigl( \sqrt{P_Y(y)} \, \phi^{X|W}_w(x) +
\sqrt{P_X(x)}\,\phi^{Y|W}_w(y) \Bigr)^2 \notag\\
&\quad= \frac12 \left( \sum_{x,y} P_Y(y)\, \phi^{X|W}_w(x)^2 + 
\sum_{x,y} P_X(x)\,
\phi^{Y|W}_w(y)^2 \right) \notag\\
&\quad= \frac12 \left(\bnorm{\bphi^{X|W}_w}^2 + \bnorm{\bphi^{Y|W}_w}^2\right) \label{eq:phitXY|W=sum}\\
&\quad\le 1. 
\label{eq:phitXY|W-bnd}
\end{align}
Hence,
\begin{equation} 
P_{X,Y|W}(\cdot,\cdot|w)\in\nbhd^{\X\times\Y}_{\sqrt{\smash[b]{2\delta(\eps)}}(1+o(1))}(P_XP_Y),
\quad w\in\W.
\label{eq:PXY|W-nbhd}
\end{equation}

Furthermore, due to \eqref{eq:XY-near-indep} and
\eqref{eq:PXY|W-nbhd}, we may apply \lemref{lem:refdist-invar} with
$P_1=P_{X,Y|W}(\cdot,\cdot|w)$, $P_2=P_{X,Y}$, $\refgen=P_XP_Y$, and
$\Pt_0=P_{X,Y}$ to
\begin{equation}
\phi^{X,Y|W}_w(x,y) \defeq \frac{P_{X,Y|W}(x,y|w) -
  P_{X,Y}(x,y)}{\sqrt{\smash[b]{2\delta(\eps)}}\, \sqrt{P_{X,Y}(x,y)}} 
\label{eq:ivXY|W-def}
\end{equation}
yielding
\begin{align} 
\phi^{X,Y|W}_w(x,y) 
&= \phit^{X,Y|W}_w(x,y) - \frac{\dtmts(y,x)}{\sqrt{\smash[b]{2\delta(\eps)}}}
  + o(1) \notag\\ 
&= \phic^{X,Y|W}_w(x,y) + o(1),
\label{eq:chg-ref}
\end{align}
where to obtain \eqref{eq:chg-ref} we have used both
\eqref{eq:phit-phic} 
and that \eqref{eq:use-delta-choice} implies
$\abs{\dtmts(y,x)}\le\delta(\eps)$.
Combining \eqref{eq:chg-ref} with \eqref{eq:phitXY|W-bnd}, we conclude
\begin{equation} 
P_{X,Y|W}(\cdot,\cdot|w)\in\nbhd^{\X\times\Y}_{\sqrt{\smash[b]{2\delta(\eps)}}(1+o(1))}(P_{X,Y}),\quad
w\in\W.
\end{equation}

As a result, for all $w\in\W$,
\begin{align} 
&D(P_{X,Y|W}(\cdot,\cdot|w) \| P_{X,Y}) \notag\\
&\qquad\qquad\qquad= \delta(\eps)\, \bnorm{\bphi^{X,Y|W}_w}^2 + o(\delta(\eps)) \label{eq:use-phixyw-def}\\
&\qquad\qquad\qquad= \delta(\eps)\, \bnorm{\bphic^{X,Y|W}_w}^2 + o(\delta(\eps)),
\label{eq:use-phiXY-small}
\end{align}
where to obtain \eqref{eq:use-phixyw-def} we have used the special
case of \lemref{lem:D-char}, 
and to obtain 
\eqref{eq:use-phiXY-small} we have used \eqref{eq:chg-ref}.
In turn, 
\begin{align}
&I(W;X,Y) \notag\\
&\quad= \sum_{w\in\W} P_W(w)\, D(P_{X,Y|W}(\cdot,\cdot|w)\|P_{X,Y}) \notag\\
&\quad= \delta(\eps) \sum_{w\in\W} P_W(w)\,
  \bnorm{\bphic^{X,Y|W}_w}^2  + o(\delta(\eps)) \label{eq:use-D-phit-form}\\ 
&\quad= \frac{\delta(\eps)}{2} \sum_{w\in\W} P_W(w) \left(
  \norm{\bphi^{X|W}_w}^2  +  
\norm{\bphi^{Y|W}_w}^2 \right) + o(\delta(\eps)),
\label{eq:IWXY-form}
\end{align}
where to obtain \eqref{eq:use-D-phit-form} we have used
\eqref{eq:use-phiXY-small}, and where to obtain \eqref{eq:IWXY-form}
we have used \eqref{eq:phitXY|W=sum}.

Hence, we seek to minimize \eqref{eq:IWXY-form} subject to the constraint
\eqref{eq:ivXY}.   To this end, let us
define
\begin{align}
\bPhit^{X|W} &\defeq \sqrt{\delta(\eps)}\, \bPhi^{X|W} \sqrt{\bP_W}\\
\bPhit^{Y|W} &\defeq \sqrt{\delta(\eps)}\, \bPhi^{Y|W} \sqrt{\bP_W},
\end{align}
where, consistent with the notation and terminology in
\defref{def:rie}, $\bPhi^{X|W}$ is a $\cardX\times\cardW$ matrix whose
$w$\/th column is $\bphi^{X|W}_w$, where $\bPhi^{Y|W}$ is a
$\cardY\times\cardW$ matrix whose $w$\/th column is $\bphi^{Y|W}_w$,
and where $\bP_W$ is a $\cardW\times\cardW$ diagonal matrix whose
$w$\/th diagonal entry is $P_W(w)$.  Then we can equivalently express
the constraint \eqref{eq:ivXY} in the form
\begin{equation}
\dtmt = \bPhit^{Y|W} \, \bigl(\bPhit^{X|W}\bigr)^\T,
\label{eq:dtmt-constr-matrix}
\end{equation}
and the objective function \eqref{eq:IWXY-form} as
\begin{align} 
I(W;X,Y) 
&= \frac12 \left( \frob{\bPhit^{X|W}}^2  + 
\frob{\bPhit^{Y|W}}^2 \right) + o(\delta(\eps)) \notag\\
&\ge  \nuclear{\dtmt} + o(\delta(\eps)) ,
\label{eq:IWXY-bound}
\end{align}
where to obtain the inequality we have used \lemref{lem:kyfan} with
\eqref{eq:dtmt-constr-matrix}.

It is straightforward to verify that the inequality in
\eqref{eq:IWXY-bound} holds with equality subject to the constraints
in \eqref{eq:Cw-alt}
when we choose the 
configuration for $W$ according to 
\begin{subequations} 
\begin{align}
\W &= \{\pm1,\dots,\pm(K-1)\} \\
\bphi^{X|W}_i &= -\bphi^{X|W}_{-i} = \sqrt{\frac{\sigma_i}{\sigmat_i\, \delta(\eps)}} \,
\bpsi^X_i,\quad 
  i=1,\dots,K-1 \label{eq:phiXW-opt}\\ 
\bphi^{Y|W}_i &= -\bphi^{Y|W}_{-i} = \sqrt{\frac{\sigma_i}{\sigmat_i\, \delta(\eps)}} \,
 \bpsi^Y_i,\quad
  i=1,\dots,K-1 \label{eq:phiYW-opt}\\ 
P_W(i) &= P_W(-i) = \frac12 \sigmat_i,\quad
  i=1,\dots,K-1, \label{eq:PWi-val}
\end{align}%
where
\begin{equation}
\sigmat_i \defeq \frac{\sigma_i}{\sum_{i'=1}^{K-1} \sigma_{i'}}.
\label{eq:sigmati-def}
\end{equation}
\label{eq:W-opt}%
\end{subequations}
Specifically, with the choices \eqref{eq:W-opt} we have
\begin{equation*} 
\bnorm{\bphi^{X|W}_w} \le 1\quad\text{and}\quad
\bnorm{\bphi^{Y|W}_w} \le 1
\end{equation*}
since 
\begin{equation*}
\sum_{i=1}^{K-1} \sigma_i = \nuclear{\dtmt}\le\eps\le\delta(\eps)
\end{equation*}
from
$P_{X,Y}\in\nbhdk_\eps^{\X\times\Y}(P_XP_Y)$ as given and the choice
\eqref{eq:delta-choice}, so the constraints 
\eqref{eq:phiXWYW-ivspace} are satisfied.  Moreover, we satisfy 
constraints\eqref{eq:phi-marg} by the symmetric construction
of our information vector sets.  And we satisfy the constraint
\eqref{eq:ivXY} by construction
since 
\begin{align*} 
\delta(\eps) \sum_{w\in\W} P_W(w) \, &\bphi^{Y|W}_w \, \bigl( \bphi^{X|W}_w
\bigr)^\T \notag\\
&= \delta(\eps) \sum_{i=1}^{K-1} \frac{\sigma_i}{\delta(\eps)} \, \bpsi^Y_i \,
\bigl( \bpsi^X_i \bigr)^\T = \dtmt,
\end{align*}
where last equality follows from \eqref{eq:dtmt-svd}.
Finally, evaluating the leading term in
\eqref{eq:IWXY-form} we obtain
\begin{multline*}
\frac{\delta(\eps)}{2}\sum_{w\in\W} P_W(w) \bigl( \norm{\bphi^{X|W}_w}^2  + 
\norm{\bphi^{Y|W}_w}^2 \bigr) \\
= \frac{\delta(\eps)}{2}\left( 2 \sum_{i=1}^{K-1}
\frac{\sigma_i}{\delta(\eps)}\right) = \sum_{i=1}^{K-1} \sigma_i ,
\end{multline*}
so the inequality in \eqref{eq:IWXY-bound} is achieved with
equality.   

It remains only to choose $\delta(\cdot)$ satisfying
\eqref{eq:delta-choice} and $\lim_{\eps\to0} \delta(\eps)=0$.  The
leading term in \eqref{eq:IWXY-form} with the configuration
\eqref{eq:W-opt} does not depend on this choice, so we focus on the
$o(\delta(\eps))$ term, which is minimized by the choice
$\delta(\eps)=\eps$, yielding \eqref{eq:CXY-full}.   In turn,
\eqref{eq:W-config-opt} is obtained by rewriting
\eqref{eq:W-opt} using \eqref{eq:phiXW-def}--\eqref{eq:phiYW-def} and
\eqref{eq:fgi-def}. 
\hfill\IEEEQED

\subsection{Proof of \corolref{corol:suffstat-common}}
\label{app:suffstat-common}

We have, for the extended model \eqref{eq:full-markov-common},
\begin{align}
&P_{X^m,Y^m|W}(x^m,y^m|w) \notag\\
&\qquad = P_{X^m}(x^m)\,P_{Y_m}(y^m) \notag\\
&\quad\qquad\qquad{} \cdot \prod_{l=1}^m 
\left(1+\sgn(w)\,\nuclear{\dtmt}^{1/2} \,
f_\abs{w}^*(x_l)\right)\notag\\
&\ \qquad\qquad\qquad\qquad{} \cdot
\left(1+\sgn(w)\,\nuclear{\dtmt}^{1/2} \,
g_\abs{w}^*(y_l)\right) \label{eq:use-Wopt-prod}\\ 
&\qquad = P_{X^m}(x^m)\,P_{Y_m}(y^m)\notag\\
&\qquad\qquad\qquad{}\cdot \left( 1 + m\,\sgn(w)\,\nuclear{\dtmt}^{1/2} r_\abs{w}^* \right) + o(\sqrt{\eps})
\label{eq:use-prod-fact-again}\\ 
&\qquad = P_{X^m,Y^m}(x^m,y^m) \notag\\
&\qquad\qquad\qquad{} \cdot \left( 1 + m\,\sgn(w)\,\nuclear{\dtmt}^{1/2} r_\abs{w}^* \right) + o(\sqrt{\eps}),
\label{eq:use-XY-close}
\end{align}
where to obtain \eqref{eq:use-Wopt-prod} we have used
\eqref{eq:W-config-opt} and \eqref{eq:full-markov-common}, to obtain
\eqref{eq:use-prod-fact-again} we have used that $X,Y$ are sub-$\eps$
dependent so \eqref{eq:sub-eps-prop} holds, together with
\factref{fact:prod-sum} and \eqref{eq:ri*-def}, and to obtain
\eqref{eq:use-XY-close} we have used that since sub-$\eps$ dependence
implies $\abs{\dtmts(y,x)}\le\eps$ for all $x\in\X$ and $y\in\Y$, 
\begin{equation*}
P_{X,Y}(x,y) = P_X(x)\,P_Y(y) + O(\eps) = P_X(x)\,P_Y(y) + o(\sqrt{\eps}) ,
\end{equation*}
whence
\begin{equation*}
P_{X^m,Y^m}(x^m,y^m) = P_{X^m}(x^m)\,P_{Y^m}(y^m) + o(\sqrt{\eps}).
\end{equation*}
Finally, substituting $P_W$ from \eqref{eq:W-config-opt} and using
\eqref{eq:use-XY-close} in
\begin{equation*} 
P_{W|X^m,Y^m}(w|x^m,y^m) =
\frac{P_{X^m,Y^m|W}(x^m,y^m|w) \, P_W(w)}{P_{X^m,Y^m}(x^m,y^m)}
\end{equation*}
yields \eqref{eq:suffstat-common}.
\hfill\IEEEQED

\subsection{Proof of \protect\corolref{corol:common-relate}}
\label{app:common-relate}

For the first part of the corollary, note that \eqref{eq:C-comp} and
\eqref{eq:CXY-full} together imply \eqref{eq:C-decomp}.   To show
\eqref{eq:C-comp}, we begin by defining
\begin{equation} 
\phi^{X,Y|W_i}_{w_i}
\defeq
  \frac{P_{X,Y|W_i}(x,y|w_i)-P_{X,Y}(x,y)}{\sqrt{2\eps}\,\sqrt{P_{X,Y}(x,y)}},\quad w_i\in\W_\circ.
\label{eq:phiXY|Wi-def}
\end{equation}
Then, since
\begin{align*}
P_{X,Y|W_i}(x,y|+1) &=
P_{X,Y|W}(x,y|i) \\ 
P_{X,Y|W_i}(x,y|-1) &= P_{X,Y|W}(x,y|{-i}).
\end{align*}
we have
\begin{subequations} 
\begin{align} 
\phi^{X,Y|W_i}_{+1} &= \phi^{X,Y|W}_{i} = \phic^{X,Y|W}_i + o(1) \\
\phi^{X,Y|W_i}_{-1} &= \phi^{X,Y|W}_{-i} = \phic^{X,Y|W}_{-i} +o(1),
\end{align}
\label{eq:phi-phit-forms}%
\end{subequations}
where $\phi^{X,Y|W}_w$ and $\phic^{X,Y|W}_w$ are as defined in
\eqref{eq:ivXY|W-def} (setting $\delta(\eps)=\eps$) and
\eqref{eq:ivcXY|W-def}, respectively.

Next, note that 
\begin{align} 
&P_{X,Y|W_i}(x,y|0) \notag\\
&\ = \frac{1}{P_{W_i}(0)} \sum_{\{j\colon
  j\neq i\}} \bigl( 1-P_{W_j}(0) \bigr)
  P_{X,Y|\{W_j\neq0\}}(x,y) \label{eq:use-partition}\\ 
&\ = \frac{1}{(1-\sigmat_i)} \sum_{\{j\colon
  j\neq i\}}  \sigmat_j\,
  P_{X,Y|\{W_j\neq0\}}(x,y) \label{eq:PXY|W-zero-form} \\
&\ = \frac{1}{2(1-\sigmat_i)} \sum_{\{j\colon
  j\neq i\}}  \sigmat_j\,
\Bigl( P_{X,Y|W}(x,y|j) \notag\\
&\ \qquad\qquad\qquad\qquad\qquad\qquad{}+ P_{X,Y|W}(x,y|{-j}) \Bigr),
\label{eq:PXY|W-nonzero-form}
\end{align}
where to obtain \eqref{eq:use-partition} we have used that the events
$\{W_i\ne0\}$, $i=1,\dots,K-1$ form a partition of sample space, 
to obtain \eqref{eq:PXY|W-zero-form} we have used \eqref{eq:PWi-val},
and to obtain \eqref{eq:PXY|W-nonzero-form} we have used
that
\begin{align*} 
&P_{X,Y|\{W_j\neq0\}}(x,y) \notag\\
&\ = \frac{P_{X,Y|W_j}(x,y|{+1}) \, P_W(j) + P_{X,Y|W_j}(x,y|{-1}) \,
  P_W(-j)}{P_W(j)+P_W(-j)} \notag\\
&\ = \frac12 \left( P_{X,Y|W}(x,y|j) + P_{X,Y|W}(x,y|{-j}) \right).
\end{align*}
Hence,
\begin{align} 
&\bphi^{X,Y|W_i}_0 \notag\\
&\quad= \frac{1}{2(1-\sigmat_i)} \sum_{\{j\colon
  j\neq i\}}  \sigmat_j\, \left( \bphi^{X,Y|W}_j +
\bphi^{X,Y|W}_{-j} \right) \label{eq:use-ne0-form}\\
&\quad= \frac{1}{2(1-\sigmat_i)} \sum_{\{j\colon
  j\neq i\}}  \sigmat_j\, \left( \bphic^{X,Y|W}_j +
\bphic^{X,Y|W}_{-j} \right) + o(1)\label{eq:use-PXYw-form}\\
&\quad= o(1),
\label{eq:phiXY|0-small}
\end{align}
where to obtain \eqref{eq:use-ne0-form} we have used
\eqref{eq:PXY|W-nonzero-form} with \eqref{eq:ivXY|W-def} (setting
$\delta(\eps)=\eps$) and
\eqref{eq:phiXY|Wi-def}, 
to obtain \eqref{eq:use-PXYw-form} we have used
\eqref{eq:phi-phit-forms}, 
and to obtain \eqref{eq:phiXY|0-small} we have used 
\eqref{eq:ivcXY|W-def}
with \eqref{eq:phiXW-opt}--\eqref{eq:phiYW-opt}
to conclude that the term in parentheses is zero, since
for $w\in\W$,
\begin{align} 
&\bphic^{X,Y|W}_w \notag\\
&\quad= 
\sgn(w)\, 
\sqrt{\frac{\sigma_\abs{w}}{2\sigmat_\abs{w}\, \eps}} \,
\left( \sqrt{P_Y(y)}\,
  \psi^Y_{\abs{w}} + \sqrt{P_X(x)}\, \psi^X_{\abs{w}} \right).
\label{eq:bphitXY|W-opt}
\end{align}

From \eqref{eq:phi-phit-forms} with 
\eqref{eq:phitXY|W-bnd}, and from \eqref{eq:phiXY|0-small}, we conclude
\begin{equation*}
P_{X,Y|W_i}(\cdot,\cdot|w_i)\in\nbhd^{\X\times\Y}_{\sqrt{2\eps}(1+o(1))}(P_{X,Y}),
\quad w_i\in\W_\circ,
\end{equation*}
whence
\begin{align} 
&D(P_{X,Y|W_i}(\cdot,\cdot|j) \| P_{X,Y})  \notag\\
&\qquad\qquad= \eps\, \bnorm{\bphi^{X,Y|W_i}_j}^2 + o(\eps) \notag\\
&\qquad\qquad= \begin{cases} 
\eps\, \bnorm{\bphic^{X,Y|W}_i}^2 + o(\eps) & j=+1 \\
\eps\, \bnorm{\bphic^{X,Y|W}_{-i}}^2 + o(\eps) & j=-1 \\
o(\eps) & j=0,
\end{cases}
\label{eq:DPXY|Wi}
\end{align}
where to obtain the first equality we have used the special case of
\lemref{lem:D-char}, and to obtain the second equality we have used
\eqref{eq:phi-phit-forms} and \eqref{eq:phiXY|0-small}.
In turn, we obtain \eqref{eq:C-comp} via
\begin{align} 
I(W_i;X,Y)
&= \sum_{j\in\W_\circ} P_{W_i}(j)\,
D(P_{X,Y|W_i}(\cdot,\cdot|j) \| P_{X,Y}) \notag\\
&= 2\,\eps\,P_W(i)\, 
\bnorm{\bphic^{X,Y|W}_i}^2 
+ o(\eps) \label{eq:use-PWi-val-a}\\
&= \sigma_i + o(\eps),
\label{eq:use-PWi-val-b}
\end{align}
where to obtain the \eqref{eq:use-PWi-val-a} we have used the first equality
in \eqref{eq:PWi-val}, and to obtain \eqref{eq:use-PWi-val-b}
we have used the second equality in \eqref{eq:PWi-val}
and that 
\begin{align} 
&\bnorm{\bphic^{X,Y|W}_w}^2 \notag\\
&\qquad=\frac{\sigma_{\abs{w}}}{2\sigmat_{\abs{w}}\eps}\,
\sum_{x,y} \Bigl( \sqrt{P_Y(y)} \, 
     \psi^X_{\abs{w}}(x) +
      \sqrt{P_X(x)}\,\psi^Y_{\abs{w}}(y)
      \Bigr)^2 \label{eq:use-bphitXY|W-opt}\\ 
&\qquad= \frac{\sigma_{\abs{w}}}{\sigmat_{\abs{w}}\,\eps}.
\label{eq:phic-norm}
\end{align}
To obtain \eqref{eq:use-bphitXY|W-opt} we have used
\eqref{eq:bphitXY|W-opt}, and to obtain \eqref{eq:phic-norm} we have
used \eqref{eq:phiXWYW-ivspace}.

Turning now to the second part of the corollary, consistent with 
\defref{def:eci}, 
\begin{equation}
C_\eps(U_i,V_i) = \min_{P_{\Wt_i|U_i,V_i}\colon U_i\markov
    \Wt_i\markov V_i} I(\Wt_i;U_i,V_i).
\label{eq:CUVi-problem}
\end{equation}
by adapting the analysis of \propref{prop:common} we used to obtain
$C_\eps(X,Y)$.  In particular, from \eqref{eq:PUiVj} we have that
the counterpart to \eqref{eq:dtms-def} is
\begin{equation*}
\dtms_i(u_i,v_i) \defeq
\frac{P_{U_i,V_i}(u_i,v_i)}{\sqrt{P_{U_i}(u_i)}\,\sqrt{P_{V_i}(v_i)}}  
= \frac12 \bigl(1+\epst^2\,\sigma_i\,u_i\,v_i\bigr) + o(\epst^2)
\end{equation*}
for $u_i,v_i\in\{-1,+1\}$, and that to \eqref{eq:dtm-def}
is
\begin{multline*}
\dtm_i \defeq
\left[\sqrt{\bP_{V_i}}\right]^{-1} \, \bP_{V_i,U_i} \,
\left[\sqrt{\bP_{U_i}}\right]^{-1} \\
= \underbrace{\frac1{\sqrt{2}} \begin{bmatrix} 1 & 1 \\ 1 & -1 
\end{bmatrix}}_{\defeq\bPsi^{V_i}} \begin{bmatrix} 1 & 0 \\ 0 & \epst^2\,\sigma_i 
\end{bmatrix} \underbrace{\begin{bmatrix} 1 & 1 \\ 1 & -1 
\end{bmatrix} \frac1{\sqrt{2}}}_{\defeq\bigl(\bPsi^{U_i}\bigr)^\T} +
o(\epst^2). 
\end{multline*}
Note too that the counterpart to \eqref{eq:dtmt-def}, i.e., 
\begin{equation*} 
\dtmt_i 
\defeq
\left[\sqrt{\bP_{V_i}}\right]^{-1} \, \left[ \bP_{V_i,U_i} -
  \bP_{V_i}\,\bP_{U_i} \right]\,
\left[\sqrt{\bP_{U_i}}\right]^{-1}
\end{equation*}
satisfies
\begin{equation*}
\nuclear{\dtmt_i} = \epst^2\,\sigma_i +o(\epst^2) \le \epst^2\,
\nuclear{\dtmt} + o(\epst^2) \le \epst^2\,\eps+o(\epst^2) ,
\end{equation*}
so
\begin{equation*} 
P_{U_i,V_i}\in\nbhdk^{\U_i\times\V_i}_{\epst^2\eps+o(\epst^2)}(P_{U_i}P_{V_i}).
\end{equation*}
Hence, the counterpart of \eqref{eq:CXY-full} in \propref{prop:common}
for the new variables $(U_i,V_i)$ is
\begin{equation*}
C_\eps(U_i,V_i) = \epst^2\, \sigma_i + o(\epst^2\eps).
\end{equation*}
as $\eps,\epst\to0$, and thus \eqref{eq:CUV} follows.
\hfill\IEEEQED

\subsection{Proof of \protect\corolref{corol:Wi-markov}}
\label{app:Wi-markov}

Since the event $W_i=j$ is the event $W=ji$ for $j\in\{-1,+1\}$, the
cases $w_i=\pm1$ in \eqref{eq:Wi-markov} follow immediately from
\corolref{corol:suffstat-common}.  The case $w_i=0$ in
\eqref{eq:Wi-markov} is then determined by the constraint that the
result is a distribution.
\hfill\IEEEQED

\subsection{Proof of \protect\corolref{corol:Wti-markov}}
\label{app:Wti-markov}

First note that \eqref{eq:markov-1} 
is readily obtained
from \eqref{eq:PUVi|XY}, exploiting that $u_i$ and $v_i$ are uniquely
determined in the cases $z_i=\pm2$.  Second, note that 
\begin{align}
&P_{\Wt_i|Z_i,X^m,Y^m}(\wt_i|z_i,x^m,y^m) \notag\\
&\quad= \vphantom{\sum^x} \smash{\sum_{\substack{\{(u_i,v_i)\colon\\ u_i+v_i=z_i\}}}}
  P_{\Wt_i|U_i,V_i}(\wt_i|u_i,v_i)\notag\\
&\quad\qquad\qquad\qquad\qquad{}\cdot
P_{U_i,V_i|Z_i,X^m,Y^m}(u_i,v_i|z_i,x^m,y^m) \label{eq:markov-int},
\end{align}
which is obtained by exploiting \eqref{eq:Zi-def} and the structure in
$\Wt_i$ implicit in \eqref{eq:eps-ci-def-uv}.  To obtain
\eqref{eq:markov-2} from \eqref{eq:markov-int} we use that
\begin{equation*} 
P_{\Wt_i|U_i,V_i}(\wt_i|u_i,v_i) 
= \frac12 \bigl( 1 + \sgn(\wt_i\,z_i)\,
\sqrt{\sigma_i} \bigr) + o(\epst\sqrt{\eps}),
\end{equation*}
which is obtained by adapting \corolref{corol:suffstat-common}, and
that the second factor in the summation in \eqref{eq:markov-int} is
unity when $z_i=\pm2$ since $u_i$ and $v_i$ are uniquely determined in
these cases. \hfill\IEEEQED

\section{Appendices for \protect\secref{sec:ace}}
\label{app:ace}

\subsection{Proof of \protect\propref{prop:sval-est-tail}}
\label{app:sval-est-tail}

Our proof makes use of two lemmas.  The first is the following vector
generalization of Bernstein's inequality
\cite[Theorem~2.4]{CandesPlan2011}.
\begin{lemma}[Bernstein Inequality (Vector Version)]
\label{lem:bernstein-vector}
For some dimension $d$, let $\Zt_1,\dots,\Zt_n \in \reals^d$ be
independent zero-mean random vectors such that for some constant $c >
0$,
\begin{equation*}
\prob{\norm{\Zt_i} \le c}=1,\qquad i=1,\dots,n.
\end{equation*}
Moreover, let $\cbar\in(0,c^2]$ be a constant such
that
\begin{equation*}
\frac{1}{n} \sum_{i=1}^n \E{\norm{\Zt_i}^2} \le \cbar.
\end{equation*}
Then, for all $0 \le \delta \leq \cbar/c$,\footnote{As noted in
  \cite{CandesPlan2011}, this bound does not depend on $d$.}
\begin{equation*}
\prob{\bbnorm{\frac{1}{n}\sum_{i=1}^n \Zt_i } \ge \delta} 
\le \expop{\frac{1}{4} - \frac{\delta^2 n}{8 \cbar}}.
\end{equation*}
\end{lemma}
The second lemma is the following.
\begin{lemma}
\label{lem:kyfan-stab}
Given dimensions $k_1$ and $k_2$ and any two matrices $\bA_1, \bA_2 \in
\reals^{k_1 \times k_2}$, we have, for every
$k\in\bigl\{1,\dots,\min\{k_1,k_2\}\bigr\}$,
\begin{equation} 
\sum_{i=1}^k \babs{ \sigma_i(\bA_1) - \sigma_i(\bA_2)} \le \sqrt{k}\, \frob{\bA_1 - \bA_2}.
\label{eq:kyfan-stab}
\end{equation}
\end{lemma} 

\begin{IEEEproof}[Proof of \protect\lemref{lem:kyfan-stab}]
We have
\begin{align}
\sum_{i=1}^k \babs{ \sigma_i(\bA_1) - \sigma_i(\bA_2)} 
&\le \sum_{i=1}^k \sigma_i(\bA_1-\bA_2) \label{eq:use-lidskii}\\
&\le \sqrt{k}\, \sqrt{\sum_{i=1}^k
  \sigma_i(\bA_1-\bA_2)^2}  \label{eq:use-cs-stab}\\
&\le \sqrt{k}\, \sqrt{\sum_{i=1}^{\min\{k_1,k_2\}}
  \sigma_i(\bA_1-\bA_2)^2} \notag \\ 
&=\sqrt{k}\, \frob{\bA_1-\bA_2}, \label{eq:use-frob-diff-def}
\end{align}
where to obtain \eqref{eq:use-lidskii} we use 
the following standard
inequality (see, e.g., \cite[Theorem~3.4.5]{hj91}):
\begin{lemma}[Lidskii Inequality]
\label{lem:lidskii}
Given dimensions $k_1$ and $k_2$ and any matrices $\bA_1, \bA_2 \in
\reals^{k_1 \times k_2}$, we have, for every $k\in\bigl\{1,
\min\{k_1,k_2\}\bigr\}$ and $1 \leq i_1 < i_2 < \cdots < i_k \leq
\min\{k_1,k_2\}$, 
\begin{equation*} 
\sum_{j=1}^{k} \babs{ \sigma_{i_j}(\bA_1) - \sigma_{i_j}(\bA_2)}
\le \kyfank{\bA_1 - \bA_2},
\end{equation*}
where $\sigma_1(\cdot)\ge \cdots \ge \sigma_{\min\{k_1,k_2\}}(\cdot)$ denote
the ordered singular values of its (matrix) argument. 
\end{lemma} 
In turn, to obtain \eqref{eq:use-cs-stab}  we use the
Cauchy-Schwarz inequality, and to obtain \eqref{eq:use-frob-diff-def} we
use the definition of the Frobenius norm. 
\end{IEEEproof}

Our proof of \propref{prop:sval-est-tail} proceeds as follows.
First, for each $i \in \{1,\dots,n\}$ let $\bZ_i$ denote a random
$\cardY\times\cardX$ matrix with $(y,x)$\/th element
\begin{equation*}
Z_i(x,y) \defeq
\frac{\kron_{\{X_i = x,\,Y_i = y\}} - P_X(x)\, P_Y(y)}{\sqrt{P_X(x)\, 
    P_Y(y)}},\quad x\in\X,\ y\in\Y.
\end{equation*}
Accordingly, the $\bZ_1,\dots,\bZ_n$ are i.i.d.\ and $\bE{\bZ_i} =
\dtmt$.
Now
\begin{equation} 
\bZt_i\defeq \bZ_i-\bE{\bZ_i} = \bZ_i-\dtmt
\label{eq:bZt-def}
\end{equation}
satisfies
\begin{align}
\bfrob{\bZt_i}^2 
&= \sum_{x\in\X} \sum_{y\in\Y} \frac{\left(\kron_{\{X_i = x,\,Y_i = y\}} -
      P_{X,Y}(x,y)\right)^2}{P_X(x)\,P_Y(y)} \notag\\ 
&\le \frac{1}{p_0^2} \sum_{x\in\X} \sum_{y\in\Y} \left(\kron_{\{X_i =
  x,\,Y_i = y\}} - P_{X,Y}(x,y)\right)^2 \label{eq:use-marg-bnd}\\ 
&= \frac{1}{p_0^2} \Biggl[ \sum_{x\in\X} \sum_{y\in\Y} \kron_{\{X_i =
    x,\,Y_i = y\}} \notag\\
&\ \ \quad\qquad{} - 2 \sum_{x\in\X} \sum_{y\in\Y} \kron_{\{X_i =
    x,\,Y_i = y\}}\, P_{X,Y}(x,y) \notag\\
&\quad\qquad\qquad{} + \sum_{x\in\X} \sum_{y\in\Y}
  P_{X,Y}(x,y)^2 \Biggr]  \label{eq:3terms}\\ 
&\le \frac{2}{p_0^2} \defeq c^2
\label{eq:Zfrob}
\end{align}
where to obtain \eqref{eq:use-marg-bnd} we have used
\eqref{eq:marg-bnd}, and where to obtain \eqref{eq:Zfrob} we have used
that in \eqref{eq:3terms} the first term within the brackets is unity,
the second is upper bounded by zero, and the third term is upper
bounded by unity since
\begin{equation}
\sum_{z\in\Z} q(z)^2 \le \sum_{z\in\Z} q(z) =1,\ \text{any
  (countable) $\Z$ and $q\in\cP^\Z$}.
\label{eq:q2-sum-bnd}
\end{equation}
Moreover, 
\begin{align}
&\frac1n \sum_{i=1}^n \E{\bfrob{\bZt_i}^2} \notag\\
&\qquad\qquad=\E{\bfrob{\bZt_1}^2} \label{eq:use-iid}\\ 
&\qquad\qquad\le \frac{1}{p_0^2} \sum_{x\in\X} \sum_{y\in\Y} 
\varop{\kron_{\{X_1=x,\,Y_1=y\}}}  \label{eq:take-exp} \\
&\qquad\qquad= \frac{1}{p_0^2} \sum_{x\in\X} \sum_{y\in\Y}
\Bigl[ P_{X,Y}(x,y) - P_{X,Y}(x,y)^2 \Bigr] \label{eq:use-bern-var}\\ 
&\qquad\qquad\le \frac1{p_0^2} \defeq \cbar,
\label{eq:cbar-bnd}
\end{align}
where to obtain \eqref{eq:use-iid} we have used that
$\bZt_1,\dots,\bZt_n$ are i.i.d., to obtain \eqref{eq:take-exp}
we take the expectation of \eqref{eq:use-marg-bnd}, to obtain
\eqref{eq:use-bern-var} we have used that 
\begin{equation*} 
\varop{\kron_{\{X_1=x,\,Y_1=y\}}} =
P_{X,Y}(x,y) - P_{X,Y}(x,y)^2
\end{equation*}
since $\kron_{\{X_1=x,\,Y_1=y\}}$ is a Bernoulli random variable, and
to obtain \eqref{eq:cbar-bnd} we have used that the second term in
\eqref{eq:use-bern-var} is upper bounded by zero.

Finally, for $0\le \delta \le
\sqrt{k/2}/p_0$ we have
\begin{align}
\prob{\sum_{i=1}^k \babs{\sigmah_i - \sigma_i} \ge \delta} 
&\le \prob{ \frob{\dtmh - \dtmt} \ge \frac{\delta}{\sqrt{k}} }
\label{eq:kyfank-frob-bnd} \\
&= \prob{ \bbfrob{\frac1n \sum_{i=1}^n \bZt_i} \ge
  \frac{\delta}{\sqrt{k}} } \label{eq:use-BZ-rel} \\
&\le \expop{\frac{1}{4} - \frac{p_0^2\, \delta^2
  n}{8 k}}, 
\label{eq:kyfank-bnd}
\end{align}
where to obtain \eqref{eq:kyfank-frob-bnd} we have
used \lemref{lem:kyfan-stab}, to obtain \eqref{eq:use-BZ-rel} we have
used that
\begin{equation}
\dtmh - \dtmt= \frac1n \sum_{i=1}^n \bZt_i,
\label{eq:Bh-Bt-bZt-ident}
\end{equation}
and to obtain \eqref{eq:kyfank-bnd} we have used
\lemref{lem:bernstein-vector} with [cf.\ \eqref{eq:Zfrob}] $c=\sqrt{2}/p_0$
and [cf.\ \eqref{eq:cbar-bnd}] $\cbar=1/p_0^2$ (and construed the
associated $\bZt_i$ as vectors).\hfill\IEEEQED

\subsection{Proof of \protect\corolref{corol:sval-est-mse}}
\label{app:sval-est-mse}

First, we have
\begin{align}
\sum_{i=1}^k \bigl|\sigmah_i-\sigma_i\bigr|
&\le \sum_{i=1}^k \bigl( \sigmah_i + \sigma_i\bigr) \label{eq:use-triangle-kyfan}\\
&= \kyfank{\dtmh} + \kyfank{\dtmt} \notag\\
&\le k \left( 1 + \spectral{\dtmh} \right) \label{eq:use-kyfank-bnd}\\
&\le k \left( 1 + \frob{\dtmh}\right), \label{eq:use-spec-frob-bnd}
\end{align}
where to obtain \eqref{eq:use-triangle-kyfan} we have used the
triangle inequality, to obtain \eqref{eq:use-kyfank-bnd} we have used
that $\kyfank{\bA}\le k\spectral{\bA}$ for any matrix
$\bA\in\reals^{k_1\times k_2}$ and $k\in\{1,\dots,\min\{k_1,k_2\}\}$,
and to obtain \eqref{eq:use-spec-frob-bnd} we have used the standard inequality
\begin{equation}
\spectral{\bA}\le \frob{\bA}\quad\text{for any matrix $\bA$}. 
\label{eq:spec-frob-bnd}
\end{equation}
In turn, 
\begin{align}
\frob{\dtmh}^2
&= \sum_{x\in\X}\sum_{y\in\Y}
\frac{\bigl(\Ph_{X,Y}(x,y)-P_X(x)\,P_Y(y)\bigr)^2}{P_X(x)\,P_Y(y)} 
\notag\\
&\le \sum_{x\in\X}\sum_{y\in\Y} \left[ \frac{\Ph_{X,Y}(x,y)^2}{P_X(x)\,P_Y(y)} +
P_X(x)\,P_Y(y) \right] \notag\\ 
&\le \sum_{x\in\X}\sum_{y\in\Y} \left[ \frac{\Ph_{X,Y}(x,y)^2}{p_0^2} +
P_X(x)\,P_Y(y) \right] \label{eq:use-p0}\\ 
&\le \frac1{p_0^2} + 1 \label{eq:use-q2}\\
&\le \frac2{p_0^2}, \label{eq:frob-dtmh-bnd}
\end{align}
where to obtain \eqref{eq:use-p0} we have used \eqref{eq:marg-bnd}, and to
obtain \eqref{eq:use-q2} we have used \eqref{eq:q2-sum-bnd}.

Next, with the event
\begin{equation*}
\cE_\delta \defeq \left\{ \sum_{i=1}^k \bigl|\sigmah_i -
  \sigma_i\bigr| \ge \delta \right\},\qquad
0\le \delta\le\frac1{p_0}{\sqrt\frac{k}2},
\end{equation*}
we have\footnote{We use $\cmp{(\cdot)}$ to denote set complement.}
\begin{align}
&\E{\left(\sum_{i=1}^k \bigl|\sigmah_i-\sigma_i\bigr|\right)^2} \notag\\
&\qquad= \E{\left(\sum_{i=1}^k \bigl|\sigmah_i-\sigma_i\bigr|\right)^2 \,\middle|\,
  \cmp{\cE_\delta}} \prob{\cmp{\cE_\delta}}  \notag\\
&\qquad\qquad{} + \E{\left(\sum_{i=1}^k
    \bigl|\sigmah_i-\sigma_i\bigr|\right)^2 \,\middle|\, 
  \cE_\delta} \prob{\cE_\delta}   \notag\\
&\qquad\le \delta^2 + k^2\left(1+\frac{\sqrt{2}}{p_0}\right)^2 
\expop{\frac14 - \frac{p_0^2\,\delta^2 n}{8k}},
\label{eq:msk-gen}
\end{align}
where to obtain the inequality we have used that
$\prob{\cmp{\cE_\delta}}\le1$, \eqref{eq:use-spec-frob-bnd} with
\eqref{eq:frob-dtmh-bnd}, and \eqref{eq:sval-est-tail} in
\propref{prop:sval-est-tail}.

To obtain the tightest bound, we optimize \eqref{eq:msk-gen} over
$\delta$, yielding \eqref{eq:sval-est-mse}.  In particular, we have
\begin{align} 
&\E{\left(\sum_{i=1}^k \bigl|\sigmah_i-\sigma_i\bigr|\right)^2} \notag\\ 
&\qquad\le \min_\delta \left(\delta^2 +
  k^2\left(1+\frac{\sqrt{2}}{p_0}\right)^2 \expop{\frac14 -
    \frac{p_0^2\,\delta^2 n}{8k}}\right) \notag\\ 
&\qquad=  \frac{8k}{p_0^2\,n} \left( 1 + \ln\left[
    k^2\left(1+\frac{\sqrt{2}}{p_0}\right)^2 \e^{1/4}\,
    \frac{p_0^2n}{8k}\right]
  \right) \label{eq:use-lem-varphi-sval}\\ 
&\qquad\le   \frac{8k}{p_0^2n} \left( \frac34 + \ln(kn) \right)
\label{eq:use-p0-bnds} \\
&\qquad= \frac{6k + 8k\ln(kn)}{p_0^2n},
\end{align}
where to obtain \eqref{eq:use-lem-varphi-sval} we 
recognize that 
the right-hand side of \eqref{eq:msk-gen} takes the form of
\eqref{eq:varphi-ab-def} with the mappings
\begin{equation} 
a = k^2\left(1+\frac{\sqrt{2}}{p_0}\right)^2\e^{1/4}, \qquad
b = \frac{n p_0^2}{8k}, \qquad
\omega = \delta^2,
\label{eq:abw-vals}
\end{equation}
and apply \lemref{lem:varphi-ab}, and 
to obtain
\eqref{eq:use-p0-bnds} we have used that $p_0+\sqrt{2}\le 2$ since
$p_0\le 1/2$ as $\min\{\cardX,\cardY\}\ge2$, and that $\ln(2)\ge 1/2$.

It remains to determine conditions under which 
\begin{equation}
\omega_*\defeq
\delta_*^2 = \frac{8k}{p_0^2n} \ln\left[
  k^2\left(1+\frac{\sqrt{2}}{p_0}\right)^2 \e^{1/4}\, \frac{p_0^2n}{8k} \right]
\label{eq:w*-def}
\end{equation}
satisfies the conditions of 
\propref{prop:sval-est-tail}, viz., 
\begin{equation}
0\le \omega_* \le \frac{k}{2p_0^2}.  
\label{eq:w-constr}
\end{equation}
Proceeding, since from \eqref{eq:w*-def} we have
\begin{equation*} 
\omega_* \le \frac{8k}{p_0^2n} \biggl( \ln(4kn) + \underbrace{\frac14
  - \ln(8)}_{<0} \biggr) < \frac{k}{2p_0^2} \left[ \frac{16}{n}
  \ln(4kn) \right] 
\end{equation*}
where to obtain the first inequality we have again used that
$p_0+\sqrt{2}\le 2$.   Hence, the second inequality in
\eqref{eq:w-constr} is satisfied when $n$ is sufficiently large that
$n\ge 16\ln(4kn)$.

Moreover, since from \eqref{eq:w*-def} we also have
\begin{equation*}
\omega_* = \frac{8k}{p_0^2n} \ln\biggl[ k \bigl(p_0+\sqrt{2}\bigr)^2
  \e^{1/4} \frac{n}{8}\biggr]
> \frac{8k}{p_0^2n} \ln\left(\frac{n}4\right),
\end{equation*}
where to obtain the inequality we have used that $k\ge 1$, $p_0>0$,
and $\e^{1/4}>1$.  Hence, the first inequality in \eqref{eq:w-constr}
is satisfied when $n\ge4$, which we note is satisfied
when our condition for satisfying the second inequality in
\eqref{eq:w-constr} is.  Indeed, satisfying $n\ge 16\ln(4kn)$ even for
$k=1$ requires $n\ge 96$.\hfill\IEEEQED

\subsection{Proof of \protect\corolref{corol:I-est-err}}
\label{app:I-est-err}

First, note that
\begin{align*}
\bbabs{\frac12\sum_{i=1}^k \bigl(\sigmah_i^2 - \sigma_i^2\bigr)}
&\le \frac12\sum_{i=1}^k \babs{\sigmah_i^2 - \sigma_i^2} \\
&= \frac12\sum_{i=1}^k \babs{\sigmah_i -
  \sigma_i}\bigl(\sigmah_i+\sigma_i\bigr) \\ 
&\le \frac12\bigl(\sigmah_1+\sigma_1\bigr)\sum_{i=1}^k \babs{\sigmah_i
  - \sigma_i} , 
\end{align*}
where
\begin{align}
\sigma_1 + \sigmah_1
&= \bspectral{\dtmt}+\bspectral{\dtmh} \notag \\
&\le 1+ \bfrob{\dtmh} \label{eq:use-spec-frob-bnd-again} \\
&\le 1 + \frac{\sqrt{2}}{p_0} \label{eq:use-corol-sval-anal}\\
&= \frac{p_0+\sqrt{2}}{p_0} \le \frac2{p_0}, \label{eq:use-p0-bnd-again}
\end{align}
whence
\begin{equation}
\bbabs{\frac12\sum_{i=1}^k \bigl(\sigmah_i^2 - \sigma_i^2\bigr)}
\le \frac1{p_0} \sum_{i=1}^k \babs{\sigmah_i - \sigma_i}.
\label{eq:I-sval-rel}
\end{equation}
To obtain \eqref{eq:use-spec-frob-bnd-again} we have used
\eqref{eq:spec-frob-bnd} and that $\dtmt$ is contractive, and to
obtain \eqref{eq:use-corol-sval-anal} we have used
\eqref{eq:frob-dtmh-bnd} in the proof of
\corolref{corol:sval-est-mse}, and to obtain
\eqref{eq:use-p0-bnd-again} we have used 
that $p_0\le1/2$ as $\min\{\cardX,\cardY\}\ge2$.

Hence, we obtain \eqref{eq:I-est-tail} from \eqref{eq:I-sval-rel}
via
\begin{align*} 
\prob{\bbabs{\frac12\sum_{i=1}^k \bigl(\sigmah_i^2 - \sigma_i^2\bigr)}
  \ge \delta}
&\le 
\prob{\frac1{p_0} \sum_{i=1}^k \babs{\sigmah_i - \sigma_i}\ge \delta}
\\
&=\prob{\sum_{i=1}^k \babs{\sigmah_i - \sigma_i}\ge p_0\, \delta } \\
&\le \expop{\frac14 - \frac{p_0^4\,\delta^2 n}{8k}}
\end{align*}
where to obtain the final inequality we have used
\eqref{eq:sval-est-tail}, which holds for 
$0 \le p_0\,\delta \le \sqrt{k/2}/p_0$.
Moreover, we obtain \eqref{eq:I-est-mse} from \eqref{eq:I-sval-rel} via
\begin{align*}
\E{\bbabs{\frac12 \sum_{i=1}^k \bigl(\sigmah_i^2-\sigma_i^2\bigr)}^2} 
&\le \E{\frac1{p_0^2} \left(\sum_{i=1}^k \babs{\sigmah_i - \sigma_i}\right)^2} \\
&\le \frac{6k+8k\ln(nk)}{p_0^4n},
\end{align*}
where to obtain the final equality we have used
\eqref{eq:sval-est-mse}.\hfill\IEEEQED

\subsection{Proof of \protect\propref{prop:svec-est-tail}}
\label{app:svec-est-tail}

Our proof makes use of two lemmas.
The first is the following matrix generalization of Bernstein's
inequality \cite[Theorem 1.6]{Tropp2012}.
\begin{lemma}[Bernstein Inequality (Matrix Version)]
\label{lem:bernstein-matrix}
For some dimensions $d_1$ and $d_2$, let
$\bZt_1,\dots,\bZt_n\in\reals^{d_1\times d_2}$ be independent,
zero-mean random matrices such that for some constant $c>0$,
\begin{equation*}
\prob{\spectral{\bZt_i}\le c} = 1,\qquad i=1,\dots,n.
\end{equation*}
Moreover, let $\cbar\in(0,c^2]$ be a constant such
that
\begin{equation*}
\max\left\{
\bbspectral{\frac1n \sum_{i=1}^n \cov\bigl(\bZt_i\bigr)},
\bbspectral{\frac1n \sum_{i=1}^n \cov\bigl(\bZt_i^\T\bigr)}\right\}
\le \cbar,
\end{equation*}
where for an arbitrary random matrix $\bW$
\begin{equation*}
\cov(\bW) \defeq \E{\bigl(\bW-\bE{\bW}\bigr)\bigl(\bW-\bE{\bW}\bigr)^\T}.
\end{equation*}
Then, for all $0\le\delta\le \cbar/c$,
\begin{equation*}
\prob{\bbspectral{\frac1n \sum_{i=1}^n \bZt}\ge \delta} \le
(d_1+d_2)\expop{-\frac{3\delta^2 n}{8\cbar}}.
\end{equation*}
\end{lemma}

The second of these lemmas is as follows.
\begin{lemma}
\label{lem:2k-norm-stab}
Given $\bA_1,\bA_2\in\reals^{k_1\times k_2}$ and
$k\in\bigl\{1,\dots,\min\{k_1,k_2\}\bigr\}$, we have
\begin{equation}
0\le \bfrob{\bA_1 \bPsi_{(k)}^{\bA_1}}^2 - \bfrob{\bA_1
    \bPsi_{(k)}^{\bA_2}}^2
\le 4k\, \bspectral{\bA_1}\,  \bspectral{\bA_1-\bA_2},
\label{eq:2k-norm-stab}
\end{equation}
where
$\bpsi_i^\bA$ denotes the
right singular vector of $\bA$ corresponding to $\sigma_i(\bA)$, and
\begin{equation*}
\bPsi_{(k)}^\bA \defeq \begin{bmatrix} \bpsi_1^\bA & \cdots &
  \bpsi_k^\bA
\end{bmatrix},
\end{equation*}
which has orthnormal columns.
\end{lemma}
\begin{IEEEproof}[Proof of \protect\lemref{lem:2k-norm-stab}]
The left-hand inequality follows immediately from \lemref{lem:eig-k}.
For the right-hand inequality, we have
\begin{align}
&\bfrob{\bA_1 \bPsi^{\bA_1}_{(k)}}^2 - \bfrob{\bA_1
      \bPsi^{\bA_2}_{(k)}}^2 \notag \\
&\ = \sum_{i=1}^k \left( \bnorm{\bA_1 \bpsi^{\bA_1}_i}^2 - \bnorm{\bA_1
      \bpsi^{\bA_2}_i}^2 \right) \notag \\
&\ \le \sum_{i=1}^k \bbabs{ \bnorm{\bA_1 \bpsi^{\bA_1}_i}^2 - \bnorm{\bA_1
      \bpsi^{\bA_2}_i}^2} \label{eq:use-triangle-prelim} \\
&\ = \sum_{i=1}^k \bbabs{\bnorm{\bA_1 \bpsi^{\bA_1}_i} - \bnorm{\bA_1
      \bpsi^{\bA_2}_i}}  \Bigl(\bnorm{\bA_1 \bpsi^{\bA_1}_i} +
  \bnorm{\bA_1 \bpsi^{\bA_2}_i}\Bigr) \notag \\ 
&\ \le 2\,\bspectral{\bA_1} \sum_{i=1}^k \bbabs{\bnorm{\bA_1
      \bpsi^{\bA_1}_i} - \bnorm{\bA_1 \bpsi^{\bA_2}_i}} \label{eq:use-spectral} \\ 
&\ \le 2\,\bspectral{\bA_1} \sum_{i=1}^k \biggl( \bbabs{\bnorm{\bA_1
    \bpsi^{\bA_1}_i} - \bnorm{\bA_2 \bpsi^{\bA_2}_i}} \notag\\
&\qquad\qquad\qquad\qquad{} + \bbabs{\bnorm{\bA_2 \bpsi^{\bA_2}_i} - \bnorm{\bA_1 \bpsi^{\bA_2}_i}} \biggr)
  \label{eq:use-triangle} \\ 
&\ \le 2\,\bspectral{\bA_1} \sum_{i=1}^k \Bigl( \babs{
  \sigma_i(\bA_1) - \sigma_i(\bA_2)} + \bnorm{(\bA_1 - \bA_2)
    \bpsi^{\bA_2}_i} \Bigr) \label{eq:use-reverse-triangle} \\  
&\ \le 2\,\bspectral{\bA_1} \sum_{i=1}^k \Bigl( \babs{
  \sigma_i(\bA_1) - \sigma_i(\bA_2)} + \bspectral{\bA_1 - \bA_2}
  \Bigr) \label{eq:use-spectral-again} \\  
&\ \le 4 k \,\bspectral{\bA_1}\, \bspectral{\bA_2 - \bA_1},
\label{eq:use-weyl}
\end{align}
where to obtain \eqref{eq:use-triangle-prelim} we have used the
triangle inequality, to obtain \eqref{eq:use-spectral} we have used
\factref{fact:frob-submult}, to obtain \eqref{eq:use-triangle} we have
again used the triangle inequality, to obtain
\eqref{eq:use-reverse-triangle} we have used that $\bnorm{\bA_1
  \bpsi^{\bA_1}_i} = \sigma_i(\bA_1)$ and $\bnorm{\bA_2
  \bpsi^{\bA_2}_i} = \sigma_i(\bA_2)$, and the (reverse) triangle
inequality, to obtain \eqref{eq:use-spectral-again} we have
again used \factref{fact:frob-submult}, and to obtain
\eqref{eq:use-weyl} we have used the following standard inequality
\cite[Corollary~7.3.5(a)]{hj12}
\cite[Theorem~1]{SVDPerturbationTheory}:
\begin{lemma}[Weyl Inequality]
\label{lem:weyl}
For every $\bA_1,\bA_2 \in \reals^{k_1\times k_2}$, we have,
with $K \defeq \min\{k_1,k_2\}$,
\begin{equation}
\max_{1 \le i \le K}{\bigl|\sigma_i\left(\bA_1\right) -
  \sigma_i\left(\bA_2\right)\bigr|} \le \bspectral{\bA_1-\bA_2}.
\label{eq:weyl}
\end{equation}
\end{lemma}
\end{IEEEproof}

Our proof of \propref{prop:svec-est-tail} proceeds as follows.
First, with $\fc_i$ as defined in \eqref{eq:fc-def} and 
$\bpsih^X_i$ as defined via \eqref{eq:psihX-def}, 
we have
\begin{align} 
&\Ed{P_Y}{\bnorm{\bEd{P_{X|Y}}{f^k_*(X)}}^2 -
  \bnorm{\bEd{P_{X|Y}}{\fc^k_*(X)}}^2} \notag\\
&\quad\qquad\qquad\qquad\qquad= 
\bfrob{\dtmt\,\bPsi^X_{(k)}}^2 -
\bfrob{\dtmt\,\bPsih^X_{(k)}}^2 \label{eq:use-ce-frob-rel}\\
&\quad\qquad\qquad\qquad\qquad\le  4k\,
\bspectral{\dtmt}\,\bspectral{\dtmt-\dtmh} \label{eq:use-lem-2k-norm-stab}
\\
&\quad\qquad\qquad\qquad\qquad\le 4k\,
\bspectral{\dtmt-\dtmh},
\label{eq:feat-diff-bnd}
\end{align}
where to obtain \eqref{eq:use-ce-frob-rel} we have used 
\eqref{eq:ce-frob-rel}, to obtain \eqref{eq:use-lem-2k-norm-stab} we
have used \lemref{lem:2k-norm-stab}, and 
to obtain \eqref{eq:feat-diff-bnd} we have used that
$\bspectral{\dtmt}\le 1$.   

Now let 
$\dtmh_i$ denote an $\cardY\times\cardX$ matrix with
$(y,x)$\/th entry 
\begin{equation*}
\dtmhs_i(x,y) \defeq
\frac{\kron_{X_i=x,\ Y_i=y}}{\sqrt{P_X(x)\,P_Y(y)}},
\end{equation*}
and  let
\begin{equation*}
\bZt_i \defeq \dtmh_i - \dtm,
\end{equation*}
which we note is consistent with the definition \eqref{eq:bZt-def} in
the proof of \propref{prop:sval-est-tail} and thus
$\E{\bZt_i}=\bzero$.  Then we have
\begin{align}
\bspectral{\bZt_i} 
&= \bspectral{\dtmh_i - \dtm} \notag\\
&\le \bspectral{\dtm} + \bspectral{\dtmh_i} \label{eq:use-spectral-triangle} \\
&= 1 + \frac{1}{\sqrt{P_X(X_i)\,P_Y(Y_i)}} \label{eq:use-spectral-dtmhi}\\
&\le 1 + \frac{1}{p_0} \defeq c, \label{eq:use-p0-svec}
\end{align}
where to obtain \eqref{eq:use-spectral-triangle} we have used the
spectral norm triangle inequality, to obtain
\eqref{eq:use-spectral-dtmhi} we have used that $\spectral{\dtm}=1$
and $\dtmh_i$ has a single nonzero entry so (with the usual abuse of
notation as discussed in footnote \ref{fn:abuse}) $\bfe_{Y_i}$ and
$\bfe_{X_i}$ are its principal left and right singular vectors,
respectively, and to obtain \eqref{eq:use-p0-svec} we have used the
definition of $p_0$.

Next, we have
\begin{align}
\bbspectral{\frac1n \sum_{i=1}^n \cov(\bZt_i)}
&= \bspectral{ \cov(\bZt_1)} \label{eq:bZi-id}\\
&= \bspectral{\bE{(\dtmh_1-\dtm)(\dtmh_1-\dtm)^\T}} \notag\\
&= \bspectral{\bE{\dtmh_1\dtmh_1^\T} -\dtm\dtm^\T} \notag\\
&\le \bspectral{\dtm\dtm^\T} +
\bspectral{\bE{\dtmh_1\dtmh_1^\T}} \label{eq:use-spectral-triangle-again}\\ 
&=  1 + \max_{y\in\Y} \sum_{x\in\X}
\frac{P_{X|Y}(x|y)}{P_X(x)} \label{eq:useEBh-form}\\ 
&\le 1 + \frac1{p_0} \max_{y\in\Y} \sum_{x\in\X}
P_{X|Y}(x|y)\label{eq:use-p0-svec-again} \\
&= 1 + \frac1{p_0},
\label{eq:cbar-calc}
\end{align}
where to obtain \eqref{eq:bZi-id} we have used that the
$\bZt_1,\dots,\bZt_n$ are identically distributed, to obtain
\eqref{eq:use-spectral-triangle-again} we have again used the triangle
inequality, to obtain \eqref{eq:useEBh-form} we have used that
$\sigma_0^2=1$ is the principal singular value of $\dtm\dtm^\T$, and
that $\dtmh_1\dtmh_1^\T$ is a diagonal matrix whose $(y,y)$\/th entry
is $\kron_{Y_1=y}/\bigl(P_X(X_1)\,P_Y(y)\bigr)$, so
$\smash{\bE{\dtmh_1\dtmh_1^\T}}$ has $(y,y)$\/th entry
\begin{align*} 
\E{\frac{\kron_{Y_1=y}}{P_X(X_1)\,P_Y(y)}} 
&= \sum_{x'\in\X,\,y'\in\Y}\!\!\!
P_{X,Y}(x',y') \frac{\kron_{y'=y}}{P_X(x')\,P_Y(y')} \\
&= \sum_{x\in\X}
\frac{P_{X,Y}(x,y)}{P_X(x)\,P_Y(y)} \\
&= \sum_{x\in\X}
\frac{P_{X|Y}(x|y)}{P_X(x)},
\end{align*}
and to obtain \eqref{eq:use-p0-svec-again} we have again used the
definition of $p_0$.
Moreover, interchanging the roles of $x$ and $y$, we have, by symmetry, 
\begin{equation} 
\bbspectral{\frac1n \sum_{i=1}^n \cov(\bZt_i^\T)}
= \bbspectral{\frac1n \sum_{i=1}^n \cov(\bZt_i)}
= 1 + \frac1{p_0} \defeq \cbar,
\label{eq:cbar-val}
\end{equation}
where to obtain the second equality we have used \eqref{eq:cbar-calc}.

Finally, we have
\begin{align}
&\probd{\fc^k_*}{\Ed{P_Y}{\bnorm{\bEd{P_{X|Y}}{f^k_*(X)}}^2 -
  \bnorm{\bEd{P_{X|Y}}{\fc^k_*(X)}}^2}
 \ge \delta} \notag\\
&\qquad\le
\prob{\bspectral{\dtmt-\dtmh} \ge
  \frac{\delta}{4k}} \label{eq:use-feat-diff-bnd}\\ 
&\qquad\le
\prob{\bbspectral{\frac1n\sum_{i=1}^n \bZt_i} \ge
  \frac{\delta}{4k}} \label{eq:use-Bht-bZt}\\ 
&\qquad\le \bigl(\cardX+\cardY\bigr)
\expop{-\frac{3n}{8}\left(\frac{1}{1+1/p_0}\right)\left(\frac{\delta}{4k}\right)^2}
\label{eq:use-bernstein-matrix} \\ 
&\qquad\le \bigl(\cardX+\cardY\bigr)
\expop{-\frac{p_0\,\delta^2\, n}{64\, k^2}},
\label{eq:use-p0-svec-bound}
\end{align}
where to obtain \eqref{eq:use-feat-diff-bnd} we have used
\eqref{eq:feat-diff-bnd}, to obtain \eqref{eq:use-Bht-bZt} we have used
\eqref{eq:Bh-Bt-bZt-ident}, 
to obtain
\eqref{eq:use-bernstein-matrix} we have used
\lemref{lem:bernstein-matrix},
and to obtain
\eqref{eq:use-p0-svec-bound} we have again used that $p_0\le1/2$ since
$\min\{\cardX,\cardY\}\ge2$. \hfill\IEEEQED

\subsection{Proof of \protect\corolref{corol:svec-est-mse}}
\label{app:svec-est-mse}

First, adapting our notation from \eqref{eq:fq-meas} for convenience,
\begin{align}
\mut_2\bigl(\fc^k_*\bigr) 
&\defeq
\Ed{P_Y}{\bnorm{\bEd{P_{X|Y}}{f^k_*(X)}}^2 -
      \bnorm{\bEd{P_{X|Y}}{\fc^k_*(X)}}^2} \notag\\
&= \bfrob{\dtmt\bPsi^X_{(k)}}^2 -
\bfrob{\dtmt\bPsih^X_{(k)}}^2 \label{eq:EtoB}\\ 
&\le \bfrob{\dtmt\bPsi^X_{(k)}}^2 \notag\\
&\le \bspectral{\dtmt}^2\, \bfrob{\bPsi^X_{(k)}}^2 \label{eq:use-submult-svec}\\
&\le \sum_{i=1}^k \bnorm{\bpsi_i^X}^2 \label{eq:use-contractive-svec} \\
&= k, \label{eq:use-orthon-svec}
\end{align}
where to obtain \eqref{eq:EtoB} we have
used  \eqref{eq:use-ce-frob-rel}, to
obtain \eqref{eq:use-submult-svec} we have used
\factref{fact:frob-submult}, to obtain \eqref{eq:use-contractive-svec}
we have used that $\bspectral{\dtmt}\le1$, and to obtain
\eqref{eq:use-orthon-svec} we have used that the singular vectors have
unit norm.

Next, with the event
\begin{equation*}
\cE_\delta \defeq \left\{ \mut_2\bigl(\fc^k_*\bigr) \ge \delta \right\},\qquad
0\le \delta\le 4k,
\end{equation*}
we have that the left-hand side of \eqref{eq:svec-est-mse} is bounded
according to
\begin{align}
\Ed{\fc^k_*}{\mut_2\bigl(\fc^k_*\bigr)^2}
&=
\Ed{\fc^k_*}{\mut_2\bigl(\fc^k_*\bigr)^2\Bigm|\cmp{\cE_\delta}}
\prob{\cmp{\cE_\delta}} \notag\\
&\qquad\qquad{}+ 
\Ed{\fc^k_*}{\mut_2\bigl(\fc^k_*\bigr)^2\Bigm|\cE_\delta}
\prob{\cE_\delta} \notag\\
&\le \delta^2 + k^2 \bigl(\cardX+\cardY\bigr) \expop{-\frac{p_0\,
    \delta^2 n}{64 k^2}}
\label{eq:svec-gen-bnd}
\end{align}
where to obtain the inequality we have used that
$\prob{\cmp{\cE_\delta}}\le1$,  \eqref{eq:use-orthon-svec}, and 
\propref{prop:svec-est-tail}.

To obtain the tightest bound, we optimize \eqref{eq:svec-gen-bnd} over
$\delta$, yielding \eqref{eq:svec-est-mse}.  In particular, we have
\begin{align}
\Ed{\fc^k_*}{\mut_2\bigl(\fc^k_*\bigr)^2}
&\le \min_\delta \left( \delta^2 + k^2 \bigl(\cardX+\cardY\bigr) \expop{-\frac{p_0\,
    \delta^2 n}{64 k^2}}
\right) \notag\\
&= \frac{64k^2}{p_0 n} \left[1+ \ln\left(k^2\bigl(\cardX+\cardY\bigr)
  \frac{p_0 n}{64k^2}\right)\right] \label{eq:use-lem-varphi-svec}\\
&= \frac{64k^2}{p_0 n} \Bigl(\ln\left[\bigl(\cardX+\cardY\bigr)
  p_0 n \right] + \bigl[1-\ln(64)\bigr] \Bigr) \notag\\
&\le \frac{64k^2}{p_0 n} \Bigl( \ln\left[\bigl(\cardX+\cardY\bigr)
  p_0 n \right] - 3\Bigr),
\end{align}
where to obtain \eqref{eq:use-lem-varphi-svec} we recognize that the
right-hand side of \eqref{eq:svec-gen-bnd} takes the form of
\eqref{eq:varphi-ab-def} with the mappings
\begin{equation} 
a = k^2\bigl(\cardX+\cardY\bigr), \qquad
b = \frac{p_0 n}{64k^2}, \qquad
\omega = \delta^2,
\label{eq:abw-vals-svec}
\end{equation}
and apply
\lemref{lem:varphi-ab}, and
to obtain the last inequality we have used that $\ln(64)\ge 4$.

It remains to impose the constraints $0\le \delta_*\le 4k$ on the
minimizer $\delta_*$, which we equivalently express in the form $0\le
\omega_* \le 16 k^2$ using \eqref{eq:abw-vals-svec}.  Substituting
\eqref{eq:varphi-ab-argmin} from \lemref{lem:varphi-ab} for $\omega_*$
and using $a$ and $b$ from \eqref{eq:abw-vals-svec}, the constraint
$\omega_*\ge0$ imposes \eqref{eq:ss-a}, viz.,
\begin{equation*}
\frac{p_0 n}{64} \ge \frac1{\bigl(\cardX+\cardY\bigr)}.
\end{equation*}
Meanwhile, the constraint $\omega_*\le 16k^2$ imposes \eqref{eq:ss-b},
viz.,
\begin{equation*}
\frac{p_0 n}{4} \ge \ln\left(\frac{p_0 n}{64}
\bigl(\cardX+\cardY\bigr)\right).
\end{equation*}
\hfill\IEEEQED

\subsection{Proof of \propref{prop:sval-est-tail-alt}}
\label{app:sval-est-tail-alt}

To obtain \propref{prop:sval-est-tail-alt}, we adapt the proof of
\propref{prop:sval-est-tail}, replacing the use of the Frobenius norm
of \lemref{lem:kyfan-stab} with the following
spectral norm bound:
\begin{lemma}
\label{lem:kyfan-stab-alt}
Given dimensions $k_1$ and $k_2$ and any two matrices $\bA_1, \bA_2 \in
\reals^{k_1 \times k_2}$, we have, for every
$k\in\bigl\{1,\dots,\min\{k_1,k_2\}\bigr\}$,
\begin{equation} 
\sum_{i=1}^k \babs{ \sigma_i(\bA_1) - \sigma_i(\bA_2)} 
\le k\,
\spectral{\bA_1 - \bA_2}.
\label{eq:kyfan-stab-alt}
\end{equation}
\end{lemma}
\begin{IEEEproof}[Proof of \lemref{lem:kyfan-stab-alt}]
We have
\begin{equation*} 
\sum_{i=1}^k \babs{ \sigma_i(\bA_1) - \sigma_i(\bA_2)} 
\le \sum_{i=1}^k \sigma_i(\bA_1 - \bA_2)
\le k\,
\spectral{\bA_1 - \bA_2},
\end{equation*}
where the second inequality follows from
\lemref{lem:lidskii}.
\end{IEEEproof}

In particular, to establish \propref{prop:sval-est-tail-alt}, we
replace \eqref{eq:kyfank-frob-bnd}--\eqref{eq:kyfank-bnd} in
\appref{app:sval-est-tail} with
\begin{align}
\prob{\sum_{i=1}^k \babs{\sigmah_i - \sigma_i} \ge \delta} 
&\le \prob{ \spectral{\dtmh - \dtmt} \ge \frac{\delta}{k} }
\label{eq:kyfank-frob-bnd-alt} \\
&= \prob{ \bbspectral{\frac1n \sum_{i=1}^n \bZt_i} \ge
  \frac{\delta}{k} } \label{eq:use-BZ-rel-alt} \\
&\begin{aligned}\le \bigl(\cardX+\cardY\bigr) \expop{-\frac{p_0\, \delta^2
  n}{4 k^2}},\\
 0\le\delta\le k,
\end{aligned}
\label{eq:kyfank-bnd-alt}
\end{align}
where to obtain \eqref{eq:kyfank-frob-bnd-alt} we use
\lemref{lem:kyfan-stab-alt}, then, as in the proof of
\propref{prop:svec-est-tail},
to obtain \eqref{eq:use-BZ-rel-alt} we
use \eqref{eq:Bh-Bt-bZt-ident}, 
and to obtain \eqref{eq:kyfank-bnd-alt}
we use \lemref{lem:bernstein-matrix} with \eqref{eq:use-p0-svec} and
\eqref{eq:cbar-val} providing $c$ and $\cbar$, respectively, and that
$p_0\le 1/2$.  \hfill\IEEEQED

\subsection{Proof of \propref{prop:svec-est-tail-alt}}
\label{app:svec-est-tail-alt}

We adapt the proof of \propref{prop:svec-est-tail}, replacing the use
of the spectral norm bound of \lemref{lem:2k-norm-stab} with the following
Frobenius norm bound:
\begin{lemma}
\label{lem:2k-norm-stab-alt}
Given $\bA_1,\bA_2\in\reals^{k_1\times k_2}$ and
$k\in\bigl\{1,\dots,\min\{k_1,k_2\}\bigr\}$, we have
\begin{equation}
0\le \bfrob{\bA_1 \bPsi_{(k)}^{\bA_1}}^2 - \bfrob{\bA_1
    \bPsi_{(k)}^{\bA_2}}^2
\le 4\sqrt{k}\, \bspectral{\bA_1}\,  \bfrob{\bA_1-\bA_2},
\label{eq:2k-norm-stab-alt}
\end{equation}
where
$\bPsi_{(k)}^\bA$ is as defined in \lemref{lem:2k-norm-stab}.
\end{lemma}
\begin{IEEEproof}[Proof of \lemref{lem:2k-norm-stab-alt}]
First, reproducing \eqref{eq:use-reverse-triangle} from the proof of
\lemref{lem:2k-norm-stab}, we have
\begin{align}
&\bfrob{\bA_1 \bPsi^{\bA_1}_{(k)}}^2 - \bfrob{\bA_1
      \bPsi^{\bA_2}_{(k)}}^2 \notag \\
&\qquad
\begin{aligned}
&\le 2\,\bspectral{\bA_1} \Biggl(\sum_{i=1}^k \babs{
  \sigma_i(\bA_1) - \sigma_i(\bA_2)} \\
&\qquad\qquad\qquad\qquad{}+ \vphantom{\sum}\smash[t]{\sum_{i=1}^k\bnorm{(\bA_1 - \bA_2)
    \bpsi^{\bA_2}_i}\Biggr)}. 
\end{aligned}
\label{eq:use-reverse-triangle-alt}
\end{align}
For the second sum in \eqref{eq:use-reverse-triangle-alt}, we have
\begin{align}
\left(\sum_{i=1}^k \bnorm{(\bA_1 - \bA_2) \bpsi^{\bA_2}_i}\right)^2
&\le k \sum_{i=1}^k \bnorm{(\bA_1 - \bA_2)
  \bpsi^{\bA_2}_i}^2 \label{eq:use-cs-alt}\\ 
&= k\, \bfrob{(\bA_1 - \bA_2)
    \bPsi^{\bA_2}_{(k)}}^2 \notag\\
&\le k\, \bfrob{\bA_1 - \bA_2}^2 
\label{eq:use-eig-k-alt}
\end{align}
where to obtain \eqref{eq:use-cs-alt} we use the Cauchy-Schwarz
inequality, and to obtain \eqref{eq:use-eig-k-alt} we use
\lemref{lem:eig-k}, recognizing that the right-hand side of
\eqref{eq:var-eig-k} is upper bounded by $\frob{\bA}^2$.  Hence, using
\lemref{lem:kyfan-stab} to bound the first term in
\eqref{eq:use-reverse-triangle-alt}, and \eqref{eq:use-eig-k-alt} to
bound the second, we obtain \eqref{eq:2k-norm-stab-alt}.
\end{IEEEproof}

To establish \propref{prop:svec-est-tail-alt}, 
starting from \eqref{eq:use-ce-frob-rel} in
\appref{app:svec-est-tail}, but using \lemref{lem:2k-norm-stab-alt}
instead of \lemref{lem:2k-norm-stab}, the bound \eqref{eq:feat-diff-bnd}
becomes
\begin{align} 
&\Ed{P_Y}{\bnorm{\bEd{P_{X|Y}}{f^k_*(X)}}^2 -
  \bnorm{\bEd{P_{X|Y}}{\fc^k_*(X)}}^2} \notag\\
&\quad\qquad\qquad\qquad\qquad= 
\bfrob{\dtmt\,\bPsi^X_{(k)}}^2 -
\bfrob{\dtmt\,\bPsih^X_{(k)}}^2 \notag\\
&\quad\qquad\qquad\qquad\qquad\le 4\sqrt{k}\,
\bfrob{\dtmt-\dtmh}.
\label{eq:feat-diff-bnd-alt}
\end{align}
In turn,
\eqref{eq:use-feat-diff-bnd}--\eqref{eq:use-p0-svec-bound} then
becomes [cf.\ \eqref{eq:svec-est-tail}]
\begin{align} 
&\probd{\fc^k_*}{\Ed{P_Y}{\bnorm{\bEd{P_{X|Y}}{f^k_*(X)}}^2 - \bnorm{\bEd{P_{X|Y}}{\fc^k_*(X)}}^2}  \ge \delta}
  \notag\\
&\qquad\le
\prob{\bfrob{\dtmt-\dtmh} \ge
  \frac{\delta}{4\sqrt{k}}} \label{eq:use-feat-diff-bnd-alt}\\ 
&\qquad= \prob{ \bbspectral{\frac1n \sum_{i=1}^n \bZt_i} \ge
  \frac{\delta}{4\sqrt{k}} } \label{eq:use-Bh-Bt-bZt-ident} \\
&\qquad
  \le \expop{\frac14 - \frac{p_0^2\, \delta^2 n}{128k}},\quad 0 \le
  \delta \le (4/p_0)\sqrt{k/2},
\label{eq:svec-est-tail-alt-pf}
\end{align}
where, as in the proof of
\propref{prop:sval-est-tail},  to obtain 
\eqref{eq:use-Bh-Bt-bZt-ident} we use \eqref{eq:Bh-Bt-bZt-ident}, and to obtain
\eqref{eq:svec-est-tail-alt-pf} we use
\lemref{lem:bernstein-vector}, with \eqref{eq:Zfrob} and
\eqref{eq:cbar-bnd} again providing $c$ and 
$\cbar$, respectively.  \hfill\IEEEQED

\subsection{Analysis of Feature Quality Measure \protect\eqref{eq:fq-meas-alt}}
\label{app:fq-alt}

To begin, we have
\begin{align}
&\bfrob{\bE{f^k_*(X)\,g^k_*(Y)^\T}-\bE{\fc^k_*(X)\,\gc^k_*(Y)^\T}}\notag\\
&\qquad= \bfrob{\bigl(\bPsi^Y_{(k)}\bigr)^\T\dtmt \bPsi^X_{(k)} 
- \bigl(\bPsih^Y_{(k)}\bigr)^\T\dtmt \bPsih^X_{(k)} } \notag\\
&\qquad= \bfrob{\bigl(\bPsi^Y_{(k)}\bigr)^\T\dtmt \bPsi^X_{(k)} 
- \bigl(\bPsih^Y_{(k)}\bigr)^\T\dtmh \bPsih^X_{(k)} \notag\\
&\qquad\qquad\qquad{} - \bigl(\bPsih^Y_{(k)}\bigr)^\T\dtmt \bPsih^X_{(k)} 
+ \bigl(\bPsih^Y_{(k)}\bigr)^\T\dtmh \bPsih^X_{(k)}} \notag\\
&\qquad =\bfrob{\bigl(\bSi_{(k)} - \bSih_{(k)}\bigr) -
\bigl(\bPsih^Y_{(k)}\bigr)^\T(\dtmt-\dtmh) \bPsih^X_{(k)}},
\label{eq:fq-alt}
\end{align}
where $\bSih_{(k)}$ is a diagonal matrix whose diagonal elements are
$\sigmah_1,\dots,\sigmah_k$.  

This measure exhibits similar sample complexity behavior to that
obtained in \secref{sec:sampcomp-features}.
To see this, note that in this case we have
\begin{align}
&\bfrob{\bigl(\bSi_{(k)} - \bSih_{(k)}\bigr) -
\bigl(\bPsih^Y_{(k)}\bigr)^\T(\dtmt-\dtmh) \bPsih^X_{(k)}} \notag\\
&\qquad\le \bfrob{\bSi_{(k)} - \bSih_{(k)}} +
  \underbrace{\bspectral{\bPsih^Y_{(k)}}}_{=1} \bfrob{(\dtmt-\dtmh) \bPsih^X_{(k)}},
\label{eq:fq-bnd}
\end{align}
where we have used, in turn, the triangle inequality for the Frobenius
norm, and \factref{fact:frob-submult}, and where the spectral norm is
unity because $\bPsih^Y_{(k)}$ has orthonormal columns.  Moreover, the
first term in \eqref{eq:fq-bnd} satisfies, using
\lemref{lem:kyfan-stab-alt},
\begin{equation} 
\bfrob{\bSi_{(k)} - \bSih_{(k)}} 
\le \sum_{i=1}^k \abs{\sigma_i-\sigmah_i} \le k\,\bspectral{\dtmt-\dtmh},
\label{eq:fq-bnd-a}
\end{equation}
while the remaining term satisfies
\begin{align} 
\bfrob{(\dtmt-\dtmh) \bPsih^X_{(k)}} 
&\le \sqrt{\sum_{i=1}^k \sigma_i\bigl(\dtmt-\dtmh\bigr)^2} \label{eq:use-eig-k-lem-fq}\\
&\le \sum_{i=1}^k \sigma_i\bigl(\dtmt-\dtmh\bigr) \label{eq:use-std-norm}\\
&\le k\,\bspectral{\dtmt-\dtmh},
\label{eq:fq-bnd-b}
\end{align}
where to obtain \eqref{eq:use-eig-k-lem-fq} we have used
\lemref{lem:eig-k}, and to obtain \eqref{eq:use-std-norm} we have used
\eqref{eq:l2-l1-bnd}.   Using \eqref{eq:fq-bnd-a} and
\eqref{eq:fq-bnd-b} in \eqref{eq:fq-bnd}, and, in turn,
\eqref{eq:fq-alt} yields
\begin{equation} 
\bfrob{\bE{f^k_*(X)\,g^k_*(Y)^\T}-\bE{\fc^k_*(X)\,\gc^k_*(Y)^\T}}
\le 2k\,\bspectral{\dtmt-\dtmh}.
\label{eq:fq-alt-bnd}
\end{equation}
Thus, we obtain a bound of the same form (to within a factor of two)
as that for the measure \eqref{eq:fq-meas}, for which we obtained
\begin{align} 
&\Ed{P_Y}{\bnorm{\bEd{P_{X|Y}}{f^k_*(X)}}^2 -
  \bnorm{\bEd{P_{X|Y}}{\fc^k_*(X)}}^2} \notag\\
&\ \qquad\qquad\qquad\qquad\qquad\qquad\qquad \le 4k\,\bspectral{\dtmt-\dtmh}.
\end{align}
As such analogous sample complexity bounds follow.

Finally, as in \secref{sec:sampcomp-compl}, we can similarly replace
the use of the spectral norm with the Frobenius norm.  In particular,
\eqref{eq:fq-bnd-a} can be replaced with
\begin{equation} 
\bfrob{\bSi_{(k)} - \bSih_{(k)}} 
\le \sum_{i=1}^k \abs{\sigma_i-\sigmah_i} \le \sqrt{k}\,\bfrob{\dtmt-\dtmh},
\label{eq:fq-bnd-a-alt}
\end{equation}
where we now use \lemref{lem:kyfan-stab} instead of
\lemref{lem:kyfan-stab-alt}.  Using \eqref{eq:fq-bnd-a-alt}, and the
simple upper bound
\begin{equation}
\bfrob{(\dtmt-\dtmh) \bPsih^X_{(k)}} 
\le \bfrob{\dtmt-\dtmh}\,\underbrace{\bspectral{\bPsih^X_{(k)}}}_{=1} 
\le \bfrob{\dtmt-\dtmh}
\end{equation}
instead of \eqref{eq:fq-bnd-b}, in \eqref{eq:fq-bnd} yields
\begin{align} 
&\bfrob{\bE{f^k_*(X)\,g^k_*(Y)^\T}-\bE{\fc^k_*(X)\,\gc^k_*(Y)^\T}} \notag\\
&\qquad\qquad\qquad\qquad\qquad\qquad\le \bigl(1+\sqrt{k}\bigr)\,\bfrob{\dtmt-\dtmh} \notag\\
&\qquad\qquad\qquad\qquad\qquad\qquad\le 2\sqrt{k}\,\bfrob{\dtmt-\dtmh},
\label{eq:fq-alt-bnd-alt}
\end{align}
the second (looser) inequality of which matches (to within a factor of two)
that for the measure \eqref{eq:fq-meas}, for which
we obtained
\begin{align} 
&\Ed{P_Y}{\bnorm{\bEd{P_{X|Y}}{f^k_*(X)}}^2 -
  \bnorm{\bEd{P_{X|Y}}{\fc^k_*(X)}}^2} \notag\\
&\quad\qquad\qquad\qquad\qquad\qquad\qquad \le
  4\sqrt{k}\,\bfrob{\dtmt-\dtmh}. 
\end{align}
As such analogous sample complexity bounds follow in this form too.

\subsection{Proof of \propref{prop:eststab}}
\label{app:eststab}

First, note that $\nbhdsf_\delta$, $\nbhdss_\delta$, and $\nbhds^k_\delta$
are non-empty as they contain $P_{X,Y}$, and bounded since $\simpXY$
is bounded in $\reals^{\cardX\times\cardY}$.
In addition, our proof makes use of the following lemma.
\begin{lemma}
\label{lem:nbhds-props}
For any $P_{X,Y}\in\relint(\simpXY)$ and $\delta>0$, let
$\nbhdsf_\delta(P_{X,Y})$, $\nbhdss_\delta(P_{X,Y})$, and
$\nbhds^k_\delta(P_{X,Y})$ be as defined in
\eqref{eq:nbhds-defs}. Then
\begin{enumerate}
\renewcommand{\theenumi}{P\arabic{enumi}}
\item \label{li:P1} $\nbhdsf_\delta(P_{X,Y}), \nbhdss_\delta(P_{X,Y}) \subseteq
  \relint(\simpXY)$ are compact sets for every $0 < \delta <
  \bmin(P_{X,Y})$, with $\bmin(\cdot)$ as defined in
  \eqref{eq:bmin-def}.
\item \label{li:P2} $\nbhdsf_\delta(P_{X,Y})\subseteq
  \nbhds^k_{4\delta\sqrt{k}}(P_{X,Y})$ 
for every $\delta > 0$.
\item \label{li:P3} $\nbhdsf_\delta(P_{X,Y})\subseteq
  \nbhdss_\delta(P_{X,Y}) \subseteq 
  \nbhds^k_{4\delta k}(P_{X,Y})$ 
for every $\delta > 0$.
\end{enumerate}
\end{lemma}

\begin{IEEEproof}[Proof of \lemref{lem:nbhds-props}]
To establish property \ref{li:P1} for $\nbhdss_\delta(P_{X,Y})$, fix
any $0 < \delta < \bmin(P_{X,Y})$, and consider the set\footnote{As in
  \appref{app:dtm-char}, we use $\bA\ge0$ to denote that all the
  entries of $\bA$ are nonnegative.}
\begin{align*} 
&\nbhdbs_\delta(P_{X,Y}) \\
&\ \defeq\Bigl\{ \bM \in \reals^{\cardY \times \cardX}
\colon \bM \geq \bzero,\  
\spectral{\bM} = 1,\ 
\spectral{\bM - \dtm} \le \delta \Bigr\} .
\end{align*}
We first show that $\nbhdbs_\delta(P_{X,Y})$ is closed. To this end, take
any sequence $\{\bM_n \in \nbhdbs_\delta(P_{X,Y}),\ n=1,2,\dots\}$ such that
$\bM_n \to \bM \in \reals^{\cardY \times \cardX}$ as $n\to\infty$. Then,
clearly $\bM \geq \bzero$ and $\spectral{\bM} = 1$ (by continuity of
the spectral norm). Moreover, we have
\begin{align*}
\spectral{\bM - \dtm} 
&\le \spectral{\bM - \bM_n} + \spectral{\bM_n - \dtm} \\
&\le \lim_{n\to\infty} \spectral{\bM - \bM_n} + \delta \\
&\le \delta,
\end{align*}
where the first inequality is the triangle inequality, the second
inequality follows from using the fact that $\bM_n \in
\nbhdbs_\delta(P_{X,Y})$ and then letting $n\to\infty$, and the final
inequality holds because $\bM_n \rightarrow \bM$. Hence,
$\nbhdbs_\delta(P_{X,Y})$ is closed.

Next, we show that $\nbhdbs_\delta(P_{X,Y}) \subseteq \dtmsett$, with
$\dtmsett$ as defined in \eqref{eq:dtmsett-def}.  Due to
\eqref{eq:dtmsett-form}, it suffices to show that $\bM>\bzero$ for
every $\bM \in \nbhdbs_\delta(P_{X,Y})$, which we obtain by noting that
for every $x\in\X$ and $y\in \Y$, we have, with $M(x,y)$ denoting the
$(y,x)$\/th entry of $\bM$,
\begin{align} 
M(x,y)
&\ge \dtms(x,y) - \abs{M(x,y) - \dtms(x,y)} \label{eq:triangle-MB}\\
&\ge  \dtms(x,y) - \spectral{\bM-\dtm} \label{eq:use-entry-specbnd} \\
&\ge \bmin(P_{X,Y}) - \delta \label{eq:use-bmin-def}\\
&> 0,
\label{eq:entrydiff-bnd}
\end{align}
where to obtain \eqref{eq:triangle-MB} we have used the triangle
inequality, to obtain \eqref{eq:use-entry-specbnd} we have used
that for an arbitrary matrix $\bA$ with entries $a_{i,j}$,
\begin{equation*}
\abs{a_{i,j}} = \abs{\bfe_i^\T\bA\bfe_j} \le \norm{\bfe_i}
\spectral{\bA} \norm{\bfe_j} = \spectral{\bA},\quad \text{all $i,j$},
\end{equation*}
with the inequality due to \lemref{lem:svd-k}, 
where to obtain \eqref{eq:use-bmin-def} we 
have used that $\bM\in\nbhdbs_\delta(P_{X,Y})$ and \eqref{eq:bmin-def},
and where to obtain \eqref{eq:entrydiff-bnd} we have used the
given constraint on $\delta$.

Now via \propref{prop:dtm-1} it follows that
$\nbhds_\delta(P_{X,Y})$ is the preimage of $\nbhdbs_\delta(P_{X,Y})
\subseteq \dtmsett$ under the DTM function $\dtm(\cdot)$, so
$\nbhds_\delta(P_{X,Y}) \subseteq \relint(\simpXY)$.  Furthermore,
since, as shown in \propref{prop:dtm-3}, the restricted DTM function
$\dtm\colon\relint(\simpXY) \mapsto \dtmsett$ is continuous,
$\nbhds_\delta(P_{X,Y})$ is closed because $\nbhdbs_\delta(P_{X,Y})$ is
closed \cite[Corollary, p.~87]{wr76}.  Since $\nbhdss_\delta(P_{X,Y})$
is also bounded, it is compact \cite[Theorem~2.41]{wr76}.

Property \ref{li:P1} for $\nbhdsf_\delta(P_{X,Y})$ is obtained in a
directly analogous manner.  In particular, since the Frobenius norm is
also continuous and satisfies the triangle inequality, it suffices to
follow the same analysis, but now with respect to the set
\begin{align*} 
&\nbhdbf_\delta(P_{X,Y}) \\
&\ \defeq\Bigl\{ \bM \in \reals^{\cardY \times \cardX}
\colon \bM \geq \bzero,\  
\spectral{\bM} = 1,\ 
\frob{\bM - \dtm} \le \delta \Bigr\} .
\end{align*}

To obtain property \ref{li:P2}, we use the fact that 
[cf.\ \eqref{eq:feat-diff-bnd-alt}]
\begin{equation*} 
\bbabs{\bfrob{\dtm\,\bPsi^X_{(k)}}^2 - \bfrob{\dtm\,\bPsih^X_{(k)}}^2} 
\le 4\sqrt{k}\,\bfrob{\dtm-\dtmh},
\end{equation*}
which is obtained in precisely the same manner as
\eqref{eq:feat-diff-bnd-alt}, from which it follows
immediately that $\nbhdsf_\delta(P_{X,Y}) \subseteq \nbhds^k_{4 \delta
  \sqrt{k}}(P_{X,Y})$.

Analogously, to obtain property \ref{li:P3}, we use the fact that
[cf.\ \eqref{eq:feat-diff-bnd}]
\begin{equation*}
\bbabs{\bfrob{\dtm\,\bPsi^X_{(k)}}^2 - \bfrob{\dtm\,\bPsih^X_{(k)}}^2} 
\le 4k\,\bspectral{\dtm-\dtmh},
\end{equation*}
which is obtained in precisely the same manner as
\eqref{eq:feat-diff-bnd}, from which it follows immediately that
$\nbhdss_\delta(P_{X,Y}) \subseteq \nbhds^k_{4 \delta k}(P_{X,Y})$.
Finally, for the remaining part of property \ref{li:P3}, we use the
standard norm inequality $\spectral{\bA}\le\frob{\bA}$, for any matrix
$\bA$, obtaining $\nbhdsf_\delta(P_{X,Y})\subseteq
\nbhdss_\delta(P_{X,Y})$.
\end{IEEEproof}

Proceeding to the proof of \protect\propref{prop:eststab}, with the
notation \eqref{eq:errexp-def}--\eqref{eq:uE-def} 
we have, via Sanov's theorem \cite[Theorem~2.1.10]{dz98}, 
\begin{align} 
\uE\bigl(\nbhdsf_\delta(P_{X,Y})\bigr) &=
E_*\bigl(\nbhdsf_\delta(P_{X,Y})\bigr) \label{eq:uE-rel-F}\\ 
\uE\bigl(\nbhdss_\delta(P_{X,Y})\bigr) &=
E_*\bigl(\nbhdss_\delta(P_{X,Y})\bigr) \label{eq:uE-rel-s}\\ 
\uE\bigl(\nbhds_\delta^k(P_{X,Y})\bigr) &\le E\bigl(\nbhds_\delta^k(P_{X,Y})\bigr), 
\label{eq:uE-rel-k}
\end{align} 
for $0<\delta<\bmin(P_{X,Y})$ with $\bmin(\cdot)$ as defined in
\eqref{eq:bmin-def}, where to obtain \eqref{eq:uE-rel-F} and
\eqref{eq:uE-rel-s} we have used that $\simpXY \backslash
\nbhdsf_\delta(P_{X,Y})$ and $\simpXY
\backslash\nbhdss_\delta(P_{X,Y})$, respectively, are open sets (with
respect to $\simpXY$), since $\nbhdsf_\delta(P_{X,Y})$ and
$\nbhdss_\delta(P_{X,Y})$ are closed according to property \ref{li:P1}
of \lemref{lem:nbhds-props}.  Hence, \eqref{eq:eststab-F} follows
according to
\begin{align} 
E\bigl(\nbhds^k_{4 \delta\sqrt{k}}(P_{X,Y})\bigr) 
&\ge \uE\bigl(\nbhds^k_{4 \delta\sqrt{k}}(P_{X,Y})\bigr)  \label{eq:eststab-1-F}\\
&\ge \uE\bigl(\nbhdsf_{\delta}(P_{X,Y})\bigr) \label{eq:eststab-2-F}\\
&= E_*\bigl(\nbhdsf_{\delta}(P_{X,Y})\bigr), \label{eq:eststab-3-F}
\end{align}
where to obtain \eqref{eq:eststab-1-F} we have used \eqref{eq:uE-rel-k},
to obtain \eqref{eq:eststab-2-F} we have used that
\begin{equation*}
\simpXY\backslash\nbhds_{4\delta \sqrt{k}}^k \subseteq
\simpXY\backslash\nbhdsf_\delta,
\end{equation*}
which follows from property \ref{li:P2} of \lemref{lem:nbhds-props},
and to obtain \eqref{eq:eststab-3-F} we have used \eqref{eq:uE-rel-F}.
Analogously, \eqref{eq:eststab-s} follows
according to
\begin{align} 
E\bigl(\nbhds^k_{4 \delta k}(P_{X,Y})\bigr) 
&\ge \uE\bigl(\nbhds^k_{4 \delta k}(P_{X,Y})\bigr)  \label{eq:eststab-1}\\
&\ge \uE\bigl(\nbhdss_{\delta}(P_{X,Y})\bigr) \label{eq:eststab-2}\\
&= E_*\bigl(\nbhdss_{\delta}(P_{X,Y})\bigr) \label{eq:eststab-3} \\
&\ge E_*\bigl(\nbhdsf_{\delta}(P_{X,Y})\bigr), \label{eq:eststab-4}
\end{align}
where to obtain \eqref{eq:eststab-1} we have used \eqref{eq:uE-rel-k},
to obtain \eqref{eq:eststab-2} we have used the first subset
relation in
\begin{equation}
\simpXY\backslash\nbhds_{4\delta k}^k \subseteq
\simpXY\backslash\nbhdss_\delta \subseteq \simpXY\backslash\nbhdsf_\delta,
\label{eq:subset-rel}
\end{equation}
which follows from property \ref{li:P3} of \lemref{lem:nbhds-props},
to obtain \eqref{eq:eststab-3} we have used \eqref{eq:uE-rel-s}, and
to obtain \eqref{eq:eststab-4} we have used \eqref{eq:uE-rel-F} and the
second subset relation in \eqref{eq:subset-rel}.
\hfill\IEEEQED

\subsection{Proof of \lemref{lem:chernoff-local}}
\label{app:chernoff-local}

Our proof makes use of the following special case of Sanov's Theorem
\cite[Theorem~2.1.10, Exercise~2.1.19]{dz98},
\cite[Theorem~2.1]{cs04}:
\begin{lemma}
\label{lem:sanov}
For every distribution 
$P_Z\in\simpZ$, and every
closed and convex subset $\cS \subseteq \simpZ$ of
probability distributions that has non-empty interior, we have that the empirical distribution $\Ph_Z$ formed from
$n$ i.i.d.\ samples of $P_Z$ satisfies
\begin{equation*}
\lim_{n\to\infty} \frac{1}{n}\log \prob{\Ph_Z \in \cS} =
-\min_{Q_Z \in \cS} D(Q_Z\|P_Z),
\end{equation*}
where the minimum is achieved by a unique distribution. 
\end{lemma}

Without loss of generality we may restrict our attention to the case
in which $\E{h(Z)} > 0$.  For any $\gamma>0$, define the sets
\begin{align*}
\cS^+_{\gamma} 
&\defeq \left\{Q_Z \in \simpZ \colon \Ed{Q_Z}{h(Z)} \ge (1+\gamma)\,
\E{h(Z)} \right\} \\ 
\cS^-_{\gamma} 
&\defeq \left\{Q_Z \in \simpZ \colon \Ed{Q_Z}{h(Z)} \le (1-\gamma)\,
\E{h(Z)} \right\}  
\end{align*}
where 
\begin{equation*}
\Ed{Q_Z}{h(Z)} \defeq \sum_{z\in\Z}{Q_Z(z)\, h(z)}.
\end{equation*}
Furthermore, since we will eventually let $\gamma \to 0$, we
may assume that
\begin{equation*}
0 < \gamma < \min\left\{ \left(\frac{\max_{z\in\Z}{h(z)}}{\E{h(Z)}}\! -\! 1\right)
, \left(1\! -\! \frac{\min_{z\in\Z}{h(z)}}{\E{h(Z)}}\right) \right\},
\end{equation*}
so that
\begin{equation*}
\min_{z\in\Z}{h(z)} < (1-\gamma)\, \E{h(Z)} < (1+\gamma)\, \E{h(Z)} <
\max_{z\in\Z}{h(z)},
\end{equation*}
where $\min_{z\in\Z}{h(z)} < \max_{z\in\Z}{h(z)}$ because
$\varop{h(Z)} > 0$. Hence, $\cS^+_{\gamma}$ and $\cS^-_{\gamma}$ are
closed and convex sets that have non-empty interiors. Using
\lemref{lem:sanov}, we have
\begin{align}
\lim_{n\to\infty} \frac{1}{n} \log \prob{\Ph_Z \in \cS^+_{\gamma}}
&= -\min_{Q_Z \in \cS^+_{\gamma}} D(Q_Z\|P_Z) \notag \\ 
&= -D(Q_Z^+\|P_Z) \label{Eq: Rate 1} \\
\lim_{n\to\infty} \frac{1}{n} \log \prob{\Ph_Z \in \cS^-_{\gamma}}
&= -\min_{Q_Z \in \cS^-_{\gamma}} D(Q_Z\|P_Z) \notag \\ 
&= -D(Q_Z^-\|P_Z), \label{Eq: Rate 2}
\end{align}
where the (unique) minimizing distributions $Q_Z^+ \in \cS^+_{\gamma}$ and
$Q_Z^- \in \cS^-_{\gamma}$ are members of the exponential
family
\begin{equation*}
Q_Z(z;\theta) = P_Z(z) \expop{\theta\, h(z) - \alpha(\theta)},\quad z\in\Z, 
\end{equation*}
with natural parameter $\theta \in \reals$.
Recall that the (infinitely differentiable)
log-partition function 
\begin{equation*}
\alpha(\theta) \defeq \log\bigl(\E{\expop{\theta\, h(Z)}}\bigr)
\end{equation*}
has derivatives
\begin{align*}
\alpha'(\theta) &=
\Ed{Q_Z(\cdot;\theta)}{h(Z)}  \\ 
\alpha''(\theta) & =
\varopd{Q_Z(\cdot;\theta)}{h(Z)} > 0,
\end{align*}
where the second
derivative is strictly positive because every element of
$\Z$ has positive probability under $Q_Z(\cdot;\theta)$.  The
minimizing distributions are $Q_Z^+ = Q_Z(\cdot;\theta_+)$ and $Q_Z^- =
Q_Z(\cdot;\theta_-)$, where the optimal parameters $\theta_+ > 0$ and $\theta_- < 0$
are chosen to satisfy (cf.\ \cite[Example~2.1]{cs04}) 
\begin{align*}
\alpha'(\theta_+) = \Ed{Q_Z(\cdot;\theta_+)}{h(Z)} 
&= (1 + \gamma) \, \E{h(Z)} , \\ 
\alpha'(\theta_-) = \Ed{Q_Z(\cdot;\theta_-)}{h(Z)} 
&= (1 - \gamma) \, \E{h(Z)}.
\end{align*}

Now assume that
\begin{equation}
\label{Eq: Local Rate}
\lim_{\gamma \to 0^+} \frac{D(Q_Z^+\|P_Z)}{\gamma^2} = \lim_{\gamma
  \to 0^+} \frac{D(Q_Z^-\|P_Z)}{\gamma^2} =
\frac12 \frac{\bigl(\E{h(Z)}\bigr)^2}{\varop{h(Z)}}
\end{equation} 
and define 
\begin{equation} 
\cS^\pm_{\gamma} 
\defeq  \cS^+_{\gamma} \cup \cS^-_{\gamma} 
= \left\{Q_Z \in \simpZ \colon
\left|\frac{\Ed{Q_Z}{h(Z)}}{\E{h(Z)}} - 1\right| \geq \gamma \right\}.
\label{eq:Spm-def}
\end{equation}
Since $\cS^+_\gamma$ and $\cS^-_\gamma$ are disjoint, we have $\prob{\Ph_Z
  \in \cS^\pm_{\gamma}} = \prob{\Ph_Z \in \cS^+_{\gamma}} +
\prob{\Ph_Z \in \cS^-_{\gamma}}$.  Hence, via the Laplace principle
it follows that
\begin{multline}
-\lim_{n \to \infty} \frac{1}{n} \log \prob{\Ph_Z \in
\cS^\pm_{\gamma}} \\
= \min\left\{D(Q_Z^+\|P_Z)\,,\, D(Q_Z^-\|P_Z) \right\},
\label{eq:almost-chernoff-local}
\end{multline}
where we have used \eqref{Eq: Rate 1} and \eqref{Eq: Rate 2}.
Applying \eqref{Eq: Local Rate} to \eqref{eq:almost-chernoff-local},
and recognizing \eqref{eq:Spm-def},
we obtain \eqref{eq:chernoff-local} as desired.

Thus, it remains only to show \eqref{Eq: Local Rate}.
To this end, consider the function 
\begin{equation*}
d(\theta) \defeq D(Q_Z(\cdot;\theta)\|P_Z),\quad \theta \in \reals.
\end{equation*}
It is straightforward to verify that
\begin{align*}
d(\theta) &= \theta\,\alpha'(\theta) - \alpha(\theta) \\
d'(\theta) & = \theta\, \alpha''(\theta) \\
d''(\theta) & = \alpha''(\theta)  + \theta\, \alpha'''(\theta) ,
\end{align*}
which means that $d(0) = d'(0) = 0$, and $d''(0)
= \alpha''(0) = \varop{h(Z)}$.  Hence, by Taylor's
theorem we have
\begin{equation}
\label{Eq: Local Approximation of KL Divergence}
\lim_{\theta \to 0} \frac{d(\theta)}{\theta^2} = \frac12 \varop{h(Z)} . 
\end{equation}

Now given any $\tau \in \reals$, there exists a unique $\theta_\tau$ such that
\begin{equation*} 
\alpha'(\theta_\tau) = \Ed{Q_Z(\cdot;\theta_\tau)}{h(Z)} = (1 + \tau)\,\E{h(Z)},
\end{equation*}
since $\alpha'$ is strictly increasing (since $\alpha''$ is
strictly positive). Next, observe that
\begin{align}
\lim_{\tau \to 0} \frac{d(\theta_\tau)}{\tau^2} 
&= \lim_{\tau \to 0} \frac{d(\theta_\tau)}{\theta_\tau^2} 
   \lim_{\tau \to 0} \frac{\theta_\tau^2}{\tau^2} \notag \\ 
&= \frac{\varop{h(Z)}}{2} \left( \lim_{\tau \to 0} \frac{\theta_\tau}{\tau} \right)^{2} 
   \notag \\  
&= \frac{\varop{h(Z)}}{2} \left(\left.\frac{\diff \theta_\tau}{\diff\tau}\right|_{\tau =
     0}\right)^{2} \notag \\ 
& = \frac{\varop{h(Z)}}{2}
   \left(\frac{\E{h(Z)}}{\alpha''(0)} \right)^{2} \notag \\ 
& = \frac{\E{h(Z)}^2}{2 \varop{h(Z)}},
\label{Eq: General Local Rate}
\end{align}
where the second equality follows from \eqref{Eq: Local Approximation
  of KL Divergence}, the fact that $\theta_\tau \to 0$ as $\tau \to 0$ (by
the continuity of the inverse of $\alpha'(\cdot)$), and the
continuity of $t \mapsto t^2$, where the third equality follows from
the definition of derivative and the fact that $\theta_0 = 0$ (since
$\alpha'(0) = \E{h(Z)}$), where the fourth equality holds because
\begin{equation*}
\left. \frac{\diff \theta_\tau}{\diff \tau} \right|_{\tau=0} = 
\left( \left. \frac{\diff \tau_\theta}{\diff\theta} \right|_{\theta=0}
\right)^{-1} 
\end{equation*}
with
\begin{equation*}
\tau_\theta = \frac{\alpha'(\theta)}{\E{h(Z)}} - 1,
\end{equation*}
and where the fifth equality holds because $\alpha''(0) =
\varop{h(Z)}$.

In turn, setting $\tau = \gamma > 0$ and $\theta_\tau = \theta_+ > 0$ yields
\begin{subequations} 
\begin{equation}
\lim_{\gamma \to 0^+} \frac{D(Q_Z^+\|P_Z)}{\gamma^2} = \lim_{\tau \to
  0^+} \frac{d(\theta_\tau)}{\tau^2} , 
\end{equation}
and setting $\tau = -\gamma < 0$ and $\theta_\tau = \theta_- < 0$ yields
\begin{equation}
\lim_{\gamma \to 0^+} \frac{D(Q_Z^-\|P_Z)}{\gamma^2} = \lim_{\tau \to
  0^-} \frac{d(\theta_\tau)}{\tau^2}.
\end{equation}%
\label{eq:lim+-}%
\end{subequations}
Finally, replacing the right-hand sides of \eqref{eq:lim+-} with \eqref{Eq:
  General Local Rate} yields \eqref{Eq: Local Rate}.
\hfill\IEEEQED

\section{Appendices for \protect\secref{sec:cf}}
\label{app:ama}

\subsection{Proof of \protect\propref{prop:attribute-match}}
\label{app:attribute-match}

From \eqref{eq:M-def}, it follows immediately that
\begin{align}
\E{M}
&= \sum_{y\in\cYh(x)} \prob{\cE_y(x)} \\
&\le \max_{\left\{\cYh(x)\subset\Y\colon \card{\cYh(x)}=l\right\}}
\sum_{y\in\cYh(x)} \prob{\cE_y(x)} \\
&= \sum_{y\in\cYh^*(x)} \prob{\cE_y(x)},
\label{eq:em-gen-bnd}
\end{align}
where
\begin{subequations} 
\begin{align} 
y_1^*(x) &= \argmax_{y\in\Y} 
\prob{\cE_y(x)} \label{eq:ystar-gen1-def} \\
y_i^*(x) &=
\!\!\!\!\smash[b]{\argmax_{y\in\Y\setminus\{y_1^*(x),\dots,y_{i-1}^*(x)\}}}
  \prob{\cE_y(x)}, \label{eq:ystar-gen-def} \\
&\quad\qquad\qquad\qquad\qquad\qquad\qquad\qquad i=2,\dots,l. \notag
\end{align}%
\end{subequations}

It remains only to evaluate the constituent event probabilities, which
are obtained as follows:
\begin{align*} 
\prob{\cE_y(x)}
&= \prob{V^k(y)=V_\circ^k(x)} \\
&= \sum_{\{v^k,v_\circ^k\colon v^k= v_\circ^k\}} 
P_{V^k|Y}(v^k|y)\, P_{V^k|X}(v_\circ^k|x) \\
&= \sum_{v^k} 
P_{V^k|Y}(v^k|y)\, P_{V^k|X}(v^k|x),
\end{align*}
wherein, using \eqref{eq:CYk-opt} and \eqref{eq:PXVi-opt},
\begin{align*} 
&P_{V^k|Y}(v^k|y)\, P_{V^k|X}(v^k|x) \notag\\
&\quad= \prod_{i=1}^k \left( P_{V_i|Y}(v_i|y)\, P_{V_i|X}(v_i|x) \right)\\
&\quad= \prod_{i=1}^k \left(  \frac{P_{Y|V_i}(y|v_i)\,P_{V_i}(v_i)}{P_Y(y)}\,
\frac{P_{X|V_i}(x|v_i)\,P_{V_i}(v_i)}{P_X(x)} \right)\\
&\quad= \frac1{2^{2k}} \prod_{i=1}^k \Bigl(
  \bigl( 1 + \eps\, v_i\, g_i^*(y) \bigr) \bigl( 1 + \eps\,
  v_i\, \sigma_i\, f_i^*(x)
  \bigr)  \Bigr) \\
&\quad= \frac1{2^{2k}} \prod_{i=1}^k \left(
   1\!+\!\eps \,v_i \bigl( \sigma_i\, f_i^*(x)\!+\!g_i^*(y) \bigr)\!+\!
   \eps^2 \sigma_i \, f_i^*(x)\, g_i^*(y) \right).
\end{align*}
Hence,
\begin{align*}
&\prob{\cE_y(x)} \notag\\
&\, = 
\frac1{2^{2k}} \prod_{i=1}^k \sum_{v_i} \Bigl( 
   1\!+\!\eps \,v_i \bigl( \sigma_i\, f_i^*(x)\!+\!g_i^*(y) \bigr)\!+\!
   \eps^2 \sigma_i \, f_i^*(x)\, g_i^*(y) \Bigr) \\
&\, = 
\frac1{2^k} \prod_{i=1}^k \bigl( 1 + \eps^2 \sigma_i \, f_i^*(x)\,
g_i^*(y) \bigr) \\
&\, = 
\frac1{2^k} \left( 1 + \eps^2 \sum_{i=1}^k \sigma_i \, f_i^*(x)\,
g_i^*(y) \right) + o(\eps^2), 
\end{align*}
the nonvanishing term of which we note is a monotonic function of 
the quantity being maximized in \eqref{eq:ystar-disc-final}. 
\hfill\IEEEQED

\section{Appendices for \protect\secref{sec:nn}}
\label{app:nn}

\subsection{Proof of \protect\propref{prop:softmax}}
\label{app:softmax}

First, without loss of
generality we impose the constraints
\begin{equation}
\bE{g(Y)}=0 \qquad\text{and}\qquad \bE{\beta(y)}=0,
\label{eq:gb-constr}
\end{equation}
since other solutions are simple reparameterizations.

It is convenient to first establish the following special case of
\propref{prop:softmax}.
\begin{lemma}
\label{lem:softmax}
Let the hypotheses of \propref{prop:softmax} be satisfied, together with the
further constraints
\begin{equation}
\bmu_S=0\qquad\text{and}\qquad \bLa_S=\bI.
\label{eq:S-norm}
\end{equation}
Then
\begin{align} 
&\min_{\Pt_{Y|S}(\cdot|s)\in\cPt^\Y_s(P_Y)}
\sum_{s\in\cS} P_S(s)\, D\bigl(P_{Y|S}(\cdot|s)\bigm\|\Pt_{Y|S}(\cdot|s)\bigr)  \notag\\
&\qquad\qquad = I(Y;S) - \frac12 \E{\bnorm{\bmu_{S|Y}(Y)}^2} + o(\eps^2),
\label{eq:D*-special}
\end{align}
and is achieved by the parameters
\begin{subequations} 
\begin{equation} 
g(y) = g_{*,S}(y) \defeq \bmu_{S|Y}(y) + o(\eps) \ \text{and}\ 
\beta(y) = \beta_{*,S}(y) \defeq o(\eps),
\label{eq:gb*-special}
\end{equation}
i.e., 
\begin{equation} 
\Pt_{Y|S}^*(y|s)
\propto P_Y(y) \, \e^{s^\T \bmu_{S|Y}(y)} \bigl(1+o(1)\bigr).
\label{eq:Pt*-special}
\end{equation}
\end{subequations}
\end{lemma}

Using \lemref{lem:softmax}, we establish \propref{prop:softmax} as follows.
First, let us assume $\bLa_S$ is nonsingular, and let
\begin{equation}
\st \defeq \bLa_S^{-1/2} \bigl( s - \bmu_S \bigr)
\label{eq:stil-def}
\end{equation}
so
\begin{equation*}
\bmu_\St = \bzero\qquad\text{and}\qquad \bLa_\St = \bI.
\end{equation*}
Then we may rewrite $\Pt^{g,\beta}_{Y|S}(y|s)$ in the form
\begin{align} 
\Pt^{g,\beta}_{Y|S}(y|s) 
&= P_Y(y)\, \expop{ \st^\T \gt(y) + \tbeta(y) - \alt(\st)} \\
&\defeq \Pt^{\gt,\tbeta}_{Y|\St}(y|\st)
\end{align}
where
\begin{align}
\gt(y) &\defeq \bLa_S^{1/2} g(y) \label{eq:gt-def}\\
\tbeta(y) &\defeq \bmu_S^\T g(y) + \beta(y) \label{eq:tbeta-def}\\
\alt(\st) &\defeq \alpha\bigl(\bmu_S +
\bLa_S^{1/2}\st\bigr), \label{eq:alt-def} 
\end{align}
and for which $\bE{\gt(Y)}=\bzero$ and $\bE{\tbeta(Y)}=0$.

Using these definitions, we have
\begin{align} 
\gt_{*,S}(y) 
&= \bEd{P_{\St|Y}(\cdot|y)}{\St} + o(\eps) \notag\\
&= \bLa_S^{-1/2} \bigl( \bmu_{S|Y}(y) - \bmu_S \bigr) + o(\eps)
\label{eq:gt*}
\end{align}
where to obtain the first equality we have used \lemref{lem:softmax},
and to obtain \eqref{eq:gt*} we have used \eqref{eq:stil-def}.
Combining \eqref{eq:gt*} with \eqref{eq:gt-def} yields
\eqref{eq:g*-softmax}.  Similarly, via \lemref{lem:softmax} we obtain
\begin{equation*} 
\tbeta_{*,S}(y) = o(\eps),
\end{equation*}
which when combined with \eqref{eq:tbeta-def} yields \eqref{eq:bt*-softmax}.
In turn, we obtain \eqref{eq:D*-softmax} via
\begin{align*} 
&\min_{\Pt_{Y|\St}(\cdot|s)\in\cPt^\Y_\st(P_Y)}
\sum_{\st\in\cS} P_\St(\st)\, D\bigl(P_{Y|\St}(\cdot|\st)\bigm\|\Pt_{Y|\St}(\cdot|\st)\bigr)  \notag\\
&\quad = I(Y;\St) - \frac12 \E{\bnorm{\bmu_{\St|Y}(Y)}^2} +
o(\eps^2) \notag\\
&\quad = I(Y;S) - \frac12\,
\E{\bnorm{\bLa_S^{-1/2}\bigl(\bmu_{S|Y}(Y)-\bmu_S\bigr)}^2} + o(\eps^2),
\end{align*}
where to obtain the first equality we have used \lemref{lem:softmax},
and to obtain the second we have used \eqref{eq:stil-def} and the
invariance of mutual information to coordinate transformations.

It remains only to establish \lemref{lem:softmax}.
\begin{IEEEproof}[Proof of \lemref{lem:softmax}]
First, note that 
\begin{equation}
D(P_{Y,S} \| \Pt_{Y|S} P_S) = I(Y;S) - \underbrace{\E{S^\T g(Y)
    -\alpha(S)}}_{\defeq \tell(g,\beta)},
\label{eq:tell-def}
\end{equation}
so we seek to maximize $\tell(g,\beta)$.  Moreover, note that since
$\chi^2$-divergence is an $f$-divergence, it satisfies a data
processing inequality \cite{ic72}, so $S,Y$ are $\eps$-dependent for
any choice of $f$ that induces $S$.

Fixing $s\in\cS$, note that
$P_{Y|S}(\cdot|s)\in\nbhd_\eps^\Y(P_Y)$ and, moreover,
$\Pt_{Y|S}^{0,0}(\cdot|s)=P_Y$.  Hence, it follows that the optimizing
$\smash{\Pt_{Y|S}^{g,\beta}(\cdot|s)}$ is such that
$\smash{\Pt_{Y|S}^{g,\beta}(\cdot|s)\in\nbhd_\eps^{\Y}(P_Y)}$, and
thus we may restrict our search to parameters $(g,\beta)$ in this
neighborhood.

In turn, defining
\begin{equation*}
\phit^{Y|S}_s(y) \defeq
\frac{\Pt_{Y|S}^{g,\beta}(y|s)-P_Y(y)}{\eps\sqrt{P_Y(y)}},
\end{equation*}
it follows from \eqref{eq:discrim} that
\begin{equation*}
\eps^2 \sum_{y\in\Y} \phit^{Y|S}_s(y)^2 
= \sum_{y\in\Y} P_Y(y) \bigl( \e^{s^\T
  g(y)+\beta(y)-\alpha(s)}-1\bigr)^2 \le \eps^2,
\end{equation*}
so 
\begin{equation*}
s^\T g(y) +\beta(y)-\alpha(s) = o(1).
\end{equation*}
Hence, for the Taylor series expansions\footnote{In this
  analysis we assume the existence of these Taylor series.} (in $\eps$)  
\begin{subequations} 
\begin{align}
g(y) &= \sum_{i=0}^2 \eps^i g^{(i)}(y) + o(\eps^2) \label{eq:taylor-g}\\
\beta(y) &= \sum_{i=0}^2 \eps^i \beta^{(i)}(y) +
o(\eps^2) \label{eq:taylor-tbeta}\\  
\alpha(s) &= \sum_{i=0}^2 \eps^i \alpha^{(i)}(s) +
o(\eps^2), \label{eq:taylor-alpha} 
\end{align}
\label{eq:taylor-gba}%
\end{subequations}
wherein $g^{(i)}(y)$, $\beta^{(i)}(y)$, and $\alpha^{(i)}(s)$ for
$i\in\{0,1,2\}$ do not
depend on $\eps$, it follows that
\begin{equation}
s^\T g^{(0)}(y) + \beta^{(0)}(y) = \alpha^{(0)}(s).
\label{eq:al0-rel}
\end{equation}
But due to \eqref{eq:gb-constr}, in
the Taylor series \eqref{eq:taylor-gba} we must also have
\begin{equation}
\bE{g^{(i)}(Y)}=0\quad\text{and}\quad \bE{\beta^{(i)}(Y)}=0,\qquad
i\in\{0,1,2\}.  
\label{eq:gbi-mean}
\end{equation}
Taking the expectation of both sides of
\eqref{eq:al0-rel} with respect to $P_Y$ then yields that
\begin{equation} 
\alpha^{(0)}(s) = 0.
\label{eq:a-zero}%
\end{equation}

Next, with 
\begin{equation}
\tau_s^{(i)}(y) \defeq s^\T g^{(i)}(y) + \beta^{(i)}(y),\qquad i=1,2,
\label{eq:tau_i-def}
\end{equation}
and using the Taylor series 
\begin{equation*}
\e^\omega=\sum_{j=0}^l \frac1{j!} \, \omega^j + o(\omega^l),
\end{equation*}
we obtain that $Z(s) \defeq \e^{\alpha(s)}$, via \eqref{eq:discrim}, can
be expressed in the form 
\begin{align*} 
Z(s) 
&= \sum_{y\in\Y} P_Y(y)\, \e^{s^\T g(y) + \beta(y)} \\
&= \Ed{P_Y}{\expop{\sum_{i=1}^2 \eps^i \tau_s^{(i)}(Y) + o(\eps^2)}} \\
&= \ev_{P_Y}\Biggl[ 
\biggl( \sum_{j=0}^2 \eps^j \frac1{j!} \tau_s^{(1)}(Y)^j+o(\eps^2) \biggr)
\\
&\qquad\qquad{}\cdot
\biggl( \sum_{j=0}^1 \eps^{2j}\frac1{j!} \tau_s^{(2)}(Y)^j+o(\eps^2) \biggr)
\Bigl(1+o(\eps^2)\Bigr)\Biggr] \\
&= 1 + \sum_{i=1}^2 \eps^i \ups_i(s) + o(\eps^2), 
\end{align*}
with
\begin{subequations} 
\begin{align}
\ups_1(s) &\defeq \bEd{P_Y}{\tau_s^{(1)}(Y)} = 0 \label{eq:ups_1-form}\\
\ups_2(s) &\defeq \Ed{P_Y}{ \frac12\, \tau_s^{(1)}(Y)^2 + \tau_s^{(2)}(Y) } =
\frac12\,\bEd{P_Y}{ \tau_s^{(1)}(Y)^2}, 
\label{eq:ups_2-form} 
\end{align}
\label{eq:ups_i-form}%
\end{subequations}
where we have used \eqref{eq:al0-rel} with \eqref{eq:a-zero} to
conclude $\tau_s^{(0)}=0$, and that
\begin{equation*} 
\bEd{P_Y}{\tau_s^{(i)}(Y)}=0,\qquad i=1,2,\dots
\end{equation*}
due to \eqref{eq:gbi-mean}.

Next, using the Taylor series
\begin{equation*}
\ln(1+\omega) = \omega - \frac12\, \omega^2 + o(\omega^2),
\end{equation*}
we obtain that $\alpha(s)=\ln Z(s)$ is of the form
\begin{align*} 
\alpha(s) &= \biggl( \sum_{i=1}^2 \eps^i \ups_i(s) \biggr)
-\frac12 \biggl( \sum_{i=1}^2 \eps^i \ups_i(s) \biggr)^2 + o(\eps^2).
\end{align*}
So in the Taylor series \eqref{eq:taylor-alpha} for $\alpha(s)$, we obtain
\begin{subequations} 
\begin{align}
\alpha^{(1)}(s) &= \ups_1(s) = 0 \label{eq:alpha_1-form}\\
\alpha^{(2)}(s) &= \ups_2(s) - \frac12\,\ups_1(s)^2 =
\ups_2(s). 
\label{eq:alpha_2-form} 
\end{align}
\label{eq:alpha_i-form}%
\end{subequations}

We write
$\tell(g,\beta)$ in \eqref{eq:tell-def}, which we seek to maximize, in
the form
\begin{align}
\tell(g,\beta) 
&= \sum_{i=0}^2 \eps^i \bE{S^\T g^{(i)}(Y) - \alpha^{(i)}(S)}
+o(\eps^2) \label{eq:use-taylor-gba}\\
&= \sum_{i=1}^2 \eps^i \bE{S^\T g^{(i)}(Y) - \alpha^{(i)}(S)} +
o(\eps^2) \label{eq:use-Ebzero}\\
&= \eps \,\bE{S^\T g^{(1)}(Y)} 
\notag\\
&\quad\qquad{}
+ \eps^2 \,\bE{S^\T g^{(2)}(Y)}- \eps^2\, \bE{\alpha^{(2)}(S)} + o(\eps^2) \label{eq:use-a1-zero} \\
&= \eps\,\bE{S^\T g^{(1)}(Y)}  -
\eps^2\,\bE{\alpha^{(2)}(S)} + o(\eps^2),
\label{eq:use-sg2-neglect}
\end{align}
where to obtain \eqref{eq:use-taylor-gba} we have  used
\eqref{eq:taylor-gba}, to obtain \eqref{eq:use-Ebzero} we have used that 
\begin{equation*}
\bE{S^\T g^{(0)}(Y)-\alpha^{(0)}(S)} = -\bE{\beta^{(0)}(Y)} = 0,
\end{equation*}
due to \eqref{eq:al0-rel} and 
\eqref{eq:gbi-mean}, to obtain \eqref{eq:use-a1-zero} we have used
\eqref{eq:alpha_1-form}, and to obtain \eqref{eq:use-sg2-neglect} we
have used that the second term in \eqref{eq:use-a1-zero} is
$o(\eps^2)$, which follows from the fact that for any $i$
\begin{equation*}
\bE{S^\T g^{(i)}(Y)} = \bE{S^\T}
\underbrace{\bE{g^{(i)}(Y)}}_{=0} + \bigO(\eps) \in
\bigO(\eps),
\end{equation*}
since 
$P_{S,Y}\in\nbhd_\eps^{\X\times\Y}(P_S P_Y)$.  

Hence, we write \eqref{eq:use-sg2-neglect} in the form
\begin{subequations}
\begin{equation}
\tell(g,\beta) = \tell_2\bigl(g^{(1)},\beta^{(1)}\bigr) + o(\eps^2),
\end{equation}
with
\begin{equation}
\tell_2\bigl(g^{(1)},\beta^{(1)}\bigr) \defeq 
\eps\, \bE{S^\T g^{(1)}(Y)} - \eps^2\,
\bE{\alpha^{(2)}(S)},
\label{eq:tell-form2}
\end{equation}
\label{eq:tell-exp}%
\end{subequations}
where we note $\tell_2(g^{(1)},\beta^{(1)})\in\bigO(\eps^2)$.  In
addition, we note that there is no dependence on $g^{(0)}$ and
$\beta^{(0)}$ in \eqref{eq:tell-form2}.  Indeed, they can be freely
chosen subject to the constraints \eqref{eq:gbi-mean}, and those
choices have no effect on the resulting
$\smash{\Pt_{Y|S}^{g,\beta}(\cdot|s)}$; for example, we may choose
\begin{equation*} 
g^{(0)}(y)=0\quad\text{and}\quad\beta^{(0)}(y)=0,\qquad\text{all $y\in\Y$}.
\end{equation*}

Proceeding, to express the second term in \eqref{eq:tell-form2} in terms of
$g^{(1)}(y)$ and $\beta^{(1)}(y)$, note that, using
\eqref{eq:ups_2-form} and \eqref{eq:tau_i-def},
\begin{align}
\ups_2(s) 
&= \frac12\,  \bEd{P_Y}{\tau_s^{(1)}(Y)^2} \notag\\
&= \frac12\,  \bEd{P_Y}{\bigl( s^\T g^{(1)}(Y) + \beta^{(1)}(Y)\bigr)^2} ,
\label{eq:ups_2-exp}
\end{align}
so
\begin{align} 
&\eps^2\,\bEd{P_S}{\alpha^{(2)}(S)} \notag\\
&\quad= \eps^2\, \bEd{P_S}{\ups_2(S)} \label{eq:use-alpha_2}\\
&\quad= \frac{\eps^2}2\,  \Ed{P_Y}{\bEd{P_S}{\bigl( S^\T g^{(1)}(Y)\bigr)^2} +
  \beta^{(1)}(Y)^2} \label{eq:use-ups_2-exp}\\
&\quad= \frac{\eps^2}2\,  \bEd{P_Y}{\bnorm{g^{(1)}(Y)}^2} +
  \frac{\eps^2}2\, \bEd{P_Y}{\beta^{(1)}(Y)^2},
\label{eq:E-alpha_2}
\end{align}
where to obtain \eqref{eq:use-alpha_2} we have used
\eqref{eq:alpha_2-form}, to obtain \eqref{eq:use-ups_2-exp} we have
used \eqref{eq:ups_2-exp} and \eqref{eq:S-norm}, and where to obtain
\eqref{eq:E-alpha_2} we have used \lemref{lem:rie} with
\eqref{eq:S-norm} (and $k_1=k$ and $k_2=1$).

Since $\beta^{(1)}(y)$ only appears in \eqref{eq:tell-form2} through the
second term in \eqref{eq:E-alpha_2}, we conclude that its optimum 
value is
\begin{equation}
\beta^{(1)}_{*,S}(y) \equiv 0.
\label{eq:tbeta_1-opt}
\end{equation}
Combining the remainder of \eqref{eq:E-alpha_2} with the first term in
\eqref{eq:tell-form2}, we then have, by the Cauchy-Schwarz inequality,
\begin{align}
\tell_2\bigl(g^{(1)},\beta_{*,S}^{(1)}\bigr) 
&= \eps\, \Ed{P_{S,Y}}{\left(S - \frac{\eps}{2}\, g^{(1)}(Y)
    \right)^\T g^{(1)}(Y)}\notag\\
&= \eps\, \Ed{P_Y}{\Ed{P_{S|Y}}{S - \frac{\eps}{2}\, g^{(1)}(Y)
    }^\T g^{(1)}(Y)}\notag\\
&\le \eps \,
  \sqrt{\Ed{P_Y}{\bbnorm{\Ed{P_{S|Y}}{S-\frac{\eps}{2}\, g^{(1)}(Y)}}^2}} 
 \notag\\
&\ \qquad\qquad\qquad {}\cdot \sqrt{\Ed{P_Y}{\bnorm{g^{(1)}(Y)}^2}}
\end{align}
where the inequality holds with equality when
\begin{equation}
g^{(1)}(Y) \propto \Ed{P_{S|Y}}{S-\frac{\eps}{2}\,g^{(1)}(Y)}
= \Ed{P_{S|Y}}{S}-\frac{\eps}{2}\,g^{(1)}(Y),
\end{equation}
for some nonnegative constant of proportionality, 
i.e., when 
\begin{equation}
g^{(1)}(Y) = c\, \Ed{P_{S|Y}}{S}
\end{equation}
for $0\le c \le 2/\eps$.  In this case,
\begin{align} 
\tell_2\bigl(g^{(1)},\beta_{*,S}^{(1)}\bigr) 
&= \eps\, c\left(1-\frac{\eps}{2}c\right) 
\Ed{P_Y}{\bnorm{\bEd{P_{S|Y}}{S}}^2} \notag\\
&= \frac12\,\bigl(1-(1-\eps\, c)^2\bigr)
\Ed{P_Y}{\bnorm{\bEd{P_{S|Y}}{S}}^2} \notag\\
&\le \frac12 \,
\Ed{P_Y}{\bnorm{\bEd{P_{S|Y}}{S}}^2},
\label{eq:ll-bnd}
\end{align}
where equality is achieved when $c=1/\eps$.  Hence, the optimum value
of $g^{(1)}(y)$ is 
\begin{equation}
g^{(1)}_{*,S}(y) = \frac1\eps \, \bmu_{S|Y}(y),
\label{eq:g_1-opt}
\end{equation}
which we note has $\bE{g^{(1)}_{*,S}(Y)}= 0$,
as our constraints \eqref{eq:gbi-mean} dictate.  In turn, substituting
the right-hand side of \eqref{eq:ll-bnd} into \eqref{eq:tell-def} via
\eqref{eq:tell-exp}, we 
obtain \eqref{eq:D*-special}
as desired.

Moreover, the corresponding $g_{*,S}(y)$ and $\beta_{*,S}(y)$ satisfy
\eqref{eq:gb*-special}
as desired, i.e., 
$\Pt_{Y|S}^*(y|s)$ takes the form \eqref{eq:Pt*-special}.
\end{IEEEproof}

\subsection{Proof of \protect\corolref{corol:softmax}}
\label{app:softmax-corol}

First, without loss of generality we impose on $f$ the
constraints \eqref{eq:cF-def}, so 
\begin{equation}
\bmu_S=\bzero\qquad\text{and}\qquad \bLa_S=\bI,
\label{eq:S-norm-repeat}
\end{equation}
Next, note that since $f$ is injective, by the invariance of mutual
information to coordinate transformations, the first term on the
right-hand side of \eqref{eq:D*-softmax} is $I(X;Y)$, which doesn't
depend on $f$.   Accordingly, we have, specializing the second term on
the right-hand side of \eqref{eq:D*-softmax} to the case
\eqref{eq:S-norm-repeat},
\begin{align*} 
f_* 
&= \argmax_{f\in\cF_k} \Ed{P_Y}{\bnorm{\bEd{\Ph_{X|Y}}{f(X)}}^2} \\
&= \argmax_{\bfVX} \bfrob{\dtmt\, \bfVX}^2,
\end{align*}
where $\dtmt$ is as defined in \eqref{eq:dtmt-def}, $\bfVX$ is the
$\cardX\times k$ matrix whose $i$\/th column is the feature vector
associated with $f_i$, the $i$\/th element of $f$, and the
maximization with respect to $\bfVX$ is subject to the constraint
\begin{equation*}
\bigl(\bfVX\bigr)^\T\bfVX=\bI,
\end{equation*}
which corresponds to \eqref{eq:cF-def}.  Accordingly, applying
\lemref{lem:eig-k}, we obtain
\begin{equation*}
\bfVX = \bPsi^X_{(k)}, 
\end{equation*}
i.e., 
\begin{equation*}
f_i(x) = f_i^*(x),\quad i=1,\dots,k,
\end{equation*}

Finally, to obtain \eqref{eq:g**-softmax} we use 
\eqref{eq:gi-condexp} in \eqref{eq:g*-softmax} with
\eqref{eq:S-norm-repeat}, and \eqref{eq:bt**-softmax} follows
immediately via \eqref{eq:bt*-softmax}.
\hfill\IEEEQED

\section{Appendices for \protect\secref{sec:gauss}}
\label{app:gauss}

\subsection{Proof of \protect\factref{fact:sv-ccm}}
\label{app:sv-ccm}

Suppose $\bM$ is $k_1\times k_2$.  Let $\bpsi_i^\mathrm{L}$ and
$\bpsi_i^\mathrm{R}$ denote the left and right singular vectors of
$\bM$ corresponding to $\sigma_i(\bM)$, for
$i=1,\dots,\min\{k_1,k_2\}$.  Then 
\begin{equation*}
\begin{bmatrix}
  \bigl(\bpsi_i^\mathrm{L}\bigr)^\T &
  \bigl(\bpsi_i^\mathrm{R}\bigr)^\T \end{bmatrix}^\T
\qquad\text{and}\qquad
\begin{bmatrix}
  \bigl(\bpsi_i^\mathrm{L}\bigr)^\T &
  -\bigl(\bpsi_i^\mathrm{R}\bigr)^\T \end{bmatrix}^\T
\end{equation*}
are eigenvectors of $\bLa$ with eigenvalues $1+\sigma_i(\bM)$ and
$1-\sigma_i(\bM)$, respectively.  The remaining
$\max\{k_1,k_2\}-\min\{k_1,k_2\}$ eigenvalues are 1 and, if $k_2>k_1$,
correspond to eigenvectors 
\begin{equation*} 
\begin{bmatrix} \bzero & \bigl(\bpsi_j^\mathrm{R}\bigr)^\T \end{bmatrix}^\T,
\end{equation*}
for $j=k_1+1,\dots,k_2$.
Likewise, if $k_1>k_2$ these unity eigenvalues correspond to
eigenvectors 
\begin{equation*}
\begin{bmatrix} \bigl(\bpsi_j^\mathrm{L}\bigr)^\T &
  \bzero \end{bmatrix}^\T,
\end{equation*}
for $j=k_2+1,\dots,k_1$.  Hence, we conclude
that the eigenvalues are nonnegative if and only if $\sigma_i(\bM)\le 1$
for $i=1,\dots,\min\{k_1,k_2\}$.  \hfill\IEEEQED

\subsection{Proof of \protect\corolref{corol:ce-gauss}}
\label{app:ce-gauss}

Applying \eqref{eq:cov-modal} we have
\begin{align*}
\bE{\bg^*(Y)\mid X}
&= \bE{\bigl(\bG^*\bigr)^\T Y\mid X} \\
&= \bigl(\bG^*\bigr)^\T \bGa_{Y|X}X  \\
&= \bigl(\bG^*\bigr)^\T \bLa_{YX} \bLa_X^{-1} X \\
&= \bigl(\bG^*\bigr)^\T \bigl(\bG^*\bigr)^{-\T} \bSi\, \bigl(\bF^*\bigr)^{-1}
\bLa_X^{-1} X \\ 
&= \bSi\, \bigl(\bPsi^X\bigr)^\T \bLa_X^{1/2} \bLa_X^{-1} X \\
&= \bSi\, \bigl(\bF^*\bigr)^\T X \\
&= \bSi\, \bff^*(X)
\end{align*}
and, analogously,
\begin{align*}
\bE{\bff^*(X)\mid Y}
&= \bE{\bigl(\bF^*\bigr)^\T X\mid Y} \\
&= \bigl(\bF^*\bigr)^\T\bGa_{X|Y}Y  \\
&= \bigl(\bF^*\bigr)^\T \bLa_{XY} \bLa_Y^{-1} Y \\
&= \bigl(\bF^*\bigr)^\T  \bigl(\bF^*\bigr)^{-\T}  \bSi\, \bigl(\bG^*\bigr)^{-1}
\bLa_Y^{-1} Y \\ 
&= \bSi\, \bigl(\bPsi^Y\bigr)^\T \bLa_Y^{1/2} \bLa_Y^{-1} Y \\
&= \bSi\, \bigl(\bG^*\bigr)^\T Y \\
&= \bSi\, \bg^*(Y).
\end{align*}
\hfill\IEEEQED

\subsection{Proof of \protect\lemref{lem:Db-invariance}}
\label{app:Db-invariance}

First, we have
\begin{align}
&(\bA\mu_P+\bc-(\bA\mu_Q+\bc))^\T \notag\\
&\qquad\qquad{}\cdot (\bA\bLa_Q\bA^\T)^{-1}(\bA\mu_P+\bc-(\bA\mu_Q+\bc)) \notag\\
&\quad = (\mu_P-\mu_Q)^\T \bA^\T 
\bA^{-\T}\bLa_Q^{-1}\bA^{-1}\bA(\mu_P-\mu_Q) \notag\\
&\quad = 
(\mu_P-\mu_Q)^\T\bLa_Q^{-1}(\mu_P-\mu_Q).
\label{eq:invar-first}
\end{align}
Second, we have
\begin{align}
&\bfrob{\bigl(\bA\bLa_Q\bA^\T\bigr)^{-1/2}\bigl(\bA\bLa_P\bA^\T\!-\!\bA\bLa_Q\bA^\T\bigr) \bigl(\bA\bLa_Q\bA^\T\bigr)^{-1/2}}^2 
  \notag\\
&\quad=\tr\Bigl(
\bigl(\bA\bLa_Q\bA^\T\bigr)^{-1}\bA\bigl(\bLa_P-\bLa_Q\bigr)\bA^\T \notag\\
&\qquad\qquad\qquad{}\cdot\bigl(\bA\bLa_Q\bA^\T\bigr)^{-1}
\bA\bigl(\bLa_P-\bLa_Q\bigr)\bA^\T \smash[t]{\Bigr)} \label{eq:use-frob-ident-a}\\
&\quad=\tr\Bigl(
\bA^{-\T}\bLa_Q^{-1}\bigl(\bLa_P-\bLa_Q\bigr) 
\bLa_Q^{-1}\bigl(\bLa_P-\bLa_Q\bigr)\bA^\T \smash[t]{\Bigr)}\notag\\
&\quad=\tr\Bigl(
\bLa_Q^{-1}\bigl(\bLa_P-\bLa_Q\bigr) 
\bLa_Q^{-1}\bigl(\bLa_P-\bLa_Q\bigr)
\smash[t]{\Bigr)} \label{eq:use-cyclic}\\ 
&\quad=\bfrob{\bLa_Q^{-1/2}\bigl(\bLa_P-\bLa_Q\bigr)\bLa_Q^{-1/2}}^2,
\label{eq:use-frob-ident-b}
\end{align}
where to obtain \eqref{eq:use-frob-ident-a} we have used
used \lemref{lem:frob-ident}, to obtain \eqref{eq:use-cyclic} we have
used the invariance of the trace operator to cyclic permutations, and 
to obtain \eqref{eq:use-frob-ident-b} we have again used
used \lemref{lem:frob-ident}.  Combining \eqref{eq:invar-first} and
\eqref{eq:use-frob-ident-b} with \eqref{eq:Db-def}, we obtain
\eqref{eq:Db-invariance}.  \hfill\IEEEQED

\subsection{Proof of \protect\lemref{lem:nbhdg-frob}}
\label{app:nbhdg-frob}

With 
\begin{equation}
\Ct \defeq \begin{bmatrix}\Zt \\ \Wt \end{bmatrix},
\label{eq:Ct-def}
\end{equation}
via \eqref{eq:C-form}, we have
\begin{equation*}
  \bLa_\Ct = \E{\Ct \Ct^\T} = \begin{bmatrix} \bI & \eps\,\bPhi^{Z|W}
  \\ \eps\,\bPhi^{Z|W} & \bI
\end{bmatrix},
\end{equation*}
Thus,
\begin{align}
\Db\bigl(\gauss(\bzero,\bLa_\Ct)\bigm\|\gauss(\bzero,\bI)\bigr)
&= \frac12 \bfrob{\bLa_\Ct-\bI}^2 \notag\\
&= \frac12 \bbfrob{\begin{bmatrix} \bzero & \eps\, \bPhi^{Z|W} \\ \eps\,
    \bigl(\bPhi^{Z|W} \bigr)^\T & \bzero
  \end{bmatrix}}^2 \notag \\
&= \eps^2\,\bfrob{\bPhi^{Z|W}}^2.
\end{align}
\hfill\IEEEQED

\subsection{Proof of \protect\lemref{lem:Db-cond-local}}
\label{app:Db-cond-local}

We obtain 
\begin{align}
&\Db(P_{Z|W}(\cdot|w)\|P_Z) \notag\\
&\qquad=\Db(P_{\Zt|\Wt}(\cdot|\wt)\|P_\Zt) \label{eq:use-coord-inv-a} \\
&\qquad= \eps^2\, \wt^\T \bigl(\bPhi^{Z|W}\bigr)^\T \bPhi^{Z|W} \wt
 \notag\\
&\quad\qquad\qquad\qquad{}+ \frac12\bbfrob{\Bigl(\bI-\eps^2\bPhi^{Z|W}\bigl(\bPhi^{Z|W}\bigr)^\T\Bigr)-\bI}^2 \label{eq:use-Db-def}\\
&\qquad=\eps^2\bnorm{\bPhi^{Z|W}\wt}^2 +
 \frac{\eps^4}2\bfrob{\bPhi^{Z|W}\bigl(\bPhi^{Z|W}\bigr)^\T}^2 \notag\\
&\qquad=\eps^2\bnorm{\bPhi^{Z|W}\wt}^2 + o(\eps^2), \notag
\end{align}
where to obtain \eqref{eq:use-coord-inv-a} we have used
\lemref{lem:Db-invariance}, and to obtain \eqref{eq:use-Db-def} we
have used \eqref{eq:Db-def} and the fact
that 
\begin{equation*}
\Zt = \eps\, \bPhi^{Z|W} \Wt + \nu_{\Wt\to\Zt},
\end{equation*}
where $\bPhi^{Z|W}$ is as defined in \eqref{eq:innov-matrix} and
\begin{equation*}
\bE{\nu_{\Wt\to\Zt} \nu_{\Wt\to\Zt}^\T} = \bI - \eps^2 \bPhi^{Z|W}
\bigl( \bPhi^{Z|W} \bigr)^\T.
\end{equation*}
\hfill\IEEEQED

\subsection{Proof of \protect\lemref{lem:Ib-identity}}
\label{app:Ib-identity}

We have
\begin{align}
&\bEd{P_W}{\Db(P_{Z|W}(\cdot|W)\|P_Z)} \notag\\
&\qquad= \eps^2\,\E{\bnorm{\bPhi^{Z|W}\Wt}^2} +
  o(\eps^2) \label{eq:use-Db-cond-local}\\   
&\qquad= \eps^2\,\bfrob{\bPhi^{Z|W}}^2  + o(\eps^2) \label{eq:use-rie-again}\\
&\qquad= \Db(P_{Z,W}\|P_ZP_W) \bigl(1 + o(1)\bigr), \label{eq:use-nbhdg-frob}
\end{align}
where to obtain \eqref{eq:use-Db-cond-local} we have used
\lemref{lem:Db-cond-local}, to obtain \eqref{eq:use-rie-again} we have
used \lemref{lem:rie} (with $k_1=\dimW$ and $k_2=1$) since 
$\Wt$ is spherically symmetric, and to obtain \eqref{eq:use-nbhdg-frob} we have
used \lemref{lem:nbhdg-frob}.
\hfill\IEEEQED

\subsection{Proof of \protect\factref{fact:logdet}}
\label{app:logdet}

Let the $i$\/th singular value of $\bA$ be $\la_i$.   Then it suffices to
note that
\begin{align*} 
\ln\bdetb{\bI - \eps^2 \bA\,\bA^\T} 
&=\ln\prod_i \bigr(1-\eps^2\la_i^2\bigl) \\
&=\ln \biggl(1-\eps^2\sum_i \la_i^2 + o(\eps^2)\biggr)\\
&= -\eps^2\sum_i \la_i^2 + o(\eps^2)\\
&= -\eps^2 \frob{\bA}^2 + o(\eps^2).
\end{align*}
\hfill\IEEEQED

\subsection{Proof of \protect\lemref{lem:D-local}}
\label{app:D-local}

We have
\begin{align}
&D\bigl(P_{Z|W}(\cdot|w)\bigm\|P_Z\bigr) \notag\\
&\qquad= D\bigl(P_{\Zt|\Wt}(\cdot|\wt)\bigm\|P_\Zt \bigr) \label{eq:use-D-invariance}\\
&\qquad= \frac12 \biggl[ \eps^2\,\bnorm{\bPhi^{Z|W}\wt}^2 \notag\\
&\qquad\qquad\qquad{}+
    \tr\biggl(\Bigl(\bI-\eps^2\bPhi^{Z|W}\bigl(\bPhi^{Z|W}\bigr)^\T\Bigr)-\bI\biggr)
    \notag\\ 
&\quad\qquad\qquad\qquad\qquad{}-\ln\Bigl|\bI-\eps^2\bPhi^{Z|W}\bigl(\bPhi^{Z|W}\bigr)^\T\Bigr|
    \biggr] \notag\\ 
&\qquad= \frac{\eps^2}2 \, \bnorm{\bPhi^{Z|W}\wt}^2 + o(\eps^2) 
\label{eq:D-nbhdg} \\
&\qquad= \frac12 \,\Db(P_{Z|W}(\cdot|w)\|P_Z) + o(\eps^2), 
\label{eq:use-Db-cond-local2}
\end{align}
where to obtain \eqref{eq:use-D-invariance} we have used the
coordinate invariance of KL divergence, to obtain \eqref{eq:D-nbhdg}
we have used \factref{fact:logdet}, and to obtain
\eqref{eq:use-Db-cond-local2} we have used
\lemref{lem:Db-cond-local}.\hfill\IEEEQED

\subsection{Proof of \protect\corolref{corol:I-local}}
\label{app:I-local}

We have
\begin{align}
I(Z;W)
&= \bEd{P_W}{D\bigl(P_{Z|W}(\cdot|W)\bigm\|P_Z\bigr)} \notag\\
&= \frac12\, \bEd{P_W}{\Db\bigl(P_{Z|W}(\cdot|W)\bigm\|P_Z\bigr)} +
o(\eps^2) \label{eq:use-lem:D-local3}\\
&= \frac12\, \Db\bigl(P_{Z,W}\bigm\|P_ZP_W\bigr)+
o(\eps^2) \label{eq:use-lem:Ib-identity} 
\end{align}
where to obtain \eqref{eq:use-lem:D-local3} we have used
\lemref{lem:D-local}, and to obtain \eqref{eq:use-lem:Ib-identity} we
have used \lemref{lem:Ib-identity}.
\hfill\IEEEQED

\subsection{Proof of \protect\lemref{lem:KL-gauss-frob}}
\label{app:KL-gauss-frob}

First, via \eqref{eq:KL-gauss} and the invariance of divergence to
coordinate tranformations we have 
\begin{align}
&D(P_{X,Y}\| Q_{X,Y}) \notag\\
&\quad = \frac12 \left[
    \tr\Bigl(\bigl(\bLa_{\Xt,\Yt}^Q\bigr)^{-1}\,\bLa_{\Xt,\Yt}^P 
  - \bI \Bigr) -
  \ln\bdetb{\bLa_{\Xt,\Yt}^P\, \bigl(\bLa_{\Xt,\Yt}^Q\bigr)^{-1}}
  \right] 
\label{eq:DPPt}
\end{align}
where
\begin{equation}
\bLa_{\Xt,\Yt}^P = \begin{bmatrix} \bI & \dtmg_P^\T \\ \dtmg_P & \bI 
\end{bmatrix} \quad\text{and}\quad \bLa_{\Xt,\Yt}^Q
= \begin{bmatrix} \bI & \dtmg_Q^\T \\ \dtmg_Q & \bI  
\end{bmatrix} .
\label{eq:bLat-def}
\end{equation}

Next, using \eqref{eq:bLat-def} with \eqref{eq:bLa-Zt-inv-2} we obtain
\begin{align*}
&\bigl(\bLa_{\Xt,\Yt}^Q\bigr)^{-1}\, \bLa_{\Xt,\Yt}^P \notag\\
&\quad = 
\begin{bmatrix} \bI+\dtmg_Q^\T\dtmg_Q & -\dtmg_Q^\T \\ -\dtmg_Q &
  \bI+\dtmg_Q\, \dtmg_Q^\T 
\end{bmatrix} \begin{bmatrix} \bI & \dtmg_P^\T \\ \dtmg_P & \bI 
\end{bmatrix} + o(\eps^2) \\
&\quad =
\begin{bmatrix}
  \bI+\dtmg_Q^\T (\dtmg_Q-\dtmg_P) 
& (\dtmg_P-\dtmg_Q)^\T \\
\dtmg_P-\dtmg_Q &
\bI+\dtmg_Q(\dtmg_Q-\dtmg_P)^\T
\end{bmatrix} + o(\eps^2),
\end{align*}
so
\begin{align} 
\tr\Bigl( \bigl(\bLa_{\Xt,\Yt}^Q\bigr)^{-1} \bLa_{\Xt,\Yt}^P \!-\!\bI
  \Bigr) 
&= 2\tr\Bigl( \dtmg_Q (\dtmg_Q-\dtmg_P)^\T
\Bigr) + o(\eps^2)\notag\\
&= 2\,\bfrob{\dtmg_Q}^2 \!-\! 2\tr\bigl( \dtmg_Q\dtmg_P^\T \bigr) +
o(\eps^2). 
\label{eq:tr-piece}
\end{align}
Finally, using \factref{fact:logdet} and the block matrix
determinant identity we obtain
\begin{subequations} 
\begin{align}
\ln\bdetb{\bLa_{\Xt\Yt}^P}
&= \ln\bdetb{\bI-\dtmg_P^\T\dtmg_P} = -\bfrob{\dtmg_P}^2 +
o(\eps^2) \\
\ln\bdetb{\bLa_{\Xt\Yt}^Q}
&= \ln\bdetb{\bI-\dtmg_Q^\T\dtmg_Q} = -\bfrob{\dtmg_Q}^2 +
o(\eps^2),
\end{align}
\label{eq:det-piece}%
\end{subequations}
so substituting \eqref{eq:tr-piece} and \eqref{eq:det-piece} in 
\eqref{eq:DPPt} yields \eqref{eq:KL-gauss-frob}.
\hfill\IEEEQED

\subsection{Nonlinear Features of Gaussian Distributions} 
\label{app:nonlinear}

For the Gaussian case, in a modal decomposition of the form
\eqref{eq:svd-gauss-alt}, only some of the features
$\ft_1^*,\gt_1^*,\ft_2^*,\gt_2^*,\dots$ are linear, which is implied
by expanding the exponentiation operator in \eqref{eq:svd-gauss-form}
using a Taylor Series.  For instance, when $\dimX=\dimY=1$, one obtains
Mehler's decomposition \cite{fgm1866}
\begin{subequations} 
\begin{equation}
P_{\Xt,\Yt}(\xt,\yt) 
= P_{\Xt}(\xt)\,P_{\Yt}(\yt) \left[ 1 + \sum_{i=1}^\infty \rho^i\,
  \hp_i(\xt)\,\hp_i(\yt) \right],
\end{equation}
where $\rho=\la_{\Xt\Yt}=\E{\Xt\Yt}$, and where
$\hp_i$ is the (scaled) $i$\/th-order Hermite polynomial
\begin{equation} 
\hp_i(\xt) \defeq \frac1{\sqrt{i!}} (-1)^i \e^{\xt^2/2}
\frac{\diff^i}{\diff \xt^i} \e^{-\xt^2/2} , \quad i=1,2,\dots.
\end{equation}%
\label{eq:hermite-exp}%
\end{subequations}
Note that $\ft_i^*=\gt_i^*=\hp_i$ satisfy
\eqref{eq:svd-gauss-alt-constr} as required, and that the features
corresponding to the dominant mode ($i=1$) are linear:
$\hp_1(\upsilon)=\upsilon$.  

For $\mindim>1$ [with \eqref{eq:Kg-def}], the modal decomposition of
the form \eqref{eq:svd-gauss-alt} is straightforward to derive as a
generalization of \eqref{eq:hermite-exp}, and involves the
corresponding multivariate Hermite polynomials---see, e.g.,
\cite{wfk45,ds72}.  However, it is important to emphasize that if
$\mindim>1$ and $k>1$, then the $k$ dominant modes need not, in
general, be linear, as the following example shows.
\begin{example}
Suppose
$\dimX=\dimY=k=2$, $(\Xt_1,\Yt_1)$ and $(\Xt_2,\Yt_2)$ are
independent, and $\rho_l=\la_{\Xt_l\Yt_l}=\bE{\Xt_l\Yt_l}$ for
$l=1,2$.  Then we have the decomposition
\begin{align*}
P_{\Xt,\Yt}(\xt,\yt) 
&= P_{\Xt_1}(\xt_1)\, P_{\Xt_2}(\xt_2)\,
P_{\Yt_1}(\yt_1)\,
P_{\Yt_2}(\yt_2) \\
&\qquad{} \cdot
\left[ 1 + \sum_{i=1}^\infty \rho_1^i\,
  \hp_i(\xt_1)\,\hp_i(\yt_1) \right] \\
&\qquad\qquad\cdot\left[ 1 + \sum_{i=1}^\infty \rho_2^i\,
  \hp_i(\xt_2)\,\hp_i(\yt_2) \right],
\end{align*}
which when expanded is of the form \eqref{eq:svd-gauss-alt}.  But if
$\rho_1^2>\rho_2$, then the dominant feature pair is
\begin{align*} 
\ft_1^*(\xt) &=\hp_1(\xt_1)=\xt_1 \\
\gt_1^*(\yt) &=\hp_1(\yt_1)=\yt_1,
\end{align*}
corresponding to singular value $\sigmat_1=\rho_1$, while the features
for the next largest singular value are
\begin{align*} 
\ft_2^*(\xt) &=\hp_2(\xt_1)=(\xt_1^2-1)/\sqrt{2}\\
\gt_2^*(\yt) &=\hp_2(\yt_1)=(\yt_1^2-1)/\sqrt{2},
\end{align*}
corresponding to singular value $\rho_1^2$, rather than the linear
features corresponding to singular value $\rho_2$.  
\end{example}

\subsection{Linear Features for NonGaussian Distributions}
\label{app:dtm-dtmg}

The following proposition directly relates the CCM $\dtmg$ from our
Gaussian analysis to the associated CDM $\dtm$ from our discrete
analysis. 

\begin{proposition}
\label{prop:dtm-dtmg}
Let $\X\subset\Xg$ and $\Y\subset\Yg$ be finite sets with probability
mass function 
$P_{X,Y}$ such that $\E{X}=0$ and $\E{Y}=0$, and let $\dtm$ be as
defined in \eqref{eq:dtm-def}.  Then
\begin{equation} 
\bPi^Y \, \dtm \, \bigl(\bPi^X\bigr)^\T = \dtmgg,
\label{eq:dtm-dtmg}
\end{equation}
where $\dtmgg$ is as defined in \eqref{eq:dtmg-def} (with the notation
refined to distinguish it from $\dtm$), and where
\begin{equation} 
\bPi^X = \bX\, \sqrt{\bP_X} \quad\text{and}\quad \bPi^Y = \bY\, \sqrt{\bP_Y},
\end{equation}
with $\bX$ and $\bY$ denoting $\dimX\times\cardX$ and $\dimY\times\cardY$
matrices whose columns are the vectors in $\X$ and $\Y$, respectively.
\end{proposition}

\begin{IEEEproof}
Given $\X=\{x_1,\dots,x_\cardX\}\subset\Xg$ and
$\Y=\{y_1,\dots,y_\cardY\}\subset\Yg$, consider (without loss of
generality) arbitrary zero-mean, unit-variance features $f\colon
\X\mapsto \reals$ and $g\colon \Y \mapsto \reals$, whose corresponding
feature vectors are $\bfvX\in\ivspaceX$ and $\bfvY\in\ivspaceY$,
respectively.

In addition, let $f_\mathrm{L}(x)=\bigl(\bfwXg\bigr)^\T x$ and
$g_\mathrm{L}(y)=\bigl(\bfwYg\bigr)^\T y$ denote the linear MMSE
approximations to $f$ and $g$, respectively, i.e.,
\begin{subequations} 
\begin{align} 
\bfwXg &= \argmin_{\bfwgeng\in\Xg} \E{\bigl(\bfwgeng^\T\, X -
  f(X)\bigr)^2} \label{eq:bxigX-def}\\
\bfwYg &= \argmin_{\bfwgeng\in\Yg} \E{\bigl(\bfwgeng^\T\, Y - g(Y)\bigr)^2}
\label{eq:bxigY-def}
\end{align}
\end{subequations}
Without loss of generality, we assume that $\bLa_X=\bI$ and
$\bLa_Y=\bI$, since if not they can be converted to this form by
linear transformation.  Then, from standard linear (MMSE) estimation
theory it follows that
\begin{subequations} 
\begin{align}
\bfwXg &= \E{ f(X)\, X^\T} 
= \underbrace{\bX \sqrt{\bP_X}}_{\defeq \bPi^X} \, \bfvX \label{eq:bPiX-def}\\
\bfwYg &= \E{\, g(Y)\, Y^\T} 
= \underbrace{\bY \sqrt{\bP_Y}}_{\defeq \bPi^Y} \, \bfvY, \label{eq:bPiY-def}
\end{align}
\end{subequations}
where $\bX$ and $\bY$ are matrices whose columns are the vectors in
$\X$ and $\Y$, respectively, and where the matrices $\bPi^X$ and
$\bPi^Y$ characterize the associated projections.

Next, note that if $f$ is linear, i.e., 
\begin{equation}
f(x)=\bfwalt^\T\, x
\label{eq:flin-def}
\end{equation}
for some
$\bfwalt$, then 
\begin{align*}
\bigl(\bPi^X\bigr)^\T\bfwalt = \sqrt{\bP_X}\, \bX^\T \bfwalt = \bfvX
\end{align*}
since $\bX^\T\bfwalt$ is a vector whose $x$\/th element is $f(x)$.

Now for any feature $f$ with feature vector $\fvX$, the
feature vector $\bfvY \defeq \dtm\, \bfvX$ corresponds to the
feature $g(y) = \E{f(X)\mid Y=y}$.   To see this, it suffices to note
that $\bfvY$ has elements
\begin{align*} 
\fvY(y) 
&= \sum_{x\in\X} \dtms(x,y)\, \fvX(x) \\
&= \sum_{x\in\X} \frac1{\sqrt{P_Y(y)}} \, P_{Y|X}(y|x) \, \sqrt{P_X(x)} \,
  \sqrt{P_X(x)}\, f(x) \\
&= \sqrt{P_Y(y)} \, \underbrace{\sum_{x\in\X} P_{X|Y}(x|y)\,
    f(x)}_{= g(y)}
\end{align*}

In turn, specializing $f$ to the case \eqref{eq:flin-def} yields
\begin{equation*}
\dtm\, \bigl(\bPi^X\bigr)^\T \bfwalt = \sqrt{\bP_Y}\, \bE{\bfwalt^\T\, X
  \bigm| Y},
\end{equation*}
and specializing the resulting $g$ in \eqref{eq:bPiY-def} with
\eqref{eq:bxigY-def} yields
\begin{align}
\bPi^Y \dtm\, \bigl(\bPi^X\bigr)^\T \bfwalt
&= \argmin_{\bfwgeng\in\Yg} \E{\bigl(\bfwgeng^\T\, Y - \bE{\bfwalt^\T\, X \bigm| Y}\bigr)^2} \notag\\
&= \argmin_{\bfwgeng\in\Yg} \E{\bigl( \bfwgeng^\T\, Y - \bfwalt^\T\, X
  \bigr)^2} \label{eq:bPi-altform}\\
&= \bLa_Y^{-1} \, \bLa_{YX} \, \bfwalt \label{eq:bPi-lls-soln} \\
&= \dtmgg \, \bfwalt \label{eq:use-dtmg-def}
\end{align}
To obtain \eqref{eq:bPi-altform} we have used
\begin{align*} 
&\E{\bigl(\bfwgeng^\T\, Y - \bfwalt^\T\, X \bigr)^2} \\
&\qquad\qquad\ = 
\E{\bigl(\bfwgeng^\T\, Y - \bE{\bfwalt^\T\, X \bigm| Y}\bigr)^2} \\
&\qquad\qquad\quad{} + \E{\bigl(\bfwalt^\T\, X - \bE{\bfwalt^\T\, X \bigm| Y}\bigr)^2} \\
&\qquad\qquad\qquad\quad{} - 2\,\ev\Bigl[\bigl(\bfwalt^\T\, X - \bE{\bfwalt^\T\, X \bigm| Y}\bigr) \\
&\qquad\qquad\qquad\qquad\qquad\cdot \bigl(\bfwgeng^\T\, Y - \bE{\bfwalt^\T\, X \bigm|
  Y}\bigr) \Bigr],  
\end{align*}
where we note that the second term does not depend on $\bfwgeng$, and that
the last term is zero due to the orthogonality property of MMSE
estimators.  In turn, to obtain \eqref{eq:bPi-lls-soln} we recognize
\eqref{eq:bPi-altform} as a linear MMSE estimation problem, whose
solution depends only on the first and second moments of $(X,Y)$, and
to obtain \eqref{eq:use-dtmg-def} we use \eqref{eq:dtmg-def}.
Finally, since \eqref{eq:use-dtmg-def} holds for all $\bfwalt$, we obtain
\eqref{eq:dtm-dtmg} as desired. 

\end{IEEEproof}

\subsection{Proof of \protect\factref{fact:gm-cov-prod}}
\label{app:gm-cov-prod}

It suffices to note that 
\begin{align*}
\bLa_{\Zt_1\Zt_3} 
&= \E{\bE{\Zt_1\Zt_3^\T|\Zt_2}} \\
&= \E{\bE{\Zt_1|\Zt_2}\,\bE{\Zt_3|\Zt_2}^\T} \\
&= \E{\bLa_{\Zt_1\Zt_2}\, \Zt_2\, \Zt_2^\T \bLa_{\Zt_3\Zt_2}^\T} \\
&= \bLa_{\Zt_1\Zt_2}\, \bLa_{\Zt_2}\, \bLa_{\Zt_2\Zt_3} \\
&= \bLa_{\Zt_1\Zt_2}\, \bLa_{\Zt_2\Zt_3}.
\end{align*}
\hfill\IEEEQED

\subsection{Proof of \protect\propref{prop:mmse-rie}}
\label{app:mmse-rie}

Without loss of generality, we restrict $S_{(k)}$ and $T_{(k)}$ so that
they are normalized with respect to $P_X$ and $P_Y$, respectively,
i.e., $\bigl(\bF_{(k)},\bG_{(k)}\bigr)\in\cL$ as defined in
\eqref{eq:cL-def}.  Accordingly, we have the representations
\eqref{eq:FG-def} in which \smash{$\bfWX$} and \smash{$\bfWY$} satisfy
\eqref{eq:orthog-gauss}.  

For the MMSE estimation of $V$ based on $S_{(k)}$, the MSE is
\begin{align}
&\mse^{V|S}\bigl(\Cg^\dimY_{\epsY}(\bLa_Y),\bF_{(k)}\bigr) \notag\\
&\quad= \tr\bigl(\bLa_V^{1/2} \bLa_{\Vt|S_{(k)}} \bLa_V^{1/2}\bigr) \notag\\
&\quad= \tr(\bLa_V) - \epsY^2\,\bfrob{\bLa_V^{1/2}\,
  \bigl(\bPhi^{Y|V}\bigr)^\T \dtmg\, \bfWX}^2
\label{eq:mse-V|S}
\end{align}
where we have used \eqref{eq:bPhiYV2XV},
so
\begin{align}
&\mseb^{V|S}\bigl(\bF_{(k)}\bigr) \notag\\
&\quad=
  \bEd{\mathrm{RIE}}{\mse^{V|S}\bigl(\Cg^\dimY_{\epsY}(\bLa_Y),\bF_{(k)}\bigr)} \notag\\
&\quad = \tr(\bLa_V) - \frac{\epsY^2}{\dimV\dimY}\,\tr(\bLa_V)\,
  \E{\bfrob{\bPhi^{Y|V}}^2} \,\bfrob{\dtmg\, \bfWX}^2 \label{eq:use-rieg}\\
&\quad\ge \tr(\bLa_V)\left[1 - \frac{\epsY^2}{\dimV\dimY}\,
\E{\bfrob{\bPhi^{Y|V}}^2} \,\sum_{i=1}^k \sigma_i^2 \right], 
\label{eq:use-eig-relax}
\end{align}
where to obtain \eqref{eq:use-rieg} we have used \lemref{lem:rie}, 
and to obtain \eqref{eq:use-eig-relax} we have used
\lemref{lem:eig-k}.  It further follows from \lemref{lem:eig-k} that
the inequality \eqref{eq:use-eig-relax} holds with equality when we
choose $\bfWX$ according to \eqref{eq:bfWX-opt}.

For the MMSE estimation of $U$ based on $S_{(k)}$, the MSE is
\begin{align}
&\mse^{U|S}\bigl(\Cg^\dimX_{\epsX}(\bLa_X),\bF_{(k)}\bigr) \notag\\
&\quad= \tr\bigl(\bLa_U^{1/2} \bLa_{\Ut|S_{(k)}} \bLa_U^{1/2}\bigr) \notag\\
&\quad= \tr(\bLa_U) - \epsX^2\,\bfrob{\bLa_U^{1/2}\,
  \bigl(\bPhi^{X|U}\bigr)^\T \bfWX}^2
\label{eq:mse-U|S}
\end{align}
so
\begin{align}
&\mseb^{U|S}\bigl(\bF_{(k)}\bigr) \notag\\
&\quad=
  \bEd{\mathrm{RIE}}{\mse^{U|S}\bigl(\Cg^\dimX_{\epsX}(\bLa_X),\bF_{(k)}\bigr)} \notag\\
&\quad = \tr(\bLa_U) - \frac{\epsX^2}{\dimU\dimX}\,\tr(\bLa_U)\,
  \E{\bfrob{\bPhi^{X|U}}^2} \,\bfrob{\bfWX}^2 \label{eq:use-rieg-again}\\
&\quad= \tr(\bLa_U)\left[1 - \frac{\epsX^2\,k}{\dimU\dimX}\,
\E{\bfrob{\bPhi^{X|U}}^2}  \right], 
\label{eq:use-eig-relax-again}
\end{align}
for any (admissible) choice of $\bF_{(k)}$, where to obtain
\eqref{eq:use-rieg-again} we have used \lemref{lem:rie}, and to obtain
\eqref{eq:use-eig-relax-again}  we have used that $\bfrob{\bfWX}^2=k$ due to
\eqref{eq:orthog-gauss}. 
Hence, the unique Pareto optimal choice of $\bF_{(k)}$ in
\eqref{eq:mse-rie-pareto} is as given by
\eqref{eq:bfWX-opt}.   

Via a symmetry argument (corresponding to interchanging the roles of
$X$ and $Y$, and $U$ and $V$, and noting that $\dtmg$ and $\dtmg^\T$
share the same singular values), it follows that
\begin{align*}
\mseb^{V|T}\bigl(\bG_{(k)}\bigr) 
&= \tr(\bLa_V)\left[1 - \frac{\epsY^2\,k}{\dimV\dimY}\,
\E{\bfrob{\bPhi^{Y|V}}^2}
\right] \notag\\
\mseb^{U|T}\bigl(\bG_{(k)}\bigr) 
&\ge \tr(\bLa_U)\left[1 - \frac{\epsX^2}{\dimU\dimX}\,
\E{\bfrob{\bPhi^{X|U}}^2} \,\sum_{i=1}^k \sigma_i^2 \right], 
\end{align*}
with equality in the latter when $\bG_{(k)}$ is given by
\eqref{eq:bfWY-opt}.  Hence, the unique Pareto optimal choice of
$\bG_{(k)}$ in \eqref{eq:mse-rie-pareto} is as given by
\eqref{eq:bfWX-opt}.

Finally, we note that we obtain \eqref{eq:mseb-opt} by recognizing that
\begin{equation*}
\bar{E}^{X|U} =
\frac{\E{\bfrob{\bPhi^{X|U}}^2}}{\dimU\dimX}
\quad\text{and}\quad
\bar{E}^{Y|V} = 
\frac{\E{\bfrob{\bPhi^{Y|V}}^2}}{\dimV\dimY}.
\end{equation*}
\hfill\IEEEQED

\subsection{Proof of \protect\propref{prop:mmse-coop}}
\label{app:mmse-coop}

For the MMSE estimation of $V$ based on $S$, the MSE is, starting from 
\eqref{eq:mse-V|S},
\begin{align}
&\mse^{V|S}\bigl(\Cg^\dimY_{\epsY}(\bLa_Y),\bF_{(k)},\bigr) \notag\\
&\ = \tr(\bLa_V) - \epsY^2\,\bfrob{\bLa_V^{1/2}\,
  \bigl(\bPhi^{Y|V}\bigr)^\T \dtmg\, \bfWX}^2 \notag\\
&\ = \tr(\bLa_V) - \epsY^2\,\bfrob{\bLa_V^{1/2}\, \bQ^{Y|V}\,\bDel^{Y|V}\,
  \bigl(\bPhit^{Y|V}\bigr)^\T \dtmg\, \bfWX}^2 \label{eq:use-bPhiYV-svd} \\
&\ \ge \tr(\bLa_V) - \epsY^2\,\bspectral{\bLa_V^{1/2}}^2\, \bspectral{\bDel^{Y|V}}^2\,\bfrob{
    \bigl(\bPhit^{Y|V}\bigr)^\T \dtmg}^2\, \bspectral{\bfWX}^2 \label{eq:use-frob-submult-4}\\ 
&\ \ge \tr(\bLa_V) - \epsY^2\,\bspectral{\bLa_V^{1/2}}^2\,
  \left(\frac{\dimY+k}{k}\right) \,\sum_{i=1}^k \sigma_i^2
     \label{eq:use-multi-attrib}\\ 
&\ \ge k - \epsY^2\,
  \left(\frac{\dimY+k}{k}\right) \,\sum_{i=1}^k \sigma_i^2,
\label{eq:use-bLaV-constr}
\end{align}
where to obtain \eqref{eq:use-bPhiYV-svd} we have expressed
$\bPhi^{Y|V}$ in terms of its SVD
\begin{equation}
\bPhi^{Y|V} = \bPhit^{Y|V} \bDel^{Y|V} \, \bigl(\bQ^{Y|V}\bigr)^\T,
\label{eq:bPhiYV-svd}
\end{equation}
where the $\dimY\times k$ matrix $\bPhit^{Y|V}$ has orthonormal columns,
i.e.,
\begin{equation} 
\bigl(\bPhit^{Y|V}\bigr)^\T \bPhit^{Y|V} = \bI,
\label{eq:PhitYV-orthog}
\end{equation}
$\bDel^{Y|V}$ is a $k\times k$ diagonal matrix, and $\bQ^{Y|V}$ is $k\times k$
orthogonal matrix.   To obtain \eqref{eq:use-frob-submult-4} we have
(repeatedly) used \factref{fact:frob-submult} and the fact that $\bQ^{Y|V}$
is orthogonal, and to obtain \eqref{eq:use-multi-attrib} we have used
that $\bfWX$ satisfies \eqref{eq:orthog-gauss},
that 
\begin{equation}
\bspectral{\bDel^{Y|V}}^2 =
\bspectral{\bPhi^{Y|V}}^2 \le \frac{\dimY+k}{k}
\label{eq:bPhiYV-constr-s}
\end{equation}
since $V$ is a Gaussian multi-attribute and so satisfies the property
of \defref{def:gauss-multi-attribute}, and applied \lemref{lem:eig-k}
with the constraint \eqref{eq:PhitYV-orthog}.  And to obtain
\eqref{eq:use-bLaV-constr} we have used the last constraint in
\eqref{eq:gauss-config-constr}, which implies that none of the
singular values of $\bLa_V$ are smaller than unity.  
Finally, it is straightforward to verify that the inequalities leading
to the right-hand side of \eqref{eq:use-bLaV-constr} hold with
equality when
\begin{subequations} 
\begin{align} 
\bLa_V &=\bI \label{eq:bLaV-I}\\
\bPhi^{Y|V} &=
\sqrt{\frac{\dimY+k}{k}}\,\bPsi^Y_{(k)} \label{eq:bPhiYV-opt}
\end{align}
\label{eq:VF-opt}%
\end{subequations}
and
\begin{equation} 
\bfWX = \bPsi^X_{(k)}. 
\label{eq:bfWX-opt-repeat}
\end{equation}

For the MMSE estimation of $U$ based on $S$, the MSE is, starting from 
\eqref{eq:mse-U|S},
\begin{align} 
&\mse^{U|S}\bigl(\Cg^\dimX_{\epsX}(\bLa_X),\bF_{(k)}\bigr) \notag\\
&\quad= \tr(\bLa_U) - \epsX^2\,\bfrob{\bLa_U^{1/2}\,
  \bigl(\bPhi^{X|U}\bigr)^\T \bfWX}^2 \notag\\
&\quad\ge \tr(\bLa_U) - \epsX^2\,\bspectral{\bLa_U^{1/2}}^2\,
  \bspectral{\bDel^{X|U}}^2\,
  \bfrob{\bigl(\bPhit^{X|U}\bigr)^\T\bfWX}^2\label{eq:use-bPhiXU-svd}\\ 
&\quad\ge \tr(\bLa_U) - \epsX^2\,\bspectral{\bLa_U^{1/2}}^2\,
  \bigl(\dimX+k\bigr) \label{eq:use-multi-again}\\ 
&\quad\ge k - \epsX^2\, \bigl(\dimX+k\bigr),
\label{eq:use-bLaU-constr}
\end{align}
where to obtain \eqref{eq:use-bPhiXU-svd} we have used
\factref{fact:frob-submult}, expressing
$\bPhi^{X|U}$ in terms of its SVD
\begin{equation}
\bPhi^{X|U} = \bPhit^{X|U} \bDel^{X|U} \, \bigl(\bQ^{X|U}\bigr)^\T,
\label{eq:bPhiXU-svd}
\end{equation}
where the $\dimY\times k$ matrix $\bPhit^{X|U}$ has orthonormal columns,
i.e., 
\begin{equation}
\bigl(\bPhit^{X|U}\bigr)^\T \bPhit^{X|U} = \bI,
\label{eq:PhitXU-orthog}
\end{equation}
$\bDel^{X|U}$ is a $k\times k$ diagonal matrix, and $\bQ^{X|U}$ is
$k\times k$ orthogonal matrix.  To obtain \eqref{eq:use-multi-again}
we have used that $\bPhit^{X|U}$ satisfies \eqref{eq:PhitXU-orthog}
and $\bfWX$ satisfies \eqref{eq:orthog-gauss}, \lemref{lem:eig-k}, and
that
\begin{equation}
\bspectral{\bDel^{X|U}}^2 = \bspectral{\bPhi^{X|U}}^2 \le \frac{\dimX+k}{k}
\label{eq:bPhiXU-constr-s}
\end{equation}
since $U$ is a Gaussian multi-attribute and so satisfies the property
of \defref{def:gauss-multi-attribute}.
To obtain \eqref{eq:use-bLaU-constr} we have used the penultimate
constraint in
\eqref{eq:gauss-config-constr}, which implies that none of the
singular values of $\bLa_U$ are smaller than unity.  
Finally, the inequalities leading to the right-hand side of
\eqref{eq:use-bLaU-constr} hold with equality when, for example,
\begin{subequations} 
\begin{align} 
\bLa_U &=\bI \label{eq:bLaU-I}\\
\bPhi^{X|U} &=
\sqrt{\frac{\dimX+k}{k}}\,\bPsi^X_{(k)} \label{eq:bPhiXU-opt} 
\end{align}
\label{eq:UG-opt}%
\end{subequations}
and $\bfWX$ is chosen according to \eqref{eq:bfWX-opt-repeat}.

Via a symmetry argument (corresponding to interchanging the roles of
$X$ and $Y$, and $U$ and $V$, and noting that $\dtmg$ and $\dtmg^\T$
share the same singular values), it follows that
$\mse^{U|T}\bigl(\Cg^\dimX_{\epsX}(\bLa_X),\bG_{(k)}\bigr)$ is
minimized by choosing \eqref{eq:UG-opt} and
\begin{equation}
\bfWY = \bPsi^Y_{(k)},
\label{eq:bfWY-opt-repeat}
\end{equation}
and that $\mse^{V|T}\bigl(\Cg^\dimY_{\epsY}(\bLa_Y),\bG_{(k)}\bigr)$ is
minimized by choosing, for example,
\eqref{eq:VF-opt} and
\eqref{eq:bfWY-opt-repeat}.  The unique pareto optimality of the
choices then follows.   Finally, \eqref{eq:opt-config-gauss} is
obtained by combining the characterizations
\begin{equation} 
\bLa_{X U} = \epsX\,\bLa_X^{1/2}\,\bPhi^{X|U}\quad\text{and}\quad
\bLa_{Y V} = \epsY\,\bLa_Y^{1/2}\,\bPhi^{Y|V}
\label{eq:bLa-XUYV-form}
\end{equation}
with \eqref{eq:bPhiXU-opt}, \eqref{eq:bPhiYV-opt}, and \eqref{eq:FG*-k-def}.
\hfill\IEEEQED

\subsection{Proof of \protect\propref{prop:ib-double-gauss}}
\label{app:ib-double-gauss}

First, since mutual information is invariant to coordinate
transformations, without loss of generality we may choose
$\bLa_U=\bLa_V=\bI$, in which case $\Ut=U$ and $\Vt=V$. 
Then, using the conditional independencies associated with the Markov
chain \eqref{eq:markov-gauss} we have
\begin{align} 
\bLa_{UV} 
&= \ev\Bigl[\ev\bigl[U\,V^\T \bigm| \Xt,\Yt\bigr]\Bigr] \notag\\
&= \ev\Bigl[\ev\bigl[U\bigm|\Xt\bigr] \, \ev\bigl[V^\T\bigm|\Yt\bigr]\Bigr] \notag\\
&= \epsX\epsY \, \E{ \bigl(\bPhi^{X|U}\bigr)^\T \Xt \, \Yt^\T \bPhi^{Y|V}}\notag\\
&= \epsX\epsY \, \bigl(\bPhi^{X|U}\bigr)^\T \dtmg^\T \bPhi^{Y|V}.
\label{eq:UV-cov}
\end{align}
Hence, with
\begin{equation*}
\Ct \defeq \begin{bmatrix} U \\ V 
\end{bmatrix}\quad\text{so}\quad \bLa_\Ct = \begin{bmatrix} \bI &
  \bLa_{UV} \\ \bLa_{UV}^\T & \bI 
\end{bmatrix},
\end{equation*}
we have
\begin{align} 
&I(U;V) \notag\\
&\quad= -\frac12 \log\detb{\bLa_{\Ct}}\notag \\
&\quad= -\frac12 \log\bdetb{\bI - \bLa_{UV}\bLa_{UV}^\T} \notag\\
&\quad= \frac{\epsX^2\epsY^2}{2}\, \bfrob{\bigl( \bPhi^{Y|V}\bigr)^\T \, \dtmg \,
  \bPhi^{X|U}}^2 + o(\epsX^2\epsY^2) \label{eq:IUV-form} \\
&\quad\le 
\frac{\epsX^2\epsY^2}{2}\, \bfrob{\dtmg \,
  \bPhit^{X|U}}^2 \notag\\
&\qquad\qquad{}\cdot
 \bspectral{\bPhit^{Y|V}}^2\, \bspectral{\bDel^{X|U}}^2 \,
\bspectral{\bDel^{Y|V}}^2 + o(\epsX^2\epsY^2) \label{eq:use-submult-thrice}\\
&\quad\le 
\frac{\epsX^2\epsY^2}{2}
\left(\frac{\dimX+k}{k}\right)\left(\frac{\dimY+k}{k}\right) \sum_{i=1}^k \sigma_i^2 + o(\epsX^2\epsY^2),
\label{eq:IUV-bnd}
\end{align}
where to obtain \eqref{eq:IUV-form} we have used \eqref{eq:UV-cov}
with \factref{fact:logdet}, to obtain \eqref{eq:use-submult-thrice} we
have used the SVDs \eqref{eq:bPhiXU-svd} and \eqref{eq:bPhiYV-svd}
with (repeatedly) \factref{fact:frob-submult}, and to obtain
\eqref{eq:IUV-bnd} we have used both \lemref{lem:eig-k} with
\eqref{eq:PhitXU-orthog}, and both \eqref{eq:bPhiXU-constr-s} and 
\eqref{eq:bPhiYV-constr-s}.
Finally, it is straightforward to verify
that the inequality \eqref{eq:IUV-bnd} is achieved with equality when
$\bPhi^{Y|V}$ and $\bPhi^{X|U}$ are chosen according to
\eqref{eq:bPhiYV-opt} and \eqref{eq:bPhiXU-opt}, respectively, which
correspond to \eqref{eq:bLaXU-opt}--\eqref{eq:bLaYV-opt}.  With these
choices, \eqref{eq:UV-cov} specializes to
\begin{align*} 
\bLa_{UV} &=
\epsX\epsY\,\sqrt{\frac{\dimX+k}{k}}\,\sqrt{\frac{\dimY+k}{k}}\,\bigl(\bPsi^X_{(k)}\bigr)^\T\dtmg^\T 
\bPsi^Y_{(k)} \\
&= \epsX\epsY\,\sqrt{\frac{\dimX+k}{k}}\,\sqrt{\frac{\dimY+k}{k}}\,\bSi_{(k)},
\end{align*}
where we have used \eqref{eq:bSi-k-def}.
\hfill\IEEEQED

\subsection{Proof of \protect\corolref{corol:ib-double-gauss}}
\label{app:ib-double-gauss-corol}

It suffices to note from \corolref{corol:I-local}
that $U$ and $V$ so constrained correspond to
$\epsX(1+o(1))$- and $\epsY(1+o(1))$-multi-attributes with 
\begin{equation} 
\epsX \defeq \eps\, \sqrt{\frac{k}{\dimX+k}}
\quad\text{and}\quad
\epsY \defeq \eps\, \sqrt{\frac{k}{\dimY+k}}.
\label{eq:epsXY-eps}
\end{equation}
In particular, the multi-attribute property
\eqref{eq:gauss-multi-attrib-cond} specialized to $U$ 
is obtained via
\begin{equation*}
\end{equation*}
\begin{align*} 
\bspectral{\bPhi^{X|U}}^2 
&= \lambda_{\max} \bigl( \bigl(\bPhi^{X|U}\bigr)^\T \bPhi^{X|U} \bigr) \notag\\
&= \frac1{\epsX^2}\lambda_{\max} \bigl( \bLa_{XU}^\T\bLa_X^{-1}\bLa_{XU} \bigr) \notag\\
&= \max_{i\in\{1,\dots,k\}} \bnorm{\bphi^{X|U_i}}^2 \notag\\
&\le \frac{\dimX+k}{k} \bigl(1+o(1)\bigr),
\end{align*}
where to obtain the second equality we have used that $\bLa_U=\bI$,
and to obtain third equality we have used that
$\bLa_{XU}^\T\bLa_X^{-1}\bLa_{XU}$ is diagonal.  To obtain the
inequality,  note that with $\epsX$ as defined, 
\begin{align*}
I(U_i;X) = \epsX^2 \bnorm{\bphi^{X|U_i}}^2 + o(\epsX^2)
\le \eps^2
= \epsX^2\left(\frac{\dimX+k}{k}\right),
\end{align*}
whence
\begin{equation*}
\bnorm{\bphi^{X|U_i}}^2 \le \frac{\dimX+k}{k} \bigl(1+o(1)\bigr).
\end{equation*}
The corresponding result for $V$ follows from symmetry.  Finally,
substituting for $\epsX$ and $\epsY$ in the right-hand side of
\eqref{eq:IUV-opt} yields \eqref{eq:IUV-opt-eps},  
\hfill\IEEEQED

\subsection{Proof of \protect\corolref{corol:suffstat-gauss}}
\label{app:suffstat-gauss}

First, \eqref{eq:use-attrib} holds from the conditional independence
in the definition of an attribute.  Next, since the variables are
jointly Gaussian, it suffices to obtain the associated second moment
characterization.  In particular, using that
$\bPhi^{X|U}=\bPsi^X_{(k)}$ and \eqref{eq:S*-gauss} we have
\begin{align} 
\E{U|X} &= \eps\,\bigl(\bPhi^{X|U}\bigr)^\T \Xt \notag\\ 
&= \eps\, S_{(k)}^* \\ 
\intertext{and}
\bLa_{U|X} &= \bI - \eps^2 \bigl(\bPhi^{X|U}\bigr)^\T \bPhi^{X|U} \notag\\
&=
\bI - \eps^2 \bigl(\bF_{(k)}^*\bigr)^\T \bLa_X \bF_{(k)}^* \notag\\
&= (1-\eps^2)\,\bI,
\end{align}
whence \eqref{eq:PUS-form}.   And
via a symmetry argument (corresponding to interchanging the roles of
$X$ and $Y$, and $U$ and $V$), we obtain \eqref{eq:PVT-form} from
\eqref{eq:PUS-form}. 

Next,
\begin{align}
p_{U|S_{(k)}^*,T_{(k)}^*,V}\bigl(u|s_{(k)}^*,t_{(k)}^*,v\bigr) 
&= \frac{p_{U,V|S_{(k)}^*,T_{(k)}^*}\bigl(u,v|s_{(k)}^*,t_{(k)}^*\bigr)}
{p_{V|S_{(k)}^*,T_{(k)}^*}(v|s_{(k)}^*,t_{(k)}^*)} \notag\\
&= \frac{p_{U,V|X,Y}(u,v|x,y)}{p_{V|X,Y}(v|x,y)} \notag\\
&= P_{U|X}(u|x) \notag\\
&= p_{U|S_{(k)}^*}\bigl(u|s_{(k)}^*\bigr).
\end{align}
verifying \eqref{eq:US-markov}, and \eqref{eq:VT-markov} follows from
symmetry considerations.

Finally, to obtain \eqref{eq:PVS-form} we 
have
\begin{align*}
\E{V|X} 
&= \eps\, \bigl(\bPhi^{Y|V}\bigr)^\T \dtmg\, \Xt \\
&= \eps\, \bigl(\bPsi^Y_{(k)}\bigr)^\T \dtmg\, \Xt \\
&= \eps\, \bSi_{(k)}\, \bigl(\bPsi^X_{(k)}\bigr)^\T \Xt \\
&= \eps\, \bSi_{(k)}\, S_{(k)}^*,
\end{align*}
and 
\begin{align*} 
\bLa_{V|X} 
&= \bI - \eps^2\, \bigl(\bPhi^{Y|V}\bigr)^\T \dtmg\, \dtmg^\T
\bPhi^{Y|V} \\ 
&= \bI - \eps^2\, \bigl(\bPsi^Y\bigr)^\T \dtmg\, \dtmg^\T \bPsi^Y \\
&= \bI - \eps^2\, \bSi_{(k)}\,\bigl(\bPsi^X\bigr)^\T
\bPsi^X\,\bSi_{(k)} \\
&= \bI - \eps^2\, \bSi_{(k)}^2.
\end{align*}
Via a symmetry argument,  we obtain \eqref{eq:PUT-form} from
\eqref{eq:PVS-form}. 
\hfill\IEEEQED

\subsection{Proof of \protect\propref{prop:ib-single-gauss}}
\label{app:ib-single-gauss}

First, since the constraints on $U$ and $V$ coincide with those of
\corolref{corol:ib-double-gauss}, from the proof of the latter we
obtain that $U$ and $V$ are $\epsX(1+o(1))$- and
$\epsY(1+o(1))$-multi-attributes with $\epsX$ and $\epsY$ as given by
\eqref{eq:epsXY-eps}.

As such, for the maximization of $I(U;Y)$ we have
\begin{align} 
I(Y;U) 
&= I(\Yt;U) \notag\\
&= \frac{\epsX^2}{2} \bfrob{\dtmg\,\bPhi^{X|U}}^2 +
o(\epsX^2) \label{eq:use-innov}\\  
&= \frac{\epsX^2}{2} \bfrob{\dtmg\,\bPhit^{X|U} \bDel^{X|U}}^2 +
o(\epsX^2) \label{eq:use-bPhiXU-Del}\\  
&\le \frac{\epsX^2}{2} \bfrob{\dtmg\,\bPhit^{X|U}}^2
\bspectral{\bDel^{X|U}}^2 + o(\epsX^2) \label{eq:use-fact-submult}\\  
&\le \frac{\epsX^2}{2}
\bfrob{\dtmg\,\bPhit^{X|U}}^2\left(\frac{\dimX+k}{k}\right) +
o(\epsX^2) \label{eq:use-multi-attrib-prop}\\   
&\le \frac{\eps^2}{2}  \sum_{i=1}^k
\sigma_i^2 + o(\eps^2), 
\label{eq:IUYg-form}
\end{align}
where to obtain \eqref{eq:use-innov} we have used
\lemref{lem:nbhdg-frob} with \corolref{corol:I-local}, to obtain 
\eqref{eq:use-bPhiXU-Del} we have used the SVD \eqref{eq:bPhiXU-svd},
to obtain \eqref{eq:use-fact-submult} we have used
\factref{fact:frob-submult}, to obtain
\eqref{eq:use-multi-attrib-prop} we have used the
\eqref{eq:bPhiXU-constr-s}, and to obtain \eqref{eq:IUYg-form} we have
used \lemref{lem:eig-k} with \eqref{eq:PhitXU-orthog}.  Finally, it is
straightforward to verify that the inequalities all hold with equality
when $\bPhi^{X|U}$ is chosen according to \eqref{eq:bPhiXU-opt}, which
corresponds to \eqref{eq:bLaXU-opt-var}.

Via a symmetry argument (corresponding to interchanging the roles of
$X$ and $Y$, and $U$ and $V$, and noting that $\dtmg$ and $\dtmg^\T$
share the same singular values), it follows that 
\begin{equation*}
I(V;X) \le \frac{\eps^2}{2}  \sum_{i=1}^k
\sigma_i^2 + o(\eps^2), 
\end{equation*}
where the inequality holds with equality when $\bPhi^{Y|V}$ is chosen
according to  
\eqref{eq:bPhiYV-opt}, which
corresponds \eqref{eq:bLaYV-opt-var}.
\hfill\IEEEQED

\subsection{Proof of \protect\propref{prop:common-gauss}}
\label{app:common-gauss}

To simplify the exposition, we first consider the case $\dimX=\dimY=\mindim$.
First, as defined in \eqref{eq:ST*-k-gauss}, $S^*\defeq S^*_{(K)}$ and
$T^*\defeq S^*_{(K)}$ are invertible transformations of $X$ and $Y$,
respectively, with [cf.\ \eqref{eq:bLa-ST*}]
\begin{equation*}
\bLa_{S^*}=\bLa_{T^*}=\bI\quad\text{and}\quad\bLa_{T^*S^*} = \bSi
\end{equation*}
and $S^*\markov X\markov W\markov Y\markov T^*$, a special case of
which is 
\begin{equation}
S^*_i\markov W\markov T^*_i, \qquad i=1,\dots,\mindim.
\label{eq:Si*-markov}
\end{equation}

Then
\begin{align}
&I(W;X,Y) \notag\\
&\quad= I(W;S^*,T^*) \label{eq:use-invariance-again}\\
&\quad= \sum_{i=1}^\mindim I(W;S^*_i,T^*_i|(S^*)^{i-1},(T^*)^{i-1}) \label{eq:use-chain-rule1}\\ 
&\quad= \sum_{i=1}^\mindim
I(S^*_i,T^*_i;W,(S^*)^{i-1},(T^*)^{i-1})  \notag\\
&\qquad\qquad\qquad\qquad{} - \underbrace{I(S^*_i,T^*_i;(S^*)^{i-1},(T^*)^{i-1})}_{=0} \label{eq:use-XYt-indep} \\ 
&\quad= \sum_{i=1}^\mindim
I(W;S^*_i,T^*_i) +
\underbrace{I((S^*)^{i-1},(T^*)^{i-1};S^*_i,T^*_i|W)}_{\ge0} 
\label{eq:use-chain-rule1a} \\   
&\quad\ge \sum_{i=1}^\mindim
I(W;S^*_i,T^*_i) \label{eq:use-chain-rule2}  \\
&\quad= \sum_{i=1}^\mindim
I(W;S^*_i) + I(W;,T^*_i|S^*_i) \label{eq:use-chain-rule3}  \\
&\quad= \sum_{i=1}^\mindim
I(W;S^*_i) + \bigl( I(S^*_i,W;T^*_i) - I(S^*_i;T^*_i) \bigr) \label{eq:use-chain-rule4}  \\ 
&\quad= \sum_{i=1}^\mindim
I(W;S^*_i) + I(W;T^*_i) - I(S^*_i;T^*_i)  \label{eq:use-Si*-markov}\\
&\quad\ge \sum_{i=1}^\mindim
I(\bE{S^*_i|W};S^*_i) + I(\bE{T^*_i|W};T^*_i) - I(S^*_i;T^*_i)  \label{eq:use-dpi}\\
&\quad\ge \sum_{i=1}^\mindim
R_{S^*_i}(D_{S^*_i}) + R_{T^*_i}(D_{T^*_i}) - I(S^*_i;T^*_i)  \label{eq:use-rdf}\\
&\quad=\frac12 \sum_{i=1}^\mindim
\log \frac{1-\sigma_i^2}{D_{S^*_i} D_{T^*_i}},  \label{eq:use-rdf-form-mi}
\end{align}
where to obtain \eqref{eq:use-invariance-again} we have used the
invariance of mutual information to coordinate transformation, to
obtain \eqref{eq:use-chain-rule1} have used chain rule of mutual
information, to obtain \eqref{eq:use-XYt-indep} we have used the chain
rule of mutual information and note that the second term is zero since
the $S^*_i,T^*_i$ are independent of $(S^*)^{i-1},(T^*)^{i-1}$, to
obtain \eqref{eq:use-chain-rule1a} we have used the chain rule of
mutual information on the first term in \eqref{eq:use-XYt-indep}, to
obtain \eqref{eq:use-chain-rule2} we neglect the second term in
\eqref{eq:use-chain-rule1a}, to obtain \eqref{eq:use-chain-rule3} we use
the chain rule of mutual information, to obtain
\eqref{eq:use-chain-rule4} we use the chain rule of information on the
second term in \eqref{eq:use-chain-rule3}, to obtain
\eqref{eq:use-Si*-markov} we have used \eqref{eq:Si*-markov}, to
obtain \eqref{eq:use-dpi} we have used the data processing inequality,
and to obtain \eqref{eq:use-rdf} we have used the definition of the
(Gaussian) rate-distortion function (see, e.g.,
\cite[Chapter~13]{ct91}).

Now
\begin{align} 
D_{S^*_i} &= \E{\bigl(S^*_i-\bE{S^*_i|W}\bigr)^2} = 1 -
\underbrace{\E{\bE{S^*_i|W}^2}}_{\defeq\delta_{S^*_i}}\\ 
D_{T^*_i} &= \E{\bigl(T^*_i-\bE{T^*_i|W}\bigr)^2} = 1 -
\underbrace{\E{\bE{T^*_i|W}^2}}_{\defeq\delta_{T^*_i}}
\end{align}
and
\begin{align}
\sigma_i^2 
&= \E{\bE{S^*_iT^*_i|W}}^2 \notag\\
&= \E{\bE{S^*_i|W}\,\bE{T^*_i|W}}^2 \notag\\
&\le \E{\bE{S^*_i|W}^2} \, \E{\bE{T^*_i|W}^2} \notag\\
&= \delta_{S^*_i}\delta_{T^*_i},
\label{eq:deltaXY-constr}
\end{align}
where we have used the Cauchy-Schwarz inequality.

Hence, the lower bound \eqref{eq:use-rdf-form-mi} is minimized by
maximizing $\bigl(1-\delta_{S^*_i})(1-\delta_{T^*_i}\bigr)$ for each
$i\in\{1,\dots,\mindim\}$, subject to the constraint
\eqref{eq:deltaXY-constr}, which is a straightforward exercise,
yielding $\delta_{S^*_i}=\delta_{T^*_i}=\sigma_i$, whence
\begin{equation*}
I(W;X,Y) \ge 
\frac12 \sum_{i=1}^\mindim
\log\left(\frac{1+\sigma_i}{1-\sigma_i}\right). 
\end{equation*}

To show that the lower bound is achieved, choose $W$ such that $W,X,Y$
are jointly Gaussian, and let $W\in\reals^\mindim$ be zero-mean with
$\bLa_W=\bI$.  Finally, choose
\begin{equation} 
\bLa_{S^* W} = \bLa_{T^* W} =\bSi^{1/2}.
\label{eq:Wg-opt-sym}
\end{equation}
Then using \factref{fact:gm-cov-prod}, we confirm that with $W$ so
defined, $S^*\markov W\markov T^*$ since
\begin{equation*}
\bLa_{S^* W} \, \bLa_{T^* W}^\T =\bSi = \bLa_{S^*T^*}.
\end{equation*}
Finally, exploiting the resulting conditional independence structure,
we have
\begin{align*}
I(W;S^*,T^*) 
&= \sum_{i=1}^\mindim I(W_i;S^*_i,T^*_i) \\
&= \sum_{i=1}^\mindim I(W_i;S^*_i) + I(W_i;T^*_i) - I(S^*_i;T^*_i) \\
&= -\frac12 \sum_{i=1}^\mindim \Bigl[ \log (1-\la_{S^*_i W_i}) + \log
(1-\la_{T^*_i W_i}) \notag\\
&\ \quad\qquad\qquad\qquad\qquad\qquad{} + \log (1-\la_{S^*_i T^*_i}) \Bigr] \\
&= \frac12 \sum_{i=1}^\mindim
\log\frac{1-\sigma_i^2}{(1-\sigma_i)(1-\sigma_i)}\\ 
&=\frac12 \sum_{i=1}^\mindim \log\frac{1+\sigma_i}{1-\sigma_i}.
\end{align*}

The extension to the case $\dimX\ne\dimY$ is straightforward.  In
particular, when $\dimX\le \dimY$, we write
\begin{equation*}
\dtmg = \bPsi^Y_- \,\bSi_- \bigl(\bPsi^X\bigr)^\T,
\end{equation*}
where the $\dimY\times \dimX$ matrix $\bPsi^Y_-$ is formed from the first
$\dimX$ columns of $\bPsi^Y$, and where the $\dimX\times \dimX$ matrix
$\bSi_-$ is formed from the first $\dimX$ rows of $\bSi$ (the rest being
all zeros).  In the associated analysis, we then replace
\eqref{eq:Wg-opt-sym} with 
\begin{align*}
\bLa_{S^* W} & =\bSi_-^{1/2} \\
\bLa_{T^* W} & =\bSi^{1/2}.
\end{align*}
Likewise, when $\dimX> \dimY$ we write
\begin{equation*}
\dtmg = \bPsi^Y \,\bSit_- \bigl(\bPsi^X_-\bigr)^\T,
\end{equation*}
where the $\dimX\times \dimY$ matrix $\bPsi^X_-$ is formed from the first
$\dimY$ columns of $\bPsi^X$, and where the $\dimY\times \dimY$ matrix
$\bSit_-$ is formed from the first $\dimY$ columns of $\bSi$ (the rest
being all zeros).  In this
case, the associated analysis replaces
\eqref{eq:Wg-opt-sym} with 
\begin{align*}
\bLa_{S^* W} & =\bSi^{1/2} \\
\bLa_{T^* W} & =\bSit_-^{1/2}.
\end{align*}
The result in both cases can be expressed in the form 
\begin{align*} 
\bLa_{XW} &=
\bLa_X^{1/2}\,\bPsi^X_{(\mindim)}\,\bSi_{(\mindim)}^{1/2} \\
\bLa_{YW} &= 
\bLa_Y^{1/2}\,\bPsi^Y_{(\mindim)}\,\bSi_{(\mindim)}^{1/2},
\end{align*}
and, in turn,
\eqref{eq:Wg-opt}.
\hfill\IEEEQED

\subsection{Proof of \protect\corolref{corol:suffstat-common-gauss}}
\label{app:suffstat-common-gauss}

With
\begin{equation*}
A^* \defeq \begin{bmatrix} S^* \\ T^* 
\end{bmatrix}
\end{equation*}
we have, as a straightforward exercise in linear algebra,
\begin{align*}
\bE{W|X,Y} 
&= \bE{W|S^*,T^*} \notag\\
&= \bLa_{W,A^*}\,\bLa_{A^*}^{-1} A^* \notag\\
&= 
\begin{bmatrix} \bSi_{(\mindim)}^{1/2} & \bSi_{(\mindim)}^{1/2} 
\end{bmatrix}
\begin{bmatrix} \bI & \bSi_{(\mindim)} \\ \bSi_{(\mindim)} & \bI 
\end{bmatrix}^{-1} \begin{bmatrix} S^*_{(\mindim)} \\ T^*_{(\mindim)} 
\end{bmatrix} \notag\\
&= \bSi_{(\mindim)}^{1/2}
\bigl(\bI+\bSi_{(\mindim)}\bigr)^{-1} \begin{bmatrix} \bI & \bI 
\end{bmatrix} \begin{bmatrix} S^*_{(\mindim)} \\ T^*_{(\mindim)} 
\end{bmatrix},
\end{align*}
and
\begin{align*} 
\bLa_{W|X,Y} 
&= \bLa_{W|S^*,T^*} \\
&= \bI - \bLa_{W|A^*}^\T \bLa_{A^*}^{-1} \bLa_{W|A^*} \\
&= \bI - 2\bSi_{(\mindim)} \bigl(\bI+\bSi_{(\mindim)}\bigr)^{-1} \\
&= \bigl(\bI-\bSi_{(\mindim)}\bigr) \bigl(\bI+\bSi_{(\mindim)}\bigr)^{-1}.
\end{align*}
\hfill\IEEEQED

\subsection{Proof of \protect\corolref{corol:Wc-opt}}
\label{app:Wc-opt}

It suffices to verify that $\Wc$ so defined has $\bLa_{\Wc X}$ and
$\bLa_{\Wc Y}$ matching \eqref{eq:Wg-opt}, i.e., that
\begin{align} 
&\begin{bmatrix}
  \bLa_{\Wc X} & \bLa_{\Wc Y}  
\end{bmatrix} \notag\\
&\qquad= \begin{bmatrix} \bLa_{\Wc U} & \bLa_{\Wc V} 
\end{bmatrix}
\begin{bmatrix} \bI & \bLa_{UV} \\ \bLa_{UV}^\T & \bI 
\end{bmatrix}^{-1} 
\begin{bmatrix} \bLa_{UX} & \bLa_{UY} \\ \bLa_{VX} &
  \bLa_{VY} \end{bmatrix} \notag\\
&\qquad\qquad= \begin{bmatrix}
  \bLa_{W X} & \bLa_{W Y}  
\end{bmatrix}
\label{eq:equiv-cond}
\end{align}
holds.
But from \eqref{eq:opt-config-gauss-var} and \eqref{eq:UV-opt-gauss}
with $\eps=1$, and using \factref{fact:gm-cov-prod} with
\eqref{eq:cov-modal}, we have
\begin{align*} 
&\begin{bmatrix} \bLa_{UX} & \bLa_{UY} \\ \bLa_{VX} &
  \bLa_{VY} \end{bmatrix} \notag\\
&\quad\qquad= 
\begin{bmatrix} \bI & \bSi_{(\mindim)} \\ \bSi_{(\mindim)} & \bI 
\end{bmatrix} \begin{bmatrix} 
\bigl(\bF^*_{(\mindim)}\bigr)^\T\bLa_X & \bzero  \\
\bzero & \bigl(\bG^*_{(\mindim)}\bigr)^\T \bLa_Y
\end{bmatrix}
\end{align*}
and $\bLa_{UV}=\bSi_{(\mindim)}$, and
from \eqref{eq:Wg-opt} we have
\begin{equation*}
\begin{bmatrix} \bLa_{W X} & \bLa_{W Y} 
\end{bmatrix} = \bSi^{1/2}_{(\mindim)} \begin{bmatrix} 
\bigl(\bF^*_{(\mindim)}\bigr)^\T\bLa_X &
\bigl(\bG^*_{(\mindim)}\bigr)^\T \bLa_Y  \end{bmatrix}.
\end{equation*}
Hence, with the choices \eqref{eq:Wc-opt-gauss}, it follows that
 \eqref{eq:equiv-cond} holds.
\hfill\IEEEQED

\subsection{Proof of \protect\lemref{lem:match-set-gauss}}
\label{app:match-set-gauss}

First, note that 
\begin{align} 
&\E{\bnorm{V(y)-V_\circ(x)}^2} \notag\\
&\quad=\E{\tr\Bigl(\bigl(V(y)-V_\circ(x)\bigr)\bigl(V(y)-V_\circ(x)\bigr)^\T\Bigr)}\notag\\  
&\quad=\tr\bLa_{V(y)-V_\circ(x)} + \bnorm{\bE{V(y)-V_\circ(x)}}^2 \label{eq:use-cov-mse-rel}\\
&\quad = \tr\bigl(\bLa_{V|Y}\bigr) + \tr\bigl(\bLa_{V|X}\bigr) \notag\\
&\qquad\quad{} + \bnorm{\bE{V|Y=y}-\bE{V|X=x}}^2,
\label{eq:use-V-forms}
\end{align}
where to obtain \eqref{eq:use-cov-mse-rel} we have used the trivial identity
$\bE{Z Z^\T} = \bLa_Z + \bE{Z} \bE{Z}^\T$,  and to obtain
\eqref{eq:use-V-forms} we have used that $V(y)$ and $V_\circ(x)$ are
independent and distributed according to $P_{V|Y}(\cdot|y)$ and
$P_{V|X}(\cdot|x)$, respectively.

Finally, substituting \eqref{eq:PVT-form} and \eqref{eq:PVS-form} from
\corolref{corol:suffstat-gauss} into \eqref{eq:use-V-forms}, we obtain
\begin{align*} 
&\E{\bnorm{V(y)-V_\circ(x)}^2} \notag\\
&\quad=2k-k\,\eps^2 - \eps^2\sum_{i=1}^k \sigma_i^2 +
  \eps^2\,\sum_{i=1}^k \bigl( g_i(y) - \sigma_i\,f_i(x) \bigr)^2
\end{align*}
and the lemma follows.
\hfill\IEEEQED

\subsection{Proof of \protect\propref{prop:attribute-match-gauss}}
\label{app:attribute-match-gauss}

First, we rewrite \eqref{eq:match-set-gauss-rel} in the equivalent form
\begin{equation}
\bigl(\bPsi^Y_{(k)}\bigr)^\T\, \yt = \bSi_{(k)}\,
\bigl(\bPsi^X_{(k)}\bigr)^\T\, \xt.
\label{eq:match-set-gauss-rel-equiv}
\end{equation}
Next, we note that the objective function $p_Y(y)$ in
\eqref{eq:attribute-match-obj} is a monotonically decreasing function of
$\norm{\yt}$, and thus we seek to find the mininum norm solution
$\yt^*(x)$ to \eqref{eq:match-set-gauss-rel-equiv}.  Via familiar
linear algebra results---see, e.g., \cite[Problem 7.3.P9]{hj12}---the
solution follows as
\begin{align} 
\yt^*(x) 
&= \bigl(\bPsi^Y_{(k)}\bigr)^{\dagger\T}
\bSi_{(k)} \, \bigl(\bPsi^X_{(k)}\bigr)^\T \, x \notag\\ 
&= \bPsi^Y_{(k)} \bSi_{(k)}\, \bigl(\bPsi^X_{(k)}\bigr)^\T \, \xt.
\label{eq:ystar-g-est}
\end{align}
Rewriting \eqref{eq:ystar-g-est} in terms of $y$ and $x$
and using standard pseudoinverse properties then yields \eqref{eq:y*-gauss}.
\hfill\IEEEQED

\subsection{Proof of \protect\propref{prop:softmax-gauss-corol}}
\label{app:softmax-gauss-corol}

We have
\begin{align} 
D\bigl(P_{X,Y}\bigm\| P_{X,Y}^{(k)}\bigr) 
&= \bfrob{\dtmg-\dtmg^{(k)}}^2 + o(\eps^2) \label{eq:use-KL-gauss-frob}\\
&\ge \sum_{i=k+1}^\mindim \sigma_i^2 + o(\eps^2), 
\label{eq:use-truncate}
\end{align}
where to obtain \eqref{eq:use-KL-gauss-frob} we have used
\lemref{lem:KL-gauss-frob} with $\dtmg$ as defined in
\eqref{eq:dtmg-def} and
\begin{equation}
\dtmg^{(k)} \defeq \bLa_Y^{-1/2} \bLa_{YX}^{(k)} \bLa_X^{-1/2},
\label{eq:Bk-def}
\end{equation}
and to obtain \eqref{eq:use-truncate} we have used \lemref{lem:ey}
together with the fact that $\rank\bigl(\dtmg^{(k)}\bigr)\le k$ since
$\rank\bigl(\bLa_{XY}^{(k)}\bigr)\le k$ and $\bLa_X$ and $\bLa_Y$ are
positive definite.  Finally, it is straightfoward to verify that the
inequality \eqref{eq:use-truncate} holds with equality when 
$\dtmg^{(k)}=\dtmg^{(k)}_*$, with $\dtmg^{(k)}_*$ as given in
\eqref{eq:Bk*-def}. \hfill\IEEEQED

\subsection{Proof of \protect\propref{prop:mmse-low-rank}}
\label{app:mmse-low-rank}

First, with the gain matrix [cf.\ \eqref{eq:gain-matrix}]
$\bGa_{Y|X}=\bLa_{YX}\,\bLa_X^{-1}$ we rewrite the optimization in
\eqref{eq:mmse-opt} as
\begin{align} 
&\argmin_{\bGat_{Y|X}\colon\rank(\bGat_{Y|X})\le k}
\Ed{P_{X,Y}}{\bnorm{Y-\bGat_{Y|X} X}^2} \notag\\
&\qquad = \argmin_{\bGat_{Y|X}\colon\rank(\bGat_{Y|X})\le k}
\Bigl(\Ed{P_{X,Y}}{\bnorm{Y-\bGa_{Y|X} X}^2} \notag\\ 
&\qquad\qquad\qquad\qquad\qquad{}+
\Ed{P_X}{\bnorm{\bigl(\bGa_{Y|X}-\bGat_{Y|X}\bigr)X}^2}\Bigr)
\label{eq:use-mmse-orthog} \\   
&\qquad = \argmin_{\bGat_{Y|X}\colon\rank(\bGat_{Y|X})\le k} 
\Ed{P_X}{\bnorm{\bigl(\bGa_{Y|X}-\bGat_{Y|X}\bigr)X}^2} \label{eq:first-term-const}\\
&\qquad = \argmin_{\bGat_{Y|X}\colon\rank(\bGat_{Y|X})\le k} 
\bfrob{\bigl(\bGa_{Y|X}-\bGat_{Y|X}\bigr)\bLa_X^{1/2}}^2,\label{eq:GYX-form}
\end{align}
where to obtain \eqref{eq:use-mmse-orthog} we have used the orthogonal
properties of the error in the MMSE estimate, and to obtain
\eqref{eq:first-term-const} we have used that the first term does not
depend on $\bGat_{Y|X}$.

In turn, with
\begin{equation}
\bA \defeq \bGa_{Y|X} \bLa_X^{1/2} = \bLa_{YX}\,\bLa_X^{-1/2} \quad\text{and}\quad
\bAt \defeq \bGat_{Y|X} \bLa_X^{1/2},
\label{eq:bAt-defs}
\end{equation}
we can rewrite \eqref{eq:GYX-form} in the form
\begin{align} 
&\min_{\bGat_{Y|X}\colon\rank(\bGat_{Y|X})\le k} 
\bfrob{\bigl(\bGa_{Y|X}-\bGat_{Y|X}\bigr)\bLa_X^{1/2}}^2 \notag\\
&\qquad\qquad\qquad\qquad\qquad= \min_{\bAt\colon\rank(\bAt)\le k} \bfrob{\bA-\bAt}^2,
\label{eq:min-At}
\end{align}
since $\rank\bigl(\bAt\bigr)\le k$ if and only if 
$\rank\bigl(\bGat_{Y|X}\bigr)\le k$
since $\bLa_X$ is nonsingular.  Hence, 
with the SVD for $\bA$ expressed in the form \eqref{eq:mmse-svd},
it follows form \lemref{lem:ey} that the minimum on the right-hand side of
\eqref{eq:min-At} is achieved by 
the choice
\begin{equation*}
\bAt = \bPsit^Y_{(k)} \bSit_{(k)}
\bigl(\bPsit^X_{(k)}\bigr)^\T 
= \bA\, \bPsit^X_{(k)}\,\bigl(\bPsit^X_{(k)}\bigr)^\T,
\end{equation*}
i.e., using \eqref{eq:bAt-defs},
\begin{align*} 
\bGat_{Y|X} &= \bLa_{YX} \bLa_X^{-1/2}
\,\bPsit^X_{(k)}\,\bigl(\bPsit^X_{(k)}\bigr)^\T
\bLa_X^{-1/2}\\
&= \bLa_{YX} \left(\bigl( \bPsit^X_{(k)}\bigr)^\T\bLa_X^{1/2}\right)^\mppi
\,\bigl(\bPsit^X_{(k)}\bigr)^\T \bLa_X^{-1/2}\\
&= \left( \bLa_X^{-1/2}\,\bPsit^X_{(k)}\, 
\bigl(\bLa_X^{1/2}\bPsit^X_{(k)}\bigr)^\mppi \bLa_{XY} \right)^\T
\end{align*}
where we obtain the second and third equalities using standard
pseudoinverse properties.   Finally, since 
\begin{equation*}
\bLa_{XY}^{(k)\circ}=
\bLa_X \bGat_{Y|X}^\T,
\end{equation*}
we obtain \eqref{eq:mmse-est-k}.  \hfill\IEEEQED

\bibliographystyle{IEEEtran}
\bibliography{learning,gww,refs}

\end{document}